\documentclass[10pt,twocolumn,letterpaper]{article}

\usepackage[accsupp]{axessibility}  % Improves PDF readability for those with disabilities.
\usepackage{wacv}              % To produce the CAMERA-READY version
\usepackage{graphicx}
\usepackage{amsmath}
\usepackage{amssymb}
\usepackage{booktabs}

\usepackage[dvipsnames]{xcolor}

\usepackage[pagebackref,breaklinks,colorlinks]{hyperref}

\usepackage{colortbl}
\usepackage{mathrsfs}
\usepackage{makecell}
\usepackage{booktabs}
\usepackage{pgfplots}
\usepackage{tikz}
\usepackage{pgfplotstable}
\usepackage{listings}
\usepackage{xcolor} % For custom colors (optional)

\lstdefinelanguage{CUDA}{
    language=C++, % Base language
    morekeywords={__global__, __device__, __shared__, __syncthreads, threadIdx, blockIdx, blockDim, gridDim}, % CUDA-specific keywords
    sensitive=true, % Case sensitivity
}

\pgfplotsset{compat=1.18}

\definecolor{iccvblue}{rgb}{0.21,0.49,0.74}
\usepackage{multirow} % For multi-row cells
\usepackage{graphicx} % Needed for \resizebox if using for width control
\usepackage{tikz-3dplot}
\usetikzlibrary{shapes,arrows}
\usetikzlibrary{calc}
\usetikzlibrary{matrix}
\usetikzlibrary{arrows,backgrounds}
\usetikzlibrary{shapes}
\usetikzlibrary{shapes.symbols}
\usetikzlibrary{shapes.geometric}
\usetikzlibrary {positioning}
\usetikzlibrary{trees}
\usetikzlibrary{shadows,calc}
\usetikzlibrary{decorations.pathreplacing}
\usetikzlibrary{decorations.markings}

\usetikzlibrary{decorations,decorations.pathmorphing}
\usetikzlibrary{patterns}
\usetikzlibrary{3d}
\usetikzlibrary{chains}
\usetikzlibrary{intersections,through,backgrounds}

\usepackage[capitalize]{cleveref}
\crefname{section}{Sec.}{Secs.}
\Crefname{section}{Section}{Sections}
\Crefname{table}{Table}{Tables}
\crefname{table}{Tab.}{Tabs.}

\def\wacvPaperID{679} % *** Enter the WACV Paper ID here
\def\confName{WACV}
\def\confYear{2025}

\usepackage{multicol}  % For multi-column tweaks if needed

\begin{document}

%%%%%%%%% TITLE - PLEASE UPDATE
% \title{\LaTeX\ Author Guidelines for \confName~Proceedings}
\title{DARB-Splatting: Generalizing Splatting with Decaying Anisotropic Radial Basis Functions}

\author{
Hashiru Pramuditha\textsuperscript{1,*} \quad
Vinasirajan Viruthshaan\textsuperscript{1,*} \quad
Vishagar Arunan\textsuperscript{1,*} \quad
Saeedha Nazar\textsuperscript{1,*} \\
Sameera Ramasinghe\textsuperscript{2} \quad
Simon Lucey\textsuperscript{2} \quad
Ranga Rodrigo\textsuperscript{1,$\dagger$} \\
\textsuperscript{1}University of Moratuwa \quad
\textsuperscript{2}University of Adelaide \quad
% Ranga Rodrigo\textsuperscript{1,$\dagger$}
}

% Footnotes with fixed symbols
\maketitle

\begingroup
\renewcommand{\thefootnote}{\fnsymbol{footnote}}
\footnotetext[1]{These authors contributed equally to this work.}
\footnotetext[2]{Correspondence: \href{mailto:ranga@uom.lk}{ranga@uom.lk}}
\endgroup

%%%%%%%%% ABSTRACT
\begin{abstract}
Splatting-based 3D reconstruction methods have gained popularity with the advent of 3D Gaussian Splatting, efficiently synthesizing high-quality novel views.
These methods commonly resort to using exponential family functions, such as the Gaussian function, as reconstruction kernels due to their anisotropic nature, ease of projection, and differentiability in rasterization. 
However, the field remains restricted to variations within the exponential family, leaving generalized reconstruction kernels  
largely underexplored, partly due to the lack of easy integrability in 3D to 2D projections.
In this light, we show that a class of decaying anisotropic radial basis functions (DARBFs), which are non-negative functions of the Mahalanobis distance, supports splatting by approximating the Gaussian function's closed-form integration advantage. 
With this fresh perspective, we demonstrate varying performances across selected DARB reconstruction kernels, achieving comparable training convergence and memory footprints, with on-par PSNR, SSIM, and LPIPS results.

% up to 34\% faster convergence during training and a 45\% reduction in memory consumption 

% across selected DARB reconstruction kernels, while maintaining comparable PSNR, SSIM, and LPIPS results. 
% We will make the code available.
\end{abstract}

%%%%%%%%% BODY TEXT
\section{Introduction}
\label{sec:intro}

3D Gaussian Splatting (3DGS) \cite{kerbl3Dgaussians} has gained widespread attention in 3D reconstruction due to its state-of-the-art (SOTA) performance and real-time rendering capabilities. 
Since then, 3DGS has seen expanding industrial applications, spanning 3D web viewers, 3D scanning, VR platforms, and more \cite{aws_lee}.  
As demand continues to rise, optimizing models for efficiency becomes crucial. Given this, the need for resource-efficient approaches in 3DGS is becoming increasingly important.

\begin{figure}[t]
    \input{images/teaser/darbf_teaser}
    % \documentclass{standalone}
% \usepackage{pgfplots}
% \pgfplotsset{compat=1.18}
% \usepackage{tikz}

\definecolor{br}{RGB}{203,65,84}
\definecolor{nb}{RGB}{0,0,128}

%\begin{document}
\scriptsize
\begin{tikzpicture}
\begin{axis}[
    name=loss,
    xlabel={Iterations},
    ylabel={Loss},
    y label style={at={(axis description cs:0.05,1)}, anchor=south, rotate=-90},
    legend pos=north east,
    grid=both,
    width=4.5cm,
    height=4cm,
    title={}
]

% Plot for loss_cosines
\addplot[color=br, mark=none] table [x=iteration, y=loss_cosines, col sep=comma] {images/teaser/graph_data.csv};
\addlegendentry{Half cosines}

% Plot for it_s_cosines
\addplot[color=nb, mark=none] table [x=iteration, y=loss_gaussians, col sep=comma] {images/teaser/graph_data.csv};
\addlegendentry{Gaussians}

\end{axis}

\begin{axis}[
    at={(loss.east)}, anchor=west,
    xshift=1cm, % Optional: Adjust spacing between the two axes    
    name=it_per_sec,
    xlabel={Iterations},
    ylabel={Iterations/s},
    y label style={at={(axis description cs:0.05,1)}, anchor=south, rotate=-90},
    legend pos=north east,
    grid=both,
    width=4.5cm,
    height=4cm,
    title={}
]

% Plot for loss_cosines
\addplot[color=br, mark=none] table [x=iteration, y=it_s_cosines, col sep=comma] {images/teaser/graph_data.csv};
\addlegendentry{Half cosines}

% Plot for it_s_cosines
\addplot[color=nb, mark=none] table [x=iteration, y=it_s_gaussians, col sep=comma] {images/teaser/graph_data.csv};
\addlegendentry{Gaussians}

\end{axis}

\end{tikzpicture}
% \end{document}
    % \caption{Representation of two different footprint (splat) functions (Gaussian and cosine) initialized for the same 3D covariance matrix $\mathbf{\Sigma}$. Cosine functions have finite support. Both functions effectively reconstruct; cosine functions yield better results in terms of training speed, leading to memory efficiency. Here cosines refer to half-cosine squares, $\cos\left(d_M^2/\xi\right),\;d_M^2<\xi \pi/2$.} 
    \caption{This figure compares two different splat functions---Gaussian and cosine (specifically, half-cosine)---initialized for the same 3D covariance matrix \( \mathbf{\Sigma} \), confined by eigenvalues $ \lambda_1, \lambda_2, \lambda_3$. Half-cosine functions, defined as \( \cos\left(\frac{d_M^2}{\xi}\right) \) for \( d_M^2 < \frac{\xi \pi}{2} \) where $\xi>0$, have finite support and exhibit effective reconstruction performance comparable to Gaussians. Notably, this non-exponential-based approach enhances training speed by 34\% and reduces memory usage by 15\%, providing a more efficient alternative.}
    \vspace{-10pt}
    \label{fig:teaser_dig}    
\end{figure}

% Paragraph 2: SOTA
Splatting plays a pivotal role in modern 3D reconstruction, enabling the representation of 3D points without relying on surface meshes. 
3DGS, which is a splatting-based technique,  represents these 3D points as a dynamically densified radiance field of 3D Gaussian primitives (ellipsoids) that act as reconstruction kernels. 
In theory, these ellipsoids are integrated along the projection direction onto the 2D image plane, resulting in 2D Gaussians or splats, a function known as the footprint function or splatting function. 
This process exploits an interesting property of Gaussians: $2\times2$ covariance matrix of the 2D Gaussian can be obtained by skipping the third row and column of the $3\times 3$ covariance matrix of the 3D Gaussian \cite{EWA_Splatting}. 
This bypasses costly integration and efficiently simplifies the rendering in practical implementation. 
The footprint function then models the spread of each splat's opacity, eventually contributing to the final pixel color.
This Gaussian-based blending effect—a smooth, continuous interpolation between splats—inspires the method’s name, ``3D Gaussian Splatting." Numerous subsequent papers \cite{Yu2024MipSplatting, lee2024c3dgs, chen2024mvsplat, chen2023gaussianeditorswiftcontrollable3d, ye2024gaussiangroupingsegmentedit} build upon this principle, although still restricted to exponential family functions (\eg, Gaussian functions) to spread each sample in image space.

Splatting methods depend on the mean (position) and covariance of 3D points, making them independent of the reconstruction kernel.
However, existing work commonly resort to the Gaussian kernel, with some variations within the exponential family (\eg, super-Gaussian \cite{Ges}, half-Gaussian \cite{3D-HGS}, Gaussian-Hermite \cite{2DGH} kernels).
However, classical signal processing and sampling theory establish that while Gaussians are useful, there exist other effective interpolators. 
In this paper we argue that the Gaussian function is not the only possible interpolator, especially when determining the pixel color, as in 3DGS. 
We see other functions outperforming them in  various contexts, with a notable example being JPEG compression \cite{125072, 952804}, which leverages Discrete Cosine Transforms (DCTs) instead of Gaussians functions for image representation. 
Additionally, Saratchandran \emph{et al.} \cite{saratchandran2024samplingtheoryperspectiveactivations} demonstrate that the $\mathrm{sinc}$ activation surpasses $\mathrm{Gaussian}$ activation in implicit neural representations \cite{mildenhall2020nerf, INRAS, INR_medical_image_segmentation, Hua_2024_CVPR}, while also suggesting several alternative activation functions. 
This raises the question: \emph{Should splatting be limited only to Gaussians or the exponential family?} This remains an underexplored area in the community, partly because costly integration processes make it computationally inefficient to use functions outside of Gaussians. 

To address these shortcomings, we generalize these kernels by introducing a broader class of functions: \textbf{Decaying Anisotropic Radial Basis Functions (DARBFs)}. These non-negative functions (\eg, modified raised cosine, half-cosine, modular sinc functions) include, but are not limited to the exponential family. 
DARBFs support anisotropic behavior as they rely on the Mahalanobis distance and are differentiable, enabling differentiable rendering (Sec.~\ref{ss:prelim_rbf}). 
Our work demonstrates the potential of moving beyond Gaussians in optimizing splatting by showcasing how selected functions contribute to training speedups (\eg, half-cosine function in Table~\ref{Table:FinalResultTable} and Fig.~\ref{fig:teaser_dig}) and memory efficiencies (\eg, inverse quadratic function in Table~\ref{Table:FinalResultTable}), while maintaining on-par visual quality in novel views.
In a field that often focuses on refining loss functions \cite{eggs, high-fold_loss}, regularizing Gaussian parameters \cite{regularization, fregs}, quantizing Gaussian parameters \cite{compGS}, we take a deeper approach by analyzing the fundamental aspects of splatting. We replace the Gaussian itself by providing a plug-and-play replacement for kernels across the DARBF spectrum.
Since our approach targets a fundamental aspect of the algorithm, it can be applied to all splatting methods built on top of the 3DGS framework.

To address the costly integration beyond Gaussians, we introduce a unique correction factor for different alternative kernels, while preserving the computational efficiency of 3DGS by approximating the Gaussian's closed-form integration shortcut. This enables feasible splatting across the diverse functions in DARBFs.

To the best of our knowledge, our method represents one of the first modern generalizations expanding splatting to non-exponential functions, while maintaining comparable rendering quality, as splatting techniques scale to repetitive industrial applications, where computational savings are increasingly critical. 

% Paragraph 5: Evaluation, proof
The contributions of our paper are as follows:
\begin{itemize}
    % \item Presenting unifying approach, where the Gaussian function is merely an instance of the broader class of DARBFs, which can be splatted. 
    % \item We present a unified approach to generalizing splatting functions across the broad class of DARBFs, where the Gaussian function is a special case within the class. 
    \item We present a generalization of splatting functions using the DARBFs class, with Gaussians as a special case.
    % \item Improved editability through alternative functions, such as cosine.
    % \item Minimum 0.02 dB difference in PSNR over the original 3DGS.
    % \item Derivations to modify the reconstruction kernel
    % \item Reduction in training time and storage with comparable performance across different DARBFs.
    % \item We showcase the impact of expanding beyond the traditional Gaussian family by using selected DARBF reconstruction kernels, achieving up to 34\% faster convergence and a 15\% reduction in memory footprint, while maintaining on-par PSNR (Sec.~\ref{sec:exp_intro}).
    \item Selected DARBF kernels offer up to 34\% faster convergence and 15\% lower memory use, with similar PSNR (Sec.~\ref{sec:exp_intro}).
    % \item We introduce a computationally feasible method using our correction factor to approximate the Gaussian's closed-form integration advantage when implementing alternative kernels, along with CUDA-based backpropagation codes (\emph{Supplementary Material}).
    \item We propose an efficient method aided through a correction factor to retain Gaussian integration benefits for other kernels, supported by CUDA-based backpropagation codes (\emph{Supplementary Material}).
    % \item CUDA-based backpropagation codes to support DARBFs efficiently
    % \item \textcolor{magenta}{Shall we mention about correlation factor here. ``A Novel Correlation factor which aligns our Integration with existing Gaussian's Closed under integration property."}
    % \textcolor{cyan}{better to mention correction factor}

\end{itemize}

\section{Related Work}
\label{sec:related_work}

% All text must be in a two-column format.
% The total allowable size of the text area is $6\frac78$ inches (17.46 cm) wide by $8\frac78$ inches (22.54 cm) high.
% Columns are to be $3\frac14$ inches (8.25 cm) wide, with a $\frac{5}{16}$ inch (0.8 cm) space between them.
% The main title (on the first page) should begin 1 inch (2.54 cm) from the top edge of the page.
% The second and following pages should begin 1 inch (2.54 cm) from the top edge.
% On all pages, the bottom margin should be $1\frac{1}{8}$ inches (2.86 cm) from the bottom edge of the page for $8.5 \times 11$-inch paper;
% for A4 paper, approximately $1\frac{5}{8}$ inches (4.13 cm) from the bottom edge of the
% page.

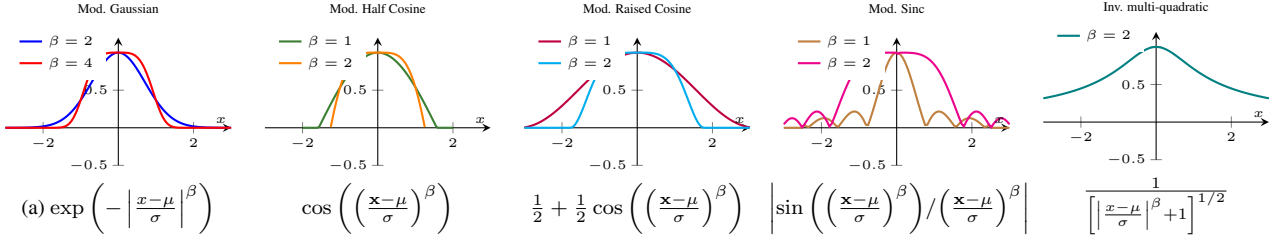
\begin{figure*}
        \tiny
    \begin{subfigure}[b]{0.195\textwidth}
        \centering
        \begin{tikzpicture}
            \begin{axis}[
                x = 0.5cm, y = 1cm,
                domain=-3.0:3.0,
                samples=100,
                ymin=-0.5, ymax=1.2,
                axis lines=middle, % Removes the box and shows only axes
                xlabel={$x$}, 
                ylabel={Mod. Gaussian},
                y label style={
                    at={(axis description cs:0.5,1.15)}, % Places ylabel at the top center
                    anchor=south
                },                   
                legend entries={$\beta=2$, $\beta=4$},
                legend style={at={(0.02,1.1)}, anchor= north west, legend columns=1, draw=none, font=\tiny},
                legend image post style={xscale=0.5},                
            ]
            \addplot[blue, thick] {exp(-abs(x) ^ 2)};
            \addplot[red, thick] {exp(-abs(x) ^ 4)};
            \end{axis}
        \end{tikzpicture}
        \caption{ $\exp \left( -  \left| \frac{x - \mu}{\sigma} \right| ^{\beta} \right)$ }
    \end{subfigure}       
    \begin{subfigure}[b]{0.195\textwidth}
        \centering
        \begin{tikzpicture}
            \begin{axis}[
                x = 0.5cm, y = 1cm,
                domain=-2.0:2.0,
                samples=100,
                ymin=-0.5, ymax=1.2,
                xmin=-3.0, xmax=3.0,
                axis lines=middle, % Removes the box and shows only axes
                xlabel={$x$}, 
                ylabel={Mod. Half Cosine},
                y label style={
                    at={(axis description cs:0.5,1.15)}, % Places ylabel at the top center
                    anchor=south
                },                   
                legend entries={$\beta=1$, $\beta=2$},
                legend style={at={(0.02,1.1)}, anchor= north west, legend columns=1, draw=none, font=\tiny},
                legend image post style={xscale=0.5},                  
            ]
            \addplot[draw=OliveGreen, thick] {cos(deg(abs(x))) * (abs(x) <= pi/2)};
            \addplot[orange, thick, domain=-1.25:1.25] {cos(deg(abs(x) ^ 2)) * (abs(x) <= (pi/2)^(1/2))};
            \end{axis}
        \end{tikzpicture}
        \caption*{$\cos \left( \left(  \frac{\mathbf{x - \mu}}{\sigma}  \right)^{\beta} \right)$}
    \end{subfigure}   
    \begin{subfigure}[b]{0.195\textwidth}
        \centering
        \begin{tikzpicture}
            \begin{axis}[
                x = 0.5cm, y = 1cm,
                domain=-3.0:3.0,
                samples=100,
                ymin=-0.5, ymax=1.2,
                axis lines=middle, % Removes the box and shows only axes
                xlabel={$x$}, 
                ylabel={Mod. Raised Cosine},
                y label style={
                    at={(axis description cs:0.5,1.15)}, % Places ylabel at the top center
                    anchor=south
                },                   
                legend entries={$\beta=1$, $\beta=2$},
                legend style={at={(0.02,1.1)}, anchor= north west, legend columns=1, draw=none, font=\tiny},
                legend image post style={xscale=0.5},                  
            ]
            \addplot[purple, thick] {(0.5 + 0.5 * cos(deg(abs(x)))) * (abs(x) <= pi)};
            \addplot[cyan, thick] {(0.5 + 0.5 * cos(deg(abs(x) ^ 2))) * (abs(x) <= pi^(1/2))};
            \end{axis}
        \end{tikzpicture}
        \caption*{$\frac{1}{2} + \frac{1}{2}\cos \left( \left( \frac{\mathbf{x - \mu}}{\sigma}  \right)^{\beta} \right)$}
    \end{subfigure} 
    \begin{subfigure}[b]{0.195\textwidth}
        \centering
        \begin{tikzpicture}
            \begin{axis}[
                x = 0.5cm, y = 1cm,
                domain=-3.0:3.0,
                samples=100,
                ymin=-0.5, ymax=1.2,
                axis lines=middle, % Removes the box and shows only axes
                xlabel={$x$}, 
                ylabel={Mod. Sinc},
                y label style={
                    at={(axis description cs:0.5,1.15)}, % Places ylabel at the top center
                    anchor=south
                },                   
                legend entries={$\beta=1$, $\beta=2$},
                legend style={at={(0.02,1.1)}, anchor= north west, legend columns=1, draw=none, font=\tiny},
                legend image post style={xscale=0.5},                  
            ]
            \addplot[brown, thick] {abs(sin(deg(4*x))/ (4*x)) * (abs(x) <= 2.5)};
            \addplot[magenta, thick] {abs(sin(deg(abs(x) ^ 2)))/(abs(x) ^ 2) * (abs(x) <= 3)};
            \end{axis}
        \end{tikzpicture}
        \caption*{$ \left|{\sin \left( \left(  \frac{\mathbf{x - \mu}}{\sigma} \right)^{\beta} \right)}/{\left(  \frac{\mathbf{x - \mu}}{\sigma} \right)^{\beta}}\right|$}
    \end{subfigure}      
    \begin{subfigure}[b]{0.195\textwidth}
        \centering
        \begin{tikzpicture}
            \begin{axis}[
                x = 0.5cm, y = 1cm,
                domain=-3.0:3.0,
                samples=100,
                ymin=-0.5, ymax=1.2,
                axis lines=middle, % Removes the box and shows only axes
                xlabel={$x$}, 
                ylabel={Inv. multi-quadratic},
                y label style={
                    at={(axis description cs:0.5,1.1)}, % Places ylabel at the top center
                    anchor=south
                },                
                legend entries={$\beta=2$, $\beta=4$},
                legend style={at={(0.02,1.1)}, anchor= north west, legend columns=1, draw=none, font=\tiny},
                legend image post style={xscale=0.5},                  
            ]
                \addplot[teal, thick] {1 / sqrt((abs(x) ^ 2) + 1)};
            \end{axis}
        \end{tikzpicture}
        \caption*{$\frac{1}{\left[\left| \frac{{x - \mu}}{\sigma} \right|^{\beta} + 1 \right]^{1/2}}$}
    \end{subfigure}
    \caption{Overview of decaying anisotropic radial basis functions (DARBFs) shown with their respective 1D curves. These
    functions decay with distance and vary in their sensitivity to direction, making them effective for capturing anisotropic features in spatial data.}
    \vspace{-7pt}
    \label{fig:darbf_plots}
\end{figure*}

\textbf{Radiance Field Representation. }Early novel view synthesis (NVS) methods leveraged light fields  \cite{light_field_rendering, lumigraph}, later enhanced by SfM \cite{sfm} and pipelines like COLMAP \cite{COLMAP} to obtain sparse reconstructions from image collections \cite{mildenhall2020nerf, kerbl3Dgaussians, mip-splatting, Instruct-nerf2nerf}.

Neural Radiance Fields (NeRFs) \cite{mildenhall2020nerf} enabled photorealistic novel view synthesis via volumetric rendering \cite{Mip-NeRF, Mip-nerf_360, Zip-nerf, Block-nerf, Nerf_in_the_dark, NeRF_in_the_Wild}, leading to extensive improvements in speed \cite{NSVF, Autoint, Derf, Kilonerf, Fastnerf}, supervision \cite{depth_prior_nerfs, Depth-supervised_nerf}, deformation \cite{Hypernerf, D-nerf, Humannerf, Nerfies}, and editing \cite{Graf, Editnerf, Fenerf, yang2021learning}.In contrast, 3D Gaussian Splatting (3DGS) adopts point-based rendering using anisotropic splats and has rapidly impacted diverse domains \cite{chen2023gaussianeditorswiftcontrollable3d, GaussianDreamer, 3dgs_survey, lei2024gart, Dynmf, huang2024point, DreamGaussian, shao2023control4d, shi2023languageembedded3dgaussians, sun20243dgstream, DreamGaussian4D, 4dgs, xie2024physgaussian, ye2024gaussiangroupingsegmentedit, GS-SLAM, LucidDreamer, CoGS}. Inspired by this, our method also uses point-based rendering but generalizes the opacity function beyond 3D Gaussians, introducing DARBFs as flexible image-space primitives, following ideas in \cite{saratchandran2024samplingtheoryperspectiveactivations}.

% %-------------------------------------------------------------------------
\textbf{Splatting. } Splatting, introduced by Westover \cite{interactive_volume_rendering}, enables point-based rendering of 3D data, such as sparse SfM points \cite{sfm} obtained from the COLMAP pipeline \cite{COLMAP}, without requiring mesh-like connectivity. Each particle’s position and shape are represented by a volume \cite{interactive_volume_rendering} that serves as a reconstruction kernel \cite{EWA_Splatting}, which is projected onto the image plane through a ``footprint function,” or ``splat function,” to spread, or splat each particle's intensity and color across a localized region. Early splatting techniques \cite{footprint_evaluation} commonly used spherical kernels for their radial symmetry to simplify calculations, although they struggled with perspective projections. To address this, elliptical kernels \cite{footprint_evaluation, EWA_Splatting}, projecting elliptical footprints on the image plane were used to better approximate elongated features through anisotropic properties, ultimately enhancing rendering quality. Westover \cite{footprint_evaluation}, further experimented with different reconstruction kernels, namely $\mathrm{sinc}$, $\mathrm{cone}$, $\mathrm{Gaussian}$, and $\mathrm{bilinear}$ functions without focusing on a specific class of functions. EWA Splatting \cite{EWA_Splatting} further refined this approach by selecting a Gaussian reconstruction kernel in 3D, which projects as an elliptical Gaussian footprint on the image plane. This motivated 3DGS to adopt Gaussians. While effective, other anisotropic radial basis functions could serve as splats but remain underexplored.
\textbf{Reconstruction Kernel Modifications.} Although 3DGS claims SOTA performance, recent work has shown further improvements by refining the reconstruction kernel. For instance, GES \cite{Ges} introduces the generalized exponential function, or the super Gaussian, by incorporating a learnable shape parameter for each point.
% However, GES does not modify the splatting function directly within the CUDA rasterizer; instead, it only approximates its effects by adjusting the scaling matrix and loss function in PyTorch, failing to fully demonstrate its superiority \cite{2DGH}. 
% In contrast, we directly modify the CUDA rasterizer's splatting function to support various DARBFs, achieving superior performance in specific cases compared to 3DGS.
Concurrent work such as 3D-HGS \cite{3D-HGS}, splits the 3D Gaussian reconstruction kernel into two halves by modeling two different opacity distributions for the same splat.
% but this approach introduces additional parameters into the CUDA rasterizer, leading to increased computational costs. 
Similarly, 2DGH \cite{2DGH} extends the Gaussian function by incorporating Hermite polynomials into a Gaussian-Hermite reconstruction kernel. 
% Although this kernel more sharply captures edges, it incurs a high memory overhead due to the increased number of parameters per primitive.
Meanwhile, triangle-based approaches \cite{Held2025Triangle}, such as Triangle Splatting, influenced by \cite{held20243dconvex}, represent scenes using explicit triangle primitives that can be rendered as differentiable splats and optimized via end-to-end gradients. Similarly, customizing differentiable rasterizer for sparse voxel rendering \cite{svraster} has been done to avoid popping artifacts in 3DGS \cite{kerbl3Dgaussians} and to achieve high rendering frame rates.
 These methods achieve impressive results in terms of quality, training time and memory footprint, but limiting to one particular class of reconstruction kernel. In our method, we preserve the general framework of point-based splatting while enhancing its expressiveness through Decaying Anisotropic Radial Basis Functions (DARBFs). By directly modifying the CUDA rasterizer, we enable a class of flexible, learnable reconstruction kernels without increasing memory overhead or requiring handcrafted priors, opening new avenues for future research. \\
\vspace{-2pt}
\textbf{Memory and Training Efficiency in Gaussian Splatting.} The unstructured nature of Gaussians in 3DGS increases memory and training costs. Prior work addresses this with refined splatting \cite{compGS}, quantization via sensitivity-aware clustering \cite{Niedermayr_2024_CVPR}, and adaptive spherical harmonics \cite{10.1145/3651282}. In contrast, we pursue kernel-level modifications, offering plug-and-replace compatibility with existing methods and opening a new path for optimization.
\section{Preliminaries}
\subsection{DARB Functions}
\label{ss:prelim_rbf}

Radial basis functions (RBFs) serve as a fundamental class of mathematical functions, which depend on the distance from a center point. However, isotropic RBFs, which are radially symmetric, often fall short in capturing anisotropic features such as local geometric details that vary directionally, resulting in inaccuracies \cite{CASCIOLA20061185}.

To address this issue, we use Anisotropic Radial Basis Functions (ARBFs) with a decaying nature, namely Decaying ARBFs (DARBFs), as we are interested in splatting, where splats decay spatially. For a particular point \(\mathbf{x} \in \mathbb{R}^n\), the Mahalanobis distance (\(d_M\)), calculated from the center \(\mathbf{\mu} \in \mathbb{R}^n\) of a 3D DARBF, is given by:
\vspace{-6pt}
\begin{equation} 
\label{eq:mahalanobis}
d_M(\mathbf{x}; \boldsymbol{\mu}, \mathbf{\Sigma}) = \left[(\mathbf{x - \mu})^T \mathbf{\Sigma}^{-1} (\mathbf{x - \mu})\right]^{\frac{1}{2}},
\end{equation}
\noindent where \(\mathbf{\Sigma}\) denotes the covariance matrix. This radially dependent anisotropy supports smooth interpolation in 3 dimensional space (n=3), playing a pivotal role in 3D reconstruction.

\subsection{Assessing DARBFs through 1D Simulations}
\label{sec:AssessDARB}

\begin{figure}[ht]
    \centering
    
    % Row 1: Gaussians
    \begin{subfigure}{0.235\textwidth}
        \includegraphics[height=1.8cm, width=4cm]{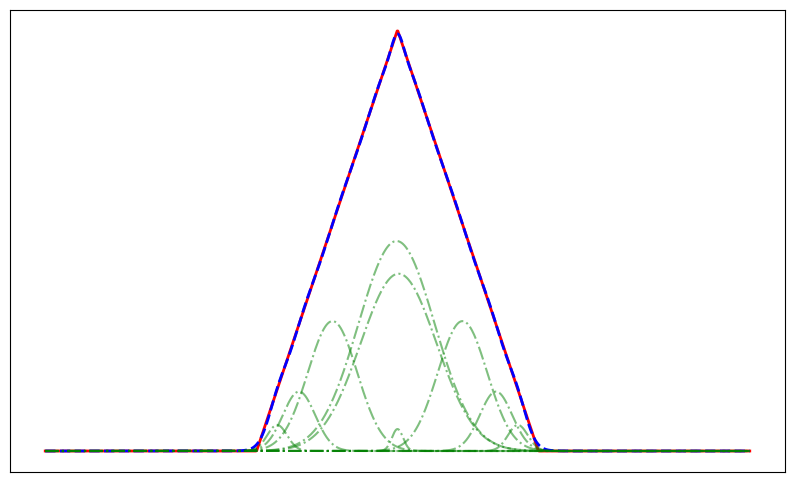}
        \caption{\scriptsize Gaussian ($N=10$) \\ Loss = 0.001}
    \end{subfigure} \hfill
    \begin{subfigure}{0.235\textwidth}
        \includegraphics[height=1.8cm, width=4cm]{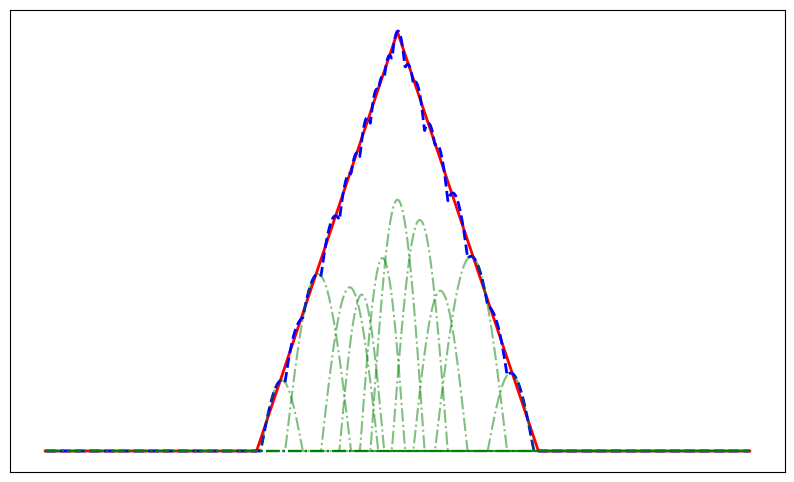}
        \caption{\scriptsize Half Cosine ($N=10$) \\ Loss = 0.003}
    \end{subfigure} \hfill

    % Row 2: Half Cosines
    \begin{subfigure}{0.235\textwidth}
        \includegraphics[height=1.8cm, width=4cm]{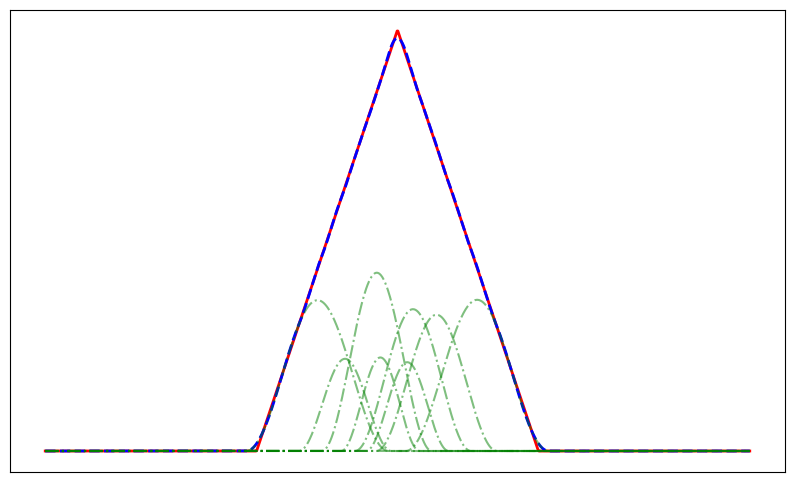}
        \caption{\scriptsize Raised Cosine ($N=8$) \\ Loss=0.0004}
    \end{subfigure}
    \begin{subfigure}{0.235\textwidth}
        \includegraphics[height=1.8cm, width=4cm]{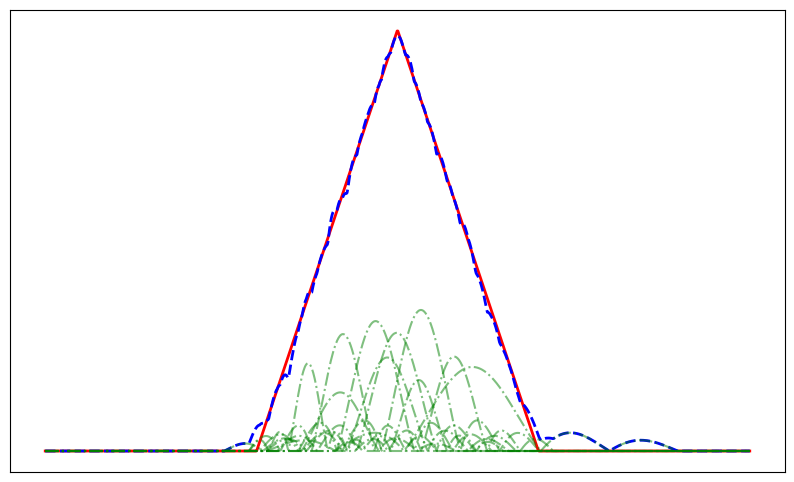}
        \caption{\scriptsize Sinc ($N=10$) \\ Loss = 0.002}
    \end{subfigure}
    
    % % Row 3: Raised Cosines
    % \begin{subfigure}{0.235\textwidth}
    %     \includegraphics[height=1.8cm, width=4cm]{images/reconstruction/triangle_raised_cosines.png}
    %     \caption{Raised Cosine ($N=5$)}
    % \end{subfigure}
    % \begin{subfigure}{0.235\textwidth}
    %     \includegraphics[height=1.8cm, width=4cm]{images/reconstruction/triangle_raised_cosines_absolute.png}
    %     \caption{Raised Cosine ($N=8$)}
    % \end{subfigure}
    
    % % Row 4: Sincs
    % \begin{subfigure}{0.235\textwidth}
    %     \includegraphics[height=1.8cm, width=4cm]{images/reconstruction/triangle_sincs.png}
    %     \caption{Sinc ($N=9$)}
    % \end{subfigure}
    % \begin{subfigure}{0.235\textwidth}
    %     \includegraphics[height=1.8cm, width=4cm]{images/reconstruction/triangle_sincs_absolute.png}
    %     \caption{Sinc ($N=10$)}
    % \end{subfigure}

    % Overall Figure Caption
    \vspace{-10pt}
    
    \caption{Simulations of 1D signal reconstruction through backpropagation across different DARBFs. Here, the target, reconstructed, and individual components are represented by red, blue, and green lines, respectively. We achieve competitive reconstruction results, validating our hypothesis that signals can be reconstructed using non-exponential kernels.}
    \vspace{-7pt}
    \label{fig:enter-label}
\end{figure}

As a simple proof of concept, we conducted 1D signal reconstruction simulations (Fig.~\ref{fig:enter-label}). Our objective was to optimize the minimum number of primitives required for reconstructing a complex shape (\eg, triangle in this case) while fine-tuning the mean and variance values to minimize the loss, computed as the difference between the target and reconstructed functions. 
The simulation results indicate that, like Gaussian functions, non-exponential functions (DARBFs) can also achieve similar performance.
% certain RBFs\textcolor{cyan}{better to mention DARBFs}, can perform comparably or even outperform Gaussians in specific cases. 
% Notably, raised cosines achieved lower reconstruction loss than Gaussian primitives. 
% These preliminary findings suggest promising alternatives to Gaussian functions and motivate further exploration of these ARBFs.
Building on this insight, we incorporated selected DARBFs into the 3DGS algorithm with adjustments in domain constraints and backpropagation. We have provided a detailed analysis of these 1D simulations, including the results for reconstructing far more complex signals, in the \emph{Supplementary Material}.

% You must include your signed IEEE copyright release form when you submit your finished paper.
% We MUST have this form before your paper can be published in the proceedings.

% Please direct any questions to the production editor in charge of these proceedings at the IEEE Computer Society Press:
% \url{https://www.computer.org/about/contact}.

% %------------------------------------------------------------------------
\section{Methodology}

\begin{figure*}[ht]
    \centering
    \input{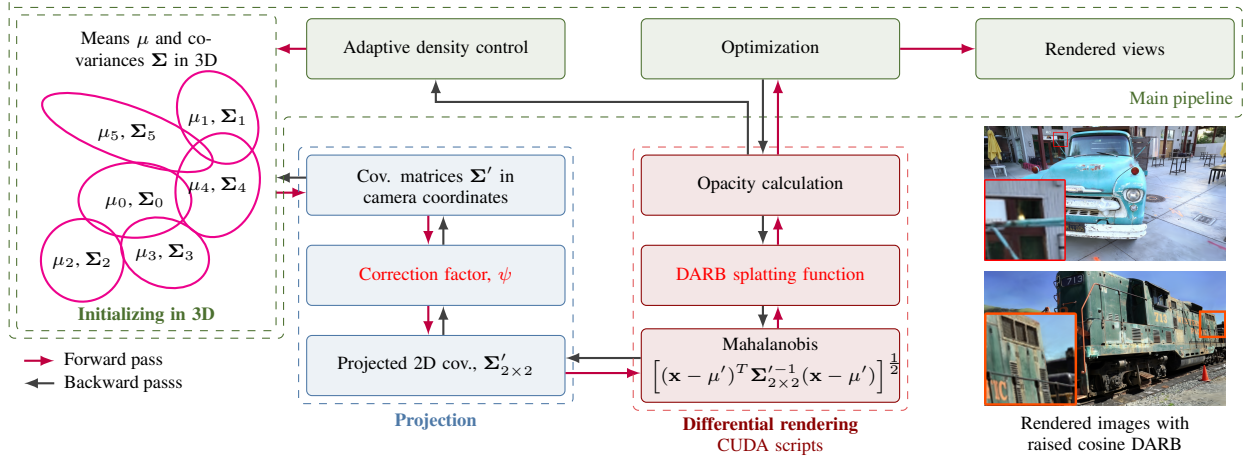}
    \caption{\textbf{Block diagram:} We use the standard optimization pipeline from 3DGS \cite{kerbl3Dgaussians} with modifications, introducing a correction factor ($\psi$) to obtain the projected 2D covariance matrix ($\Sigma'_{2 \times 2}$) compatible with splatting for DARBFs within the existing framework. Our changes within the pipeline are highlighted in red text.}
    \vspace{-10pt}
    \label{fig:DARBSPipeline}
\end{figure*}

We present DARBFs as a plug-and-play replacement for existing Gaussian kernels, offering improvements in training time and memory efficiency. The pipeline for DARB splatting is presented in Fig.~\ref{fig:DARBSPipeline}.

Splatting-based reconstruction methods like 3DGS represent scenes by placing anisotropic Gaussian kernels at 3D points from an SfM pipeline, such as COLMAP. These kernels are transformed into the camera frame and projected onto 2D planes, creating splats. The pixel error between views and rasterized splats drives backpropagation to optimize the 3D Gaussians. This pipeline relies on four key properties: (1) differentiable rasterization, (2) anisotropic kernel behavior, (3) easy projectability, and (4) rapid decay. 

A reconstruction kernel being a function of Mahalanobis distance $d_M$ (Eq.~\ref{eq:mahalanobis}), yields the first two properties \cite{Lassner_pulsar}. Non-negative DARBFs, being rapidly decaying and a function of the Mahalanobis distance  $d_M$ ~\cite{bishop2006pattern}, satisfy all four. 
In Sec. \ref{sec:DARBSplatting} we propose a method to approximate the projection of DARBFs, thereby satisfying the third property. The DARBF class can be explicitly defined as:
\begin{equation}
\label{eq:darbf_class}
K(\mathbf{x}; \boldsymbol{\mu}, \mathbf{\Sigma})
=
F\!\left(d_M(\mathbf{x}; \boldsymbol{\mu}, \mathbf{\Sigma})\right)
\end{equation}
Here, \(F:\mathbb{R}_{\ge 0}\to\mathbb{R}_{\ge 0}\) is a decaying, differentiable (finite- or fast-decay) radial profile.

\subsection{DARB-Splatting}
\label{sec:DARBSplatting}

\textbf{DARB Representation. }Similar to 3DGS,
% which represents each 3D Gaussian using a set of parameters---a 3D mean ($\mu$) for its center point, a 3D covariance matrix ($\Sigma$) to describe its spread and orientation, an opacity value to control transparency, and spherical harmonic coefficients to encode lighting and colour information \cite{kerbl3Dgaussians}---
we use the same parameterization to represent 3D DARBFs, as the Gaussian function is an instance of the DARBF class. 
In Table~\ref{tab:functions}, we present the generic mathematical expressions of selected reconstruction kernel functions.

\begin{table}[h!]
    % \label{tab:equations}
    \centering
    \scriptsize
    % \caption{Overview of decaying anisotropic radial basis functions (DARBFs) in 3D, with their limits (Sec.~\ref{sec:DARBSplatting}). Incorporating the Mahalanobis distance \(d_M\) (Eq. \ref{eq:mahalanobis}) enables precise control over each function’s main lobe, similar to 3D Gaussians, for capturing localized 3D features. The scaling factor $\xi(>0)$ controls the spread of the covariance, while \(\beta\) controls the main lobe's roll-off. Here, ``modified" refers to the generalized version (\eg, modified half-cosine with generic $\beta$ and half-cosine $\beta=2$)}
    \resizebox{\columnwidth}{!}{ % Resize table to fit column width
    \begin{tabular}{@{}llr@{}}
        \toprule
        \textbf{Function  $F(d_M)$} & \textbf{Expression} & \textbf{Domain} \\
        \midrule
        Modified Gaussian &  $\exp \left( -\frac{1}{\xi} \left(  d_M  \right)^{\beta} \right)$ & $d_M \geq 0 $ \\
        \midrule
        Modified half-cosine & $\cos \left( \frac{1}{\xi}\left(  d_M  \right)^{\beta} \right) $ & \makecell[c]{$\frac{1}{\xi} \left(  d_M  \right)^{\beta} \leq \frac{\pi}{2}$} \\
        \midrule
        Modified raised cosine & $ 0.5 + 0.5 \cdot \cos \left( \frac{1}{\xi} \left(  d_M  \right)^{\beta} \right)$ & $\frac{1}{\xi} \left(  d_M  \right)^{\beta} \leq  {n \pi}$ \\
        \midrule
        Modified modular sinc & \renewcommand{\arraystretch}{3} $\dfrac{\left| \sin\left( \frac{1}{\xi} \left(  d_M  \right)^{\beta} \right) \right|}{\frac{1}{\xi} \left(  d_M  \right)^{\beta}}$& $\frac{1}{\xi} \left(  d_M  \right)^{\beta} \leq  \frac{(n+ 1) \pi}{2}$  \\
        \midrule
        Inv. multiquadric & $\dfrac{1}{ \left[\frac{1}{\xi} \left(  d_M  \right)^{2} + 1 \right]^{\frac{1}{2}}} $ & $d_M \geq 0$ \\
                \midrule

         Parabola & \renewcommand{\arraystretch}{3} $ 1 -  \frac{1}{\xi} \left(  d_M  \right)^2 $& $\frac{1}{\xi} \left(  d_M  \right)^2 \leq  1 $  \\
        \bottomrule
    \end{tabular}
    }
    \vspace{-10pt}
    \caption{Overview of decaying anisotropic radial basis functions (DARBFs) $F(d_M)$ (Eq.   \ref{eq:darbf_class}) in 3D, with their limits (Sec.~\ref{sec:DARBSplatting}). Incorporating the Mahalanobis distance \(d_M\) (Eq.  \ref{eq:mahalanobis}) enables precise control over each function’s main lobe, similar to 3D Gaussians, for capturing localized 3D features. The scaling factor $\xi(>0)$ controls the spread of the covariance, while \(\beta\) controls the main lobe's roll-off. Here, ``modified" refers to the generalized version (\eg, modified half-cosine with generic $\beta$ and half-cosine $\beta=2$)}
    \label{tab:functions}
\end{table}

\noindent\textbf{DARB Projection. }To project 3D DARBs to DARB-splats, we use the method proposed by EWA Splatting \cite{EWA_Splatting, ye2023mathematicalsupplementtextttgsplatlibrary}. To project 3D mean ($\mu$) in world coordinate system onto the 2D image plane, we use the conventional perspective projection. However, the same projection cannot be used to project the 3D covariance ($\Sigma$) in world coordinates to the 2D covariance ($\Sigma'_{2 \times 2}$) in image space. Hence, first, the 3D covariance ($\Sigma$) is projected onto camera coordinate frame which results in a projected $3 \times 3$ covariance matrix ($\Sigma'$). Given a viewing transformation $W$, the projected covariance matrix $\Sigma'$ in camera coordinate frame can be obtained as 
$
    \Sigma' = J W \Sigma W^T J^T
$, 
where $J$ denotes the Jacobian of the affine approximation of the projective transformation. Importantly, integrating a normalized 3D Gaussian along one coordinate axis results in a normalized 2D Gaussian itself. From that, the EWA Splatting  \cite{EWA_Splatting} shows that the 2D covariance matrix ($\Sigma'_{2 \times 2}$) of the 2D Gaussian can be easily obtained by taking the upper-left $2\times 2$ sub-matrix of $\Sigma'$, specifically by skipping the third row and third column (Eq.~\ref{eq:2x2covProjection}).

 %%%%%%%%%%%%%%%%%%%%%%%55555
% % \textcolor{cyan}{we canmove this paragraph to supplementry if spaceneeded.}
However, this sub-matrix shortcut applies only to Gaussian reconstruction kernels due to their inherent properties. Our core argument questions whether the integration described above is essential for the splatting process.
While the Gaussian’s closed form integration is convenient, we propose interpreting the 3D covariance as a parameter that represents the 2D covariances from all viewing directions.
When viewing a 3D Gaussian from a particular direction, we may think 
that the opacity would be distributed across the volume defined by each 3D Gaussian, and that alpha blending composites these overlapping volumes. 
But in reality, the 3D Gaussian is splatted into 2D and then only the opacity is distributed. 
In practice, however, the opacity distribution takes place after projecting 3D Gaussians to 2D splats. Thus, when a ray approaches, it only considers the Gaussian value derived from the 2D covariance of the splat.
Additionally, the Gaussian value is not calculated for each point in the splat using \textbf{normalized} Gaussian function.
As a simple example, consider two 3D Gaussians that are nearly identical in all properties except for the variance along the z-axis ($(\sigma'_z)^2$). When these two Gaussians are projected into 2D, we expect the 3D Gaussian with a higher $\sigma'_z$ to have greater opacity values. In contrast, 3DGS yields a similar matrix ($\Sigma'_{2 \times 2}$) for both covariances, while exerting the same influence on opacity distribution along the splats. This further strengthens our argument for treating $\Sigma'$ as a parameter to represent $\Sigma'_{2 \times 2}$.
%%%%%%%%%%%%%%%%%%%%%%

Generally, integrating a 3D DARB along a certain axis does not yield the same DARBF in 2D, nor does it result in a closed-form solution.
Therefore, we established the relationship between $\Sigma'$ and $\Sigma'_{2 \times 2}$ for DARBFs through simulations. 
For clarity, we provide the derivations for the half-cosine splat here. Detailed derivations and corresponding CUDA scripts for selected DARBFs are provided in the \emph{Supplementary Material}. The 3D half-cosine kernels is as follows:
% \vspace{-10pt}
\begin{equation}
P_{i,j,k} = 
\begin{cases} 
\cos\left( \frac{\pi}{18} (d_M)^2 \right), & \text{if } \ (d_M)^2 < 9, \\
0, & \hfill \text{otherwise}.
\end{cases}
\end{equation}
where $d_M$ (Eq.~\ref{eq:mahalanobis}) is the Mahalanobis distance calculated between the kernel centers, retrieved from the point cloud as $P_{i,j,k}$ and the surrounding points in the 3D space. We then calculate the projected $\Sigma'_{2 \times 2}$ by summing values along a certain axis, similar to integrating along the same axis. 
% Then we find the total summation of the values along one axis and projected the summation to the splatting axis.
% After the projection into 2D, we find the covariance matrix $\hat{\Sigma'}_{2 \times 2}$ .
As $\Sigma_{3 \times 3}$ is positive semi-definite to maintain strict decomposition, it is a symmetric matrix; $\Sigma'_{2 \times 2}$ are also symmetric.
\vspace{-6pt}
\begin{equation}
\Sigma' = 
\begin{pmatrix}
a & b & c \\
b & d & e \\
e & d & f \\
\end{pmatrix}
\rightarrow
\psi \begin{pmatrix}
a & b \\
b & d \\
\end{pmatrix}
= \Sigma'_{2 \times 2}.
\label{eq:2x2covProjection}
\end{equation}

\noindent\textbf{Correction Factor ($\psi$).} $\psi$ accommodates our DARBFs to work efficiently with the existing 3DGS pipeline. We acknowledge that most of the functions are not closed under integration like Gaussians. To approximate this closed-form integration, we formulate an optimization problem that minimizes the difference between the covariance proposed by EWA Splatting \cite{EWA_Splatting}, and the actual projected covariance, which we compute using Simpson's rule \cite{burden2010numerical} along an axis.
% Proposed by EWA splatting 
We solve this problem using a multilayer perceptron (MLP) and integrate these precomputed 
% primitive-specific consts
% mention psi to all 
% simpson's rule algo also mention 
% we have given freedom to these approaches and seen that a simple const works
weights into the CUDA pipeline.

Thus, incorporating the influence of \( \psi \) allows us to approximate the projected covariance matrix \( \Sigma'_{2 \times 2} \), aligning it with the expected covariance and achieving comparable visual quality. However, this introduces a slight computational overhead. Therefore, we further evaluated this approach as a regression problem to approximate a pre-computed scalar \( \psi \) (derived in the \emph{Supplementary Material}) that can efficiently represent these weights without compromising visual performance. We have extensively ablated this. %(Sec. ~\ref{sec:ablation}).

\begin{table}[h]
    \label{tab:params}
    \centering
    \scriptsize 
    \renewcommand{\arraystretch}{1.2} % Adjust row height
    \setlength{\tabcolsep}{10pt} % Adjust column spacing
    \begin{tabular}{ccccccc}
        \hline
        \textbf{Kernel} & \textbf{G} & \textbf{HC} & \textbf{RC} & \textbf{Sinc} & \textbf{IMQ} & \textbf{P} \\
        \hline
        \textbf{$\xi$} & 2 & 18/$\pi$ & 2.5/$\pi$ & 3/$\pi$ & 1 & 9 \\
        \hline
        \textbf{\( \psi \)} & 1 & 1.34 & 0.66 & 1.18 & 1.61 & 1.3 \\
        \hline
    \end{tabular}
    \vspace{-10pt}
    \caption{Correction factor (\( \psi \)) values and selected scaling factor ($\xi$) values for DARBF primitives. Here, G: Gaussian, HC: Half-cosine, RC: Raised Cosine, Sinc: modular sincs, IMQ: Inverse Multiquadric, and P: Parabolic functions (Table.~\ref{tab:functions}).}
    \label{tab:kernels_cf}
\end{table}

Similarly, based on such empirical results, we modified the existing CUDA scripts of 3DGS to implement each DARBF separately with its respective $\psi$ value to ensure comparable results. Additionally, we introduce a scaling factor $\xi$ to match the extent of our function closely with the Gaussian, allowing a fair comparison of DARBF performance. Although Gaussian functions are spatially infinite, 3DGS \cite{kerbl3Dgaussians} uses a bounded Gaussian to limit the opacity distribution in 2D, thereby reducing unnecessary processing time. For projected 2D splats, they calculate the maximum radius as $R = 3 \cdot \sqrt{\max\{\lambda_1, \lambda_2\}}
 $ , where $\lambda_1$ and $\lambda_2$ denote the eigenvalues of the 2D covariance matrix $\Sigma'_{2 \times 2}$. Using this radius \(R\), we obtain a similar extent for DARBFs as Gaussians with the help of $\xi$. Taking that into consideration, for $\xi = \frac{18}{\pi}$, the 2D half-cosine splat is defined as follows:
% \vspace{-6pt}
\begin{equation}
    % w = \cos\left( \frac{ \left( x - \boldsymbol{\mu'} \right)^T \boldsymbol(\Sigma'_{2 \times 2})^{-1} \left(x - \boldsymbol{\mu'} \right)}{\xi} \right),
    w = F(d_M) = \cos\left( \frac{1}{\xi} (d_M)^2 \right),
\label{eq:HalfCosine}
\end{equation}
\vspace{-10pt}
\begin{equation}
    d_M = \left[(\mathbf{x - \mu})^T \boldsymbol(\Sigma'_{2 \times 2})^{-1} (\mathbf{x - \mu})\right]^{\frac{1}{2}},
\label{2DSplattedFunction}
\end{equation}

where \(x \in \mathbb{R}^2\) is the position vector, \(\mu' \in \mathbb{R}^2\) is the projected mean,  and \(\Sigma'_{2 \times 2} \in \mathbb{R}^{2 \times 2}\) represents the covariance matrix.
Here, \(R_{\text{cosine}}\) will be \(\frac{\sqrt{2 \pi \xi \max(\lambda_1, \lambda_2)}}{2},\) indicating 100\% extent, as we consider the entire central lobe. However, for a 2D Gaussian with an extent of 99.7\% (up to three standard deviations), the radius \(R_{\text{Gaussian}}\) is given by \(3\sqrt{\max(\lambda_1, \lambda_2)}.\) To match the extent of both functions, we select $\xi(>0)$ such that \(\frac{\sqrt{2 \pi \xi \max(\lambda_1, \lambda_2)}}{2} = 3 \sqrt{\max(\lambda_1, \lambda_2)}\), which implies \(\xi = \frac{18}{\pi}\). 

Within this bounded region, we use our footprint function (Eq.~\ref{eq:HalfCosine}) to model the opacity, similar to the approach in 3DGS. Following this, we apply alpha blending to determine the composite opacity  \cite{kerbl3Dgaussians, ye2023mathematicalsupplementtextttgsplatlibrary} for each pixel, eventually contributing to the final color of the 2D rendered image.

\subsection{Backpropagation and Error Calculation}

To support backpropagation (BP), we must account for the differentiability of DARBFs. As usual, the BP process begins with a loss function that combines the same losses discussed in the original work \cite{kerbl3Dgaussians}: an $\mathcal{L}_1$ loss and an  Structural Similarity Index Measure (SSIM) loss, which guide the 3D DARBs to optimize its parameters. 
% similar to 3DGS \cite{kerbl3Dgaussians}. The final loss function is defined as:}
% \vspace{-10pt}
\begin{equation}
\mathcal{L} = (1 - \lambda) \mathcal{L}_1 + \lambda \mathcal{L}_{\text{D-SSIM}}
\end{equation}
where $\lambda$ denotes each loss term's contribution to the final image loss. After computing the loss, we perform backpropagation, where we explicitly modify the differentiable Gaussian rasterizer from 3DGS \cite{kerbl3Dgaussians} to support respective gradient terms for each DARBF.

%%%%%%%%%%%%%%%%
The following example demonstrates these modifications for the 3D half-cosine function (Eq.~\ref{eq:HalfCosine}) with $\xi = \frac{18}{\pi}\ $.

\begin{equation}
\footnotesize
% \frac{dw}{d (x - \boldsymbol{\mu'} )} = \frac{-2 \, (\Sigma'_{2 \times 2})^{-1} \, (x - \mu') \, \sin\left( \frac{\left( x - \boldsymbol{\mu'} \right)^T (\Sigma'_{2 \times 2})^{-1} \left( x - \boldsymbol{\mu'} \right)}{\xi} \right)}{\xi}
\frac{dw}{d (x - \boldsymbol{\mu'} )} = F'(d_M)\frac{d(d_M)}{d(x - \boldsymbol{\mu'})}
\end{equation}
% \vspace{-10pt}
\begin{equation}
\footnotesize
% \frac{dw}{d ((\Sigma'_{2 \times 2})^{-1} )} = \frac{- (x - \mu') \, \left( x - \boldsymbol{\mu'} \right)^T \, \sin\left( \frac{\left( x - \boldsymbol{\mu'} \right)^T (\Sigma'_{2 \times 2})^{-1} \left( x - \boldsymbol{\mu'} \right)}{\xi} \right)}{\xi}
\frac{dw}{d ((\Sigma'_{2 \times 2})^{-1} )} = F'(d_M)\frac{d(d_M)}{d((\Sigma'_{2 \times 2})^{-1})}
\end{equation}

\vspace{1pt}

% \scriptsiz
% \textcolor{red}{Where $F'(d_M) = - \sin\left(\frac{\left( x - \boldsymbol{\mu'} \right)^T (\Sigma'_{2 \times 2})^{-1} \left( x - \boldsymbol{\mu'} \right)}{\xi} \right)$, $\frac{d_M}{d ((\Sigma'_{2 \times 2})^{-1} )} = \frac{(x - \mu') \, \left( x - \boldsymbol{\mu'} \right)^T}{\xi}$ and $\frac{d_M}{d (x - \boldsymbol{\mu'} )} = \frac{2 \, (\Sigma'_{2 \times 2})^{-1} \, (x - \mu')}{\xi}$ }}

Where
\begin{equation}
\scriptsize
F'(d_M) = - \sin\!\left(\frac{(x - \boldsymbol{\mu'})^T (\Sigma'_{2 \times 2})^{-1} (x - \boldsymbol{\mu'})}{\xi} \right)
\end{equation}
\vspace{-10pt}
\begin{equation}
\scriptsize
\frac{\partial d_M}{\partial (\Sigma'_{2 \times 2})^{-1}} 
= \frac{(x - \boldsymbol{\mu'})(x - \boldsymbol{\mu'})^T}{\xi}, \quad
\frac{\partial d_M}{\partial (x - \boldsymbol{\mu'})} 
= \frac{2 (\Sigma'_{2 \times 2})^{-1}(x - \boldsymbol{\mu'})}{\xi}
\end{equation}

%%%%%%%%%

\begin{table*}[ht!]
\centering
\scriptsize
\setlength{\tabcolsep}{3pt}
\renewcommand{\arraystretch}{1.1}
\begin{tabular}{@{}lcccccccccccccccr@{}}
\toprule
\multirow{2}{*}{\textbf{Function}} & \multicolumn{5}{c@{}}{\textbf{Mip-NeRF 360} } & \multicolumn{5}{c}{\textbf{Tanks \& Temples} } & \multicolumn{5}{c}{\textbf{Deep Blending}} \\
\cmidrule(lr){2-6}
\cmidrule(lr){7-11}
\cmidrule(lr){12-16}
 & SSIM$\uparrow$ & PSNR$\uparrow$ & LPIPS$\downarrow$ & Train & Memory & SSIM$\uparrow$ & PSNR$\uparrow$ & LPIPS$\downarrow$ & Train & Memory & SSIM$\uparrow$ & PSNR$\uparrow$ & LPIPS$\downarrow$ & Train & Memory \\
\midrule
3DGS (7k) \cite{kerbl3Dgaussians}    & 0.770 & 25.60 & 0.279 & 4m 43s$^*$ &523 MB &0.767 & 21.20 & 0.28 & 5m 05s$^*$ &270 MB & 0.875 & 27.78 & 0.317 & 3m 22s$^*$ &368 MB\\
3DGS (30K) \cite{kerbl3Dgaussians}  & 0.815 & 27.21 & 0.214 & 30m 33s$^*$ &734 MB & 0.841 & 23.14 & 0.183 & 19m 46s$^*$ &411 MB & 0.903 & 29.41 & 0.243 & 26m 29s$^*$ & 676 MB\\
\hline
GES \cite{Ges}   & 0.794 & 26.91 & 0.250 & 19m 9s &377 MB & 0.836 & 23.35 & 0.198 & 15m 26s$^*$ &222 MB & 0.901 & 29.68 & 0.252 & 22m 44s$^*$ &399 MB\\
\hline
3DGS-updated (7K) & \cellcolor{orange!35}0.769 & \cellcolor{orange!35}25.95 & \cellcolor{orange!35}0.281 & 3m 24s & 504 MB & \cellcolor{orange!35}0.781 & \cellcolor{orange!35}21.78 & \cellcolor{orange!35}0.261 & 2m 02s & 293 MB& \cellcolor{orange!35}0.879 & \cellcolor{orange!35}28.42 & \cellcolor{orange!35}0.305 & 3m 14s & 462 MB\\
3DGS-updated (30K)& \cellcolor{red!25}0.813 & \cellcolor{red!25}27.45 & \cellcolor{orange!35}0.218 & 19m 06s & 633 MB& \cellcolor{orange!35}0.847 & \cellcolor{red!25}23.77 & \cellcolor{orange!35}0.173 & 11m 24s & 371 MB & \cellcolor{orange!35}0.901 & \cellcolor{red!25}29.66 & \cellcolor{orange!35}0.242 & 20m 12s & 742 MB\\
\midrule
Raised cosine (7K) & \cellcolor{red!25}0.773 & \cellcolor{red!25}26.01 & \cellcolor{red!25}0.273 & \cellcolor{yellow!25}3m 13s& 513 MB& \cellcolor{red!25}0.788 & \cellcolor{red!25}21.88 & \cellcolor{red!25}0.251 & \cellcolor{yellow!25}1m 59s & 304 MB & \cellcolor{red!25}0.882 & \cellcolor{red!25}28.47 & \cellcolor{red!25}0.300 & \cellcolor{yellow!25}2m 57s & 401 MB \\
Raised cosine (30K)& \cellcolor{red!25}0.813 & \cellcolor{red!25}27.45 & \cellcolor{red!25}0.214 & 19m 36s & 645 MB & \cellcolor{red!25}0.851 & \cellcolor{orange!35}23.64 & \cellcolor{red!25}0.166 & 12m 06s & 364 MB & \cellcolor{orange!35}0.901 & \cellcolor{orange!35}29.63 & \cellcolor{red!25}0.240 & 20m 01s & 767 MB \\
\midrule
Half cosine (7K) & 0.737 & 25.46 & 0.316 &  \cellcolor{red!25}2m 52s & \cellcolor{orange!35}400 MB & 0.749 & 21.07 & 0.295 &  \cellcolor{red!25}1m 42s & \cellcolor{yellow!25}244 MB & 0.872 & 27.938 & 0.320 &  \cellcolor{red!25}2m 42s & \cellcolor{yellow!25}355 MB \\
Half cosine (30K)& 0.790 & 27.04 & 0.247 &  \cellcolor{orange!35}16m 15s & \cellcolor{orange!35}524 MB & 0.824 & 23.11 & 0.201 &  \cellcolor{red!25}9m 11s & 338 MB & 0.900 & 29.38 & 0.253 &  \cellcolor{red!25}17m 35s & \cellcolor{yellow!25}634 MB \\
\midrule
Modular Sinc (7K) & \cellcolor{yellow!25}0.751 & \cellcolor{yellow!25}25.74 \cellcolor{orange!20}& 0.302 & 3m 20s & 439 MB & \cellcolor{yellow!25}0.767 & \cellcolor{yellow!25}21.59 & \cellcolor{yellow!25}0.276 & 2m 04s & 255 MB & \cellcolor{yellow!25}0.878 & \cellcolor{yellow!25}28.39 & \cellcolor{yellow!25}0.308 & 3m 17s & 399 MB \\
Modular Sinc (30K)& \cellcolor{orange!35}0.801 & \cellcolor{orange!35}27.30 & 0.234 & 18m 50s & 622 MB & \cellcolor{yellow!25}0.836 & \cellcolor{yellow!25}23.51 & 0.189 & 16m 08s & \cellcolor{yellow!25}333 MB & \cellcolor{orange!35}0.901 & \cellcolor{red!25}29.66 & 0.247 & 20m 47s & 682 MB \\
\midrule
Parabola (7K) & 0.7460 & 25.63 & 0.3080 & \cellcolor{orange!35}3m 8s & \cellcolor{yellow!25}423 MB & 0.7602 & 21.33 & 0.2818 & \cellcolor{orange!35}1m 48s & \cellcolor{orange!35}241 MB & 0.8773 & 28.26 & 0.3126 & \cellcolor{orange!35}2m 51s & \cellcolor{orange!35}347 MB \\
Parabola (30K)& \cellcolor{yellow!25}0.7971 & \cellcolor{yellow!25}27.15 & 0.2399 & \cellcolor{yellow!25}17m 23s & \cellcolor{yellow!25}542 MB & 0.831 & 23.28 & 0.1935 & \cellcolor{yellow!25}10m 24s & \cellcolor{orange!35}317 MB & \cellcolor{red!25}0.902 & \cellcolor{yellow!25}29.53 & 0.2489 & \cellcolor{yellow!25}18m 57s & \cellcolor{orange!35}516 MB \\
\midrule
Inverse MQ (7K) & 0.7443 & 25.25 & \cellcolor{yellow!25}0.3008 & 3m 19s & \cellcolor{red!25}281 MB & 0.76 & 21.365 & 0.2765 & 2m 6s & \cellcolor{red!25}175 MB & 0.874 & 28.09 & 0.3095 & 3m 10s & \cellcolor{red!25}248 MB \\
Inverse MQ (30K) & 0.7932 & 26.82 & \cellcolor{yellow!25}0.2258 & \cellcolor{red!25}16m & \cellcolor{red!25}365 MB & 0.8305 & 23.343 & \cellcolor{yellow!25}0.1875 & \cellcolor{orange!35}10m 13s & \cellcolor{red!25}223.5 MB & \cellcolor{yellow!25}0.9005 & 29.504 & \cellcolor{yellow!25}0.2435 & \cellcolor{orange!35}18m 10s & \cellcolor{red!25}364 MB \\
\midrule
\bottomrule
\end{tabular}
\caption{
% This table presents a comparative analysis of various methods across different datasets. 
\textbf{Splatting results.}
The best values are highlighted in red and second-best values in orange and the third-best values in yellow. Note that, $^*$ denotes the difference in computation power that is used to implement our model and previous models (3DGS \cite{kerbl3Dgaussians}, GES \cite{Ges}), hence we used a speed-up factor \emph{(see Supplementary Materials)}, to obtain fair results compared to the training time in the original publications. Our DARBFs were implemented based on the \emph{updated} 3DGS codebase. 
%Note that, even though 3DGS has shown state-of-the-art results, 
Notice, half-cosine shows a significant reduction in training time, while inverse multiquadric shows a significant reduction in memory footprint.
Additionally, the raised cosine kernel achieves modest gains in PSNR, SSIM, LPIPS and training duration.  
We evaluate the performance of each model by considering the results at 7K and 30K iteration checkpoints (here, Inverse MQ means inverse multiquadric function and Parabola means \( 1 - \frac{1}{9}(d_M)^2 \)).}
\vspace{-10pt}
\label{Table:FinalResultTable}
% \textcolor{orange}{RR: State what the reader must conclude by going through this table.}}
\end{table*}

% % ------------------------------------------------------------------------
\section{Experiments}
\label{sec:exp_intro}

\textbf{Experimental Settings.} We anchor our contributions on the recently \emph{updated codebase} of 3DGS as of \emph{October 2024} \cite{kerbl3Dgaussians}, adjusting the CUDA scripts to support a range of DARBFs.
% We implemented our method using the code provided by 3DGS \cite{kerbl3Dgaussians}.
For a fair comparison, we use the same scenes as the 3DGS paper, including both bounded indoor and large outdoor environment scenes from various datasets \cite{Mip-nerf_360, Tanks&Temples2017, DeepBlending2018}. We utilized the same COLMAP \cite{COLMAP} initialization provided by the official dataset for all tests conducted to ensure fairness. Furthermore, we retained the original hyperparameters, adjusting only the opacity learning rate to 0.02 for improved results. All experiments and evaluations were conducted, and further verified on a single NVIDIA GeForce RTX 4090 GPU.

We compare our work with SOTA splatting-based methods that use exponential family reconstruction kernels, such as the \emph{original} 3DGS, \emph{updated codebase} of 3DGS, and GES \cite{Ges} as benchmarks. We also evaluated the DARBFs using the standard and frequently used PSNR, LPIPS and SSIM metrics similar to these benchmarks (Table~\ref{Table:FinalResultTable}). 
%%%%%%%%%%%%%%%%%%%%%%%%%%%%%%%%%%%%%%%%%%%%%%%%%

\subsection{Quantitative Results and Discussion}

% \begin{figure*}
%      % Column headings
%     \parbox{0.24\textwidth}{\centering {Ground Truth}} 
%     \parbox{0.24\textwidth}{\centering {Raised cosines (Ours)}} 
%     \parbox{0.24\textwidth}{\centering {3DGS}} 
%     \parbox{0.24\textwidth}{\centering {Half-cosine (Ours)}} \\
%     \includegraphics[width=0.24\textwidth]{images/evaluation/truck_gt.png} 
%     \includegraphics[width=0.24\textwidth]{images/evaluation/truck_rc.png} 
%     \includegraphics[width=0.24\textwidth]{images/evaluation/truck_gs.png} 
%     \includegraphics[width=0.24\textwidth]{images/evaluation/truck_cs.png}
    
%     \caption{\textbf{Rendered views. } We compare our proposed methods (raised cosine and half-cosine) with 3DGS baseline using the TRUCK scene from the Tanks \& Temples \cite{Tanks&Temples2017} dataset. Our approach shows modest improvements, such as the appearance of a light bulb in the Raised Cosine rendering, which is absent in the Gaussian rendering. While improved reconstruction is beyond our scope, we achieve significantly better training and memory efficiency without compromising visual quality (see Table \ref{Table:FinalResultTable}).} 
%     \label{fig:VisualComparisons}
% \end{figure*}

\begin{figure}[h]
    \centering
    \begin{subfigure}[b]{0.45\linewidth}
        \centering
        \includegraphics[width=\linewidth]{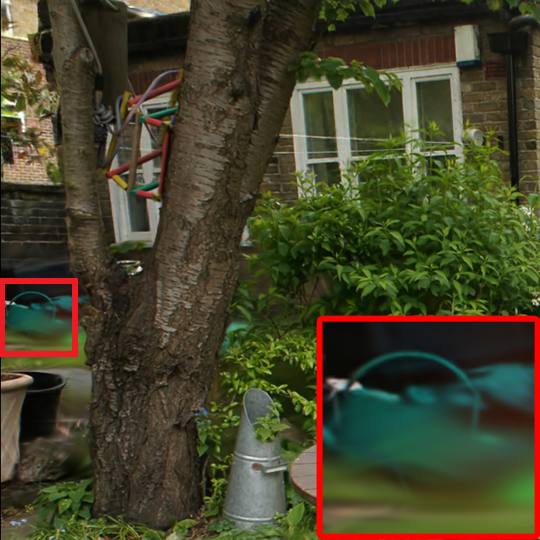}
        \caption{The famous blurred, blunt-peaked artifacts of Gaussians due to its sharp peak.
}
        \label{fig:cosine_sample}
    \end{subfigure}
    \hfill
    \begin{subfigure}[b]{0.45\linewidth}
        \centering
        \includegraphics[width=\linewidth]{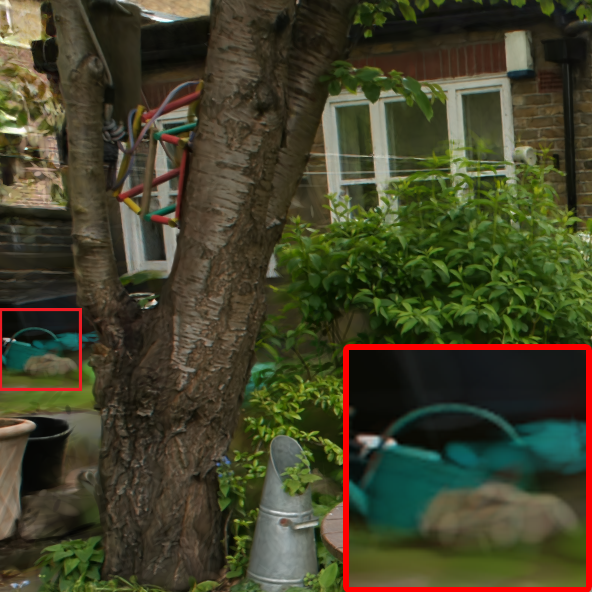}
        \caption{Round-shaped artifacts of half-cosines due to its lengthy blunt peak.}
        \label{fig:another_image}
    \end{subfigure}
    \caption{Comparision of rendered views between different DARBFs from the Garden scene (Mip-NeRF 360 dataset). This aligns with the inherent shape of half-cosines due to its lengthy blunt peak as graphically modeled in Fig.~\ref{fig:teaser_dig} \& ~\ref{fig:gaussian_cosine}, helps reduce artifacts caused by the needle-shaped Gaussians in certain cases.}
    \vspace{-10pt}
    \label{fig:side_by_side}
\end{figure}

Table~\ref{Table:FinalResultTable} shows that different functions showcase different capabilities in terms of training and memory efficiency. 
% various aspects, such as training time, memory efficiency, and visual quality. 
We do not aim for superior reconstruction gains. 
However, results in Table~\ref{Table:FinalResultTable} show that the modified raised cosine function ($\xi = \frac{2.5}{\pi}\ $) achieves modestly improved performance compared to Gaussians, particularly in LPIPS and SSIM metrics (visual comparisons are available in \emph{Supplementary Material}). 
Meanwhile, half-cosines, modular sincs, and parabolic functions achieve on-par PSNR, SSSIM, LPIPS results. This also demonstrates that certain DARBFs, despite not belonging to the exponential family, can compete effectively with previous benchmarked work \cite{Ges}. \\
\noindent\textbf{Reduced memory footprint.} Table~\ref{Table:FinalResultTable} shows that the inverse multiquadric (Inverse MQ) function reduces memory usage by over a significant 45\%, while half-cosines achieve a 14\% reduction on average compared to Gaussians. Fig.~\ref{fig:gaussian_cosine} illustrates this with 1D plots of half-cosine (\(\beta=2\)
%,\(\xi=36\) 
for modified half-cosine in Table~\ref{tab:functions}), Gaussians, and Inverse MQ. A single primitive with the same variance spans a larger region with higher opacity in cosine and Inverse MQ functions, reducing the need for multiple splats to achieve the same accumulated value.
% \textcolor{red}{Histogram plots in Fig.~\ref{fig:opacity_histograms} further compare Gaussians and half-cosines, showing that Gaussians result in a higher number of high-opacity primitives, leading to increased memory usage.} 
This is further illustrated by the visual comparison of artifact shapes in the Gaussian and half-cosine reconstructed scenes shown in Fig.~\ref{fig:side_by_side}. A similar conclusion applies to Inverse MQ.

\begin{center}
    \begin{tikzpicture}
        \begin{axis}[
            x = 0.5cm, y = 1cm,
            domain=-5.0:5.0,
            samples=100,
            ymin=0, ymax=1.2,
            axis lines=middle, % Removes the box and shows only axes
            xlabel={$x$}, 
            y label style={
                at={(axis description cs:0.5,1.15)}, % Places ylabel at the top center
                anchor=south
            },                   
            legend entries={Gaussian, Half-cosine , Inverse MQ},
            legend style={at={(0.2,1.1)}, anchor=north east, legend columns=1, draw=none, font=\scriptsize},
            legend image post style={xscale=0.5},                  
        ]
        % Gaussian function
        \addplot[brown, thick] {exp(-(x)^2) * (abs(x) <= 3)};
        % Cosine function
        \addplot[magenta, thick] {cos(deg(2 * pi * x^2 / 36)) * (abs(x) <= 3)};
        \addplot[teal, thick] {1 / sqrt((abs(x) ^ 2) + 1) * (abs(x) <=5)};
        \end{axis}
    \end{tikzpicture}

    % \begin{tikzpicture}
    %         \begin{axis}[
    %             x = 0.5cm, y = 1cm,
    %             domain=-3.0:3.0,
    %             samples=100,
    %             ymin=-0.5, ymax=1.2,
    %             axis lines=middle, % Removes the box and shows only axes
    %             xlabel={$x$}, 
    %             ylabel={Inv. multi-quadratic},
    %             y label style={
    %                 at={(axis description cs:0.5,1.1)}, % Places ylabel at the top center
    %                 anchor=south
    %             },                
    %             legend entries={$\beta=2$, $\beta=4$},
    %             legend style={at={(0.02,1.1)}, anchor= north west, legend columns=1, draw=none, font=\tiny},
    %             legend image post style={xscale=0.5},                  
    %         ]
    %             \addplot[teal, thick] {1 / sqrt((abs(x) ^ 2) + 1)};
    %         \end{axis}
    %     \end{tikzpicture}

    \captionof{figure}{Comparison of various reconstruction kernels with the same variance, using parameters as specified in Table~\ref{tab:kernels_cf}. Notice the blunt peak in the half-cosine curve, the wide-spread area under the inverse MQ curve, and the sharp peak in the Gaussian curve, all of which contribute to their differing performances.}
    \label{fig:gaussian_cosine}
\end{center}

\noindent\textbf{Training efficiency.} Table~\ref{Table:FinalResultTable} shows that half-cosines (\( \xi = \frac{18}{\pi} \)) achieve a 15\% reduction in training time on average compared to Gaussians (\emph{updated codebase}) and 12-34\% speedups across different scenes. Although cosine computations are more computationally intensive than Gaussians, this improvement stems from the pre-computed scalar correction factor, lower primitive count, and steeper roll-off in half-cosines.
The training loss and speed plots, \emph{(see Supplementary Materials)} further demonstrate this by comparing across DARBFs. 
Despite similar training loss curves to 3DGS, half-cosine splatting outperforms Gaussians in terms of training speed.
This can be attributed to half-cosines leveraging fewer, higher-span primitives with a steeper roll-off (Fig.~\ref{fig:gaussian_cosine}), which enhances gradient updates and enables faster convergence. Additionally, do note that this training efficiency is irrespective of hardware performance \emph{(see Supplementary Materials)}.

% \begin{table}[h]
%     \centering
%     \scriptsize
%     \begin{tabular}{c|cc|cc}
%         \toprule
%          {\textbf{Scene}}& \multicolumn{2}{c|}{\textbf{RTX 3090}} & \multicolumn{2}{c}{\textbf{RTX 4090}} \\
%         (Mip-NeRF 360 dataset) & Gaussian & HC & Gaussian & HC \\
%         \midrule
%         Kitchen & 32m 15s  & 29m 59s  & 21m 53s  & 19m 41s  \\
%         Room    & 28m 32s  & 25m 28s  & 19m 48s  & 17m 32s  \\
%         \bottomrule
%     \end{tabular}
%     \caption{Performance comparison of Gaussian and half-cosine (HC) methods on RTX 3090 and RTX 4090 for different environments. Even with upgraded computational capabilities, our results indicate that half-cosine kernels achieve better training speeds compared to Gaussian kernels.}
%     \label{tab:performance}
% \end{table}

\noindent\textbf{Convergence experiments.} We acknowledge that the objective function remains Lipschitz continuous, with smoothness varying across reconstruction kernels. Table~\ref{tab:comparisonwithLR_main} shows that neither kernel performs optimally at extreme learning rates, and while cosine kernels induce less smoothness than Gaussians, their steeper roll-off and sparse gradients enable faster convergence at optimal rates. 
As this is the first study exploring the behavior of non-exponential kernels in this context, what we established here, that DARBFs can successfully replace Gaussians in 3DGS, is, in our opinion, significant. We defer a detailed mathematical analysis on convergence to future work.

\begin{table}[h]
    \centering
    \scriptsize
    \begin{tabular}{c|cc|cc|cc}
        \toprule
        \textbf{LR} & \multicolumn{2}{c|}{\textbf{PSNR}} & \multicolumn{2}{c|}{\textbf{Training Time}} & \multicolumn{2}{c}{\textbf{Memory}} \\
        & Gaussian & HC & Gaussian & HC & Gaussian & HC \\
        \midrule
        0.01 & 31.49  & 31.34  & 22m 31s  & 23m 26s  & 255.61 MB  & 246.48 MB  \\
        0.02 & 31.75  & 31.34   & 19m 48s  & 17m 32s  & 293.61 MB  & 282.65 MB  \\
        0.1  & 31.1   & 29.38   & 13m 31s  & 11m 20s  & 364 MB     & 391.76 MB \\
        0.2  & 30.5   & 26.54   & 10m 34s  & 8m 26s   & 241.42 MB & 315.93 MB \\
        \bottomrule
    \end{tabular}
    \caption{Comparison of Gaussian and half-cosine (HC) methods based on PSNR, Training Time, and Memory usage for different learning rates.}
    \label{tab:comparisonwithLR_main}
\end{table}

These experiments show that users can select sharp-peaked kernels (Gaussians, raised-cosines) for detailed scenes to ensure finer reconstruction, whereas for large-scale scenes with memory and inference constraints, blunt-peaked widely-spanned kernels like half-cosines and inverse multiquadrics are preferable.

\subsection{Ablations}

\textbf{Choosing Decay in ARBFs. }We validated various reconstruction kernels through low-dimensional simulations (Fig.~\ref{fig:enter-label}), showing that normalized, decaying functions focus energy near the origin (Sec. \ref{sec:DARBSplatting}), ensuring each sample's influence on the reconstruction is strongest its point, aiding accurate signal reconstruction. In contrast, the instability caused by unbounded growing functions led to their omission in favor of DARBFs.

\textbf{Correction factor.} As discussed in (Sec.~\ref{sec:DARBSplatting}), we introduce \( \psi \), a novel correction factor, to approximate the closure property of Gaussians. We evaluated three cases (Table~\ref{table:comparison_correctionfac}): (1) without \( \psi \), (2) with matrix \( \psi \) from an MLP, and (3) with scalar \( \psi \) precomputed from closed-form regression experiments (scalar \( \psi \) is more computationally efficient, hence we select it). 

Table~\ref{tab:comparison} shows that these modifications significantly improved DARBF performance, achieving comparable results to the \emph{updated} 3DGS codebase on the Mip-NeRF 360 \cite{Mip-nerf_360} dataset. 

\begin{table}[h!]
    \centering
    % \caption{Performance comparison of different types of correction factors (\( \psi \)) on Gaussian and half-cosine (HC) kernels across selected scenes from the Mip-NeRF 360 dataset. The results show that the scalar correction factor approximation from precomputed closed-form regression achieves comparable results to the MLP-derived correction factor. Additionally, observe how HC without \(\psi\) underperforms the same with \(\psi\). }
    \scriptsize
    \setlength{\tabcolsep}{8pt}
    \begin{tabular}{@{}lccccr@{}}
        \hline
        \rowcolor{white}
        Method & SSIM$\uparrow$ & PSNR$\uparrow$ & LPIPS$\downarrow$ \\
        \hline
        Gaussian w/o  $\psi$  & 0.81 & 27.45 & 0.22 \\
        HC w/o  $\psi$  & 0.76 & 25.89 & 0.28 \\
        HC w/  $\psi$ (MLP) & 0.794 & 27.17 & 0.24 \\
        HC w/  $\psi$ (Scalar) & 0.795 & 27.16 & 0.238 \\

        \hline
    \end{tabular}
    % \caption{Comparison of various existing methods with SSIM, PSNR, LPIPS and training time metrics.}
    \caption{Performance comparison of different types of correction factors (\( \psi \)) on Gaussian and half-cosine (HC) kernels across selected scenes from the Mip-NeRF 360 dataset. The results show that the scalar correction factor approximation from precomputed closed-form regression achieves comparable results to the MLP-derived correction factor. Additionally, observe how HC without \(\psi\) underperforms the same with \(\psi\). }
    \label{table:comparison_correctionfac}
\end{table}
\vspace{-10pt}
\begin{table}[h]
    \scriptsize
    \centering
     \setlength{\aboverulesep}{0pt} % Remove extra space above rules
    \setlength{\belowrulesep}{0pt}
    % \renewcommand{\arraystretch}{1.1}
    % \caption{Comparison of PSNR, SSIM, and LPIPS metrics for Mip-NeRF 360 scenes with \( \psi \), showing that its incorporation enhances kernel performance and is useful to producing comparable results to 3DGS.}
    \scriptsize
    \begin{tabular}{lcccccc}
        \toprule
        & \multicolumn{3}{c}{\textbf{with $\psi$}} & \multicolumn{3}{c}{\textbf{without $\psi$}} \\
        \cmidrule(lr){2-4} \cmidrule(lr){5-7}
        \textbf{Scenes} & {PSNR} & {SSIM} & {LPIPS} & {PSNR} & {SSIM} & {LPIPS} \\
        \midrule
        3DGS-updated  & --- & --- & --- &  \cellcolor{red!25}27.45 &  \cellcolor{red!25}0.81 &  \cellcolor{orange!25}0.22 \\
        Raised Cosine  &  \cellcolor{red!25}27.45 &  \cellcolor{red!25}0.81 &  \cellcolor{red!25}0.21 & 26.69 & 0.78 & \cellcolor{orange!25}0.25 \\
        Half Cosine    & \cellcolor{orange!25}27.04 & \cellcolor{orange!25}0.79 & \cellcolor{orange!25}0.25 & 25.89 & 0.76 & 0.28 \\
        % Sinc    & 21.27 & 0.82 & 0.27 & 17.93 & 0.82 & 0.11 \\
        % \midrule
        % \textbf{Mean} & \textbf{31.28} & \textbf{0.93} & \textbf{0.09} & \textbf{30.20} & \textbf{0.92} & \textbf{0.10} \\
        \bottomrule
    \end{tabular}
    \caption{Comparison of PSNR, SSIM, and LPIPS metrics for Mip-NeRF 360 scenes with \( \psi \), showing that its incorporation enhances kernel performance and is useful to producing comparable results to 3DGS.}
    \vspace{-10pt}
    \label{tab:comparison}

\end{table}

% % ------------------------------------------------------------------------
\color{black}

\section{Conclusion} 
To the best of our knowledge, we are the first to generalize splatting techniques with DARB-Splatting, extending beyond the conventional exponential family. In introducing this new class of functions, DARBFs, we highlight the distinct performances of each function. 
We establish the relationships between 3D covariance and 2D projected covariance, providing an effective approach for understanding covariance transformations under projection.
Our modified CUDA codes for each DARBF are available to facilitate further research and exploration in this area. 

\noindent \textbf{Limitation.} While we push the boundaries of splatting functions with DARBFs, the useful properties and capabilities of each function for 3D reconstruction, within a classified mathematical framework, remain to be fully explored. Future research could explore utility applications and the incorporation of existing signal reconstruction algorithms, such as the Gram-Schmidt process, with DARBFs.  

% We acknowledge and thank 3DGS for providing the original codebase, which served as a foundation for our modifications.

\section{Acknowledgments}
The authors acknowledge the University of Moratuwa for conference publication support. 
They also thank the University of Moratuwa Alumni Batch ’96 and the Accelerating Higher Education Expansion and Development (AHEAD) Operation of the Ministry of Higher Education, Sri Lanka, funded by the World Bank, for generous computational resources.

\label{sec:exp_intro}

%%%%%%%%% REFERENCES
{\small
\bibliographystyle{ieee_fullname}
\bibliography{egbib}
}

%\newpage
\clearpage
\setcounter{section}{0} 
\renewcommand*{\thesection}{\Alph{section}}
\maketitlesupplementary
% \author{
%         Vishagar Arunan\textsuperscript{1*} \quad
%         Saeedha Nazar\textsuperscript{1*} \quad
%         Hashiru Pramuditha\textsuperscript{1*} \quad
%         Vinasirajan Viruthshaan\textsuperscript{1*} \\
%         Sameera Ramasinghe\textsuperscript{2} \quad
%         Simon Lucey\textsuperscript{2} \quad
%         Ranga Rodrigo\textsuperscript{1} \\
%         \textsuperscript{1}University of Moratuwa \quad
%         \textsuperscript{2}University of Adelaide \\
%         % {\tt\small \{vishagararunan.20, saeedhanazar.20, hashirupramuditha.20, viruthshaan.20\}@cse.mrt.ac.lk} \\
%         % {\tt\small \{sameera.ramasinghe, simon.lucey\}@adelaide.edu.au, ranga@cse.mrt.ac.lk}
%     }

\section{DARB-Splatting}
We generalize the reconstruction kernel to include non-exponential functions by introducing a broader class of Decaying Anisotropic Radial Basis Functions (DARBFs). 
One of the main reasons why non-exponential functions have not been widely explored is the advantageous integration property of the Gaussian function, which simplifies the computation of the 2D covariance of a splat. However, through Monte Carlo experiments, we demonstrate that DARBFs can also exhibit this desirable property, even though most of them lack a closed-form solution for integration. Additionally, the integration of DARBFs does not generally relate to other DARBFs.

We also explained (Sec. \ref{sec:DARBSplatting}), using an example, that in 3DGS \cite{kerbl3Dgaussians}, the opacity contribution for each pixel from a reconstruction kernel (3D) is derived from its splats (2D), rather than from their 3D volume. Based on this, we consider the covariance in 3D as a variable representing the 2D covariances in all directions. In the next section, we describe the Monte Carlo experiments we conducted, supported by mathematical equations.

In surface reconstruction tasks \cite{CASCIOLA20061185}, DARBFs leverage principal component analysis (PCA) of the local covariance matrix to identify directionally dependent features and orient the 3D ellipsoids accordingly. This approach enables DARBFs to model local anisotropies in the data and reconstruct surfaces, preserving fine details more effectively than isotropic models. Furthermore, the decaying nature of these functions, as presented in Table \ref{tab:functions}, results in a more localized influence, effectively focusing the function within a certain radius. 
This localization is advantageous in 3D reconstruction, where only neighboring points contribute significantly to a given point in space, ensuring smooth blending. 
Therefore, we can conclude that all DARBFs are suitable for splatting.
Although there are many DARBFs, we focus on a selected few here due to limited space. 
In the following sections, we elaborate on the mathematical formulation of these selected DARBFs, outline their computational implementation, and demonstrate their utility in accurately modeling complex opacity distributions in the context of 3D scene reconstruction.

%%%%%%%%%%%%%%%%%%%%%%%%%%%%%%%%%%%%%%%%%%%%

\subsection{Assessing DARBFs in Simulations}
\label{sec:AssessDARB}

\textbf{Motivation. }We evaluate and compare the performance of our DARB reconstruction kernels against conventional Gaussian kernels through simulations in lower-dimensional settings. We generated various synthetic 1D functions across a specified range, including \textit{square, exponential, truncated-sinusoid, Gaussian, and triangular pulses}, as well as some irregular functions to simulate 
real-world signal variability. These signals were reconstructed using Gaussian functions and various DARBFs to evaluate the number of primitives required for accurate reconstruction and the cost difference between original and reconstructed signals.\\

This pipeline optimizes the number of primitives needed for signal approximation across different reconstruction kernels.
Our approach focuses on three key parameters: position, covariance (for signal spread), and amplitude (for signal strength). The model configurations comprises a mixture network of \textit{Gaussian, half-cosine, raised-cosine, and modular sinc} components, each with a predefined variable number of components.

\textbf{Training Recipe. }The model was trained for a fixed number of epochs with respect to a mean squared error loss function, which minimized reconstruction error between the predicted and target signals. Unlike the original Gaussian Splatting algorithm \cite{kerbl3Dgaussians}, which calculates loss in the projected 2D image space with reconstruction in 3D space, we simplified by calculating the loss and performing the reconstruction directly in 1D because, 1D signal projection cannot be represented in 0D, as this would be meaningless. Finally, a parameter sweep was conducted to identify the optimal configuration with the lowest recorded loss, indicating the optimal number of components. We provide an extensive experimental analysis of 1D simulations in Sec.~\ref{sec:sup_results_1dsim}.
%%%%%%%%%%%%%%%%%%%%%%%%%%%%%%%%%%%%%%%%%%%%

\subsection{Mathematical Expressions of Monte Carlo Experiments}
\label{sec:MonteCarloExperiments}

When it comes to 3DGS \cite{kerbl3Dgaussians}, we initially start with a $3\times3$ covariance matrix in the 3D world coordinate system. By using,
\begin{equation}
    \Sigma' = J W \Sigma W^T J^T
\end{equation}
we obtain a $3\times3$ covariance matrix ($\Sigma'$) in the camera coordinate space. According to the integration property mentioned in the EWA Spatting paper \cite{EWA_Splatting}, this $\Sigma'$ can be projected into a $2 \times 2$ covariance matrix in the image space ($\Sigma'_{2\times2}$) by simply removing the third row and column of $\Sigma'$.
However, the same process does not apply to other DARBFs. For instance, there is no closed-form solution for the marginal integration of the half-cosine and raised-cosine functions used in this paper. To simplify the understanding of this integration process and address this issue, we conducted the following experiment.

\textbf{Experimental Setup.} First, we introduce our 3D point space with \(x, y, z\) coordinates in equally spaced intervals for \(N\) number of points, a random mean vector ($\mu$) and a random $3\times3$ covariance matrix ($\Sigma$). Based on a predetermined limit specific for each DARBF (we will discuss about this in Sec.~\ref{sec:DARBFLimits}), we calculate the density/power assigned by the DARB kernel at a particular point $\boldsymbol{x}$ in 3D space as follows:
\begin{equation}
    P = \cos\left( \frac{ 2 \pi \left( \boldsymbol{x} - \boldsymbol{\mu} \right)^T \boldsymbol(\Sigma)^{-1} \left(\boldsymbol{x} - \boldsymbol{\mu} \right)}{36} \right),
\label{eq:SimCosineKernel}
\end{equation}
where $\boldsymbol{x} = \begin{bmatrix} x_i&y_i&z_i \end{bmatrix}$. As for the integration, we take the sum of these \(P\) ($P \in \mathbb{R}^{N \times N \times N}$) matrices along one dimension (for instance, along \(z\) axis) and name it \(total\_density\), which can be obtained as follows:
\begin{equation}
\text{total\_density} \, (x, y) = \sum_z P(x, y, z)
\end{equation}

where $total\_density\in \mathbb{R}^{N \times N}$. This will, for example, integrate the cosine kernel in Eq.~\ref{eq:SimCosineKernel} along \(z\)-direction, collapsing into a 2D density in XY plane. Following this integration, these total densities will be normalized as follows:
\begin{equation}
\text{total\_density}_{\text{normalized}} = \frac{\text{total\_density}}{\max(\text{total\_density})}
\end{equation}
This normalization step does not change the typical covariance relationship. To compare with other functions' projection better, we use this normalization, so that the maximum of total density will be equal one.

For the visualization of these 2D densities, we create a 2D mesh grid by using only \(x, y\) coordinate matrices called $coords\in \mathbb{R}^{N^2 \times 2}$. At the same time, we flatten the 2D density matrix and get a density grid as $density\in \mathbb{R}^{N^2\times1}$.

If the dimensions of \(x, y\) coordinates are different, we need to repeat the $coords$ and $density$ arrays separately to perfectly align each 2D coordinate for its corresponding density value. Since we use the same dimension for \(x, y\) coordinates, we can skip this step. By using these $coords$ and $density$ matrices, we then calculate the weighted covariance matrix as follows:
\noindent
\[
\begin{array}{cc}
\bar{x} = \frac{\sum_{i=1}^{N'} w_i x_i}{\sum_{i=1}^{N'} w_i} &
\bar{y} = \frac{\sum_{i=1}^{N'} w_i y_i}{\sum_{i=1}^{N'} w_i}
\end{array}
\]
where \( N' = N^2 \) and $w_i$ denote the corresponding parameters from the $density$ matrix. Finally, the vector \(\bar{\boldsymbol{m}}\) is given by:
\begin{equation}
\bar{\boldsymbol{m}} =
\begin{bmatrix}
\bar{x} \\
\bar{y}
\end{bmatrix}_{2 \times 1}
\end{equation}
By using the above results, we can determine the projected $2\times2$ covariance matrix ($\Sigma'_{2\times2}$) as follows:
\[
\Sigma'_{2 \times 2} = 
\begin{bmatrix}
\sigma_{xx} & \sigma_{xy} \\
\sigma_{xy} & \sigma_{yy}
\end{bmatrix}
\]
where $\sigma_{xx}$, $\sigma_{yy}$ and $\sigma_{xy}$ terms can be determined as:
\[
\begin{aligned}
    \sigma_{xx} &= \frac{\sum_{i=1}^{N} w_i (x_i - \bar{x})^2}{\sum_{i=1}^{N} w_i} 
    \qquad &
    \sigma_{yy} &= \frac{\sum_{i=1}^{N} w_i (y_i - \bar{y})^2}{\sum_{i=1}^{N} w_i} \\
    \sigma_{xy} &= \frac{\sum_{i=1}^{N} w_i (x_i - \bar{x})(y_i - \bar{y})}{\sum_{i=1}^{N} w_i}
\end{aligned}
\]
We can express this entire operation in matrix form as follows:
\begin{equation}
\Sigma'_{2 \times 2} = 
\frac{1}{\sum_{i=1}^{N'} w_i} 
\sum_{i=1}^{N'} w_i \left( \boldsymbol{x}_i - \bar{\boldsymbol{m}} \right) \left( \boldsymbol{x}_i - \bar{\boldsymbol{m}} \right)^T
\end{equation}
\[
\text{where} \quad \boldsymbol{x}_i = 
\begin{bmatrix}
x_i \\
y_i
\end{bmatrix}
\]
Based on this simulation, we received the projected $2\times2$ covariance matrix ($\Sigma'_{2\times2}$) for different DARBFs and identified that they are not integrable for volume rendering \cite{EWA_Splatting}. Simply saying, we cannot directly obtain the first two rows and columns of $\Sigma'_{2\times2}$ from the first two rows and first tow columns of the 3D covariance matrix $\Sigma'_{3\times3}$ as they are. 
\textbf{Correction Factor ($\psi$).} Since most of the DARB functions do not have closed form integration, 

But we noticed that there is a common ratio between the values of these two matrices. To resolve this issue, we introduce a correction factor $\psi$ as a scalar to multiply with the $\Sigma'_{2\times2}$ matrix. This scalar holds different values for different DARBFs since their density kernels act differently.

\subsection{Determining the Boundaries of DARBFs}
\label{sec:DARBFLimits}

From the 1D signal reconstruction simulations (Sec.~\ref{sec:AssessDARB} and Sec.~\ref{sec:sup_results_1dsim}) and the splatting results (Sec.~\ref{sec:sup_results_1dsim}), we demonstrate that strictly decaying functions can represent the scenes better. Therefore, we use the limits for each function to ensure the strictly decaying nature, while also considering the size of each splat. By restricting to a single pulse, we achieve a more localized representation, resulting in better quality. Incorporating more pulses allows them to cover a larger region compared to one pulse (with the same $\xi$ and $\beta$), leading to memory reduction, albeit with a tradeoff in quality.

In 3DGS \cite{kerbl3Dgaussians}, opacity modeling happens after the projection of the 3D covariance ($\Sigma$) in world coordinate space onto 2D covariance ($\Sigma'_{2\times2}$) in image space. By using the inverse covariance ($\Sigma'^{-1}_{2\times2}$) and the difference between the center of the splat ($\mu'$) and the coordinates of the selected pixel, they introduce the Mahalnobis component (Eq.~\ref{eq:mahalanobis}) within the Gaussian kernel to model the opacity distribution across each splat. When determining the final color of a particular pixel, they have incorporated a bounding box mechanism to identify the area which a splat can have the effect when modeling the opacity, so that they can do the tile-based rasterization using the computational resources efficiently.

Since the Gaussian only has a main lobe, we can simply model the opacity distribution across a splat using the bounding box mentioned in Sec.~\ref{sec:DARBSplatting}. This bounding box, determined by the radius $R = 3 \cdot \sqrt{\max\{\lambda_1, \lambda_2\}}$, where $\lambda_1$ and $\lambda_2$ denote the eigenvalues of the 2D covariance matrix $\Sigma'_{2 \times 2}$ (Sec.~\ref{sec:DARBSplatting}), will cover most of the function (main lobe), affecting the opacity modeling significantly. However, in our DARBFs, we have multiple side lobes which can have an undesirable effect on this bounding box unless the range is specified correctly. If these side lobes are included within the bounding box, each splat will have a ring effect in their opacity distributions.
 
To avoid this ring effect, we identified the range of the horizontal spread of the main lobe of each DARBF in terms of their Mahalanobis distance component and introduced a limit in opacity distribution to carefully remove the effects from their side lobes. In our Monte Carlo experiments (Sec.~\ref{sec:MonteCarloExperiments}), we used this limit in the 3D DARB kernel, as we directly perform the density calculation in 3D and the projection onto 2D image space afterwards. For example, let us consider the 3D raised cosine as follows:
 \begin{equation}
w = 0.5 + 0.5 \cos\left(\frac{2 \pi d_M^{2}}{5}\right)
\end{equation}
where $\xi = \frac{5}{2\pi}$ and $\beta = 2$ according to the standard expression mentioned in Table~\ref{tab:functions}. To avoid the side lobes and only use the main lobe, we assess the necessary range that we should consider with the Gaussian curves in 1D and chose the following limit (according to Table~\ref{tab:functions}):

\begin{equation}
d_M^{2} < 6.25 = \left(\pi\times\frac{5}{2\pi}\right)^2
\end{equation}

If the above limit is not satisfied by the Mahalanobis component, the density value will be taken as zero for those cases. As in the Table~\ref{tab:functions}, this limit will be different for different DARBFs since each DARBF shows different characteristics regarding their spread, main lobe and side lobes. Applying these limits will help to consider the 100$\%$ support of the main lobe of each DARBF into the bounding box.

Even though we apply these limits on the 3D representation of each kernel and calculate the densities, in our reconstruction pipeline, we use these limits on DARB splats (2D) similar to 3DGS \cite{kerbl3Dgaussians}. In our experiments, our main target was to identify the relationship between the 2D covariance $\Sigma'_{2\times2}$ and the 3D sub-matrix $\Sigma'_{3\times3}$ (Sec.~\ref{sec:MonteCarloExperiments}), and implement these limits on 3D DARB kernels.

\section{DARB-Splatting Implementation}
\label{sec:DARBFMathExpressions}
Here, we present the reconstruction kernel (3D) and splat (2D) functions (footprint functions) for selected DARBFs, along with their respective derivative term modifications related to backpropagation, in both mathematical expressions and CUDA codes. These modifications have been incorporated into the splatting pipeline and CUDA code changes. In the code, \( d \) denotes \( x - \boldsymbol{\mu}' \) , and $con = \begin{bmatrix} a & b \\ b & c \end{bmatrix}$ denotes the inverse of the 2D covariance matrix \((\Sigma'_{2 \times 2})^{-1}\) in the mathematical form. For each DARBF, we clearly show $\xi$ and $d_M$ in the code. We use a unique correction factor $\psi$ for each DARBF, determined through Monte Carlo experiments, to compute $\Sigma'_{2 \times 2}$ from $\Sigma'$.

\subsection{Raised Cosine Splatting}

Here, a single pulse of the raised cosine signal is selected for enhanced performance. The 3D raised cosine function is as follows: 

\begin{equation}
0.5 + 0.5 \cos \left( \frac{1}{\xi} \left( d_M \right) \right), \quad \frac{1}{\xi} \left( d_M \right) \leq \pi
\end{equation}

The raised cosine splat function is as follows: 
\begin{equation}
    w = 0.5 + 0.5  \cos\left( \frac{\sqrt{\left( x - \boldsymbol{\mu'} \right)^T \boldsymbol{\Sigma'}_{2 \times 2}^{-1} \left( x - \boldsymbol{\mu'} \right)}}{\xi} \right), 
\end{equation}
\[
\frac{\sqrt{\left( x - \boldsymbol{\mu'} \right)^T \boldsymbol{\Sigma'}_{2 \times 2}^{-1} \left( x - \boldsymbol{\mu'} \right)}}{\xi} \leq \pi
\]

Modifications in derivative terms related to the raised cosine splat during backpropagation are provided next. 

\begin{tiny}
\begin{equation}
\frac{\partial w}{\partial (x - \boldsymbol{\mu'})} = 
-\frac{0.5}{\xi} \sin\left( \frac{\sqrt{\left( x - \boldsymbol{\mu'} \right)^T \boldsymbol{\Sigma'}_{2 \times 2}^{-1} \left( x - \boldsymbol{\mu'} \right)}}{\xi} \right) \cdot 
\frac{\boldsymbol{\Sigma'}_{2 \times 2}^{-1} (x - \boldsymbol{\mu'})}{\sqrt{\left( x - \boldsymbol{\mu'} \right)^T \boldsymbol{\Sigma'}_{2 \times 2}^{-1} \left( x - \boldsymbol{\mu'} \right)}}.
\end{equation}
\end{tiny}
\begin{tiny}
\begin{equation}
\frac{\partial w}{\partial \boldsymbol{\Sigma'}_{2 \times 2}^{-1}} = -\frac{0.5}{\xi} \sin\left( \frac{\sqrt{\left( x - \boldsymbol{\mu'} \right)^T \boldsymbol{\Sigma'}_{2 \times 2}^{-1} \left( x - \boldsymbol{\mu'} \right)}}{\xi} \right) \cdot 
\frac{\left( x - \boldsymbol{\mu'} \right) \left( x - \boldsymbol{\mu'} \right)^T}{\sqrt{\left( x - \boldsymbol{\mu'} \right)^T \boldsymbol{\Sigma'}_{2 \times 2}^{-1} \left( x - \boldsymbol{\mu'} \right)}}.
\end{equation}
\end{tiny}

The following CUDA code modifications were implemented to support raised cosine splatting. 

\begin{lstlisting}[
    caption={Modifications in forward propagation in CUDA rasterizer for raised cosine splatting.},
    label={lst:vector_add},
    basicstyle=\ttfamily\scriptsize % Use a smaller font
]
float correctionFactor =  0.655f; // The correction factor we introduce 
cov *= correctionFactor ; // Multiply the 3D covariance matrix by the correction factor

float ellipse = con_o.x * d.x * d.x + con_o.z * d.y * d.y + 2.f * con_o.y * d.x * d.y; // dM**2
if (ellipse < 0 || ellipse > 6.25f) // Limit set on 2D ellipse to restrict to one pulse
    continue;
float ellipse_root = sqrt(ellipse); // dM
float ellipseFactor =  M_PI * ellipse_root / 2.5f; // Here scaling factor, xi = 5/(2 * math_pi) 
const float Cosine = 0.5f + .5f * cos(ellipseFactor);
const float alpha = min(0.99f, con_o.w * Cosine); // w * raised cosine
\end{lstlisting}

\begin{lstlisting}[
    caption={Modifications in backward propagation in CUDA rasterizer for raised cosine splatting.},
    label={lst:vector_add},
    basicstyle=\ttfamily\scriptsize % Use a smaller font
]
float correctionFactor = 0.655f; 
cov2D *= correctionFactor;

// Gradients of loss w.r.t. entries of 2D covariance matrix,
// given gradients of loss w.r.t. conic matrix (inverse covariance matrix).
// e.g., dL / da = dL / d_conic_a * d_conic_a / d_a
dL_da = correctionFactor * denom2inv * (-c * c * dL_dconic.x + 2 * b * c * dL_dconic.y + (denom - a * c) * dL_dconic.z);
dL_dc = correctionFactor *denom2inv * (-a * a * dL_dconic.z + 2 * a * b * dL_dconic.y + (denom - a * c) * dL_dconic.x);
dL_db = correctionFactor *denom2inv * 2 * (b * c * dL_dconic.x - (denom + 2 * b * b) * dL_dconic.y + a * b * dL_dconic.z);

float ellipse = (con_o.x * d.x * d.x + con_o.z * d.y * d.y + 2.f * con_o.y * d.x * d.y);
if (ellipse < 0 || ellipse > 6.25f) // limit set on 2D ellipse to restrict to one pulse
    continue;
float ellipse_root = sqrt(ellipse);   // dM
float ellipseFactor =  M_PI *ellipse_root / 2.5f; 
const float Cosine = .5f + .5f* cos(ellipseFactor); // raised cosine 
const float alpha = min(0.99f, con_o.w * Cosine); // w * raised cosine

// Helpful reusable temporary variables
const float dL_dCosine = con_o.w * dL_dalpha;
const float sindL_dCosine = sin(ellipseFactor)* dL_dCosine;
const float factor1 =  - M_PI * sindL_dCosine /5.f / ellipse_root;
const float factor2 = factor1 * 0.5 ;
const float dL_ddelx = ( con_o.x * d.x + con_o.y * d.y)* factor1;
const float dL_ddely = ( con_o.z * d.y + con_o.y * d.x)* factor1;

atomicAdd(&dL_dmean2D[global_id].x, dL_ddelx * ddelx_dx);
atomicAdd(&dL_dmean2D[global_id].y, dL_ddely * ddely_dy);
// Update gradients w.r.t. 2D covariance (2x2 matrix, symmetric)
atomicAdd(&dL_dconic2D[global_id].x, d.x * d.x * factor2);
atomicAdd(&dL_dconic2D[global_id].y, d.x * d.y * factor2);
atomicAdd(&dL_dconic2D[global_id].w, d.y * d.y * factor2);
// Update gradients w.r.t. opacity of the Cosine
atomicAdd(&(dL_dopacity[global_id]), Cosine * dL_dalpha);
\end{lstlisting}

\subsection{Half-cosine Squared Splatting}

The 3D half-cosine square function we selected is as follows: 
\[
\cos \left( \frac{1}{\xi} \left( d_M \right)^{2} \right), \quad \frac{1}{\xi} \left( d_M \right)^{2} \leq \frac{\pi}{2}
\]

The corresponding half-cosine squared splat function is as follows: 
\begin{equation}
    w = \cos\left( \frac{ \left( x - \boldsymbol{\mu'} \right)^T \boldsymbol(\Sigma'_{2 \times 2})^{-1} \left(x - \boldsymbol{\mu'} \right)}{\xi} \right),
\label{eq:HalfCosineSquared}
\end{equation}
\[
\frac{\sqrt{\left( x - \boldsymbol{\mu'} \right)^T \boldsymbol{\Sigma'}_{2 \times 2}^{-1} \left( x - \boldsymbol{\mu'} \right)}}{\xi} \leq \frac{\pi}{2}
\]

Adjustments to the derivative terms associated with the half-cosine squared splat during backpropagation are given below. 

\begin{equation}
\footnotesize
\frac{dw}{d (x - \boldsymbol{\mu'} )} = \frac{-2 \, (\Sigma'_{2 \times 2})^{-1} \, (x - \mu') \, \sin\left( \frac{\left( x - \boldsymbol{\mu'} \right)^T (\Sigma'_{2 \times 2})^{-1} \left( x - \boldsymbol{\mu'} \right)}{\xi} \right)}{\xi}
\end{equation}
\begin{equation}
\footnotesize
\frac{dw}{d ((\Sigma'_{2 \times 2})^{-1} )} = \frac{- (x - \mu') \, \left( x - \boldsymbol{\mu'} \right)^T \, \sin\left( \frac{\left( x - \boldsymbol{\mu'} \right)^T (\Sigma'_{2 \times 2})^{-1} \left( x - \boldsymbol{\mu'} \right)}{\xi} \right)}{\xi}
\end{equation}

The following CUDA code changes were made to support half-cosine squared splatting.

\begin{lstlisting}[
    caption={Modifications in forward propagation in CUDA rasterizer for half-cosine squared splatting.},
    label={lst:vector_add},
    basicstyle=\ttfamily\scriptsize % Use a smaller font
]
float correctionFactor =  1.36f;  // The correction factor we introduce 
cov *= correctionFactor ; // Multiply the 3D covariance matrix by the correction factor

float ellipse = con_o.x * d.x * d.x + con_o.z * d.y * d.y + 2.f * con_o.y * d.x * d.y; // dm_squared
if (ellipse < 0.f ||  ellipse > 9.f ) // limit set on 2D ellipse to restrict half cosine pulse
    continue;
float ellipseFactor = 	0.175f *ellipse; // 2 * M_PI * ellipse / 36;
const float Cosine = cos(ellipseFactor);
const float alpha = min(0.99f, con_o.w * Cosine); // w * cosine
\end{lstlisting}

\begin{lstlisting}[
    caption={Modifications in backward propagation in CUDA rasterizer for half-cosine squared splatting.},
    label={lst:vector_add},
    basicstyle=\ttfamily\scriptsize % Use a smaller font
]
float correctionFactor = 1.36f; 
cov2D *= correctionFactor;

// Gradients of loss w.r.t. entries of 2D covariance matrix,
// given gradients of loss w.r.t. conic matrix (inverse covariance matrix).
// e.g., dL / da = dL / d_conic_a * d_conic_a / d_a
dL_da = correctionFactor * denom2inv * (-c * c * dL_dconic.x + 2 * b * c * dL_dconic.y + (denom - a * c) * dL_dconic.z);
dL_dc = correctionFactor *denom2inv * (-a * a * dL_dconic.z + 2 * a * b * dL_dconic.y + (denom - a * c) * dL_dconic.x);
dL_db = correctionFactor *denom2inv * 2 * (b * c * dL_dconic.x - (denom + 2 * b * b) * dL_dconic.y + a * b * dL_dconic.z);

float ellipse = con_o.x * d.x * d.x + con_o.z * d.y * d.y + 2.f * con_o.y * d.x * d.y;
if (ellipse < 0 || ellipse > 9.f) // limit set on 2D ellipse
    continue;
float ellipseFactor = 0.174f *ellipse ; // 2* M_PI *ellipse / 36.f;
const float Cosine = cos(ellipseFactor); // half cosine square
const float alpha = min(0.99f, con_o.w * Cosine);

// Helpful reusable temporary variables
const float dL_dCosine = con_o.w * dL_dalpha;
const float sindL_dCosine = -0.349* sin(ellipseFactor) * dL_dCosine;
const float halfsindL_dCosine = 0.5f * sindL_dCosine;
const float dCosine_ddelx =  ( con_o.x * d.x + con_o.y * d.y)* sindL_dCosine ;
const float dCosine_ddely = ( con_o.z * d.y + con_o.y * d.x)* sindL_dCosine;

// Update gradients w.r.t. 2D mean position of the half cosine squared
atomicAdd(&dL_dmean2D[global_id].x,  dCosine_ddelx * ddelx_dx);
atomicAdd(&dL_dmean2D[global_id].y,  dCosine_ddely * ddely_dy);
// Update gradients w.r.t. 2D covariance (2x2 matrix, symmetric)
atomicAdd(&dL_dconic2D[global_id].x,  d.x * d.x * halfsindL_dCosine);
atomicAdd(&dL_dconic2D[global_id].y, d.x * d.y * halfsindL_dCosine );
atomicAdd(&dL_dconic2D[global_id].w,  d.y * d.y *halfsindL_dCosine);
// Update gradients w.r.t. opacity of the half cosine squared
atomicAdd(&(dL_dopacity[global_id]), Cosine * dL_dalpha);
\end{lstlisting}

%%%%%%%%%%%%%%%%%%%%%%%%%%%%%%%%%%%%%%%%%%%%%%%%%%%%%%%%%%%%%%

\subsection{Sinc Splatting}

Here, the sinc function refers to a single pulse of the modulus sinc function. This configuration was selected due to improved performance. The corresponding 3D sinc function is provided below:
\[
\left| \dfrac{ \sin\left( \frac{1}{\xi} \left( d_M \right) \right) }{\frac{1}{\xi} \left( d_M \right)}\right|, \quad \frac{1}{\xi} \left( d_M \right) \leq \pi
\]

The related sinc splat function is as follows:
\begin{equation}
    w = \left| \dfrac{ \sin\left( \frac{ \sqrt{\left( x - \boldsymbol{\mu'} \right)^T (\Sigma'_{2 \times 2})^{-1} \left(x - \boldsymbol{\mu'} \right)}}{\xi} \right) }{ \left( \frac{ \sqrt{\left( x - \boldsymbol{\mu'} \right)^T (\Sigma'_{2 \times 2})^{-1} \left(x - \boldsymbol{\mu'} \right)}}{\xi} \right)} \right|,
\end{equation}
\[
\frac{\sqrt{\left( x - \boldsymbol{\mu'} \right)^T \boldsymbol{\Sigma'}_{2 \times 2}^{-1} \left( x - \boldsymbol{\mu'} \right)}}{\xi} \leq \pi
\]

The modifications to the derivative terms related to the sinc splat in backpropagation are outlined below. 

\begin{equation}
\footnotesize
\frac{\partial w}{\partial (x - \boldsymbol{\mu'} )} = \operatorname{sgn}\left( \frac{\sin(A)}{A} \right) 
\cdot \frac{A \cos(A) - \sin(A)}{A^2} 
\cdot \frac{2 (\Sigma'_{2 \times 2})^{-1} (x - \mu')}{\xi},
\end{equation}

% \vspace{-10pt}
\begin{equation}
\footnotesize
\frac{\partial w}{\partial (\Sigma'_{2 \times 2})^{-1}} = 
\operatorname{sgn}\left( \frac{\sin(A)}{A} \right) 
\cdot \frac{A \cos(A) - \sin(A)}{A^2} 
\cdot \frac{(x - \mu')(x - \mu')^T}{\xi},
\end{equation}
where $\quad A = \frac{(x - \mu')^T (\Sigma'_{2 \times 2})^{-1} (x - \mu')}{\xi}.$
\\

The following CUDA code changes were made to support the sinc splatting described here.  

\begin{lstlisting}[
    caption={Modifications in forward propagation in CUDA rasterizer for sinc splatting.},
    label={lst:vector_add},
    basicstyle=\ttfamily\scriptsize % Use a smaller font
]
float correctionFactor =  1.18f; // The correction factor we introduce 
cov *= correctionFactor ; // Multiply the 3D covariance matrix by the correction factor

float constA = 3.0f/ M_PI;  
float ellipse = con_o.x * d.x * d.x + con_o.z * d.y * d.y + 2.f * con_o.y * d.x * d.y; // dM**2
if (ellipse <=  0 || ellipse > 9.f) // limit set on 2D ellipse to restrict to one pulse
    continue;
float ellipse_root = sqrt(ellipse); // dM
float ellipse_rootdivA = ellipse_root / constA;
float alpha = min(0.99f, con_o.w * fabs(sin(ellipse_rootdivA) / ellipse_rootdivA)); // w * modulus sinc
\end{lstlisting}

\begin{lstlisting}[
    caption={Modifications in backward propagation in CUDA rasterizer for sinc splatting.},
    label={lst:vector_add},
    basicstyle=\ttfamily\scriptsize % Use a smaller font
]
float correctionFactor = 1.18f; 
cov2D *= correctionFactor;

// Gradients of loss w.r.t. entries of 2D covariance matrix,
// given gradients of loss w.r.t. conic matrix (inverse covariance matrix).
// e.g., dL / da = dL / d_conic_a * d_conic_a / d_a
dL_da = correctionFactor * denom2inv * (-c * c * dL_dconic.x + 2 * b * c * dL_dconic.y + (denom - a * c) * dL_dconic.z);
dL_dc = correctionFactor *denom2inv * (-a * a * dL_dconic.z + 2 * a * b * dL_dconic.y + (denom - a * c) * dL_dconic.x);
dL_db = correctionFactor *denom2inv * 2 * (b * c * dL_dconic.x - (denom + 2 * b * b) * dL_dconic.y + a * b * dL_dconic.z);

float constA =3.0f/ M_PI;
float ellipse = con_o.x * d.x * d.x + con_o.z * d.y * d.y + 2.f * con_o.y * d.x * d.y;
if (ellipse <=  0 || ellipse > 9.f ) // limit set on 2D ellipse to restrict to one pulse
    continue;
float ellipse_root = sqrt(ellipse);  // dM
float ellipse_rootdivA = ellipse_root / constA;
const float sinellipse_rootdivA = sin(ellipse_rootdivA);
const float G =  fabs(sinellipse_rootdivA / ellipse_rootdivA); // modulus sinc
const float alpha = min(0.99f, con_o.w * G); // w * modulus sinc

// Helpful reusable temporary variables
const float dL_dG = con_o.w * dL_dalpha;
// Original equation:
// cospart = cos(ellipse_rootdivA) * sin(ellipse_rootdivA) * fabs(1 / ellipse_root) / ellipse_root / fabs(sin(ellipse_rootdivA))
// Simplified equation using copysignf for clarity:
const float cospart = cos(ellipse_rootdivA ) * copysignf(1.0f, sinellipse_rootdivA) / ellipse ;
const float sinpart = constA *fabs(sinellipse_rootdivA) / ellipse / ellipse_root  ;
const float commonpart = (cospart - sinpart)* dL_dG;
const float commonpartdiv2 = commonpart  * 0.5f;
const float dG_ddelx = commonpart * (d.x * con_o.x + d.y * con_o.y);
const float dG_ddely = commonpart * (d.x * con_o.y + d.y * con_o.z);
// Update gradients w.r.t. 2D mean position of the modulus sinc
atomicAdd(&dL_dmean2D[global_id].x, dG_ddelx * ddelx_dx);
atomicAdd(&dL_dmean2D[global_id].y,  dG_ddely * ddely_dy);
// Update gradients w.r.t. 2D covariance (2x2 matrix, symmetric)
atomicAdd(&dL_dconic2D[global_id].x, commonpartdiv2 * d.x * d.x );
atomicAdd(&dL_dconic2D[global_id].y, commonpartdiv2 * d.x * d.y );
atomicAdd(&dL_dconic2D[global_id].w, commonpartdiv2 * d.y * d.y );
// Update gradients w.r.t. opacity of the modulus sinc
atomicAdd(&(dL_dopacity[global_id]), G * dL_dalpha);
\end{lstlisting}

%%%%%%%%%%%%%%%%%%%%%%%%%%%%%%%%%%%%%%%%%%%%%%%%%%%%%%%%

\subsection{Inverse Multiquadric Splatting}
Here, we define the inverse multiquadric formulation based on the following 3D inverse multiquadric function:
\[
\dfrac{1}{\left[ \frac{1}{\xi} \left( d_M \right)^2 + 1 \right]}, \quad d_M \geq 0
\]

The corresponding inverse multiquadric splat function is as follows: 

\begin{equation}
    w = \dfrac{1}{\left[ \frac{1}{\xi} \left( x - \boldsymbol{\mu'} \right)^T \boldsymbol(\Sigma'_{2 \times 2})^{-1} \left(x - \boldsymbol{\mu'} \right) + 1 \right]}
\end{equation}

The changes done to the derivative terms related to the inverse multiquadric splat during backpropagation are detailed below.

\begin{equation}
\footnotesize
\frac{\partial w}{\partial z} = 
-\frac{2}{\xi \left( \frac{1}{\xi} z^T (\Sigma'_{2 \times 2})^{-1} z + 1 \right)^2} 
(\Sigma'_{2 \times 2})^{-1} z.
\end{equation}
% \vspace{-10pt}
\begin{equation}
\footnotesize
\frac{\partial w}{\partial (\Sigma'_{2 \times 2})^{-1}} = 
- \frac{1}{\left( \frac{1}{\xi} z^T (\Sigma'_{2 \times 2})^{-1} z + 1 \right)^2} \cdot \frac{1}{\xi} z z^T.
\end{equation}
where $z = (x - \mu').$

The CUDA code modifications provided below were implemented to enable the inverse multiquadric splatting described in this section.

\begin{lstlisting}[
    caption={Modifications in forward propagation in CUDA rasterizer for inverse multiquadric splatting.},
    label={lst:vector_add},
    basicstyle=\ttfamily\scriptsize % Use a smaller font
]
float correctionFactor =  1.38f; // The correction factor we introduce 
cov *= correctionFactor ; // Multiply the 3D covariance matrix by the correction factor

float ellipse = con_o.x * d.x * d.x + con_o.z * d.y * d.y + 2.f * con_o.y * d.x * d.y ; // dm**2
if (ellipse <=  0)
    continue;
if (ellipse  >= 9.f) // limit set on 2D ellipse
    continue;
float alpha = min(0.99f, con_o.w * (1.f / (ellipse + 1.f))); // w * inverse quadric
\end{lstlisting}

\begin{lstlisting}[
    caption={Modifications in backward propagation in CUDA rasterizer for inverse multiquadric splatting.},
    label={lst:vector_add},
    basicstyle=\ttfamily\scriptsize % Use a smaller font
]
float correctionFactor = 1.38f; 
cov2D *= correctionFactor;

// Gradients of loss w.r.t. entries of 2D covariance matrix,
// given gradients of loss w.r.t. conic matrix (inverse covariance matrix).
// e.g., dL / da = dL / d_conic_a * d_conic_a / d_a
dL_da = correctionFactor * denom2inv * (-c * c * dL_dconic.x + 2 * b * c * dL_dconic.y + (denom - a * c) * dL_dconic.z);
dL_dc = correctionFactor *denom2inv * (-a * a * dL_dconic.z + 2 * a * b * dL_dconic.y + (denom - a * c) * dL_dconic.x);
dL_db = correctionFactor *denom2inv * 2 * (b * c * dL_dconic.x - (denom + 2 * b * b) * dL_dconic.y + a * b * dL_dconic.z);

float ellipse = con_o.x * d.x * d.x + con_o.z * d.y * d.y + 2.f * con_o.y * d.x * d.y ;
if (ellipse <=  0)
    continue;
if (ellipse >= 9.f) // limit set on 2D ellipse
    continue;
const float G = (1.f / (ellipse + 1.f)); // inverse quadric 
const float alpha = min(0.99f, con_o.w * G);

// Helpful reusable temporary variables
const float dL_dG = con_o.w * dL_dalpha;
const float ellipsepow =  - dL_dG * 2.f / pow(ellipse + 1.f, 2.f);
const float halfellipsepow  = 0.5f * ellipsepow;
const float dG_ddelx = ellipsepow * (d.x * con_o.x + d.y * con_o.y);
const float dG_ddely = ellipsepow * (d.x * con_o.y + d.y * con_o.z);
// Update gradients w.r.t. 2D mean position of the IQF
atomicAdd(&dL_dmean2D[global_id].x,  dG_ddelx * ddelx_dx);
atomicAdd(&dL_dmean2D[global_id].y,  dG_ddely * ddely_dy);
// Update gradients w.r.t. 2D covariance (2x2 matrix, symmetric)
atomicAdd(&dL_dconic2D[global_id].x, halfellipsepow * d.x * d.x );
atomicAdd(&dL_dconic2D[global_id].y, halfellipsepow * d.x * d.y );
atomicAdd(&dL_dconic2D[global_id].w, halfellipsepow * d.y * d.y );
// Update gradients w.r.t. opacity of the IQF
atomicAdd(&(dL_dopacity[global_id]), G * dL_dalpha);
\end{lstlisting}

%%%%%%%%%%%%%%%%%%%%%%%%%%%%%%%%%%%%%%%%%%%%%%%%%%%%%%%%%%
\subsection{Parabolic Splatting}
Here, we define the parabolic formulation based on the following 3D parabolic function:
\[
1 - \frac{1}{\xi} (d_M)^2, \quad d_M \geq 0
\]

The corresponding parabolic splat function is as follows: 

\begin{equation}
    w = \left[1 - \frac{1}{\xi} \left( x - \boldsymbol{\mu'} \right)^T \boldsymbol(\Sigma'_{2 \times 2})^{-1} \left(x - \boldsymbol{\mu'} \right) \right]
\end{equation}

The changes done to the derivative terms related to the parabolic splat during backpropagation are detailed below.

\begin{equation}
\footnotesize
\frac{\partial w}{\partial z} = -\frac{2}{\xi} (\Sigma'_{2 \times 2})^{-1} z.
\end{equation}
% \vspace{-10pt}
\begin{equation}
\footnotesize
\frac{\partial w}{\partial (\Sigma'_{2 \times 2})^{-1}} = -\frac{1}{\xi} zz^T
\end{equation}
where $z = (x - \mu').$

The CUDA code modifications provided below were implemented to enable the parabolic splatting described in this section.

\begin{lstlisting}[
    caption={Modifications in forward propagation in CUDA rasterizer for parabolic splatting.},
    label={lst:vector_add},
    basicstyle=\ttfamily\scriptsize % Use a smaller font
]
float correctionFactor = 1.3f;
cov *= correctionFactor ; 

float ellipse = (con_o.x * d.x * d.x + con_o.z * d.y * d.y + 2.f * con_o.y * d.x * d.y); //dM**2

if (ellipse < 0 || ellipse > 9.f)
	continue;

const float parabola =  1. - ellipse / 9.; 
const float alpha = min(0.99f, con_o.w * parabola);
\end{lstlisting}

\begin{lstlisting}[
    caption={Modifications in backward propagation in CUDA rasterizer for parabolic splatting.},
    label={lst:vector_add},
    basicstyle=\ttfamily\scriptsize % Use a smaller font
]
float correctionFactor = 1.3f;
cov2D *= correctionFactor ; 

// Gradients of loss w.r.t. entries of 2D covariance matrix,
// given gradients of loss w.r.t. conic matrix (inverse covariance matrix).
// e.g., dL / da = dL / d_conic_a * d_conic_a / d_a
dL_da = denom2inv * (-c * c * dL_dconic.x + 2 * b * c * dL_dconic.y + (denom - a * c) * dL_dconic.z);
dL_dc = denom2inv * (-a * a * dL_dconic.z + 2 * a * b * dL_dconic.y + (denom - a * c) * dL_dconic.x);
dL_db = denom2inv * 2 * (b * c * dL_dconic.x - (denom + 2 * b * b) * dL_dconic.y + a * b * dL_dconic.z);

float ellipse = (con_o.x * d.x * d.x + con_o.z * d.y * d.y + 2.f * con_o.y * d.x * d.y);


if (ellipse < 0 || ellipse > 9.f)
	continue;

const float parabola =  1. - ellipse / 9.;
const float alpha = min(0.99f, con_o.w * parabola);

// Helpful reusable temporary variables
const float dL_dP = con_o.w * dL_dalpha;
const float dP_ddelx = ( con_o.x * d.x + con_o.y * d.y)* -2. / 9.;
const float dP_ddely =( con_o.z * d.y + con_o.y * d.x)* -2. / 9.;

atomicAdd(&dL_dmean2D[global_id].x, dL_dP  * dP_ddelx * ddelx_dx);
atomicAdd(&dL_dmean2D[global_id].y, dL_dP  * dP_ddely * ddely_dy);
// Update gradients w.r.t. 2D covariance (2x2 matrix, symmetric)

atomicAdd(&dL_dconic2D[global_id].x, dL_dP  * d.x * d.x / -9.f);
atomicAdd(&dL_dconic2D[global_id].y, dL_dP  * d.x * d.y / -9.f);
atomicAdd(&dL_dconic2D[global_id].w, dL_dP  * d.y * d.y / -9.f);

// Update gradients w.r.t. opacity of the Cosine
atomicAdd(&(dL_dopacity[global_id]), parabola * dL_dalpha);
\end{lstlisting}

%%%%%%%%%%%%%%%%%%%%%%%%%%%%%%%%%%%%%%%%%%%%%%%%%%%%%%

\section{Utility Applications of DARB-Splatting}

\subsection{Enhanced Quality}

\begin{figure*}
     % Column headings
    \parbox{0.32\textwidth}{\centering {Ground Truth}} 
    \parbox{0.32\textwidth}{\centering {Gaussians}} 
    \parbox{0.32\textwidth}{\centering {Raised cosines (ours)}}\\[-0.5em]
    % \parbox{0.24\textwidth}{\centering {Half-cosine Squares (Ours)}} \\

    % Counter
    \includegraphics[width=0.32\textwidth]{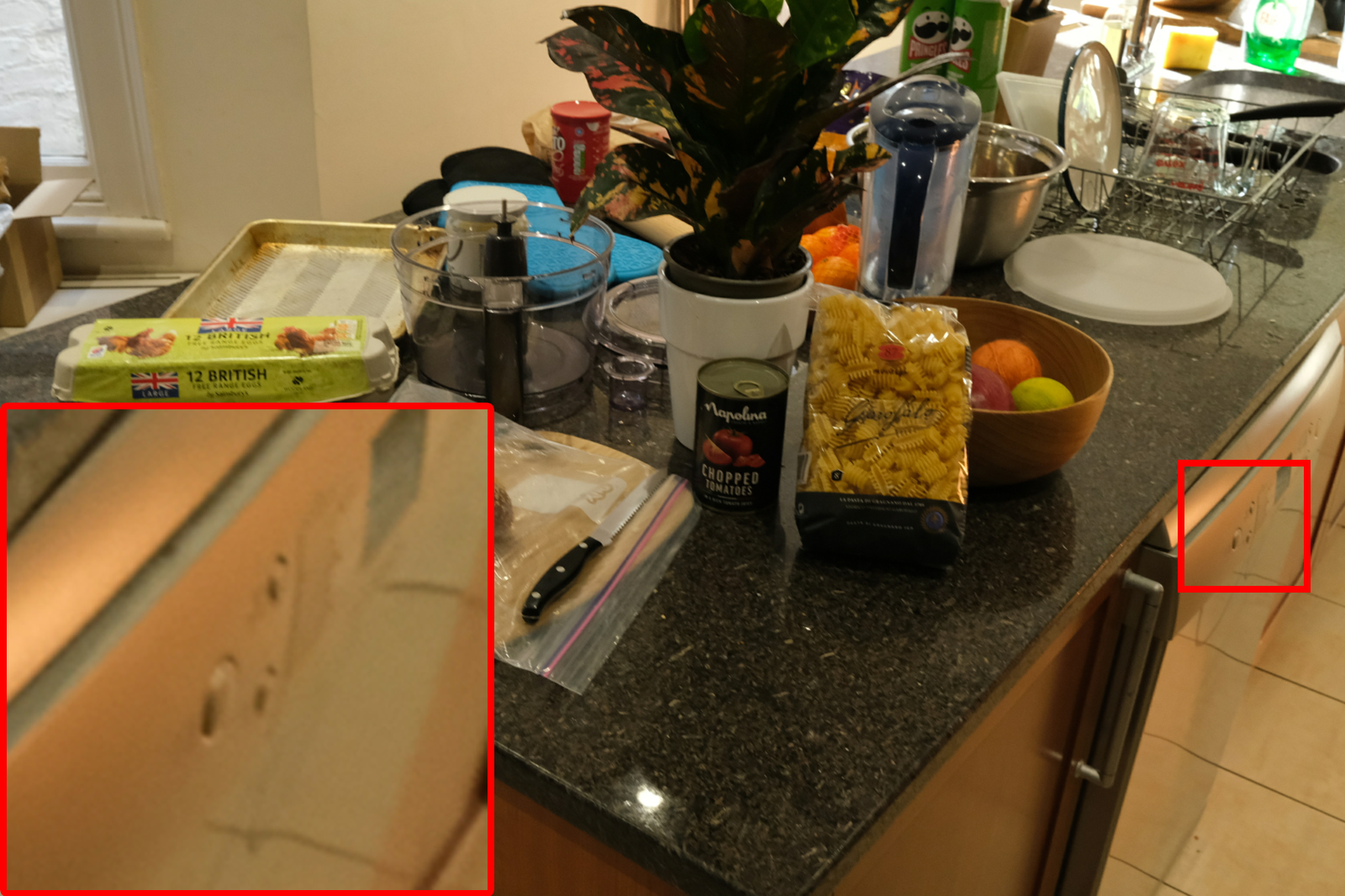} 
    \includegraphics[width=0.32\textwidth]{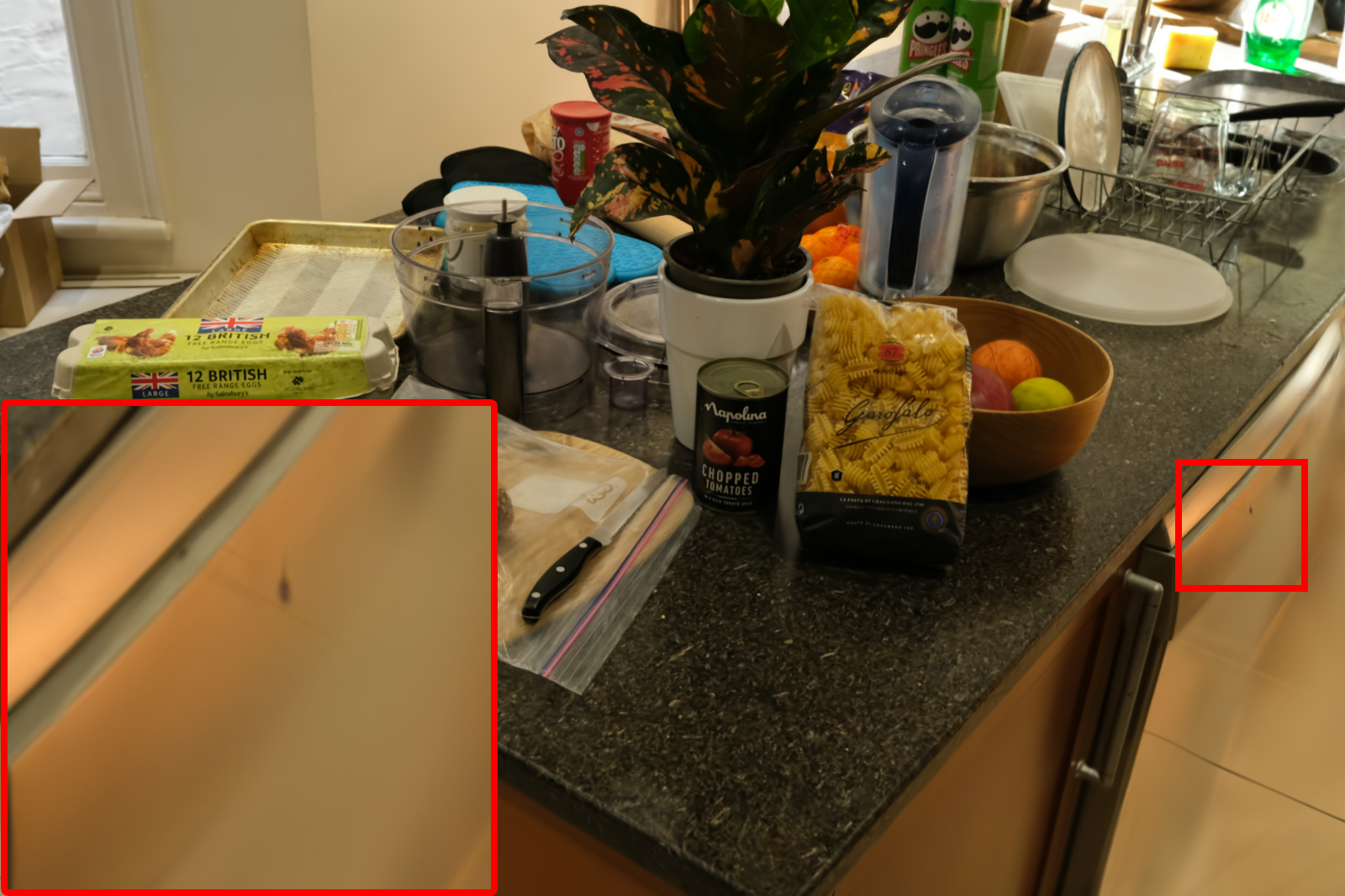} 
    \includegraphics[width=0.32\textwidth]{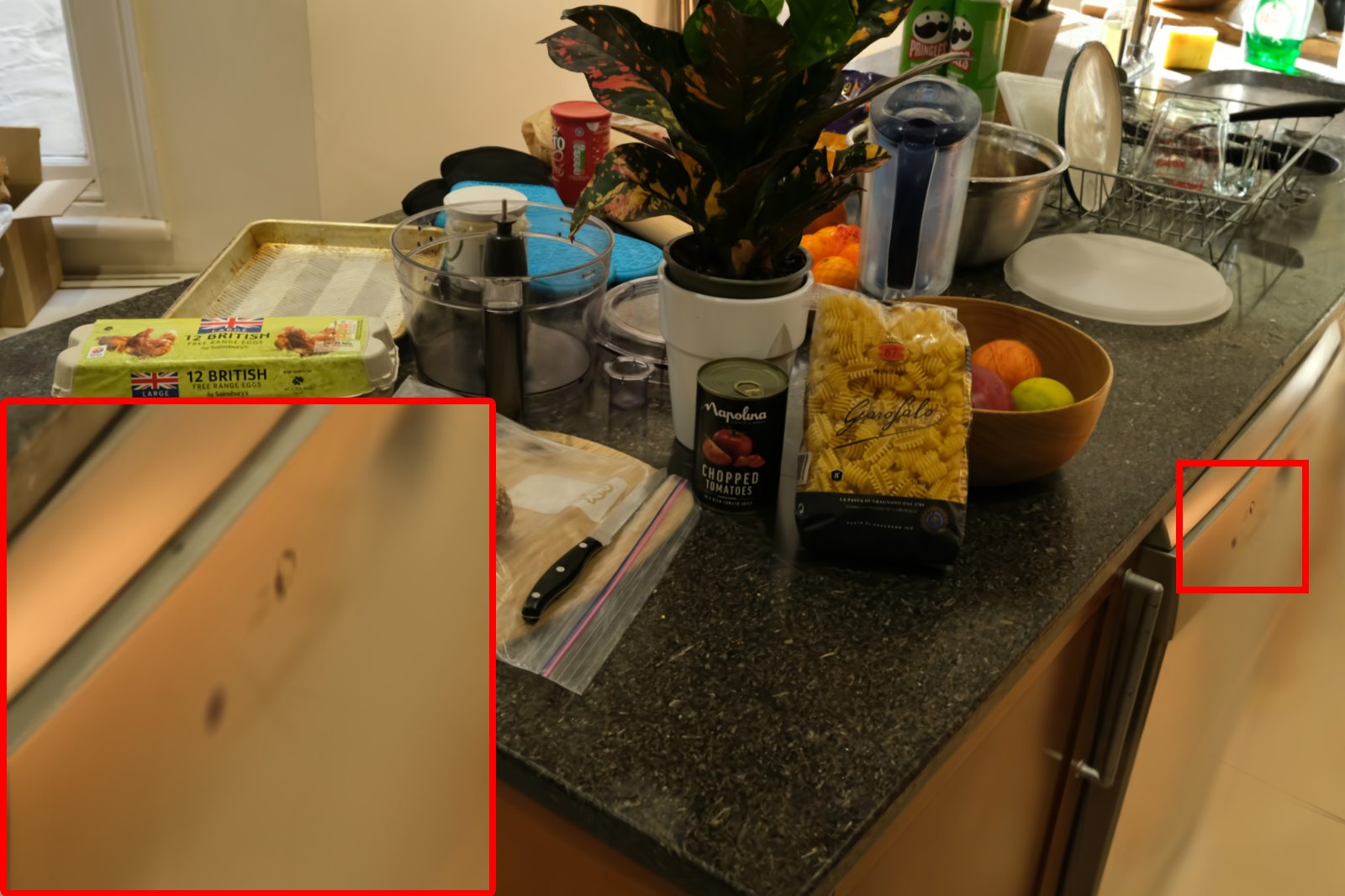}

    % Truck
    \includegraphics[width=0.32\textwidth]{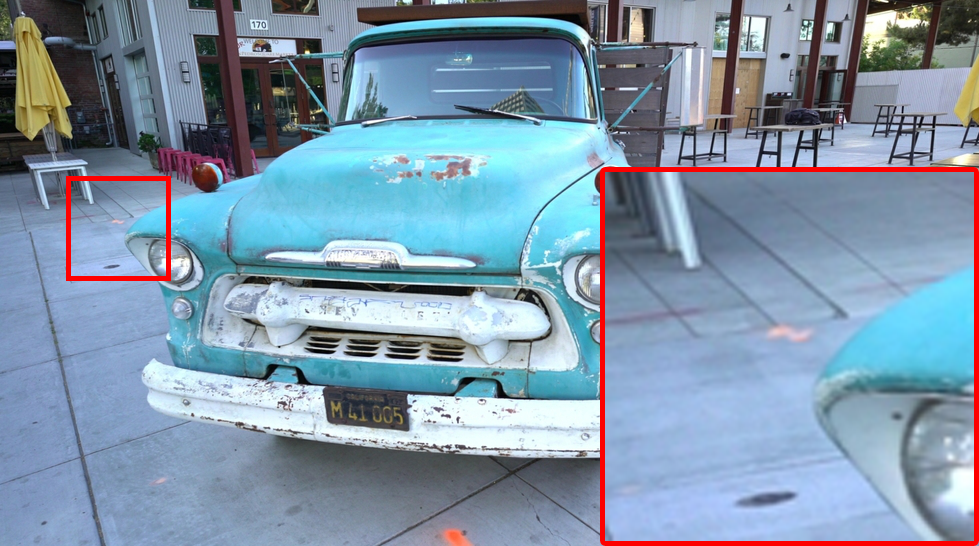} 
    \includegraphics[width=0.32\textwidth]{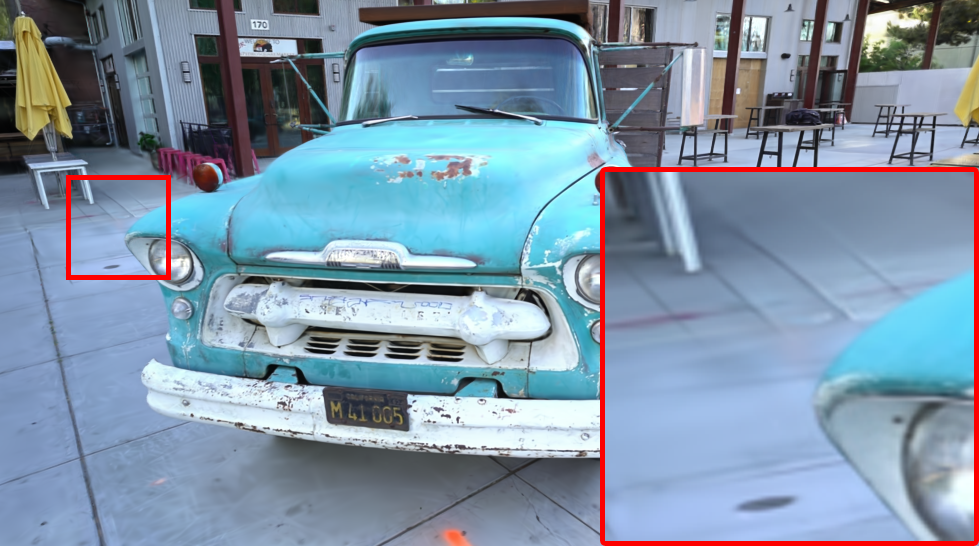} 
    \includegraphics[width=0.32\textwidth]{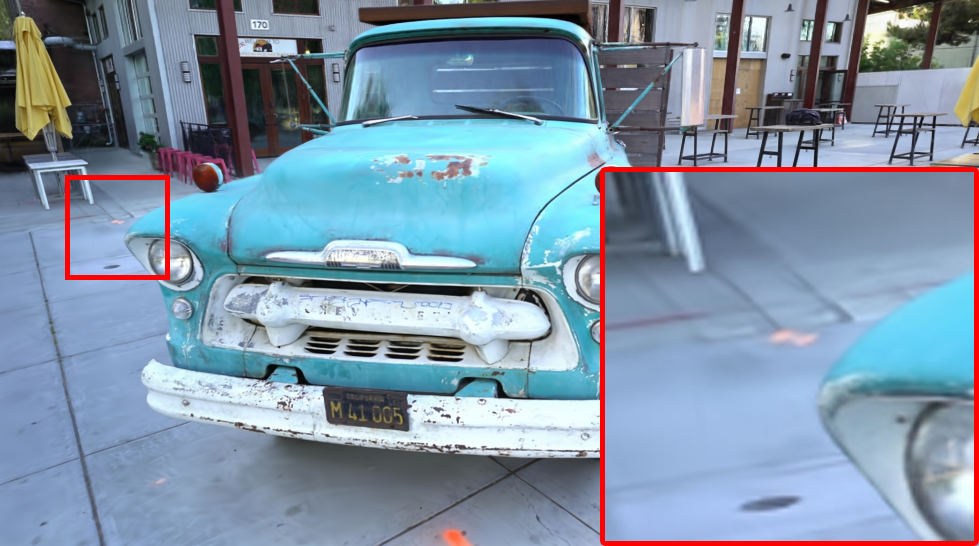} 

    % DrJohnson
    \includegraphics[width=0.32\textwidth]{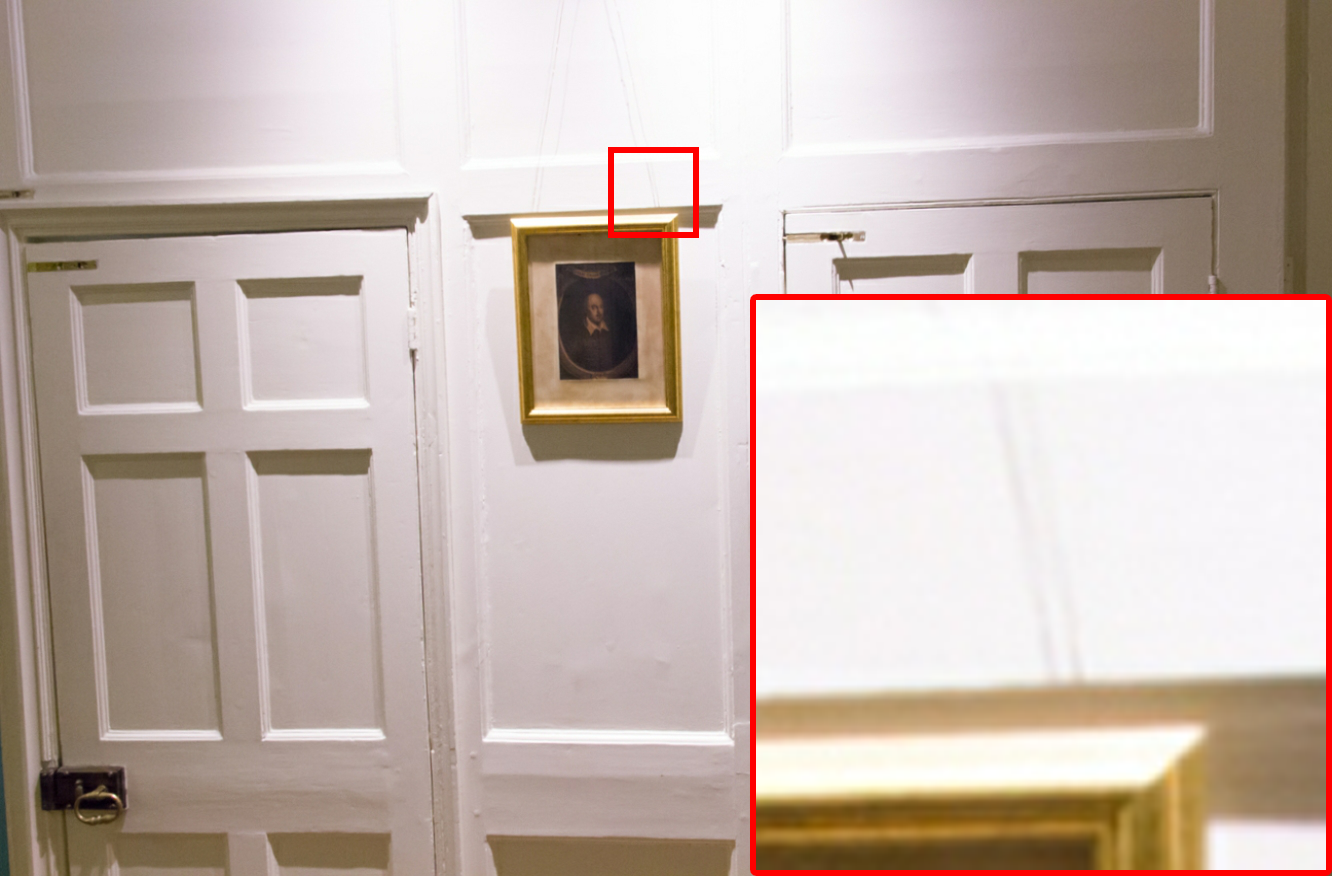} 
    \includegraphics[width=0.32\textwidth]{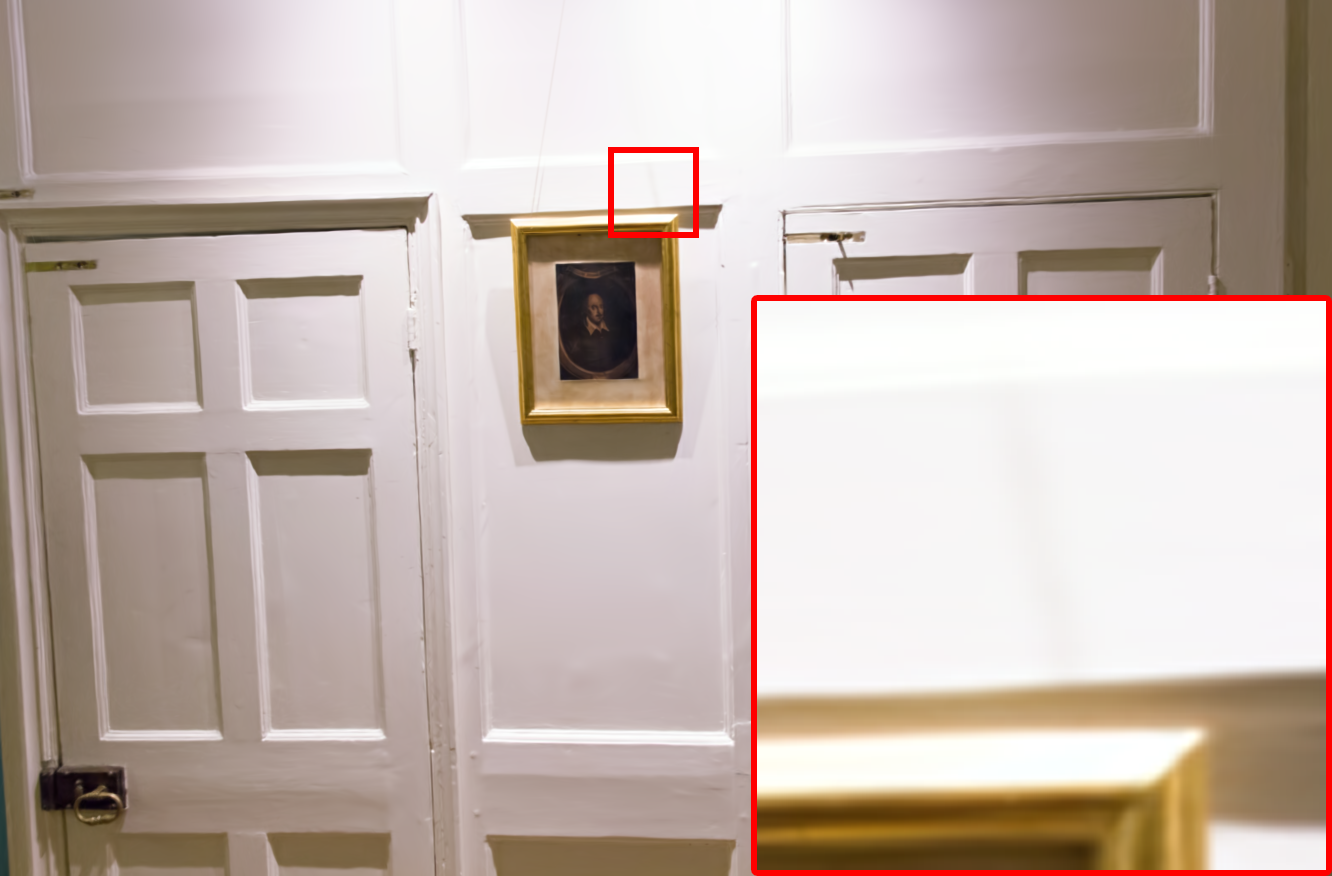} 
    \includegraphics[width=0.32\textwidth]{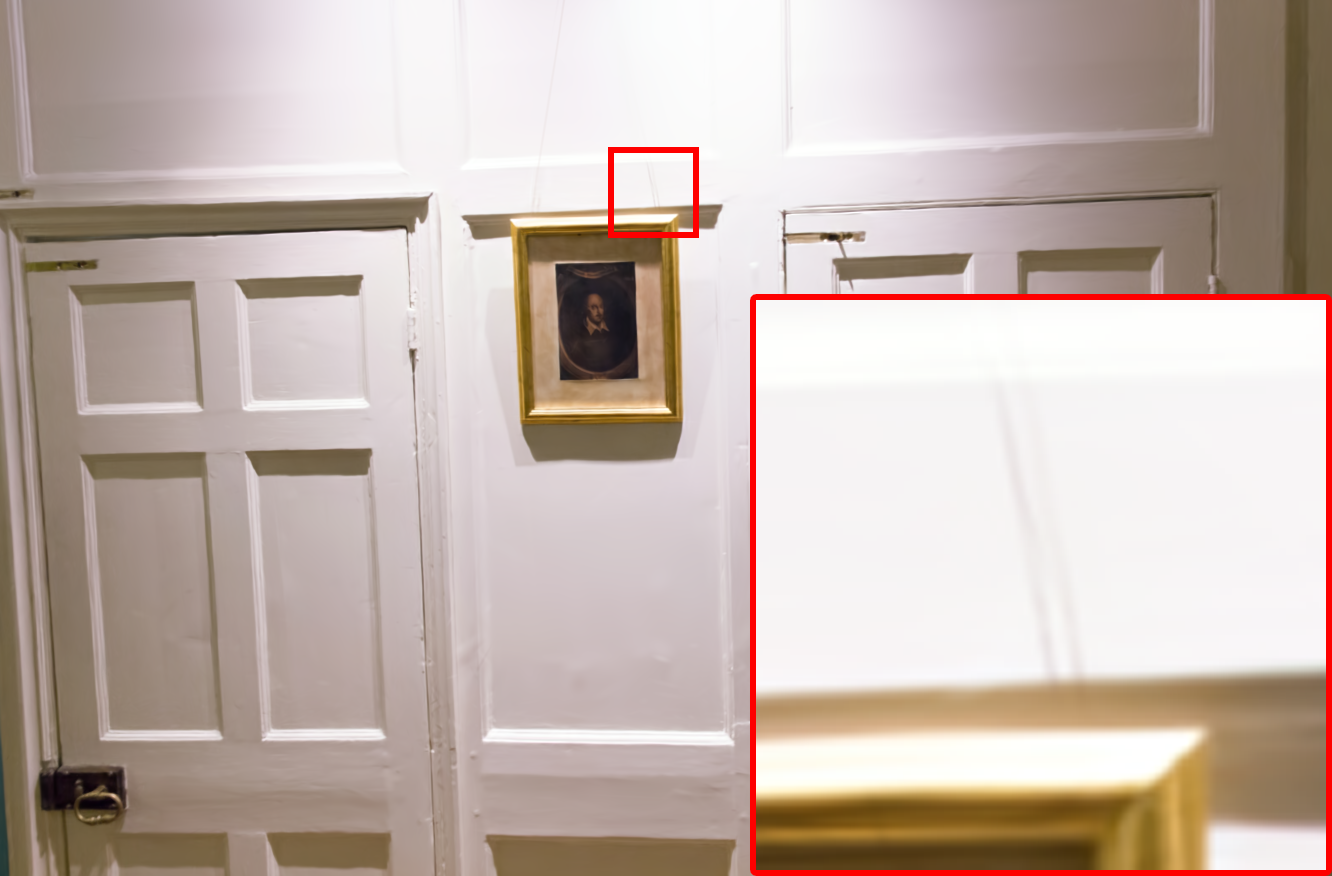} 
    
    % Bonsai
    \includegraphics[width=0.32\textwidth]{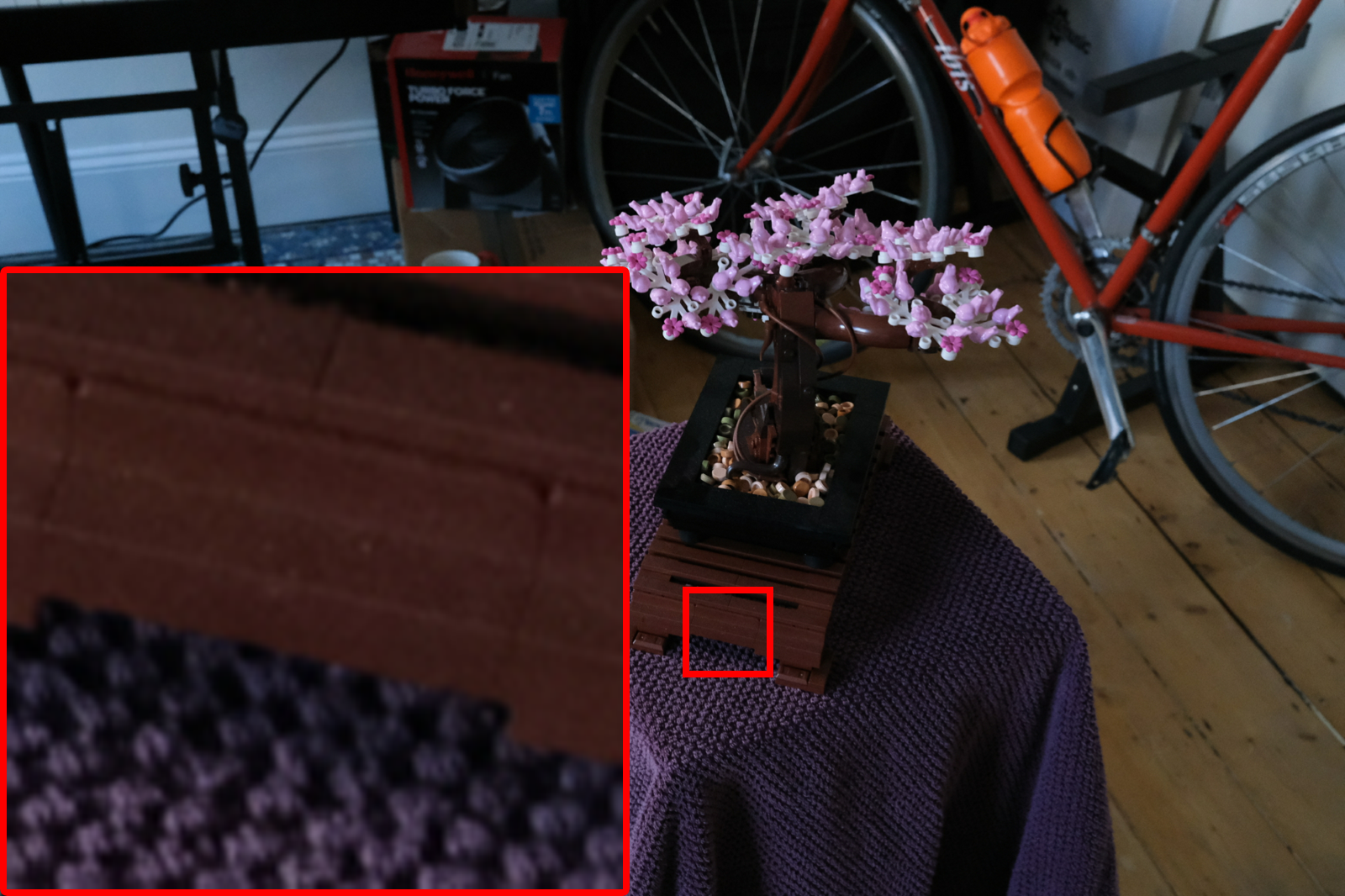} 
    \includegraphics[width=0.32\textwidth]{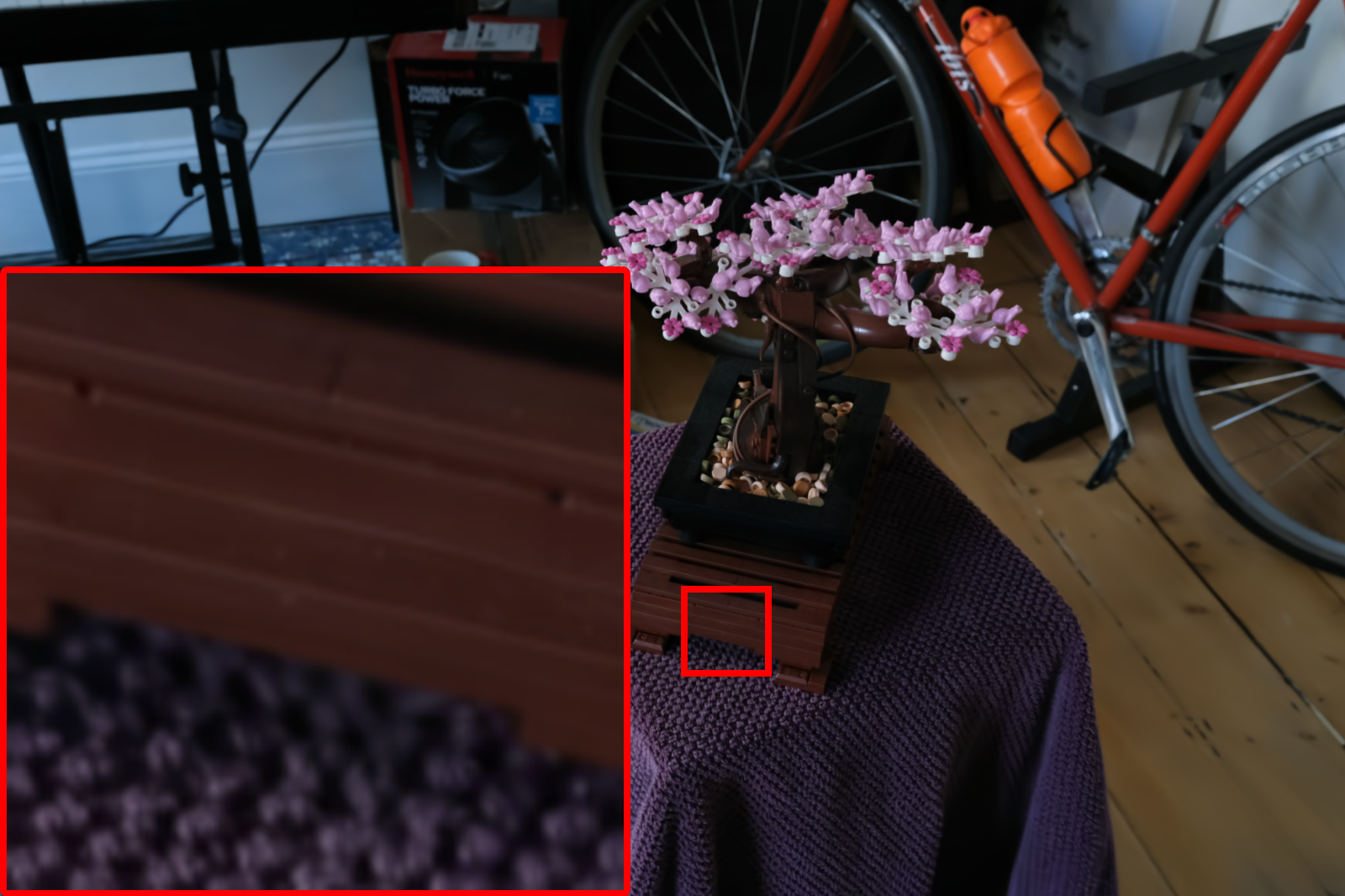} 
    \includegraphics[width=0.32\textwidth]{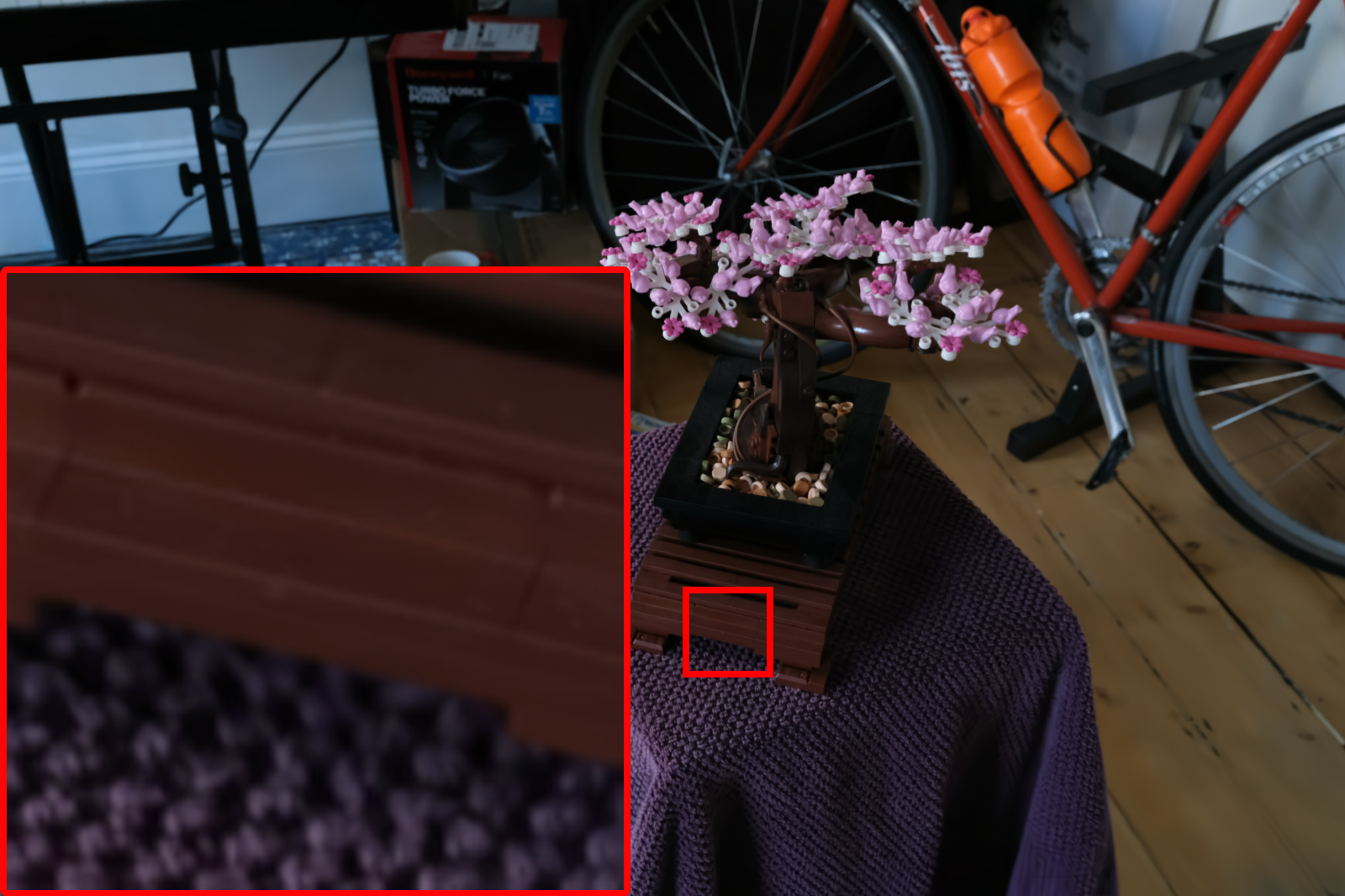} 
    
    \caption{\textbf{Qualitative Visualization Across 3DGS and raised cosine splatting.} Displayed are side-by-side comparisons across the Counter, Truck, DrJohnson, Bonsai scenes (top to bottom) repsectively from Mip-NeRF 360, Tanks\&Temples and Deep Blending datasets. In the Counter scene, raised cosines outperform Gaussians by better reconstructing the buttons, rendering them more prominently, whereas Gaussians struggle to achieve this even after full training. Similarly, in the Truck scene, raised cosines successfully reconstruct the orange mark on the floor, a detail that Gaussians fail to capture. In the Dr. Johnson scene, our method renders the string of the picture frame with greater clarity, closely resembling the ground truth imagery, while Gaussians fail to achieve the same level of detail. Lastly, in the Bonsai scene, the edges of the pot are more accurately represented by raised cosines as we can see its shadows, producing results that are closer to the ground truth image compared to those achieved with Gaussians. These examples highlight the advantages of raised cosine splatting in capturing finer details in 3D reconstruction compared to Gaussians.}
    \label{fig:SupVisualComparisons}
\end{figure*}

In terms of splatting, despite Gaussians providing SOTA quality, we demonstrate that the raised cosine function can deliver modestly improved visual quality compared to Gaussians. Our 1D simulations, presented in Sec.~\ref{sec:AssessDARB} and Sec.~\ref{sec:sup_results_1dsim}, and the qualitative visual comparisons demonstrated in Fig.~\ref{fig:SupVisualComparisons}, illustrate this effectively.

Across the selected DARBFs, only the raised cosine outperforms the Gaussian in terms of quality, albeit by a small margin. The others fail to surpass the Gaussian in terms of quality. The primary reason is that exponentially decaying functions ensure faster blending compared to relatively flatter functions. However, these functions have other utilities, which we will discuss next. 

%%%%%%%%%%%%%%%%%%%%%%%%%%%%%%%%%%%%%%%%%%%%%%%%%%%%
%%%%%%%%%%%%%%%%%%%%%%%%%%%%%%%%%%%%%%%%%%%%%%%%%%%%

\subsection{Reduced Training Time}

According to Fig.~\ref{fig:gaussian_cosine}, which shows the half-cosine square with \(\beta = 2\) and \(\xi = 36\), along with the Gaussian 1D plot, we can observe that for a single primitive with the same variance, the cosine function can provide higher opacity values. Instead of requiring multiple splats to composite to determine the final pixel color, the cosine function can achieve the same accumulated value required with fewer primitives. Although the cosine function's computation is more time-intensive compared to the Gaussian calculations, the overall training time is reduced due to the lower number of primitives required. 

\begin{figure*}
    % Column headings
    \begin{center}
        \parbox{0.425\textwidth}{\centering \textbf{Train Loss}}%
        \parbox{0.425\textwidth}{\centering \textbf{Train Speed}}
    \end{center}

    % Row 3: Kitchen
    \centering
    \begin{minipage}[c]{0.015\textwidth}
        \centering
        \rotatebox{90}{\textit{Kitchen}}
    \end{minipage}%
    \begin{minipage}[c]{0.425\textwidth}
        \centering
        \includegraphics[width=\textwidth]{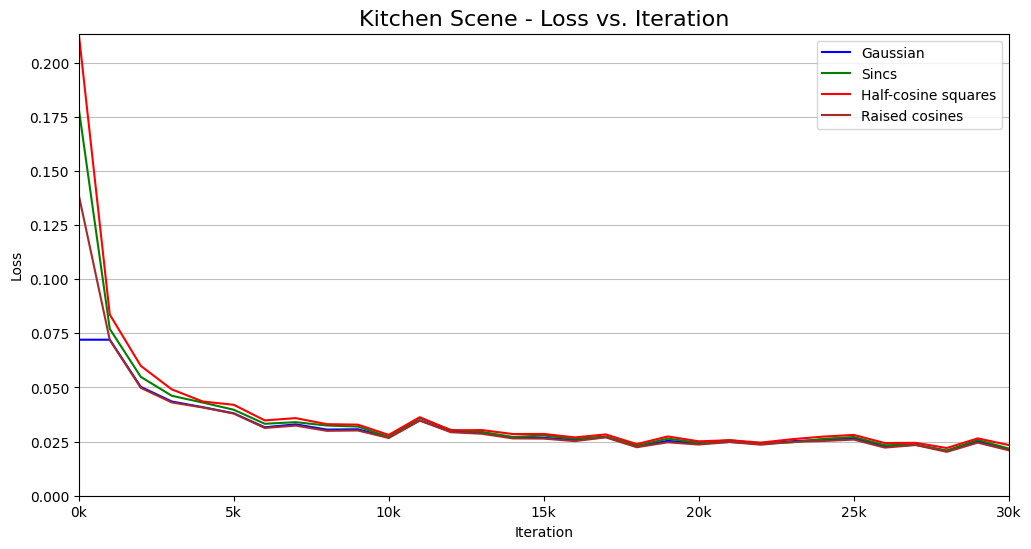}
    \end{minipage}%
    \begin{minipage}[c]{0.425\textwidth}
        \centering
        \includegraphics[width=\textwidth]{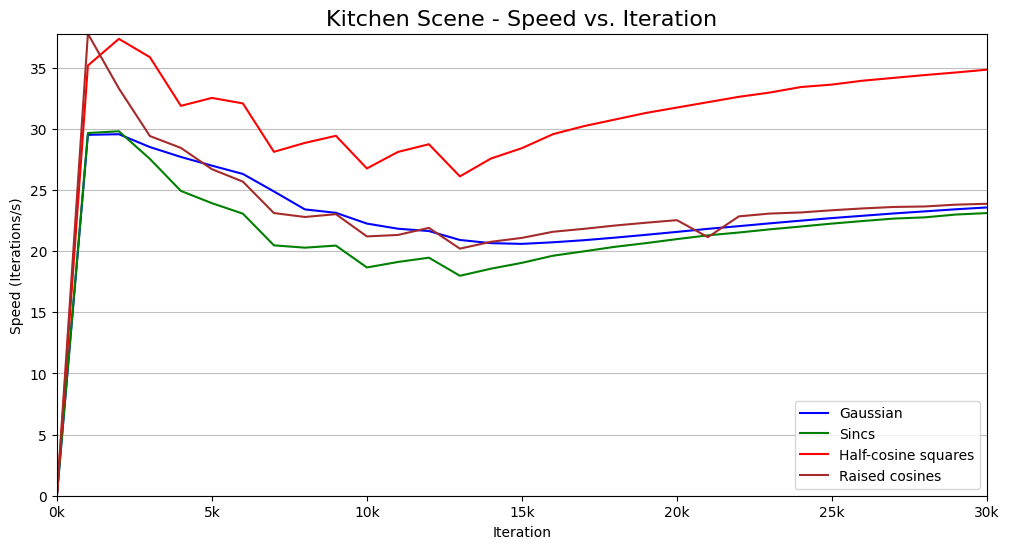}
    \end{minipage}\\[0.5em]
    
    % Row 1: Room
    \centering
    \begin{minipage}[c]{0.015\textwidth}
        \centering
        \rotatebox{90}{\textit{Room}}
    \end{minipage}%
    \begin{minipage}[c]{0.425\textwidth}
        \centering
        \includegraphics[width=\textwidth]{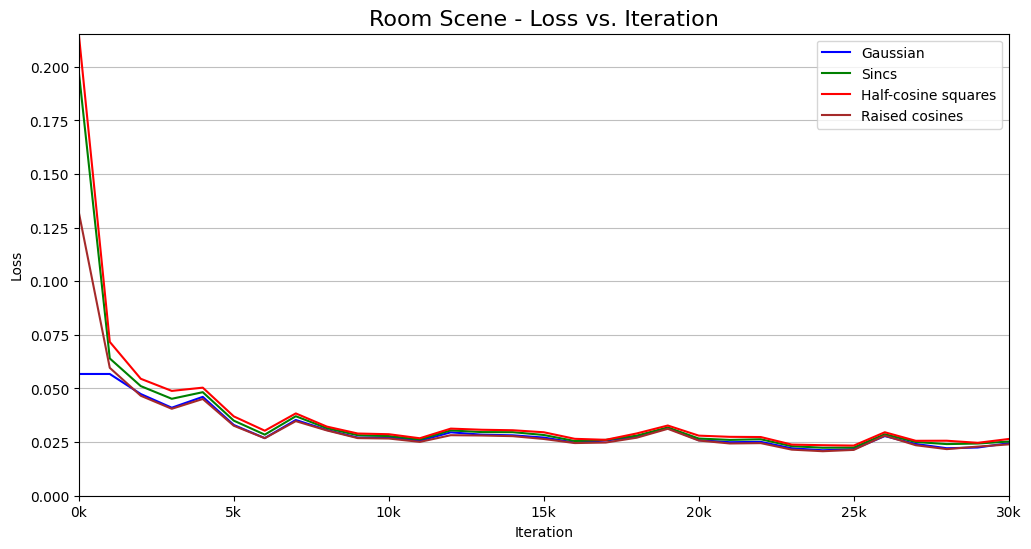}
    \end{minipage}%
    \begin{minipage}[c]{0.425\textwidth}
        \centering
        \includegraphics[width=\textwidth]{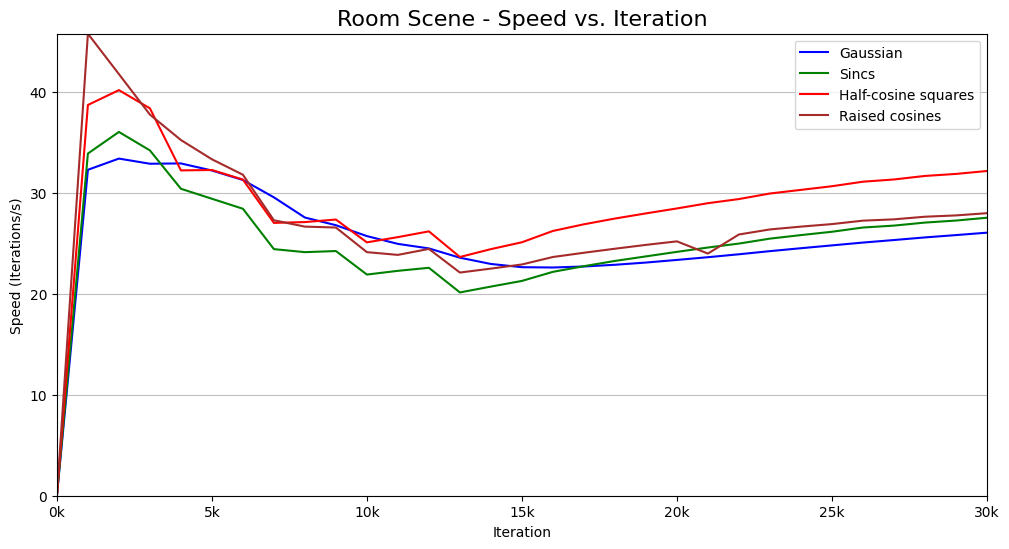}
    \end{minipage}\\[0.5em]
    
    % Row 2: Counter
    \centering
    \begin{minipage}[c]{0.015\textwidth}
        \centering
        \rotatebox{90}{\textit{Counter}}
    \end{minipage}%
    \begin{minipage}[c]{0.425\textwidth}
        \centering
        \includegraphics[width=\textwidth]{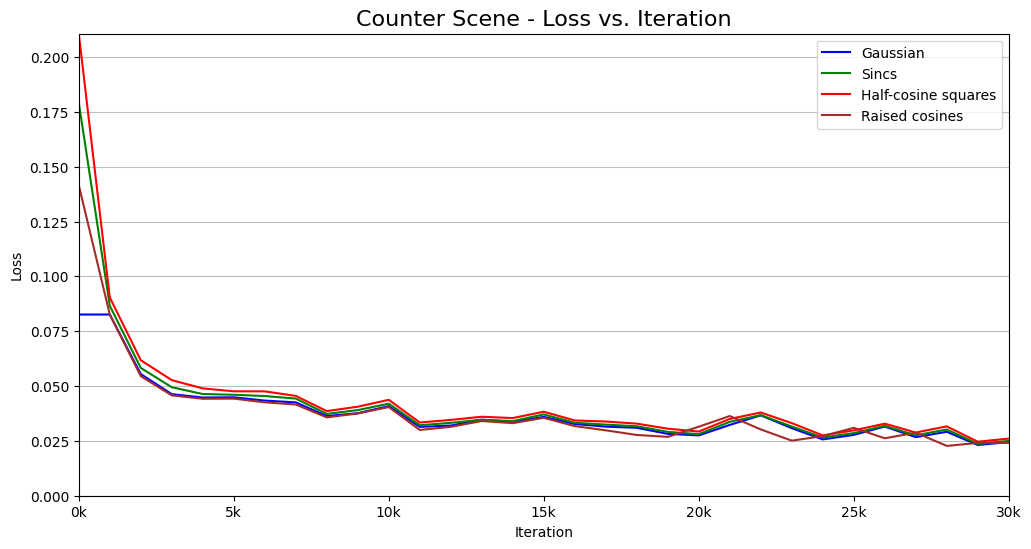}
    \end{minipage}%
    \begin{minipage}[c]{0.425\textwidth}
        \centering
        \includegraphics[width=\textwidth]{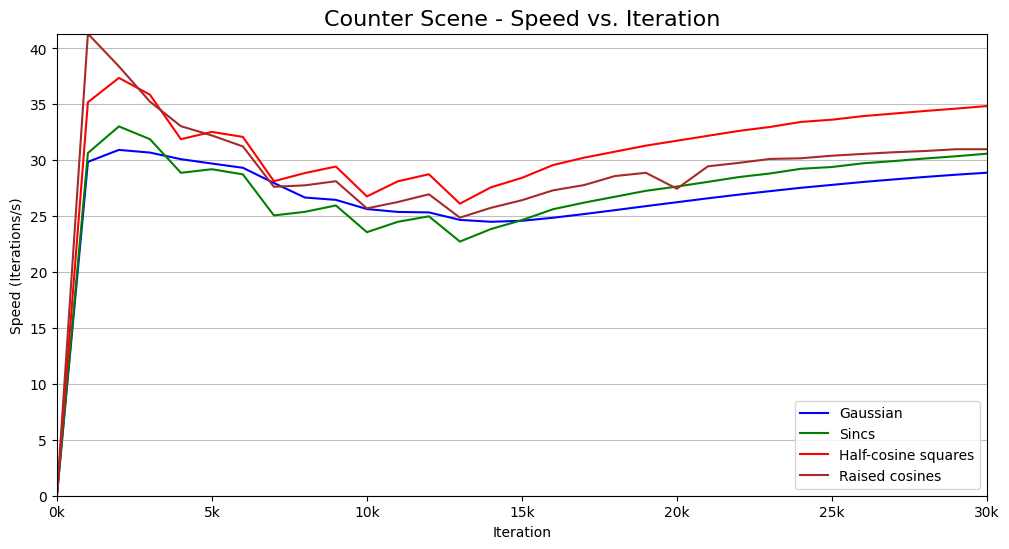}
    \end{minipage}\\[0.5em]

    % Row 2: Flowers
    \centering
    \begin{minipage}[c]{0.015\textwidth}
        \centering
        \rotatebox{90}{\textit{Flowers}}
    \end{minipage}%
    \begin{minipage}[c]{0.425\textwidth}
        \centering
        \includegraphics[width=\textwidth]{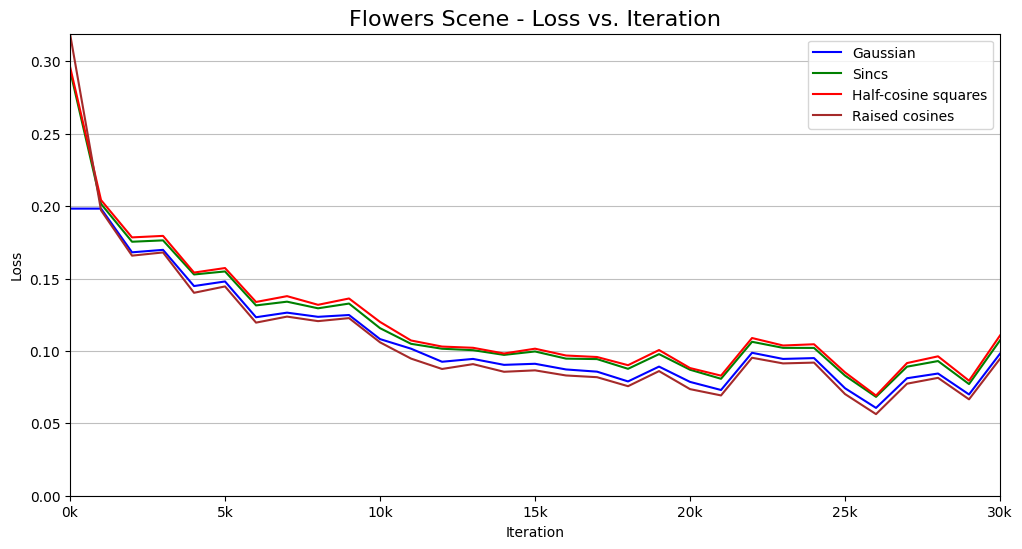}
    \end{minipage}%
    \begin{minipage}[c]{0.425\textwidth}
        \centering
        \includegraphics[width=\textwidth]{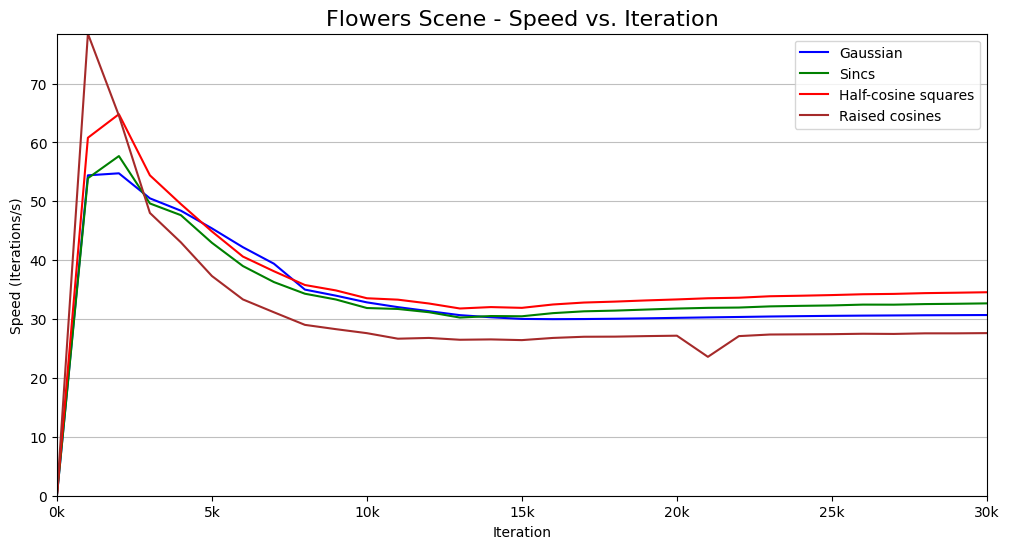}
    \end{minipage}\\[0.5em]

    % Row 4: Bonsai
    \centering
    \begin{minipage}[c]{0.015\textwidth}
        \centering
        \rotatebox{90}{\textit{Bonsai}}
    \end{minipage}%
    \begin{minipage}[c]{0.425\textwidth}
        \centering
        \includegraphics[width=\textwidth]{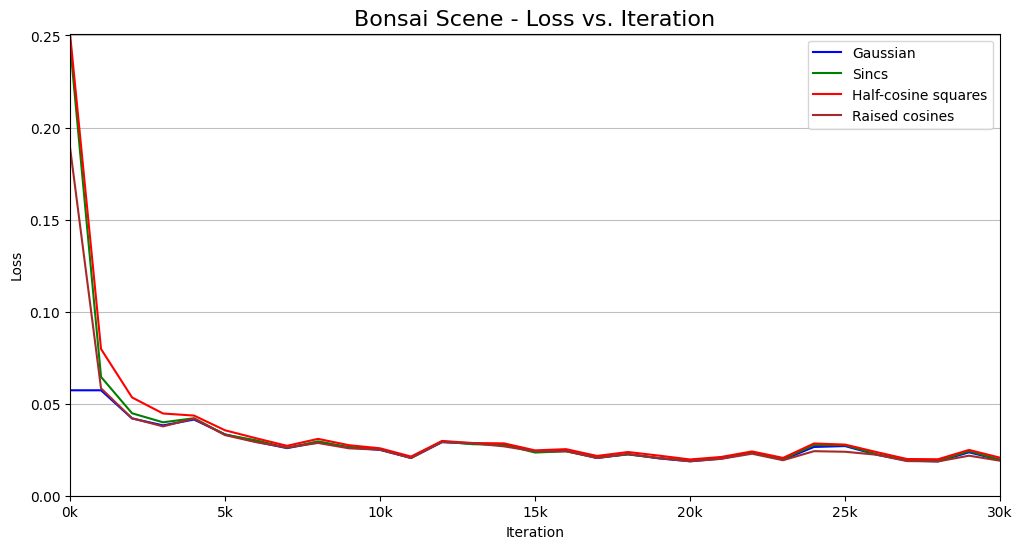}
    \end{minipage}%
    \begin{minipage}[c]{0.425\textwidth}
        \centering
        \includegraphics[width=\textwidth]{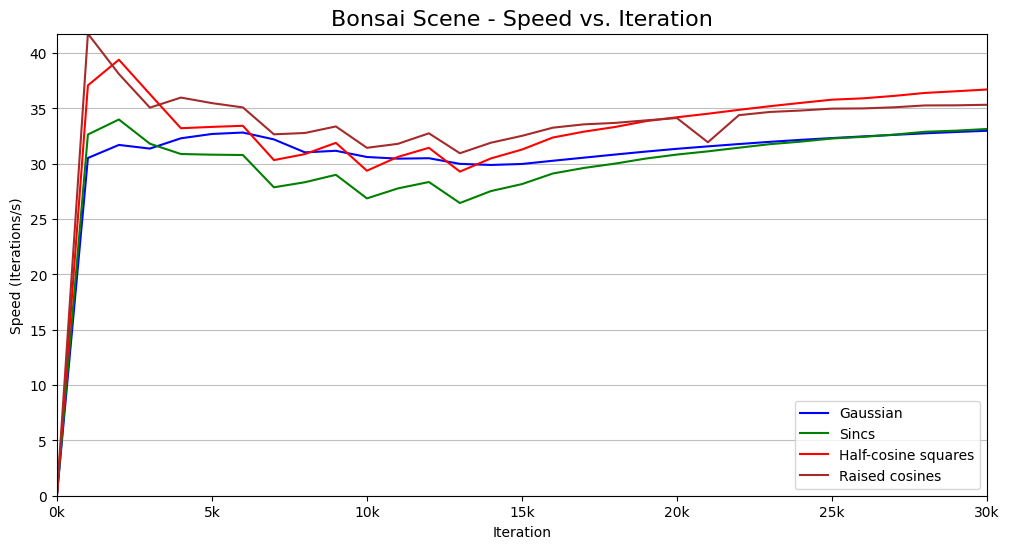}
    \end{minipage}\\[0.5em]
    
    \caption{Training loss and speed curves across different scenes reveal significant performance differences. Specifically, the superior convergence speed of half-cosine squared splatting stands out compared to other selected DARBFs, particularly the Gaussian function. Although all the selected functions exhibit similar loss curves, notable variations are observed in their respective training speed curves across various scenes. These differences can be attributed to the inherent characteristics of each scene, which influence the training dynamics of the functions.}
    \label{fig:speed_1}
\end{figure*}

\begin{figure*}
    % Column headings
    \begin{center}
        \parbox{0.425\textwidth}{\centering \textbf{Train Loss}}%
        \parbox{0.425\textwidth}{\centering \textbf{Train Speed}}
    \end{center}
    
    % Row 1: Bicycle
    \centering
    \begin{minipage}[c]{0.015\textwidth}
        \centering
        \rotatebox{90}{\textit{Bicycle}}
    \end{minipage}%
    \begin{minipage}[c]{0.425\textwidth}
        \centering
        \includegraphics[width=\textwidth]{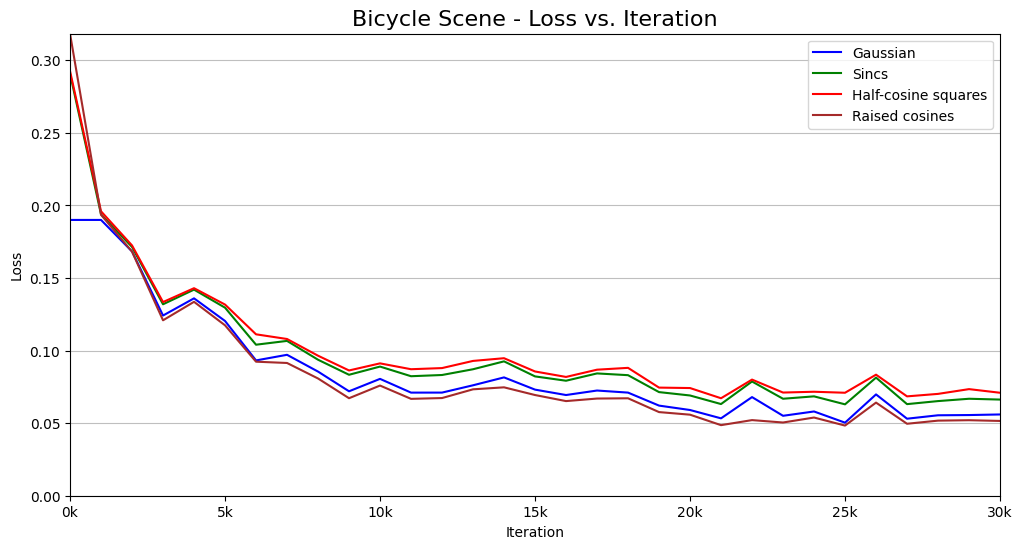}
    \end{minipage}%
    \begin{minipage}[c]{0.425\textwidth}
        \centering
        \includegraphics[width=\textwidth]{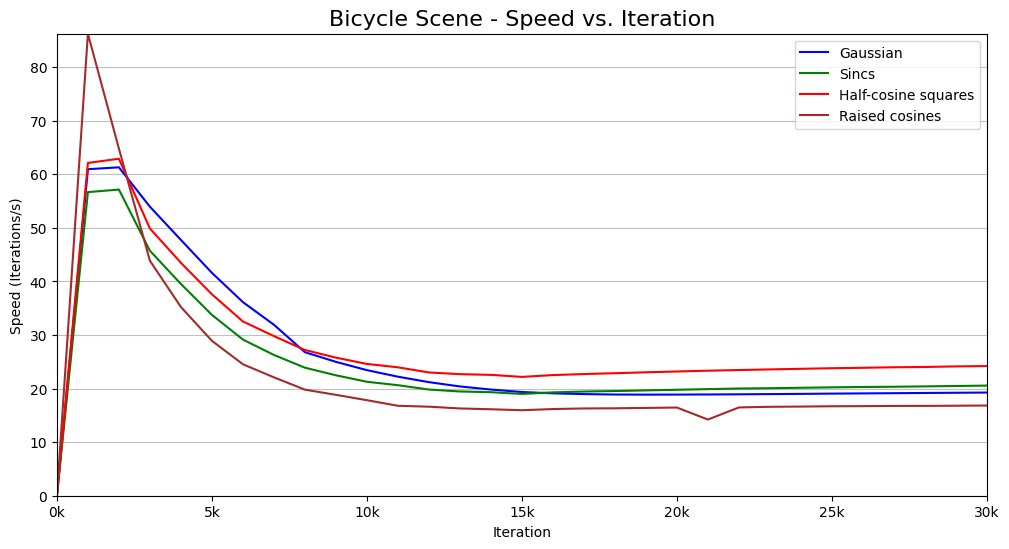}
    \end{minipage}\\[0.5em]

   % Row 3: Garden
   \centering
    \begin{minipage}[c]{0.015\textwidth}
        \centering
        \rotatebox{90}{\textit{Garden}}
    \end{minipage}%
    \begin{minipage}[c]{0.425\textwidth}
        \centering
        \includegraphics[width=\textwidth]{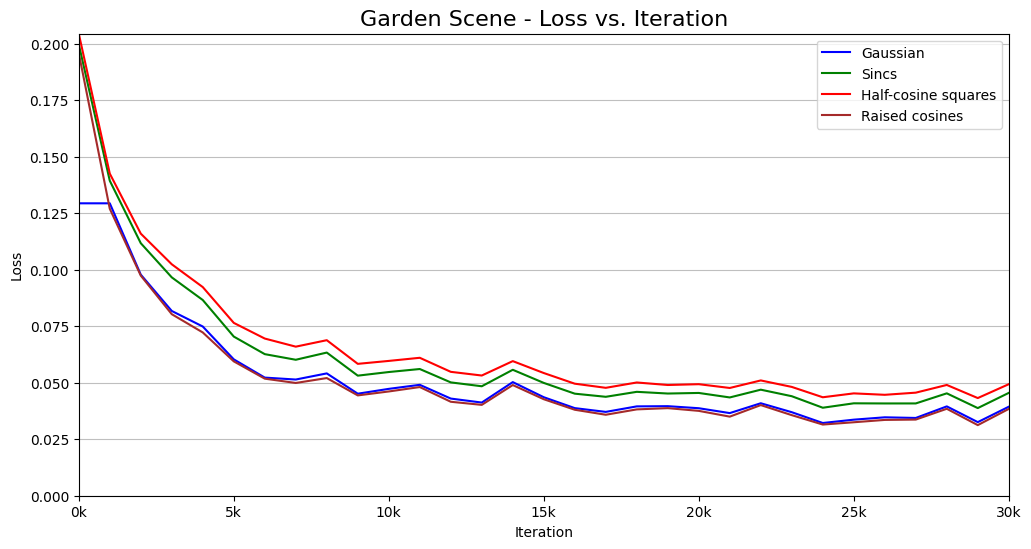}
    \end{minipage}%
    \begin{minipage}[c]{0.425\textwidth}
        \centering
        \includegraphics[width=\textwidth]{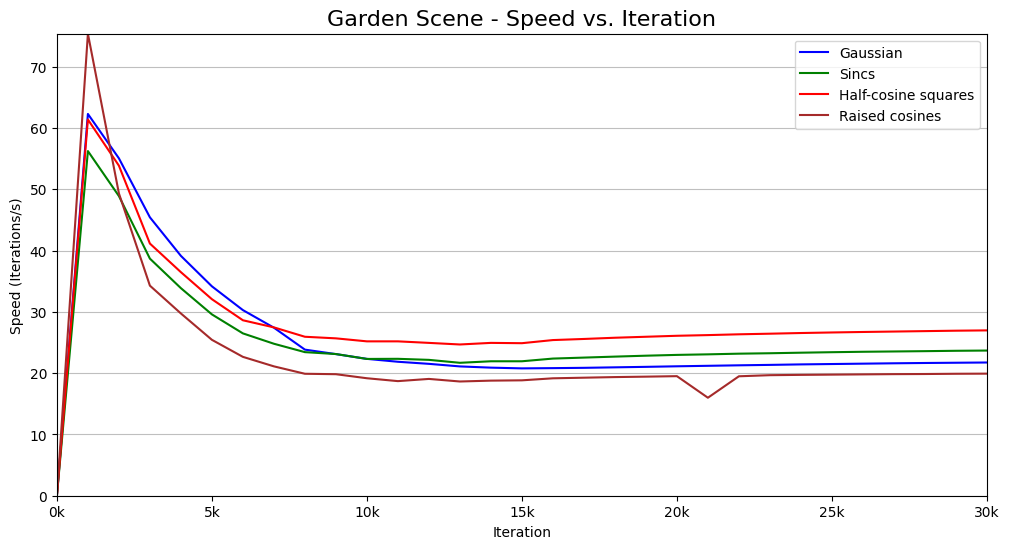}
    \end{minipage}\\[0.5em]

    % Row 4: Train
    \centering
    \begin{minipage}[c]{0.015\textwidth}
        \centering
        \rotatebox{90}{\textit{Train}}
    \end{minipage}%
    \begin{minipage}[c]{0.425\textwidth}
        \centering
        \includegraphics[width=\textwidth]{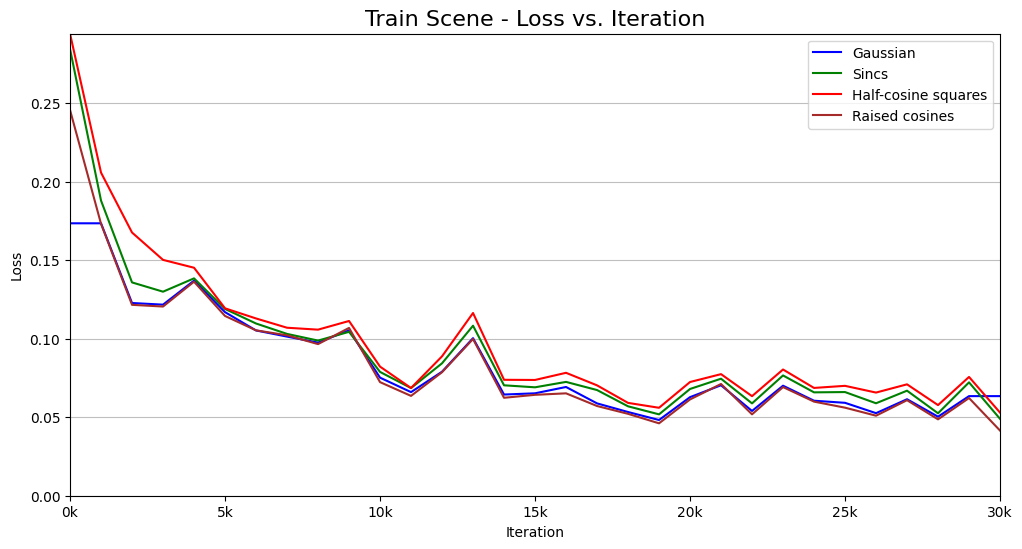}
    \end{minipage}%
    \begin{minipage}[c]{0.425\textwidth}
        \centering
        \includegraphics[width=\textwidth]{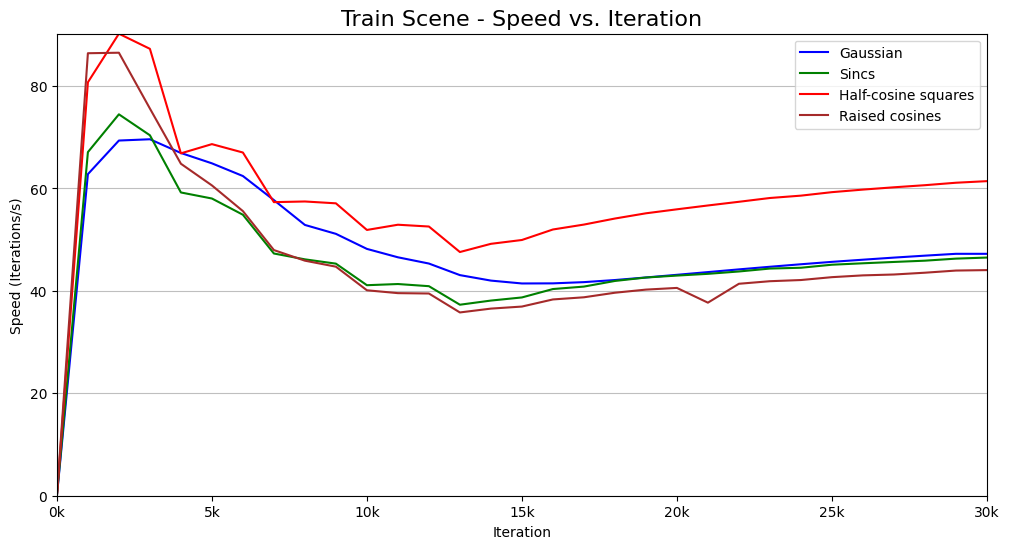}
    \end{minipage}\\[0.5em]

    % Row 5: Truck
    \centering
    \begin{minipage}[c]{0.015\textwidth}
        \centering
        \rotatebox{90}{\textit{Truck}}
    \end{minipage}%
    \begin{minipage}[c]{0.425\textwidth}
        \centering
        \includegraphics[width=\textwidth]{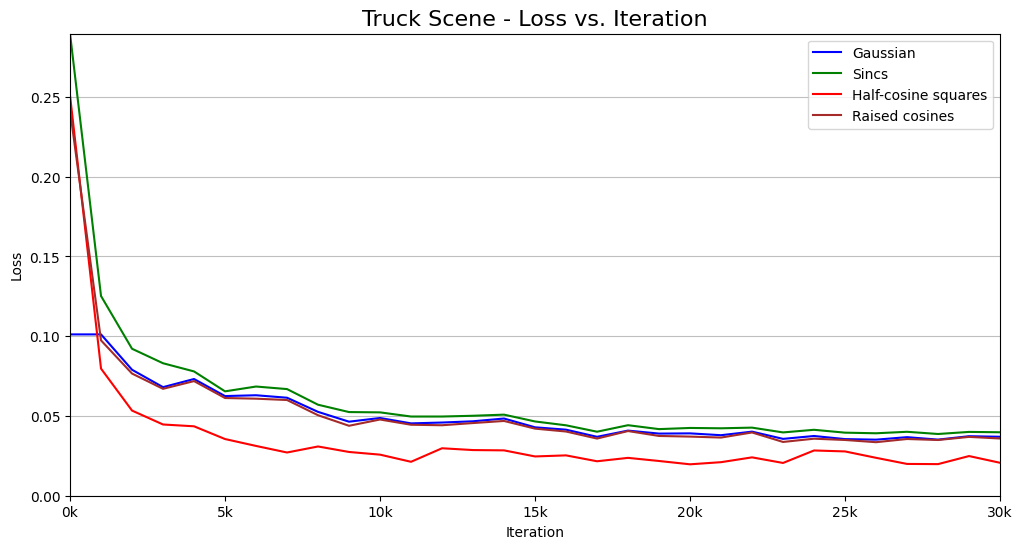}
    \end{minipage}%
    \begin{minipage}[c]{0.425\textwidth}
        \centering
        \includegraphics[width=\textwidth]{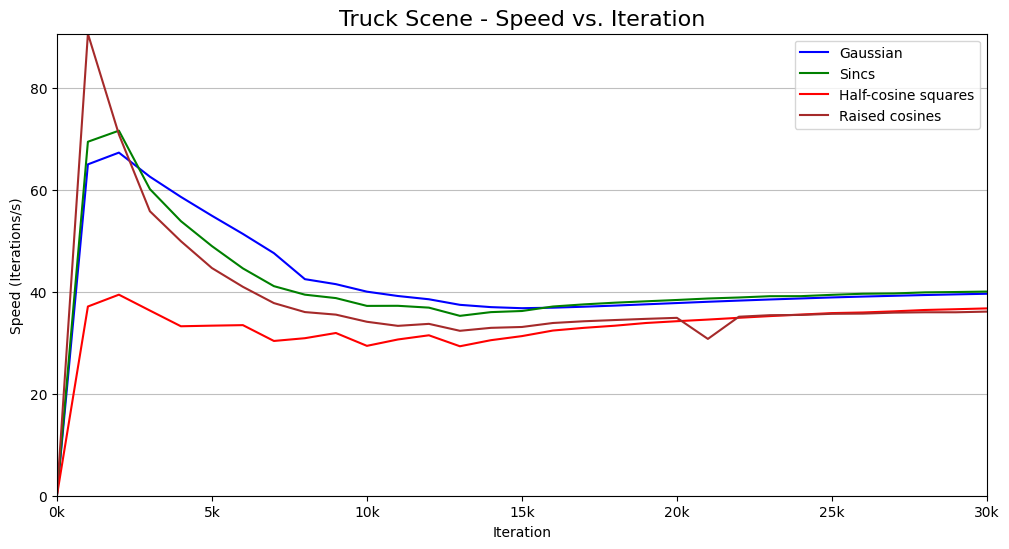}
    \end{minipage}\\[0.5em]

    % Row 5: Treehill
    \centering
    \begin{minipage}[c]{0.015\textwidth}
        \centering
        \rotatebox{90}{\textit{Treehill}}
    \end{minipage}%
    \begin{minipage}[c]{0.425\textwidth}
        \centering
        \includegraphics[width=\textwidth]{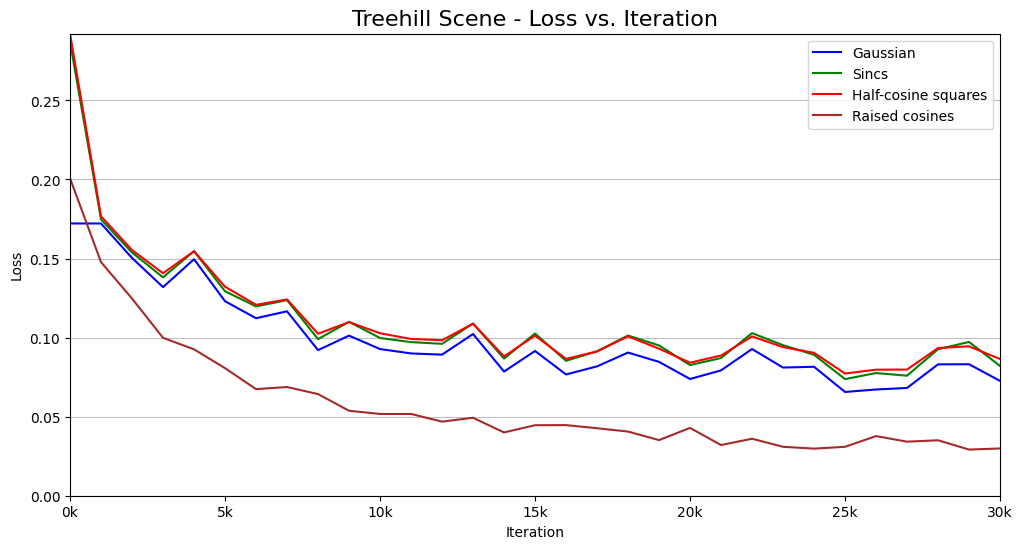}
    \end{minipage}%
    \begin{minipage}[c]{0.425\textwidth}
        \centering
        \includegraphics[width=\textwidth]{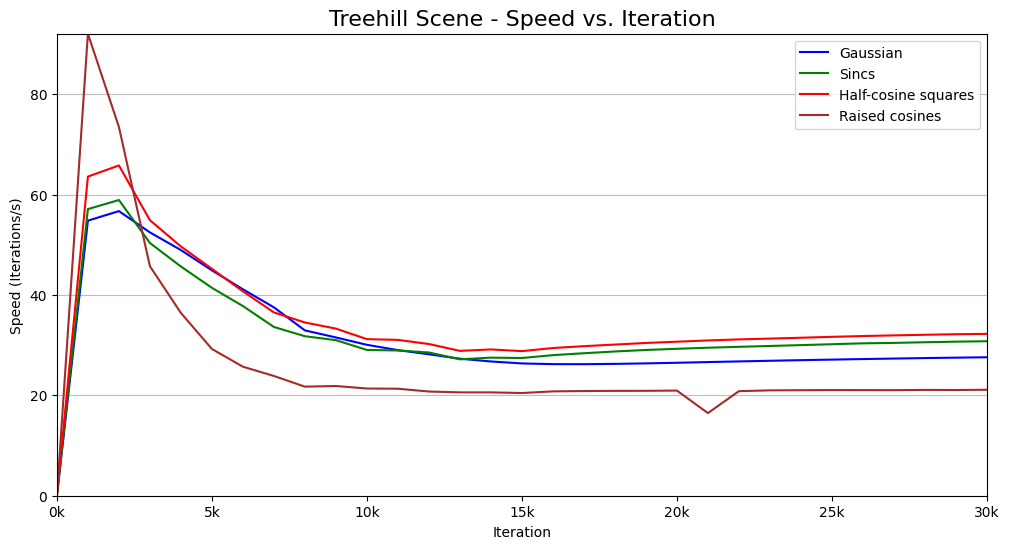}
    \end{minipage}\\[0.5em]
    
    \caption{Training loss and speed curves across different scenes reveal significant performance differences. As an exception, in the Truck scene, we observe that the speed curves of Gaussians and raised cosines overlap and outperform half-cosine squares. However, when considering the overall performance across all scenes, half-cosine squares demonstrates superior efficiency in training time.}
    \label{fig:speed_2}
\end{figure*}

This is further illustrated in Figures~\ref{fig:speed_1} and~\ref{fig:speed_2}, which provide a detailed analysis of the training loss and speed curves across various dataset scenes. Overall, despite having similar training loss curves with 3DGS, half-cosine splatting demonstrates superior performance compared to Gaussians.

\subsection{Reduced Memory Usage}
As previously mentioned, half cosine squared splatting specifically requires fewer primitives compared to Gaussians. This results in lower memory usage, as they provide higher opacity values across most of the regions they cover. In contrast, Gaussians require more primitives to achieve a similar accumulated opacity coverage. By using fewer half cosine squared primitives, we can achieve the desired color representation in the image space more efficiently. Similarly, sinc splatting and inverse multiquadric splatting also consume lesser memory compared to Gaussians. The results in Table ~\ref{table:SupTable1} and Table ~\ref{table:SupTable2} further showcase this.

\section{Extended Results and Simulations}
\label{sec:sup_results_1dsim}

\textbf{Extended Results. } As mentioned in our paper, we trained our models on a single NVIDIA GeForce RTX 4090 GPU and recorded the training time. Since the benchmark models from other papers \cite{kerbl3Dgaussians, Ges} were trained on different GPUs, we applied a scaling factor to ensure a fair comparison of training times with the \emph{original} papers' results. According to \cite{lambdaGPUbenchmarks}, the relative training throughput of the RTX 4090 GPU and other GPU models (specifically, the RTX 3090 and RTX A6000 GPUs) can be determined with respect to a 1xLambdaCloud V100 16GB GPU. By dividing the training time data by these values, we have presented our training time results in a fair and comparable manner in Table~\ref{Table:FinalResultTable}.

A detailed breakdown of our results across every scene in the Mip-NeRF 360 \cite{Mip-nerf_360}, Tanks\&Temples \cite{Tanks&Temples2017}, and Deep Blending \cite{DeepBlending2018} datasets, along with their average values per dataset, is provided in Table~\ref{table:SupTable1} and Table~\ref{table:SupTable2}. These results pertain to the selected DARB-Splatting algorithms, namely raised cosine splatting (3DRCS), half-cosine squared splatting (3DHCS), sinc splatting (3DSS), and inverse multiquadric splatting (3DIQS). Key evaluation metrics, including PSNR, SSIM, LPIPS, memory usage, and training time for both 7k and 30k iterations, are analyzed in detail. These are presented alongside the results from implementing the \emph{updated codebase} of 3DGS on our single NVIDIA GeForce RTX 4090 GPU to ensure a fair comparison. Our pipeline is anchored on this \emph{updated codebase}, which produces improved results compared to those reported in the original 3DGS paper \cite{kerbl3Dgaussians}. As shown in the tables, each DARB-Splatting algorithm demonstrates unique advantages in different utilities.

\begin{table}[h!]
    \centering
    \scriptsize
    %\renewcommand{\arraystretch}{1.1}
    % \caption{Comparison of 1D simulation results with Gaussian components vs. other reconstruction kernel components. Note that $^*$ indicates that we achieved better reconstruction with fewer components ($N=8$) in terms of mean squared error, using our modified raised cosine pulses compared to Gaussians.}
    % \label{tab:functions_1D_results}
    \begin{tabular}{@{}lccr@{}}
        \toprule
        \textbf{Function} & \textbf{Loss ($N=5$)} & \textbf{Loss ($N=10$)} \\
        \midrule
        Gaussian & \cellcolor{orange!15} 0.0002 & \cellcolor{orange!35} 0.0001 \\
        Modified half cosine & 0.0014 & 0.0003 \\
        Modified raised cosine & \cellcolor{orange!15} 0.0002 & \cellcolor{red!35} 0.00004$^*$ \\
        Modified sinc (modulus) & 0.0030 & 0.0002 \\
        \bottomrule
    \end{tabular}
    \caption{Comparison of 1D simulation results with Gaussian components vs. other reconstruction kernel components. Note that $^*$ indicates that we achieved better reconstruction with fewer components ($N=8$) in terms of mean squared error, using our modified raised cosine pulses compared to Gaussians.}
    \label{tab:functions_1D_results}
\end{table}

\begin{figure*}
     % Column headings
    \parbox{0.33\textwidth}{\centering {N=1}} 
    \parbox{0.33\textwidth}{\centering {N=5}} 
    \parbox{0.33\textwidth}{\centering {N=10}} \\
    \includegraphics[width=0.33\textwidth]{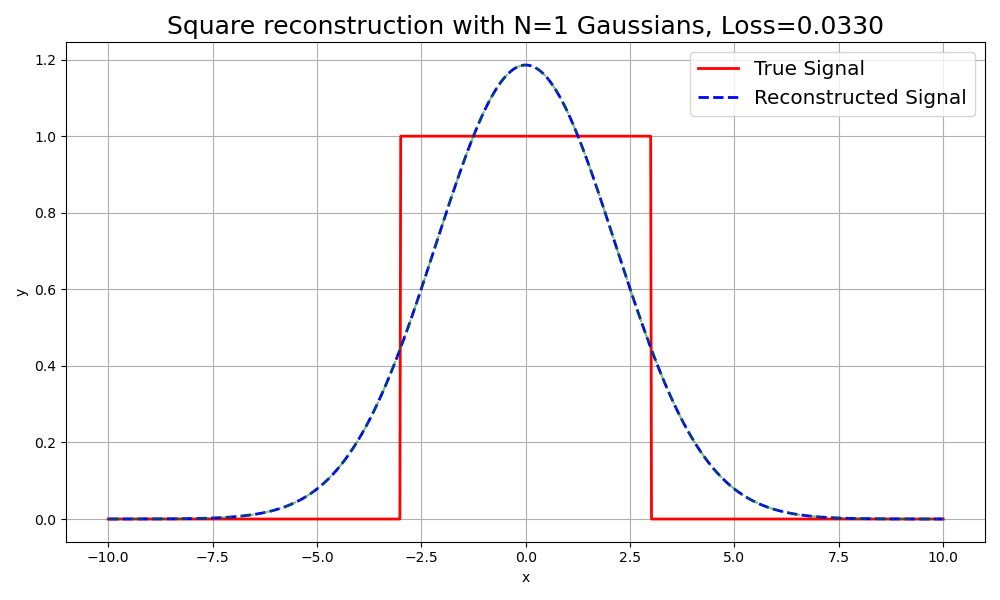} 
    \includegraphics[width=0.33\textwidth]{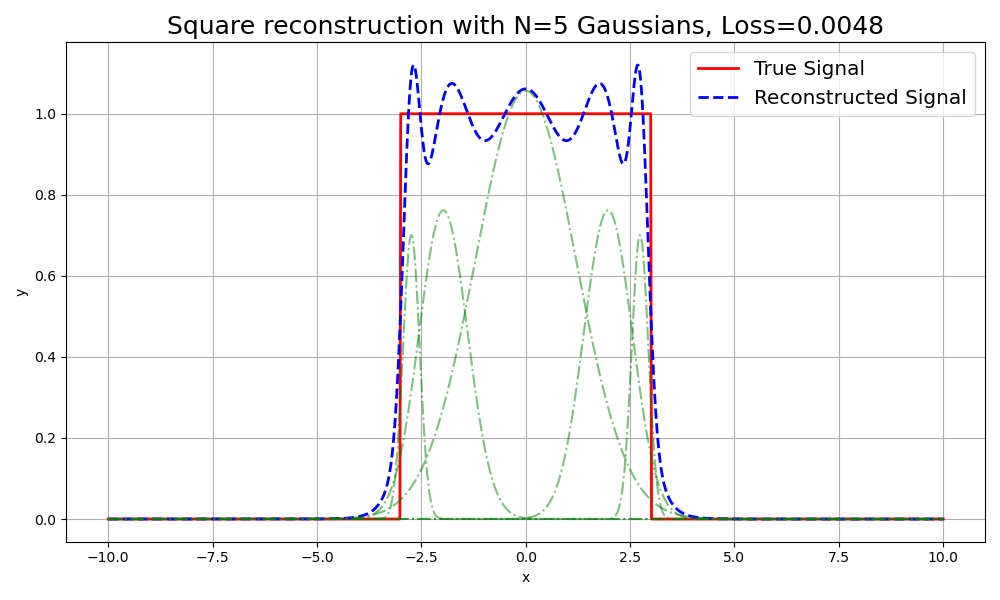} 
    \includegraphics[width=0.33\textwidth]{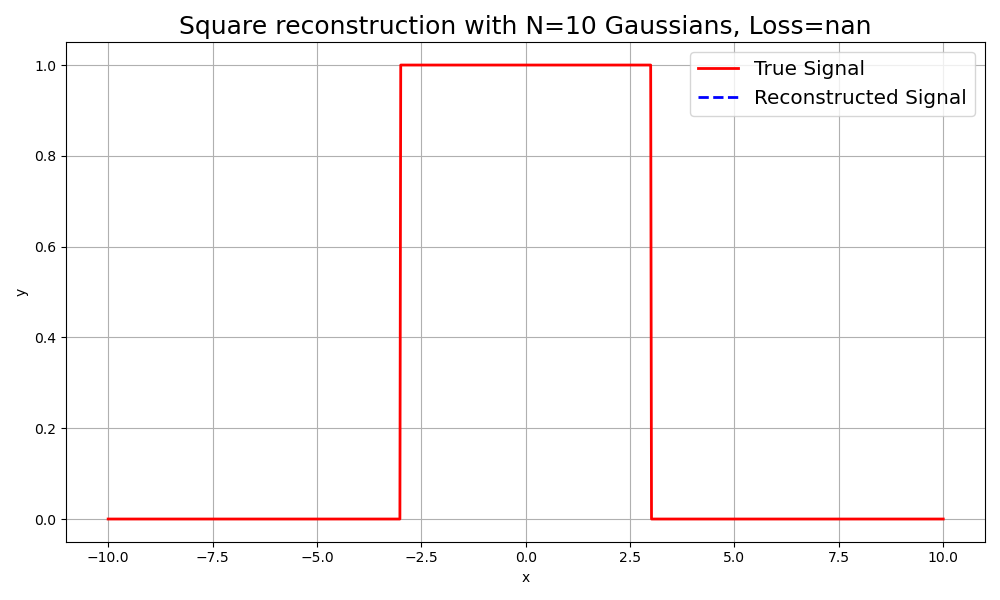} \\
    \includegraphics[width=0.33\textwidth]{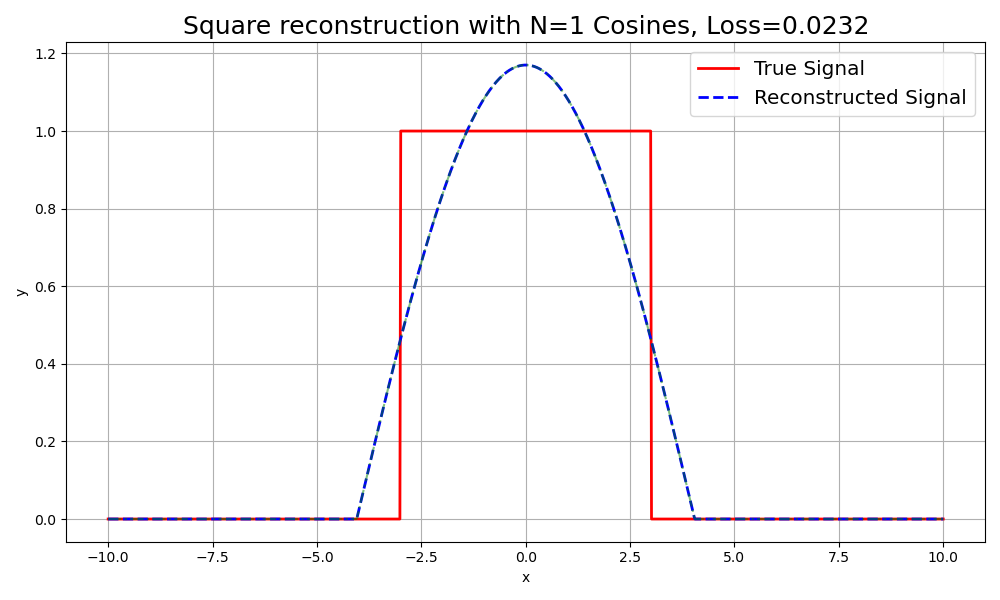} 
    \includegraphics[width=0.33\textwidth]{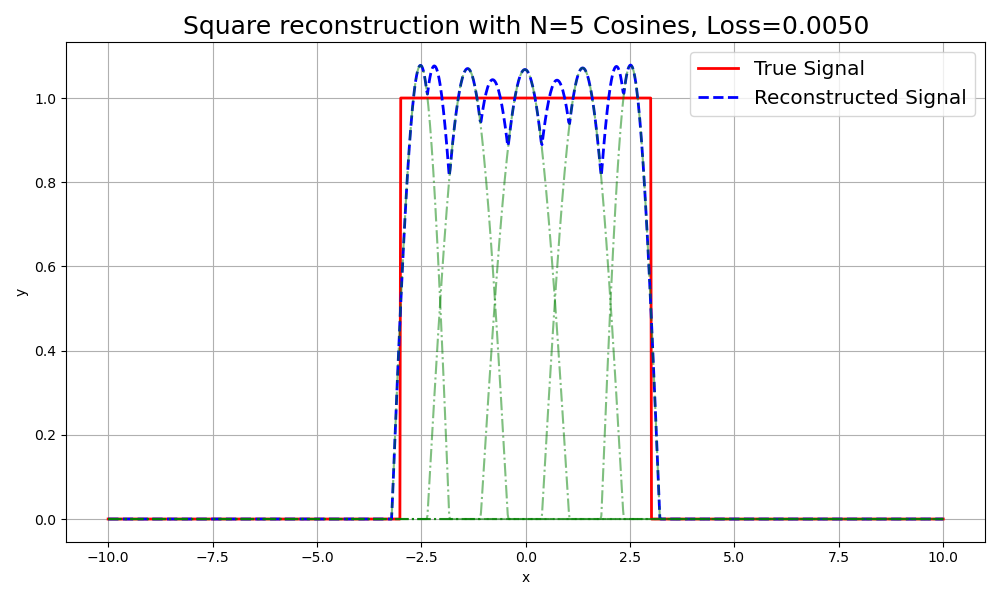} 
    \includegraphics[width=0.33\textwidth]{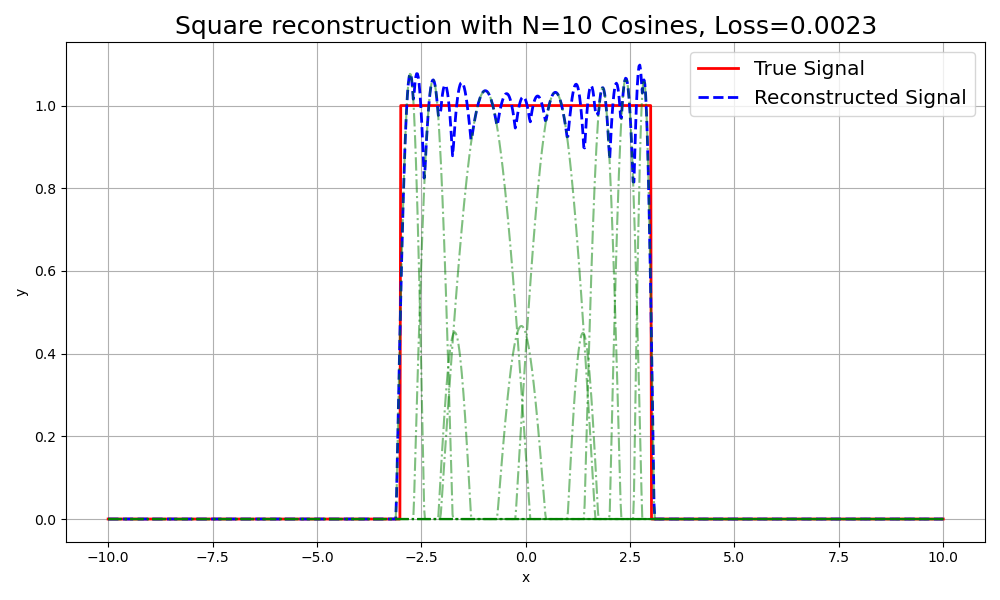} \\
    \includegraphics[width=0.33\textwidth]{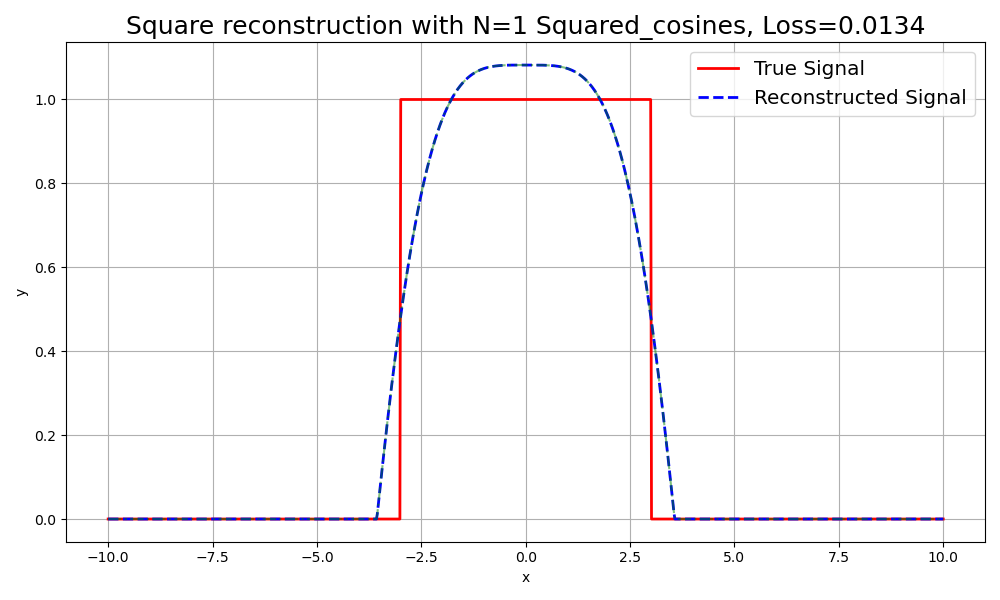} 
    \includegraphics[width=0.33\textwidth]{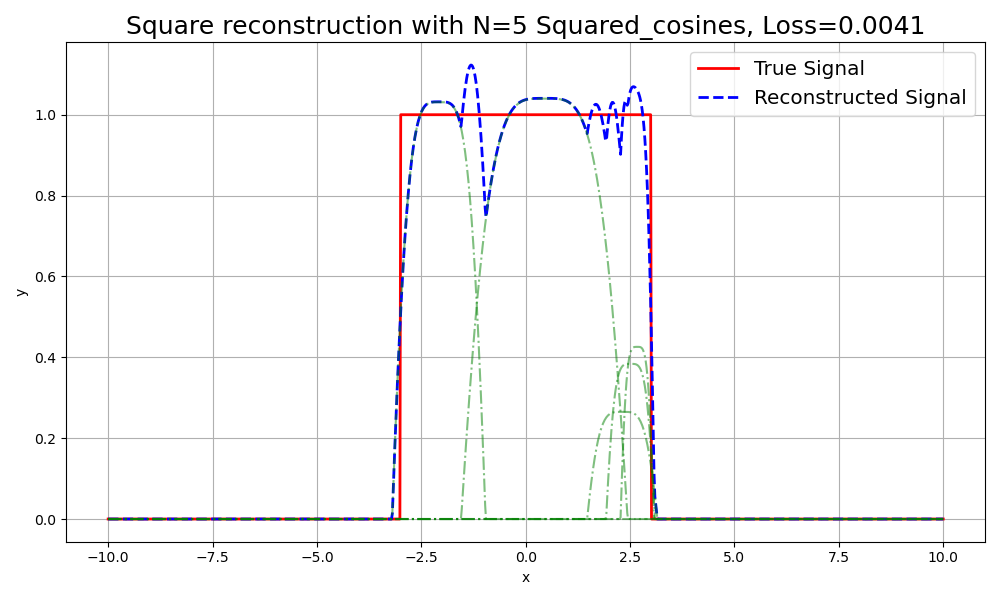} 
    \includegraphics[width=0.33\textwidth]{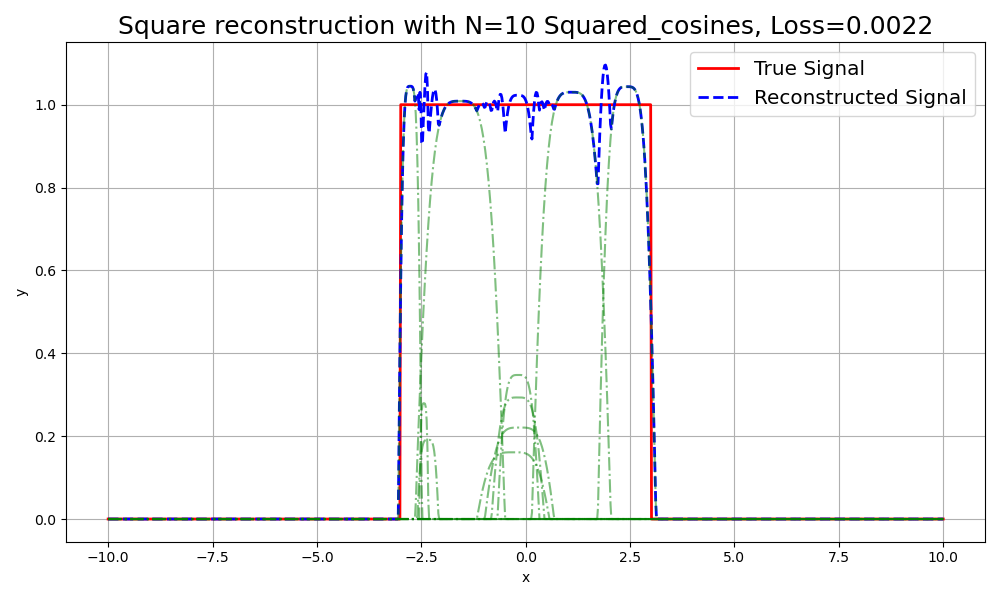} \\
    \includegraphics[width=0.33\textwidth]{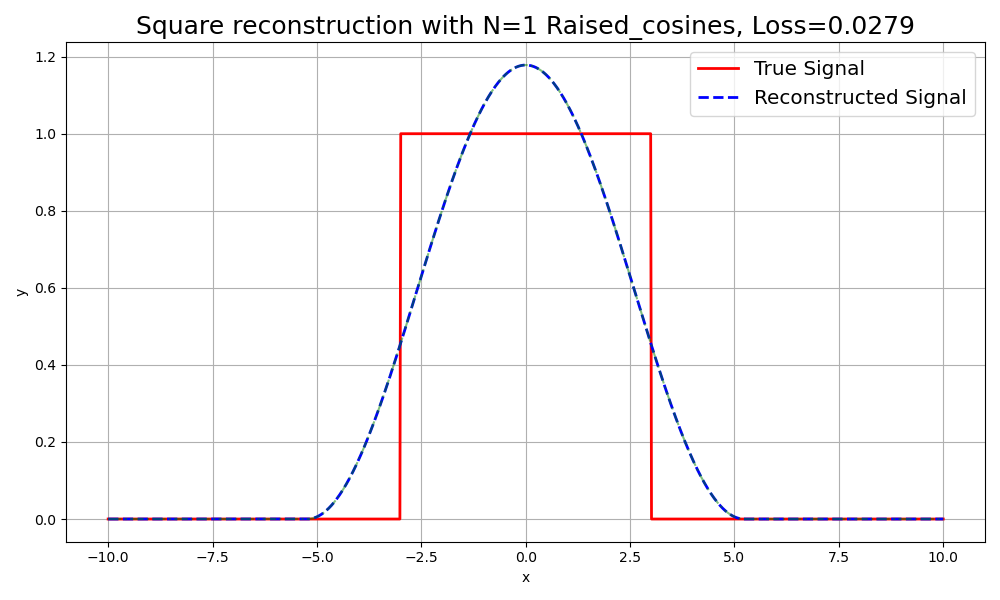} 
    \includegraphics[width=0.33\textwidth]{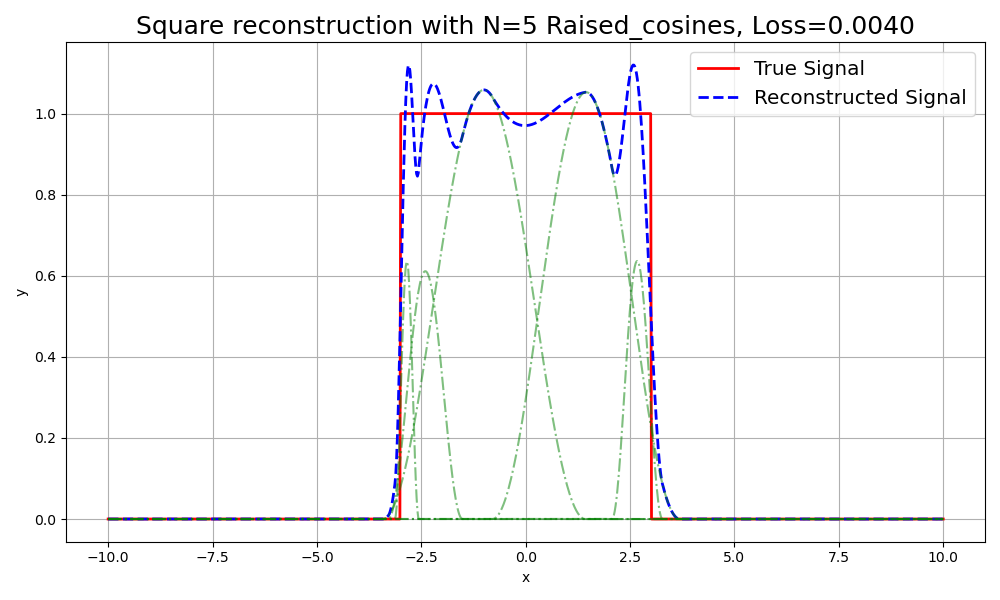} 
    \includegraphics[width=0.33\textwidth]{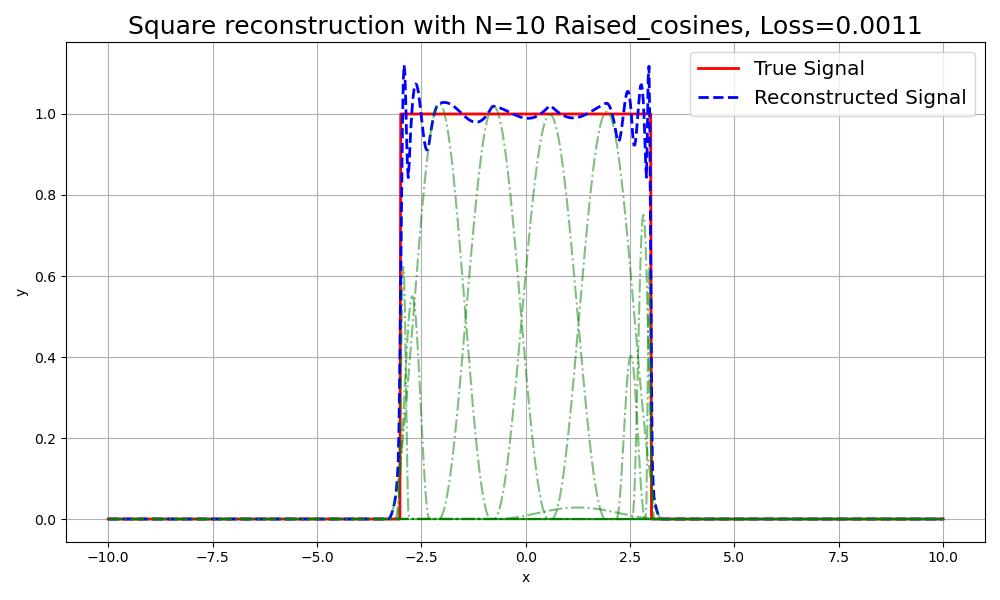} \\
    \includegraphics[width=0.33\textwidth]{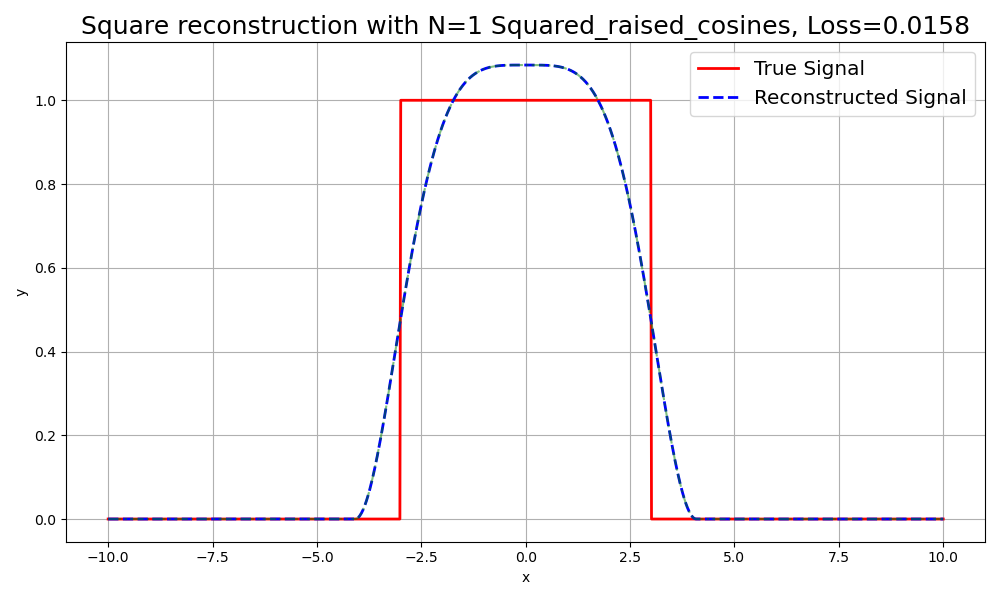} 
    \includegraphics[width=0.33\textwidth]{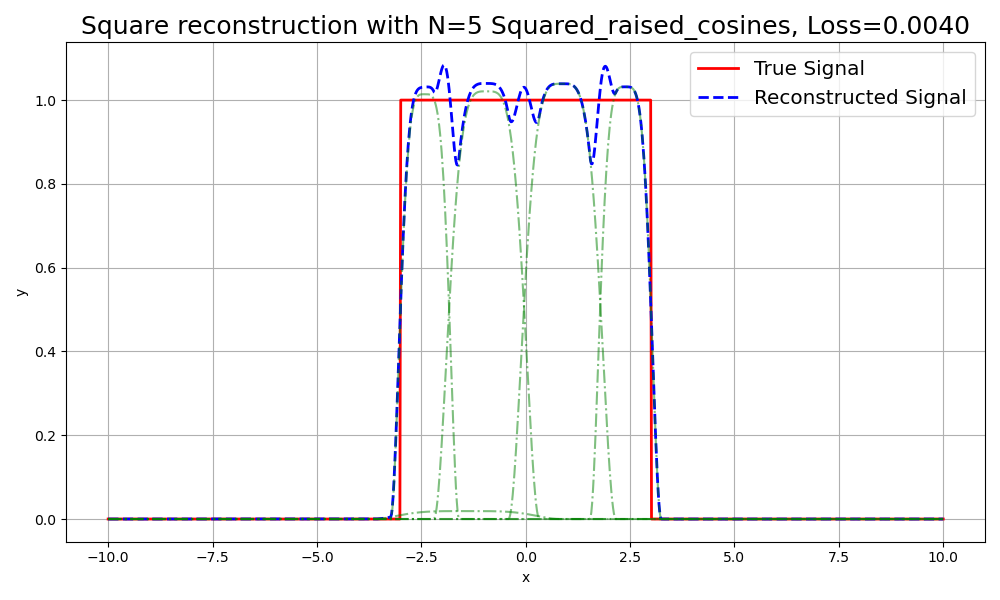} 
    \includegraphics[width=0.33\textwidth]{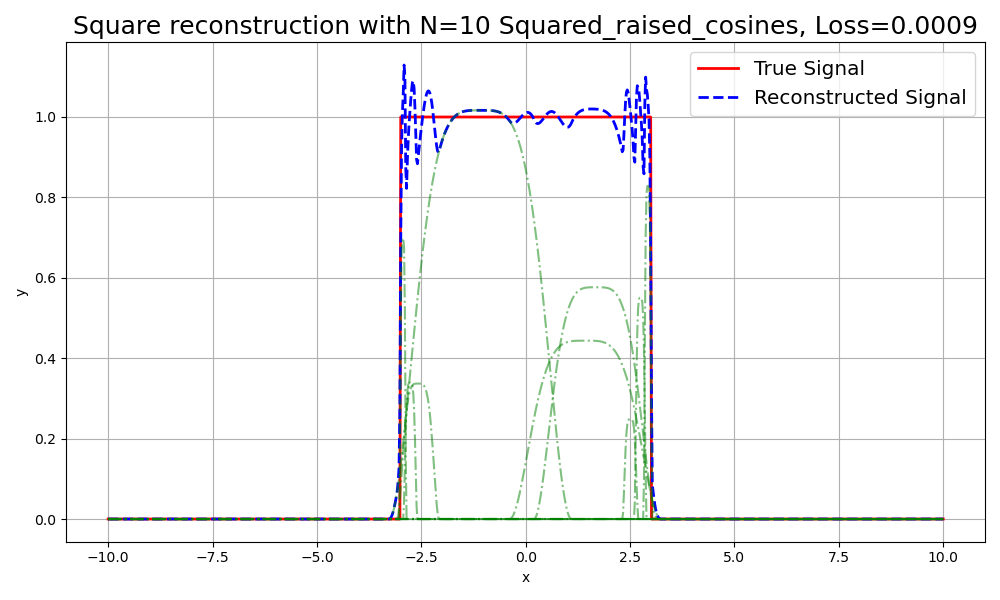} \\
    \includegraphics[width=0.33\textwidth]{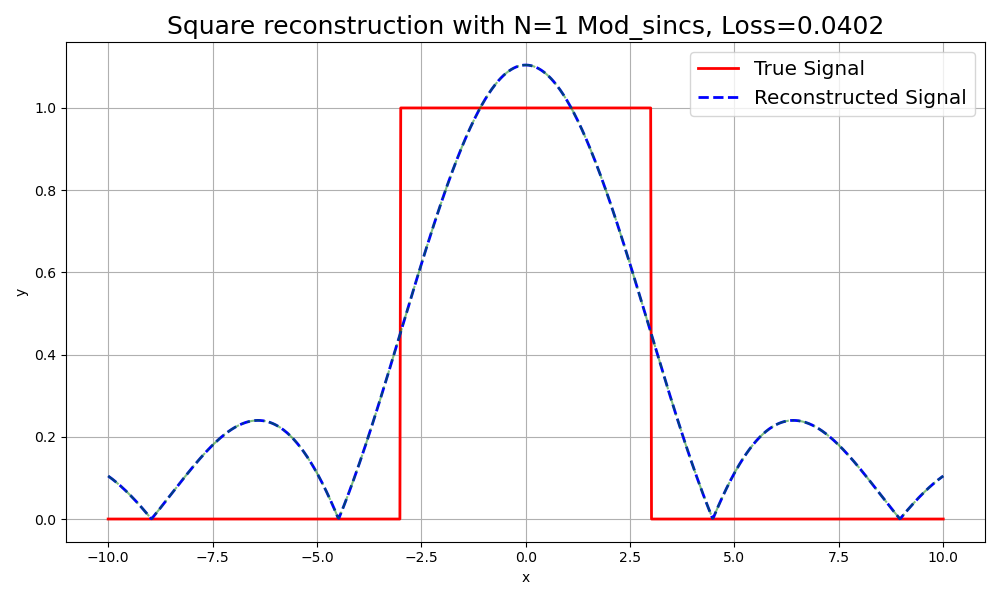} 
    \includegraphics[width=0.33\textwidth]{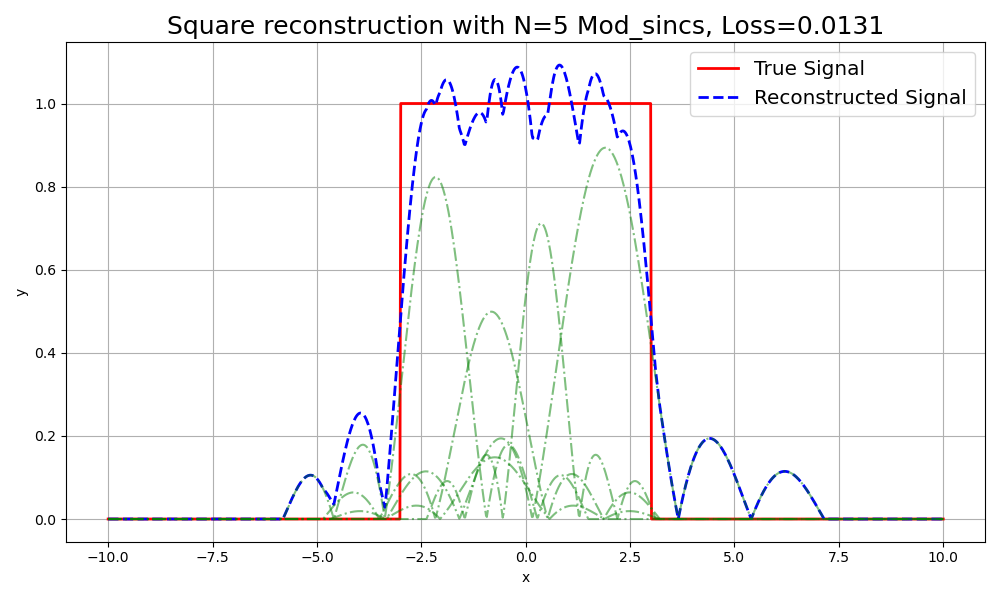} 
    \includegraphics[width=0.33\textwidth]{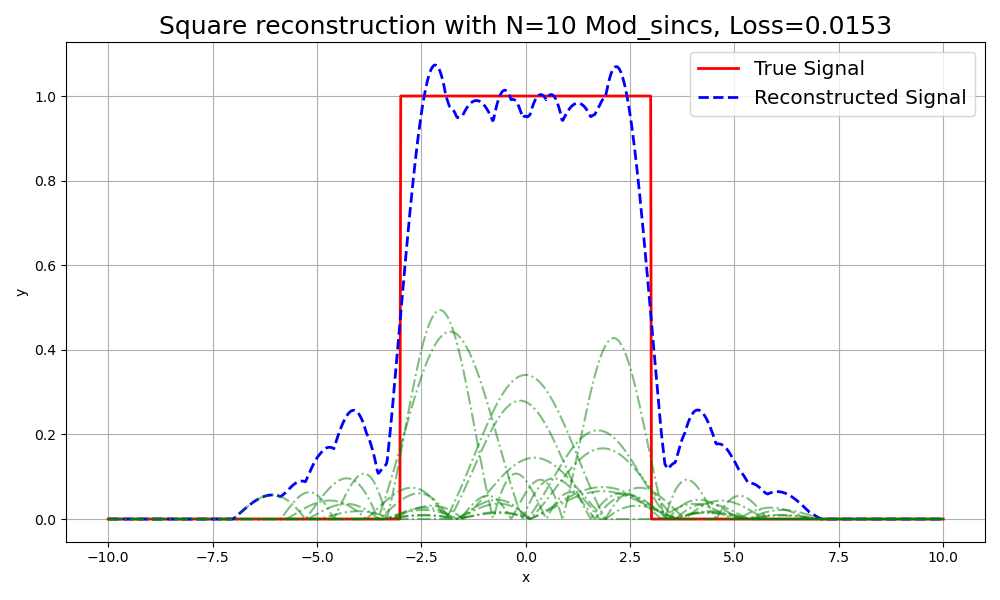} \\
    \caption{Visualization of 1D simulations for different splatting methods with varying primitives (N=1, N=5, N=10) for a square pulse. Each row corresponds to a specific splatting method: Gaussian, Cosine, Squared Cosine, Raised Cosine, Squared Raised Cosine, and Modulated Sinc. The columns represent the number of primitives used.}
    \label{fig:1DSim1}
\end{figure*}

\begin{figure*}
     % Column headings
    \parbox{0.33\textwidth}{\centering {N=1}} 
    \parbox{0.33\textwidth}{\centering {N=5}} 
    \parbox{0.33\textwidth}{\centering {N=10}} \\
    \includegraphics[width=0.33\textwidth]{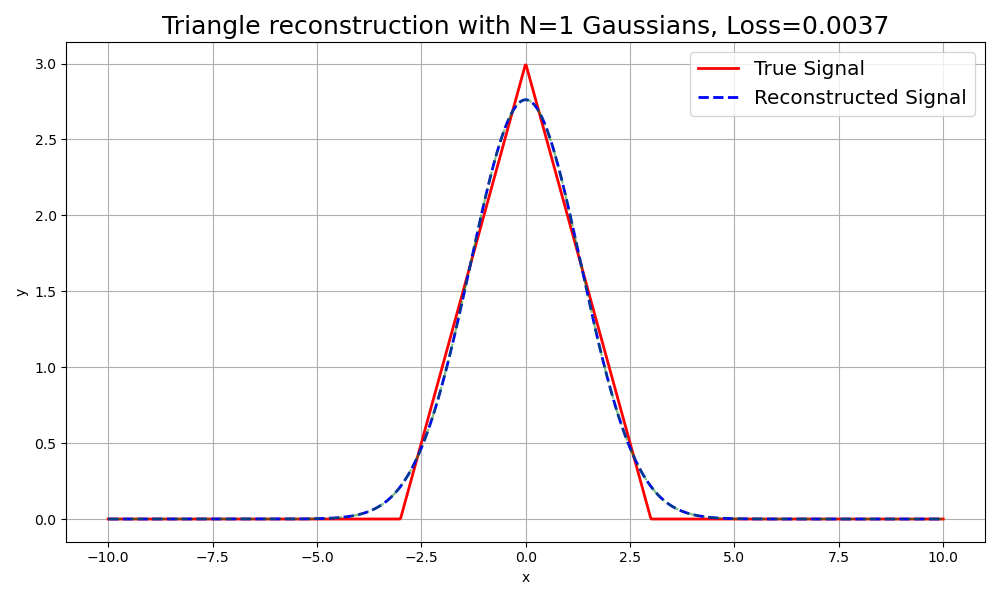} 
    \includegraphics[width=0.33\textwidth]{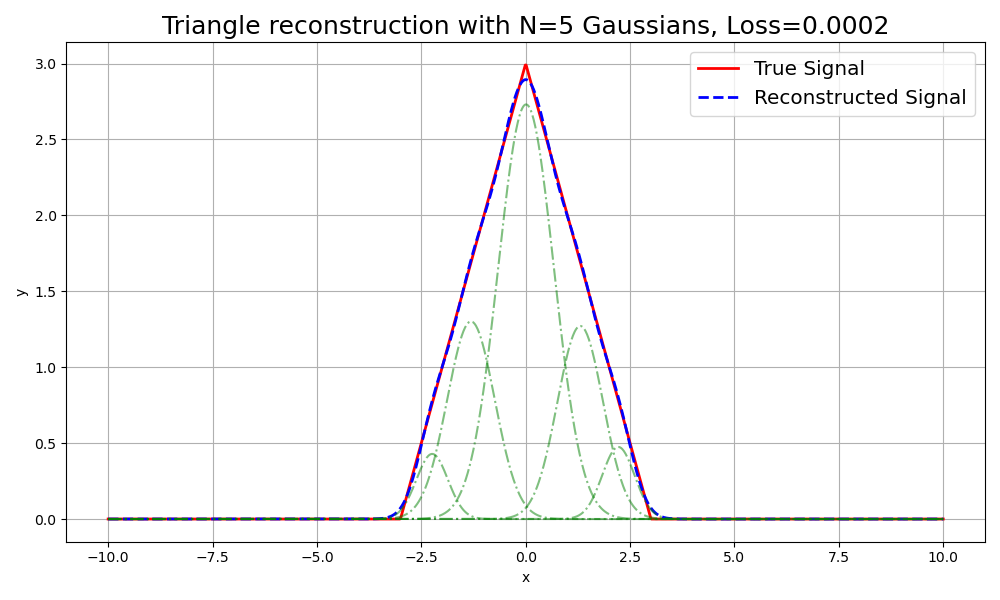} 
    \includegraphics[width=0.33\textwidth]{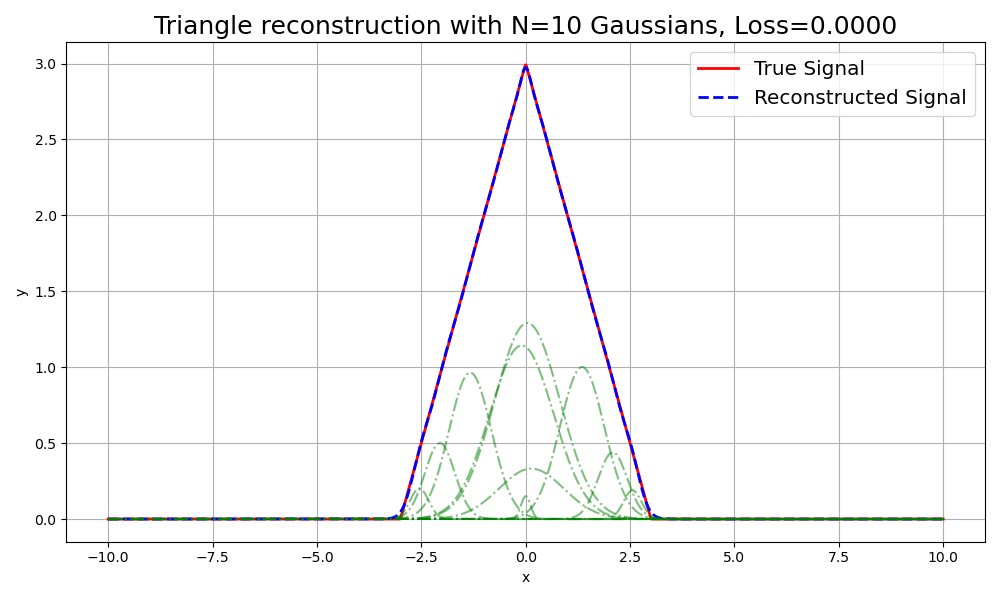} \\
    \includegraphics[width=0.33\textwidth]{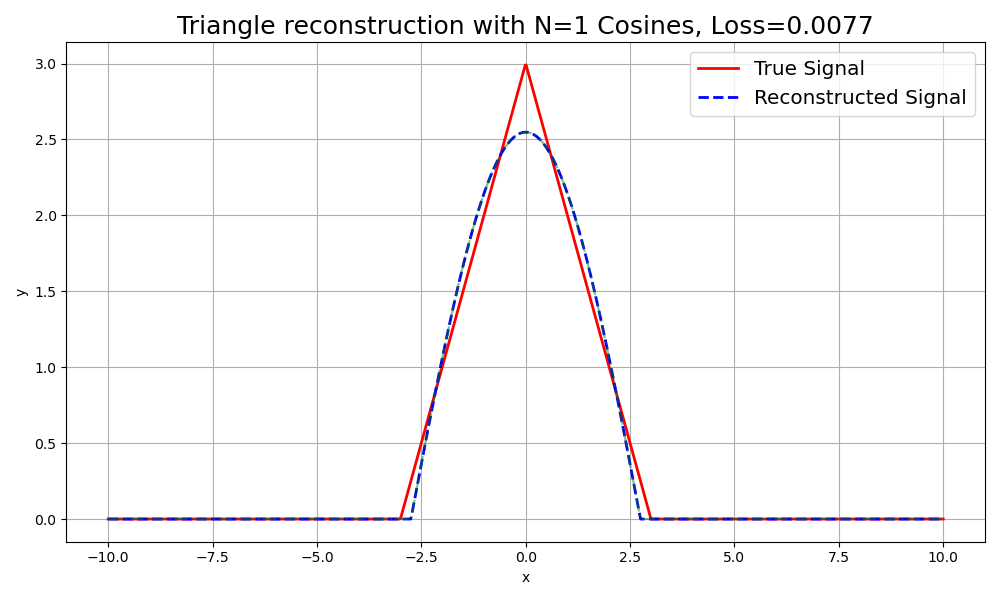} 
    \includegraphics[width=0.33\textwidth]{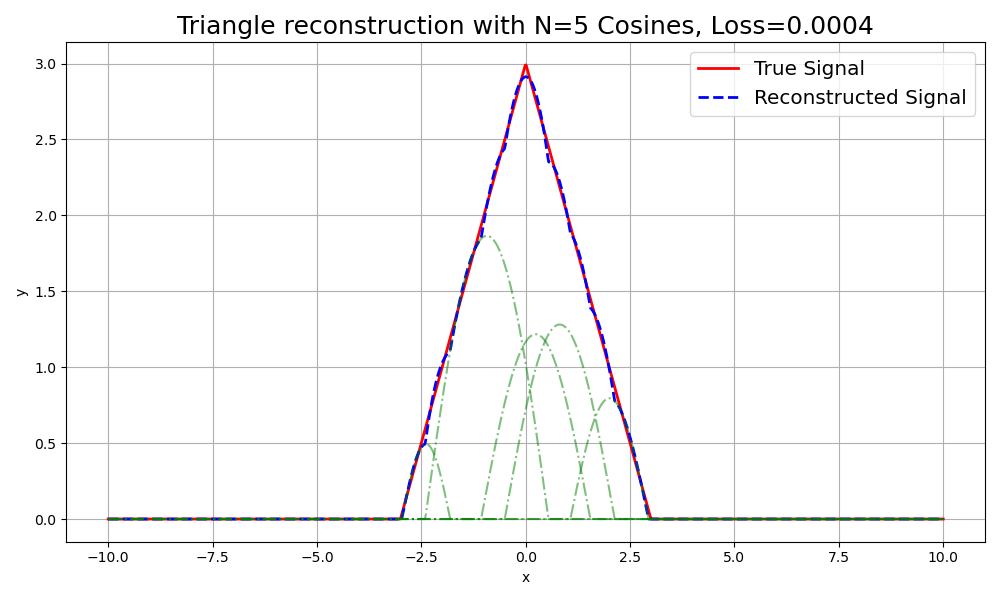} 
    \includegraphics[width=0.33\textwidth]{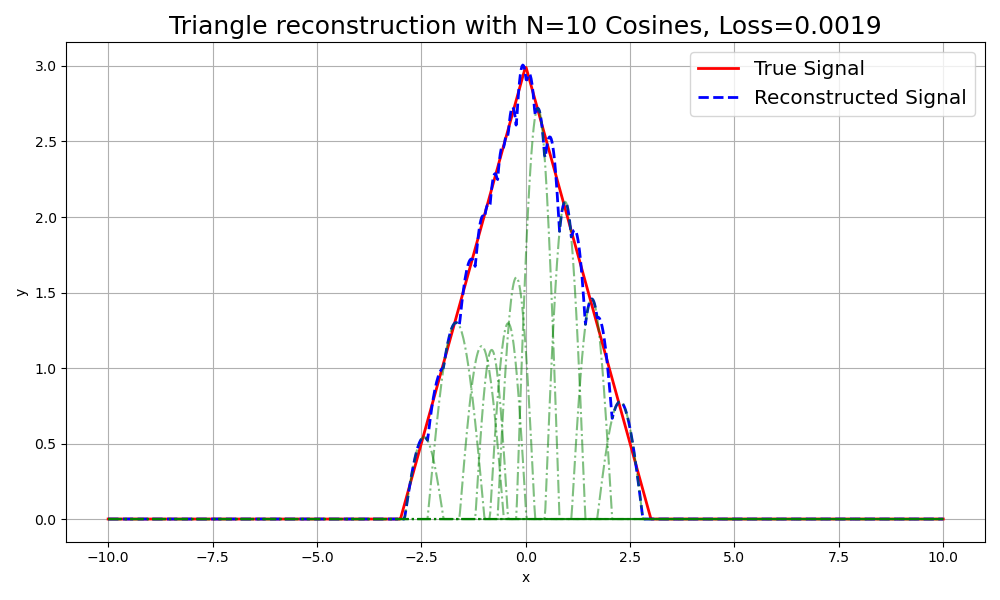} \\
    \includegraphics[width=0.33\textwidth]{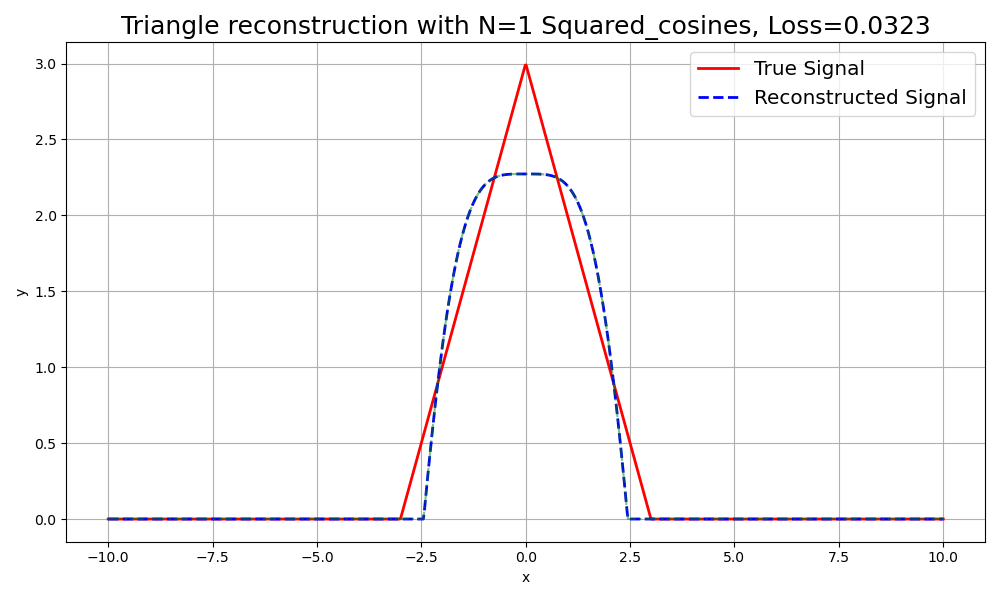} 
    \includegraphics[width=0.33\textwidth]{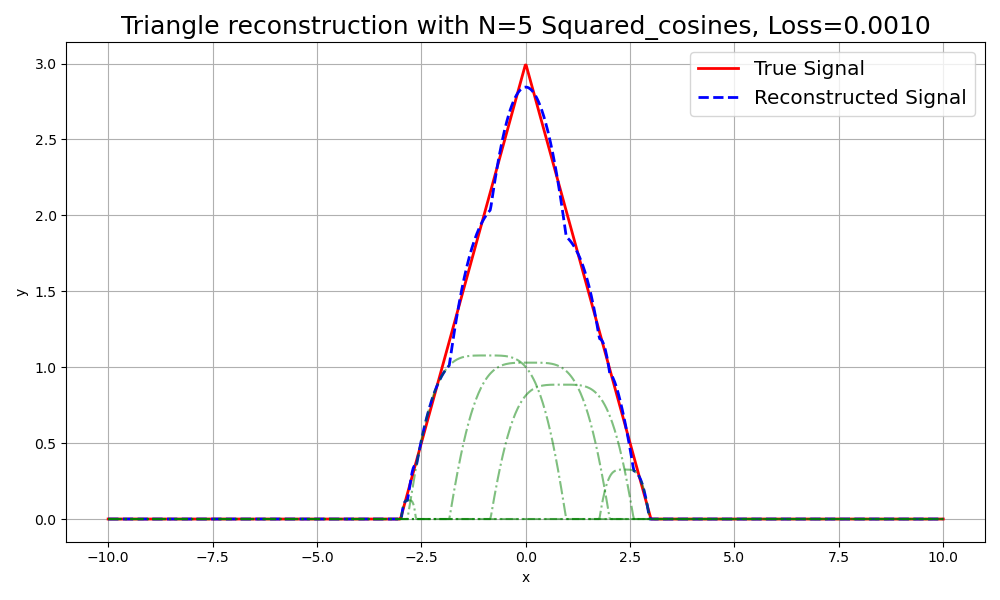} 
    \includegraphics[width=0.33\textwidth]{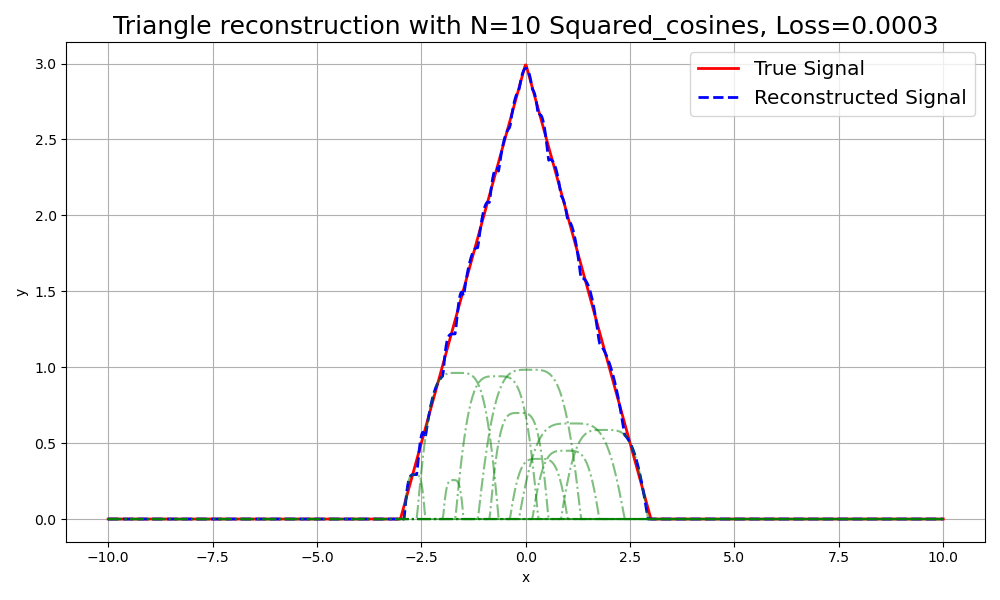} \\
    \includegraphics[width=0.33\textwidth]{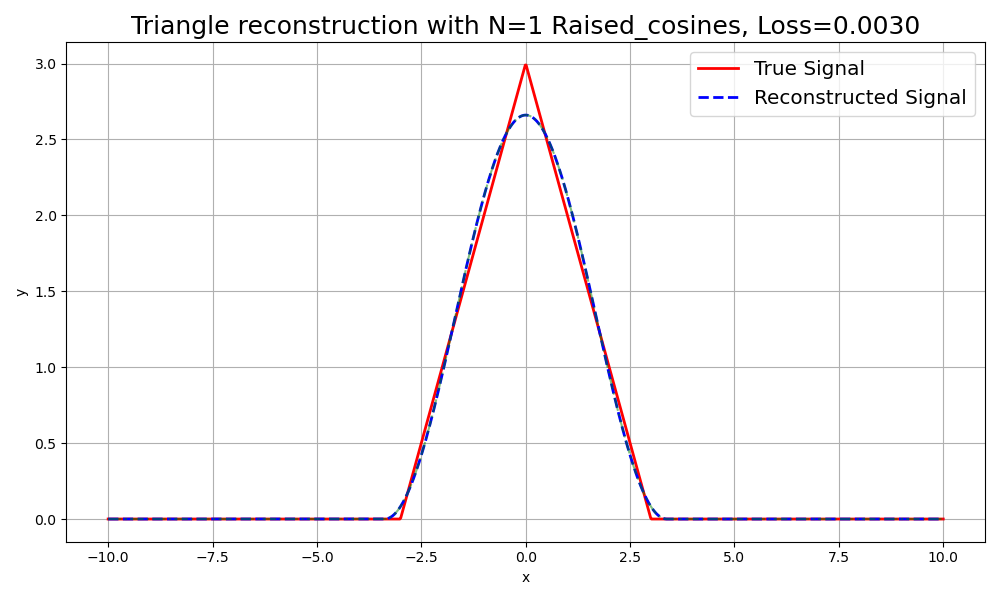} 
    \includegraphics[width=0.33\textwidth]{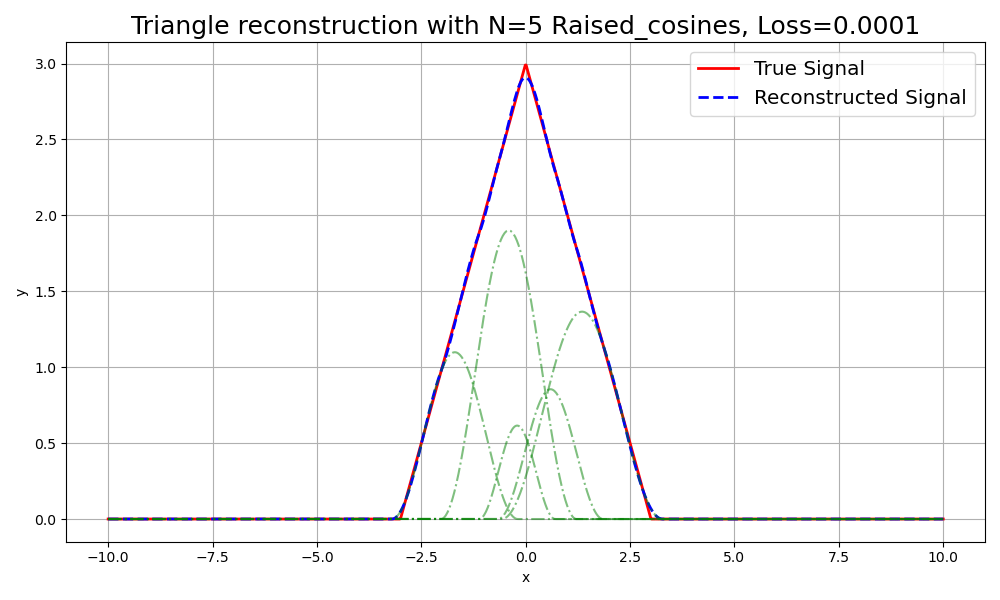} 
    \includegraphics[width=0.33\textwidth]{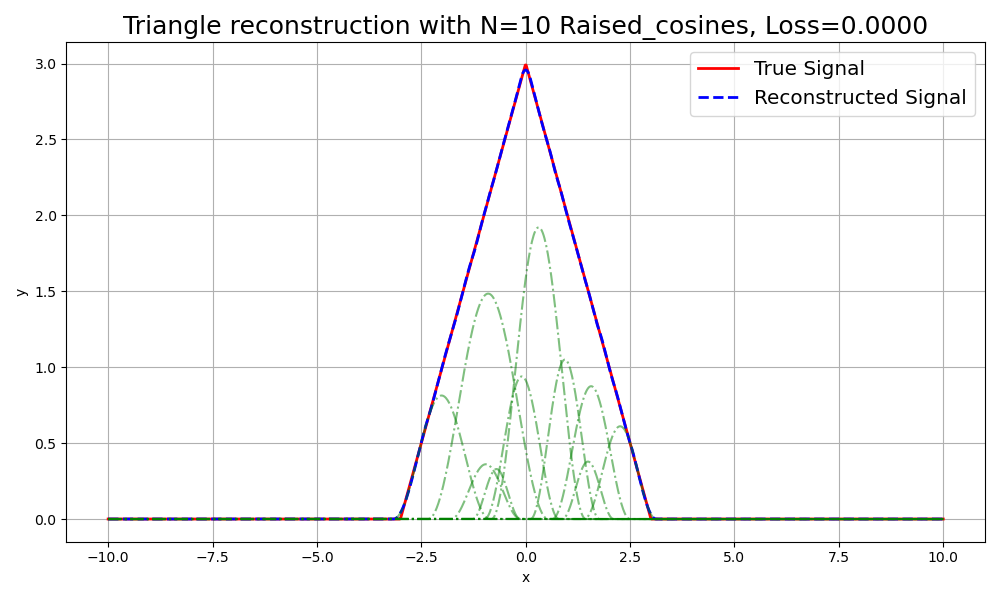} \\
    \includegraphics[width=0.33\textwidth]{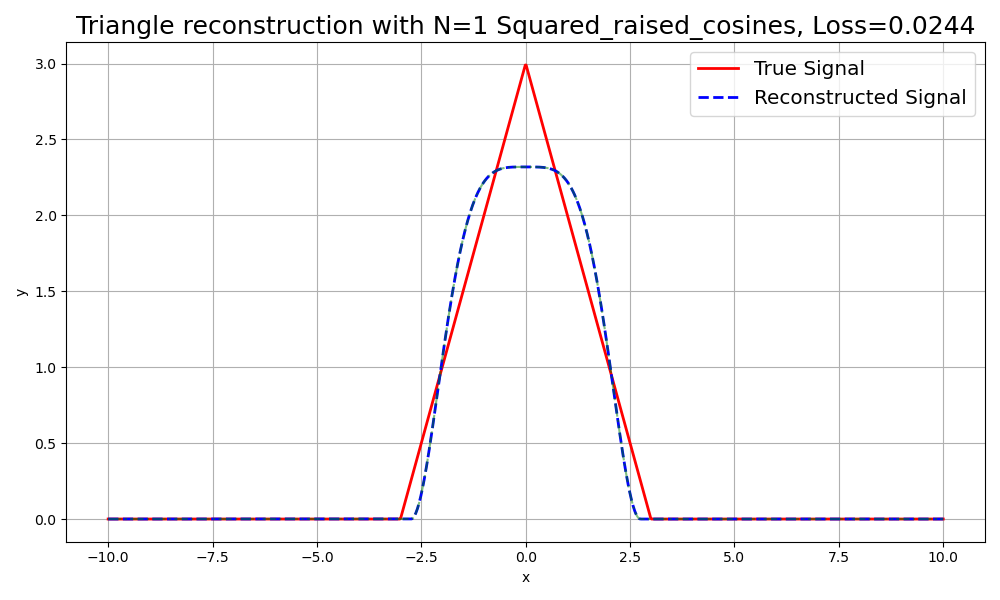} 
    \includegraphics[width=0.33\textwidth]{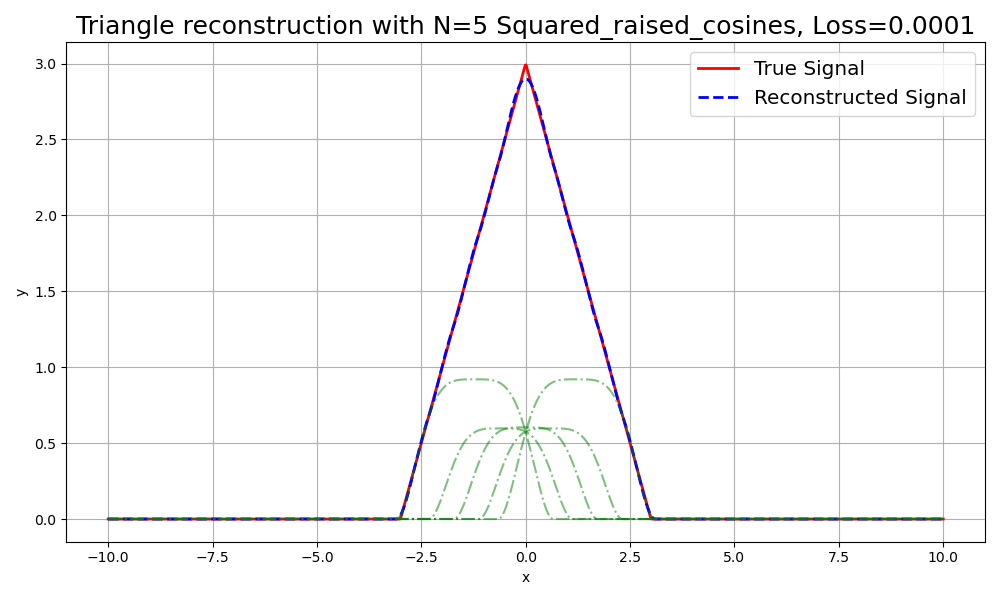} 
    \includegraphics[width=0.33\textwidth]{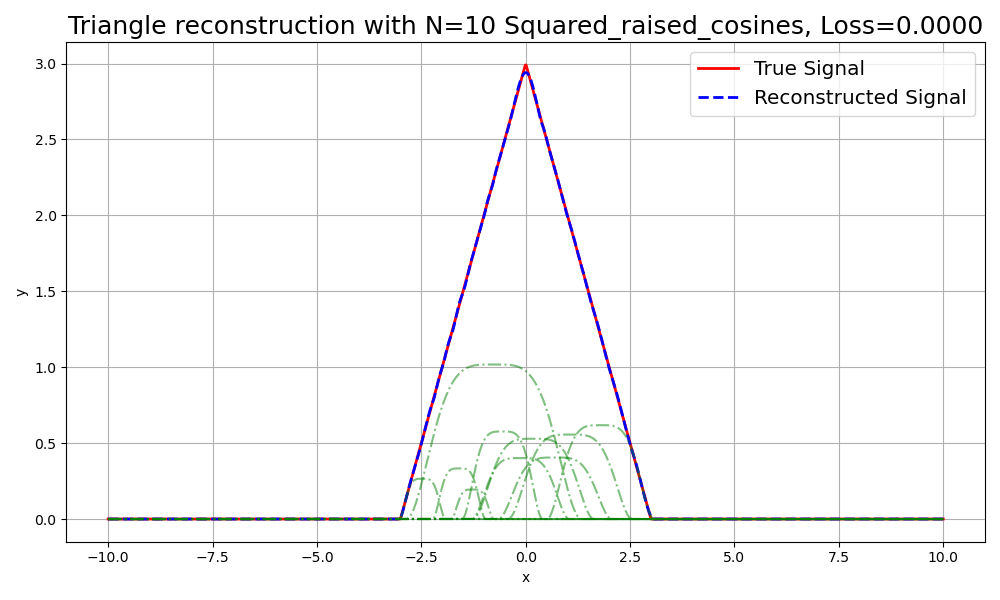} \\
    \includegraphics[width=0.33\textwidth]{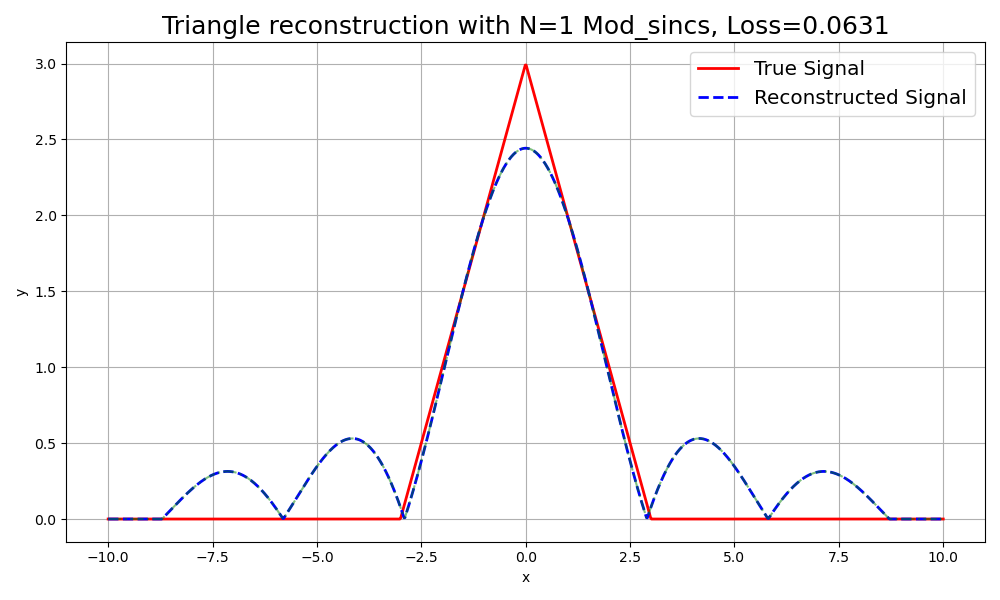} 
    \includegraphics[width=0.33\textwidth]{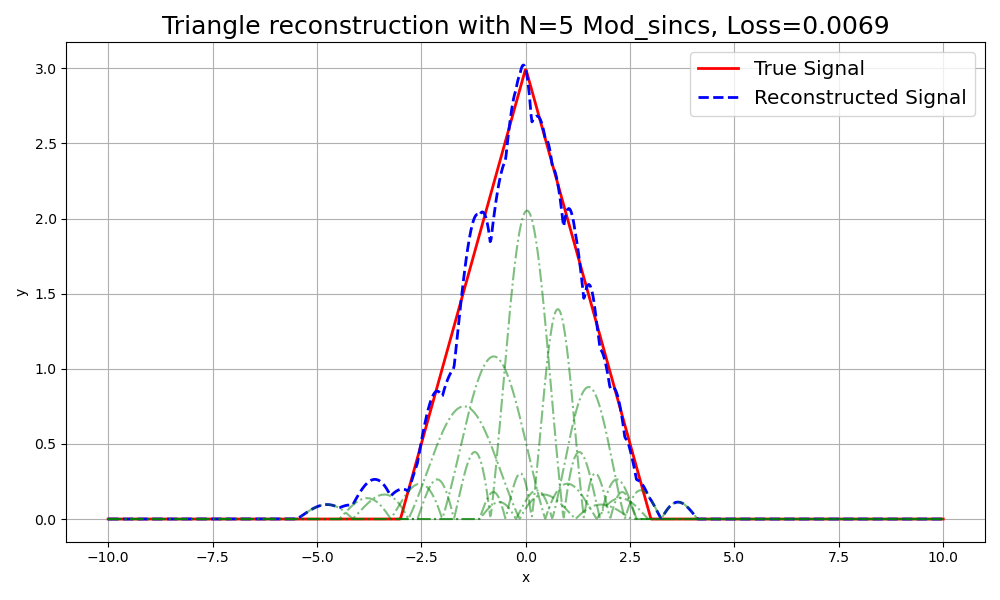} 
    \includegraphics[width=0.33\textwidth]{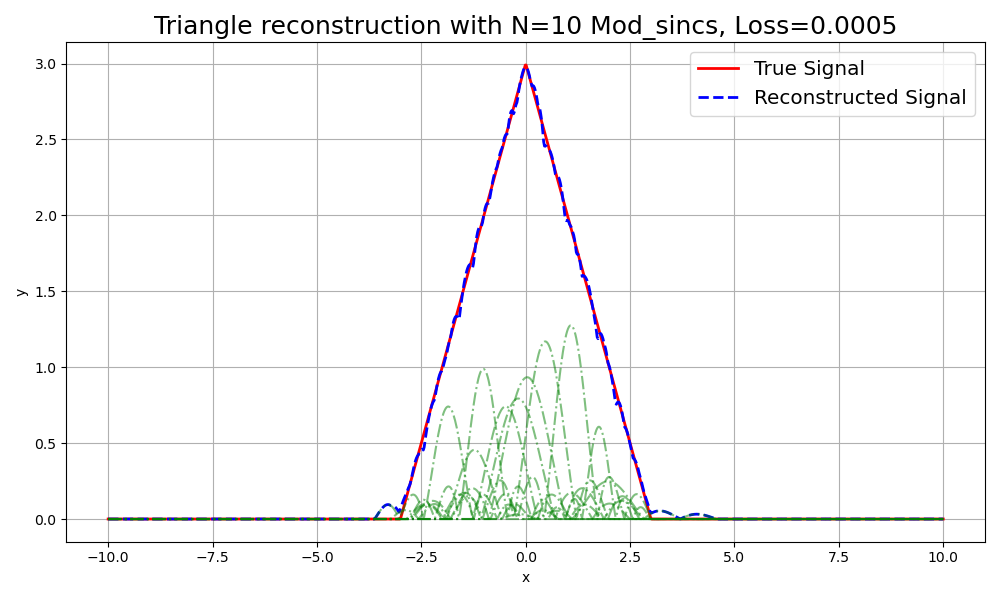} \\
    \caption{Visualization of 1D simulations for different splatting methods with varying primitives (N=1, N=5, N=10) for a triangle pulse. Each row corresponds to a specific splatting method: Gaussian, Cosine, Squared Cosine, Raised Cosine, Squared Raised Cosine, and Modulated Sinc. The columns represent the number of primitives used.}
    \label{fig:1DSim2}
\end{figure*}

\begin{figure*}
     % Column headings
    \parbox{0.33\textwidth}{\centering {N=1}} 
    \parbox{0.33\textwidth}{\centering {N=5}} 
    \parbox{0.33\textwidth}{\centering {N=10}} \\
    \includegraphics[width=0.33\textwidth]{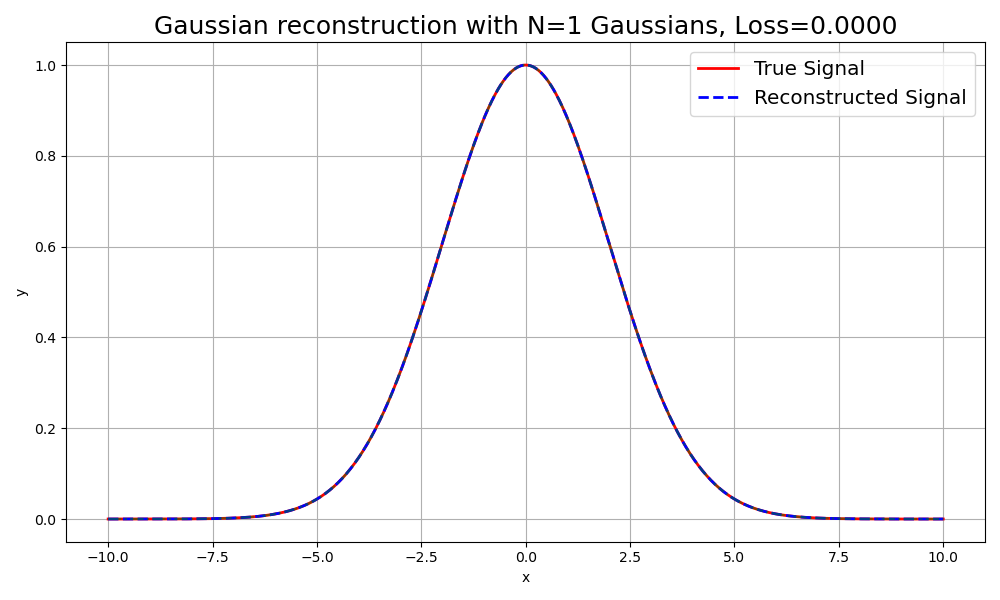} 
    \includegraphics[width=0.33\textwidth]{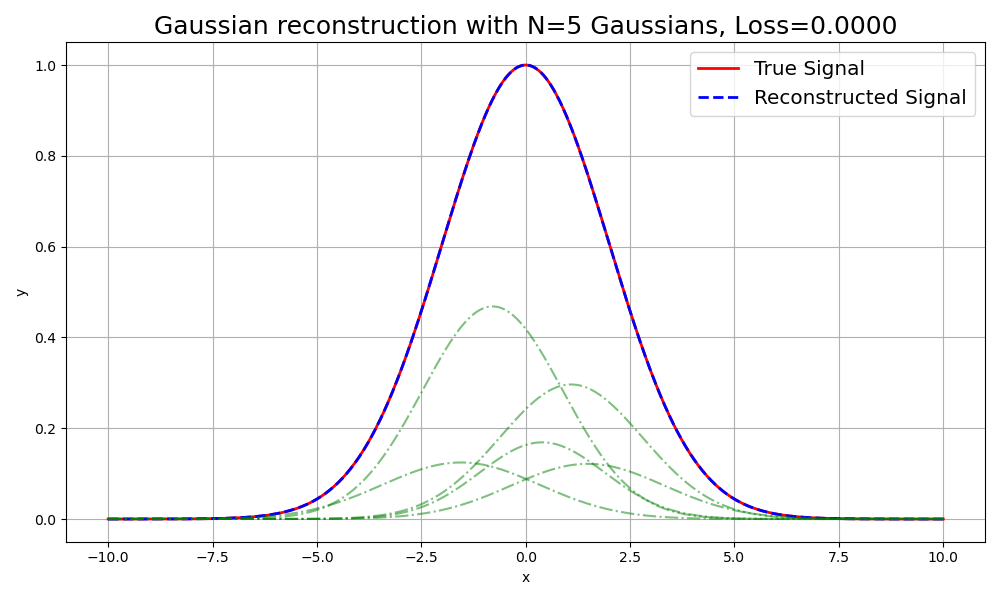} 
    \includegraphics[width=0.33\textwidth]{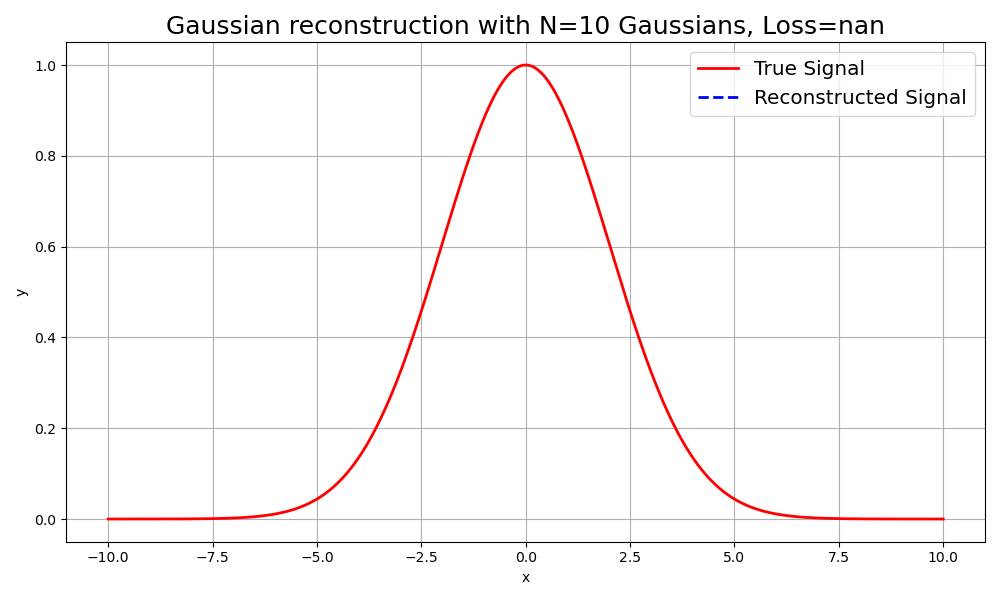} \\
    \includegraphics[width=0.33\textwidth]{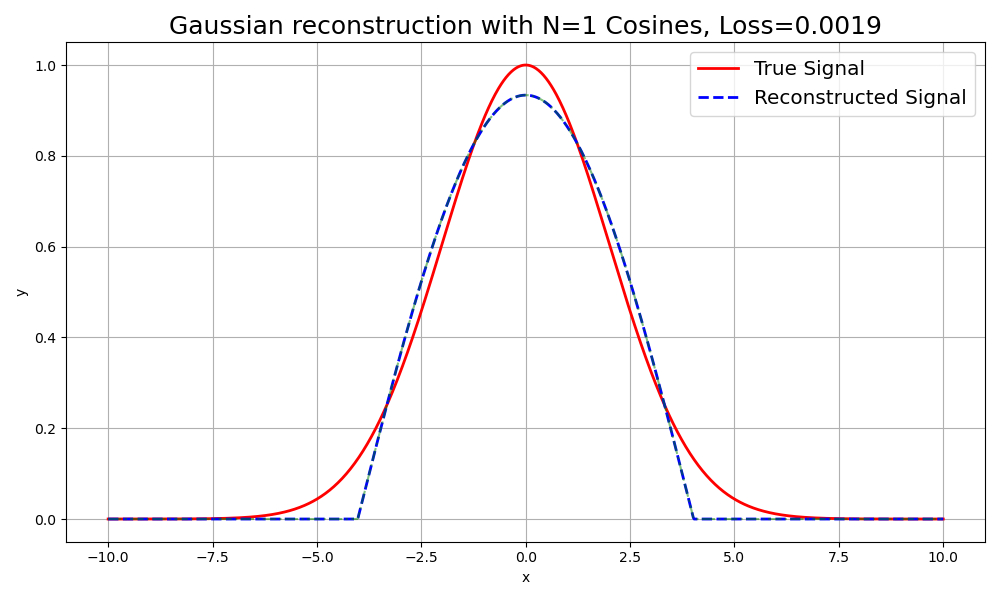} 
    \includegraphics[width=0.33\textwidth]{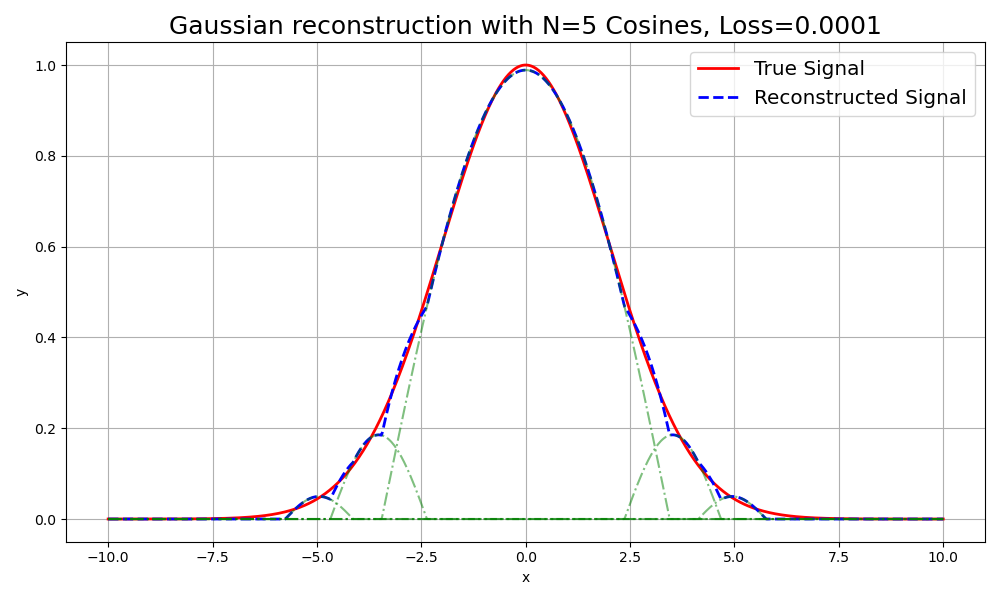} 
    \includegraphics[width=0.33\textwidth]{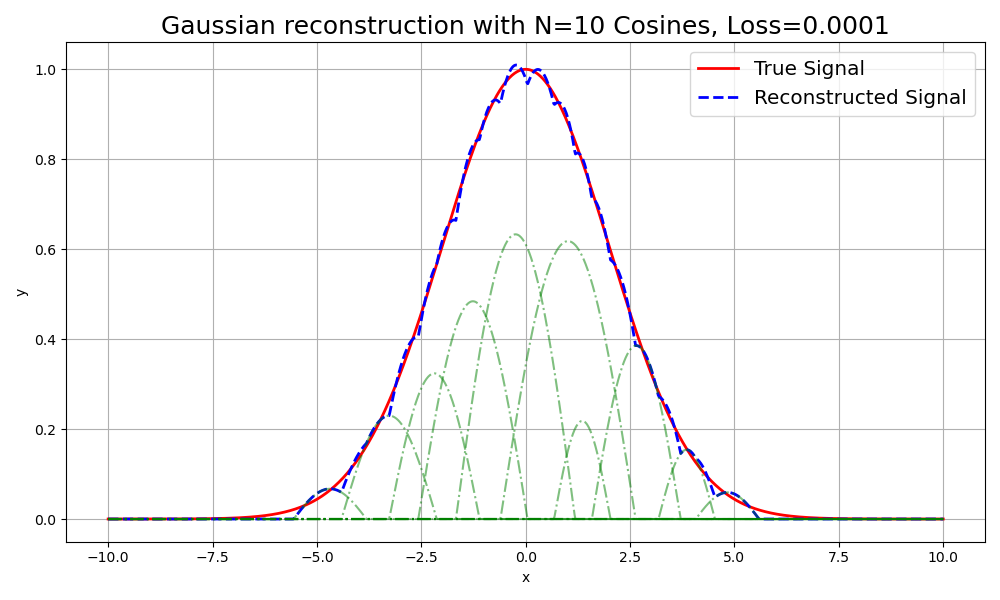} \\
    \includegraphics[width=0.33\textwidth]{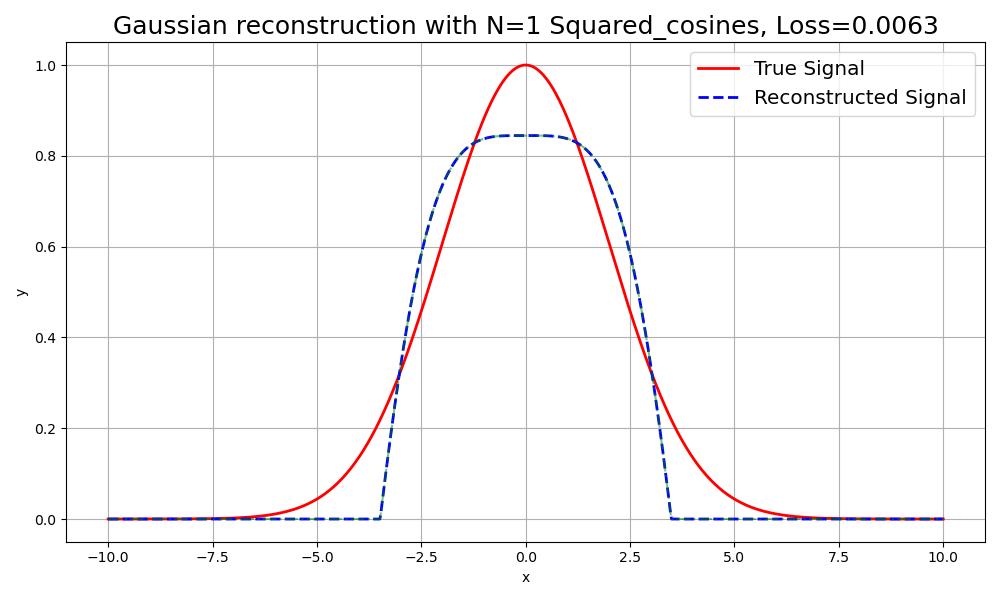} 
    \includegraphics[width=0.33\textwidth]{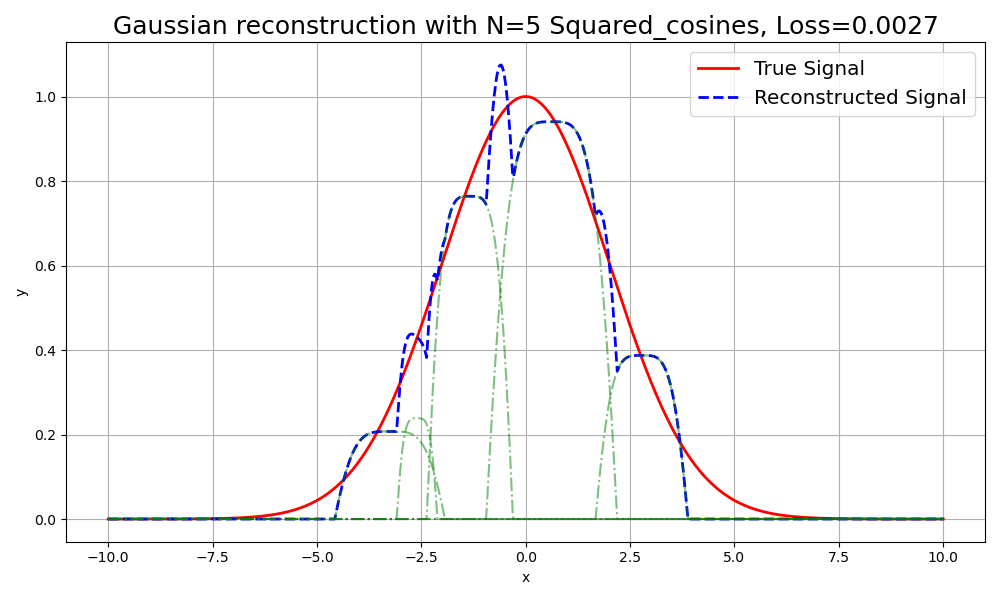} 
    \includegraphics[width=0.33\textwidth]{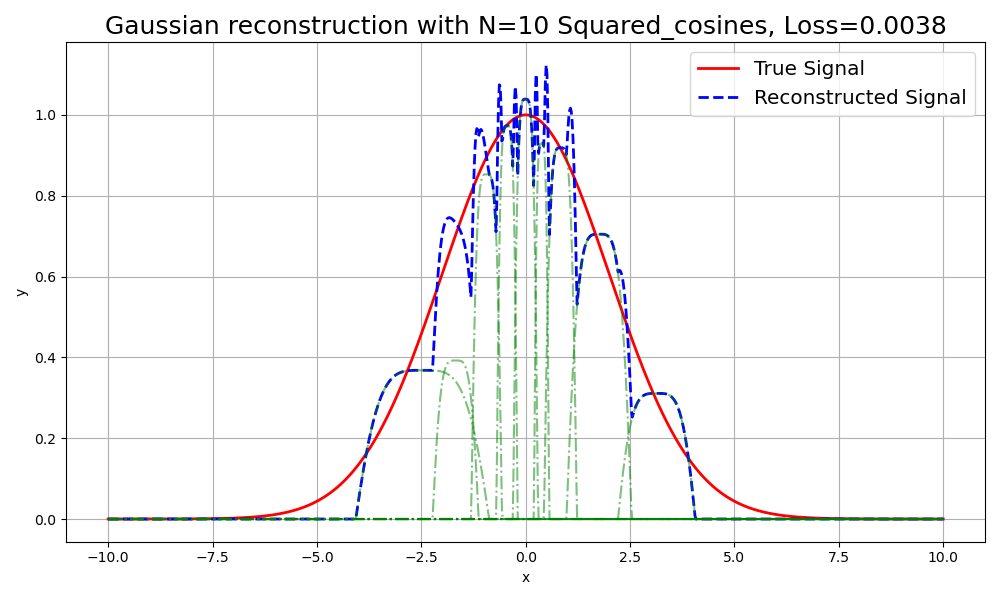} \\
    \includegraphics[width=0.33\textwidth]{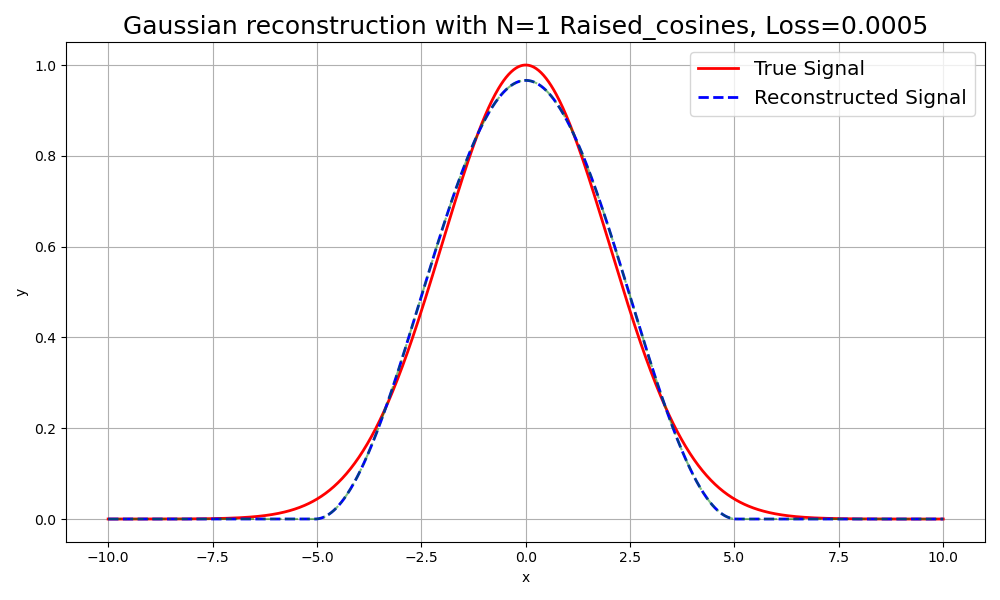} 
    \includegraphics[width=0.33\textwidth]{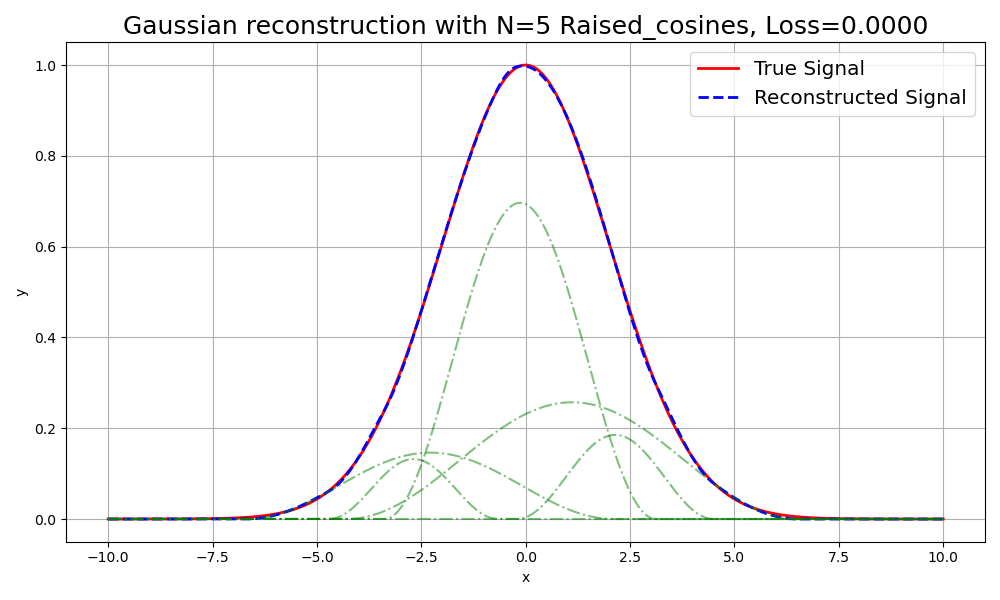} 
    \includegraphics[width=0.33\textwidth]{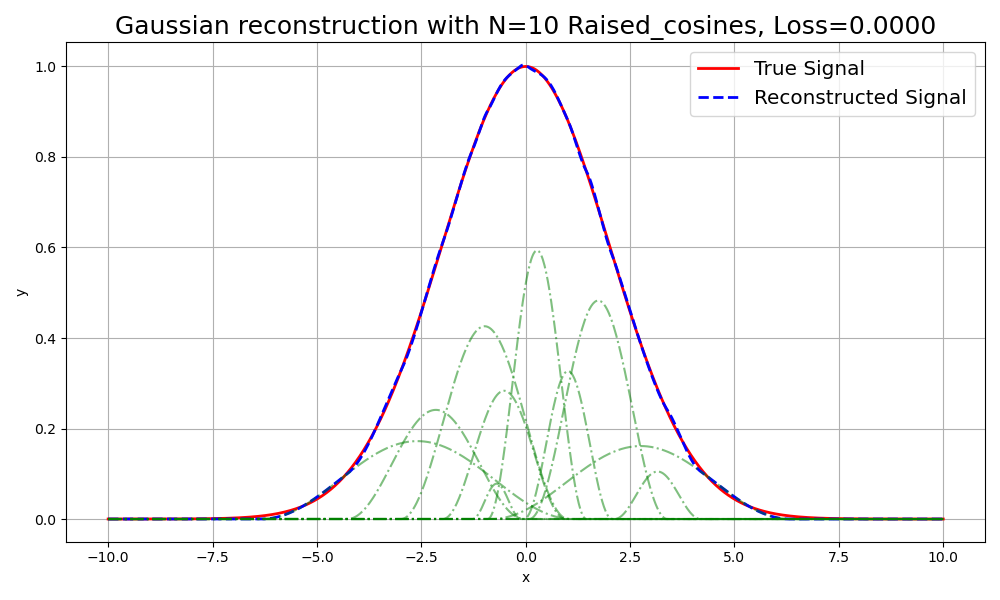} \\
    \includegraphics[width=0.33\textwidth]{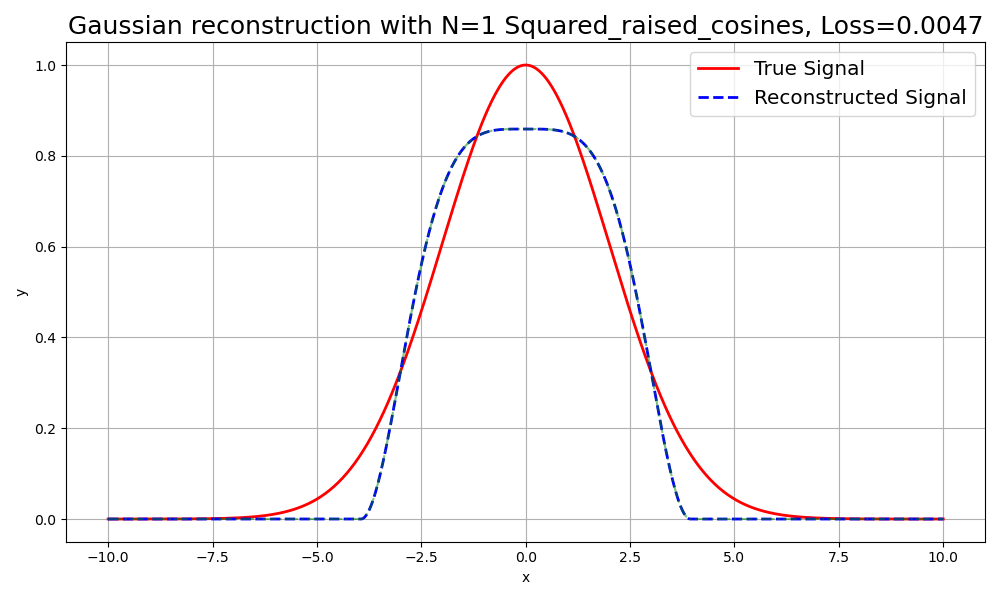} 
    \includegraphics[width=0.33\textwidth]{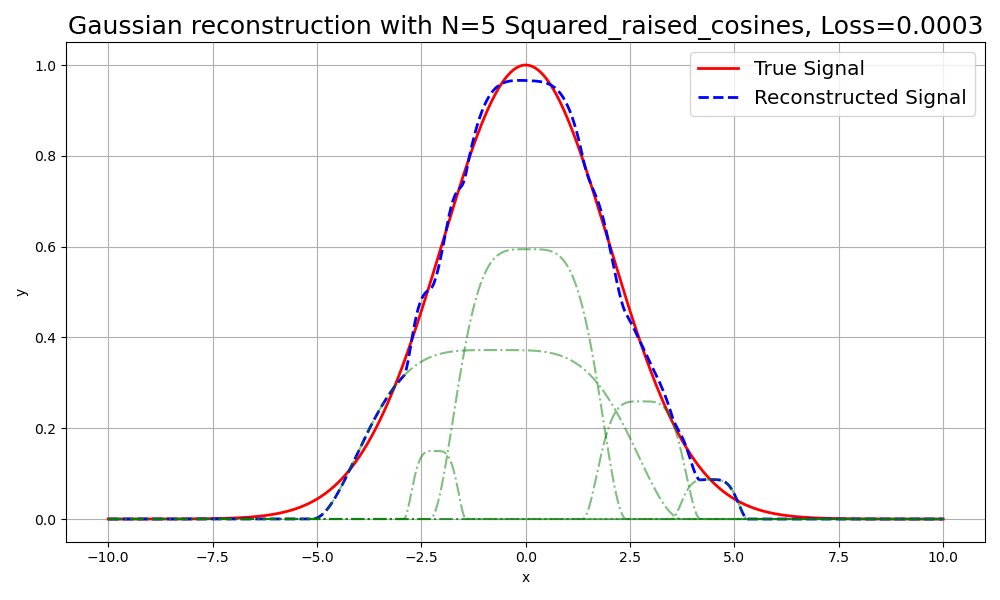} 
    \includegraphics[width=0.33\textwidth]{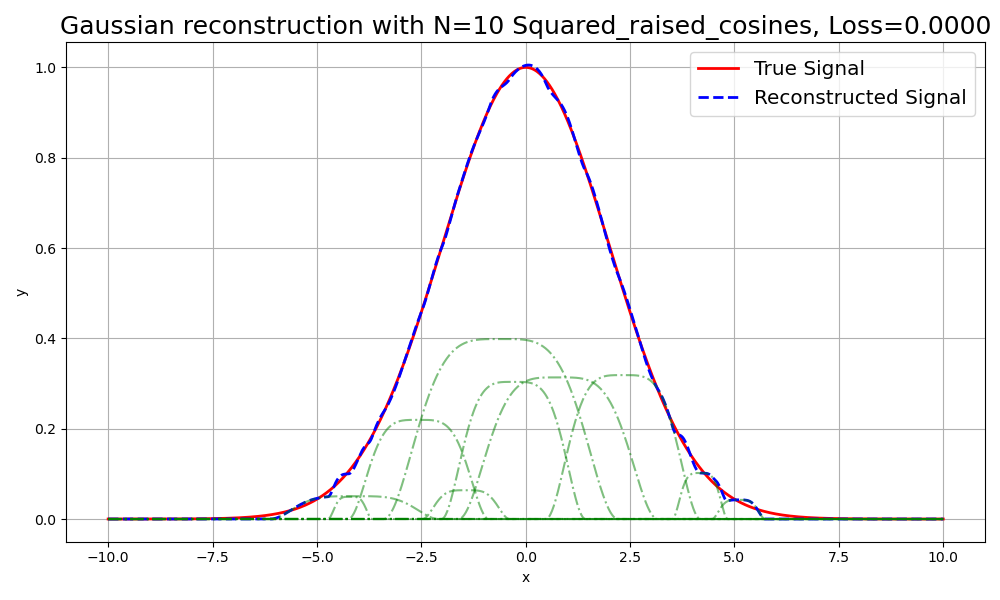} \\
    \includegraphics[width=0.33\textwidth]{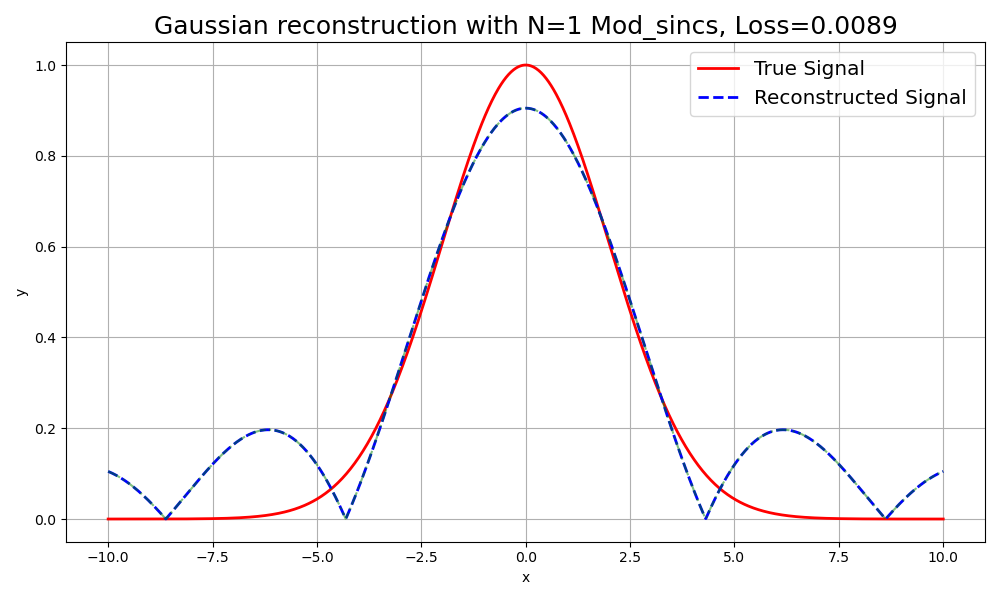} 
    \includegraphics[width=0.33\textwidth]{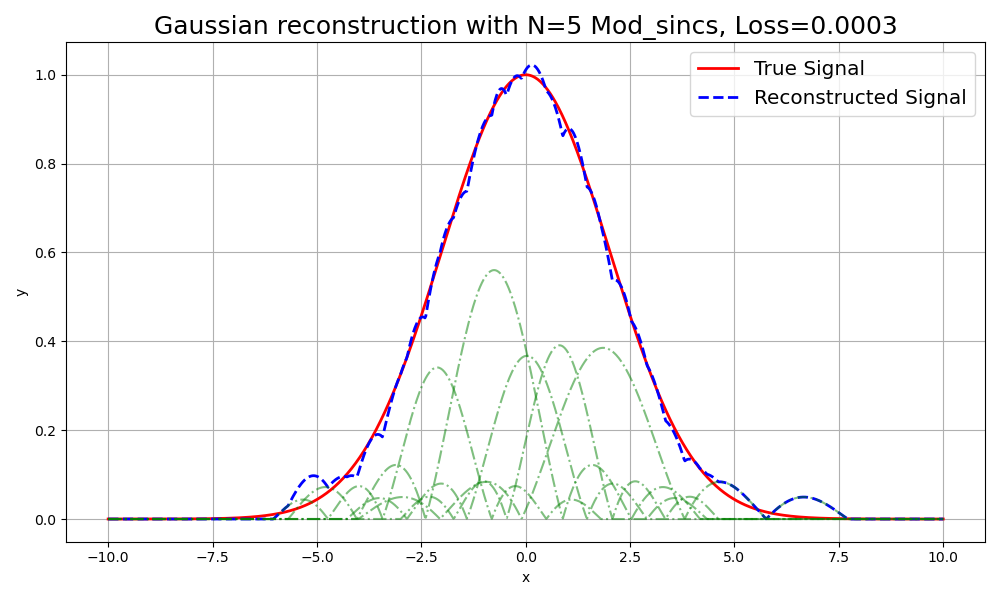} 
    \includegraphics[width=0.33\textwidth]{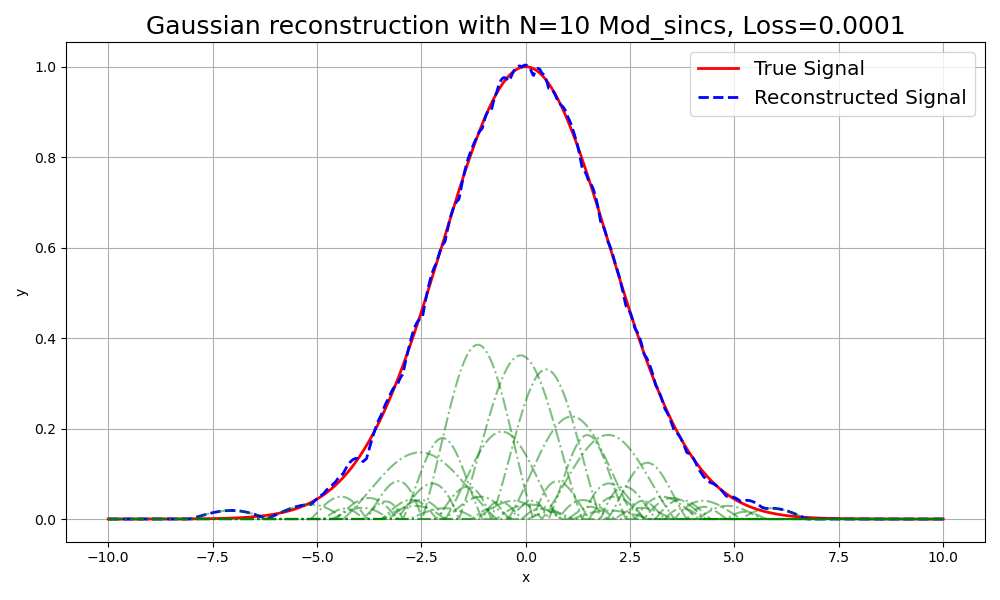} \\
    \caption{Visualization of 1D simulations for different splatting methods with varying primitives (N=1, N=5, N=10) for a Gaussian. Each row corresponds to a specific splatting method: Gaussian, Cosine, Squared Cosine, Raised Cosine, Squared Raised Cosine, and Modulated Sinc. The columns represent the number of primitives used.}
    \label{fig:1DSim3}
\end{figure*}

\begin{figure*}
     % Column headings
    \parbox{0.33\textwidth}{\centering {N=1}} 
    \parbox{0.33\textwidth}{\centering {N=5}} 
    \parbox{0.33\textwidth}{\centering {N=10}} \\
    \includegraphics[width=0.33\textwidth]{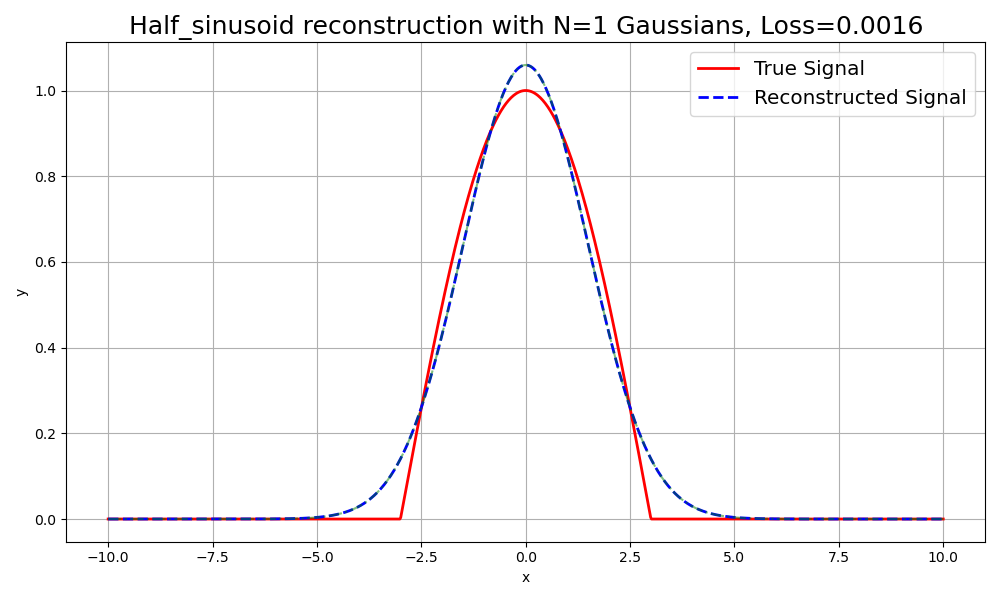} 
    \includegraphics[width=0.33\textwidth]{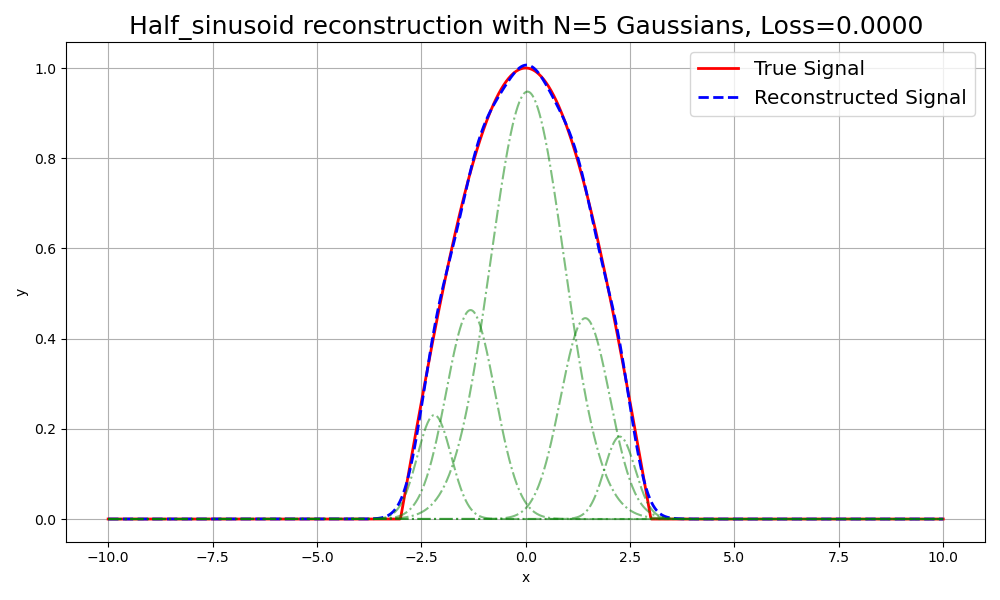} 
    \includegraphics[width=0.33\textwidth]{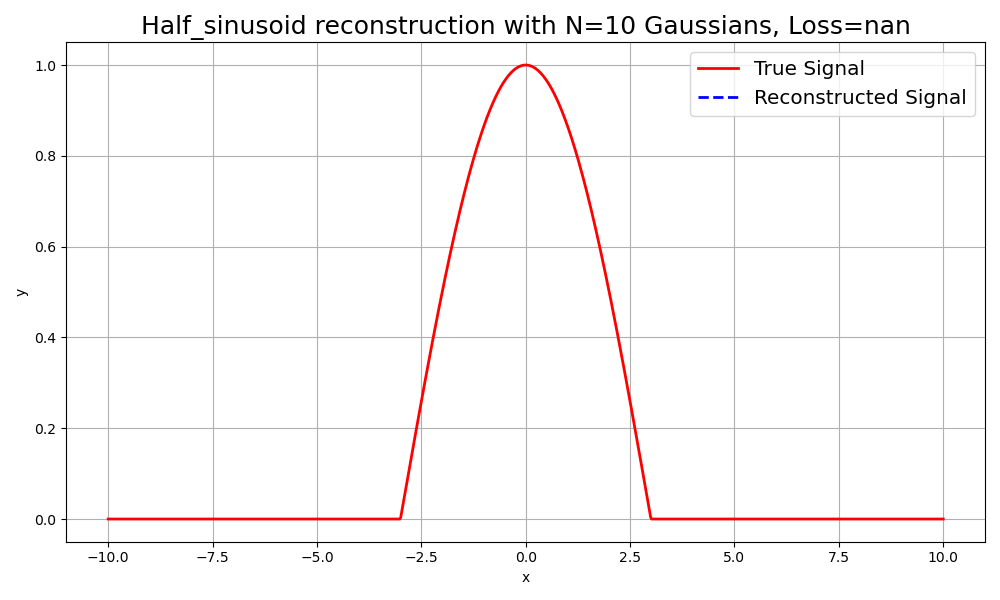} \\
    \includegraphics[width=0.33\textwidth]{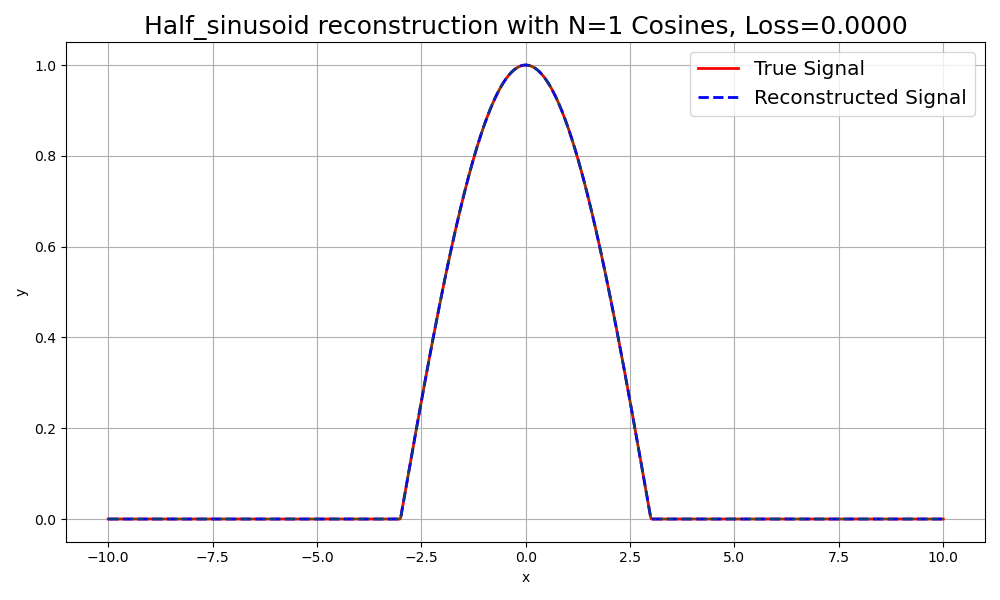} 
    \includegraphics[width=0.33\textwidth]{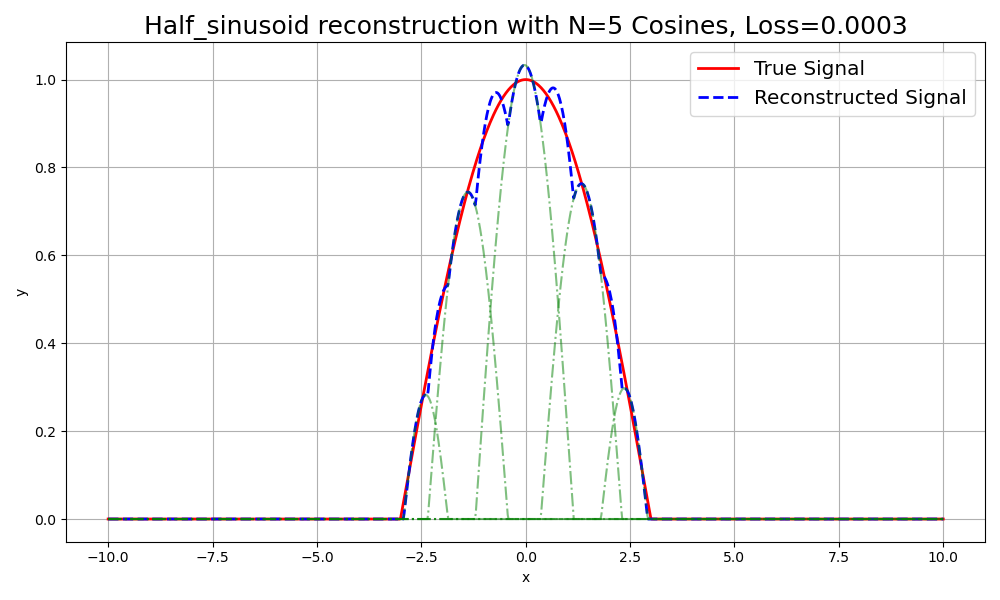} 
    \includegraphics[width=0.33\textwidth]{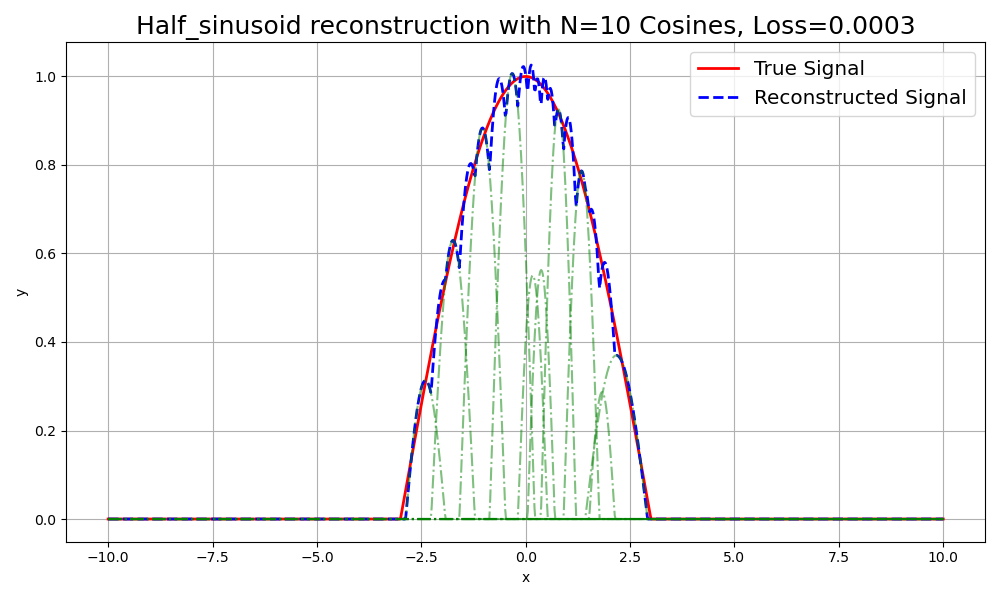} \\
    \includegraphics[width=0.33\textwidth]{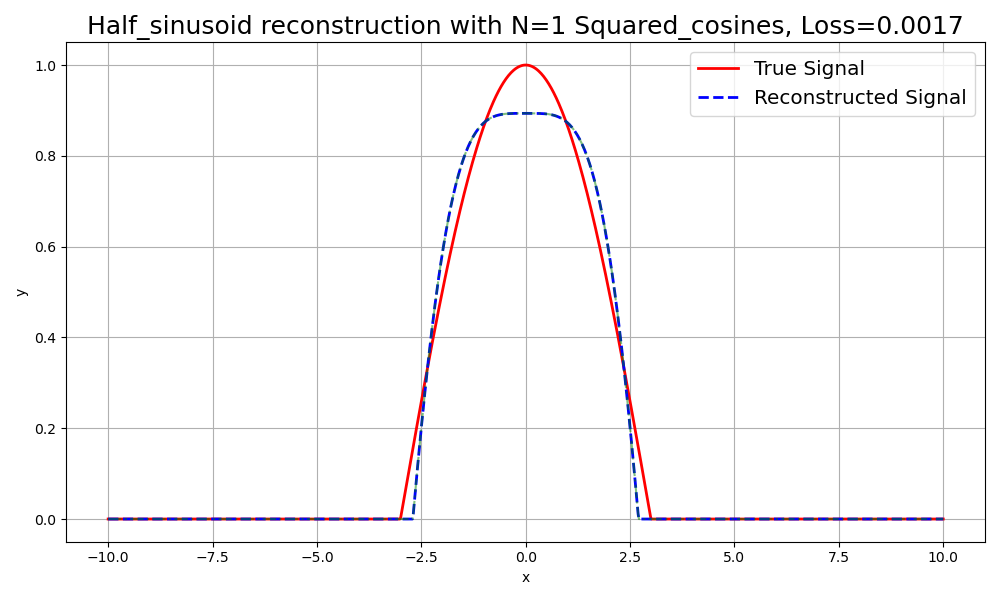} 
    \includegraphics[width=0.33\textwidth]{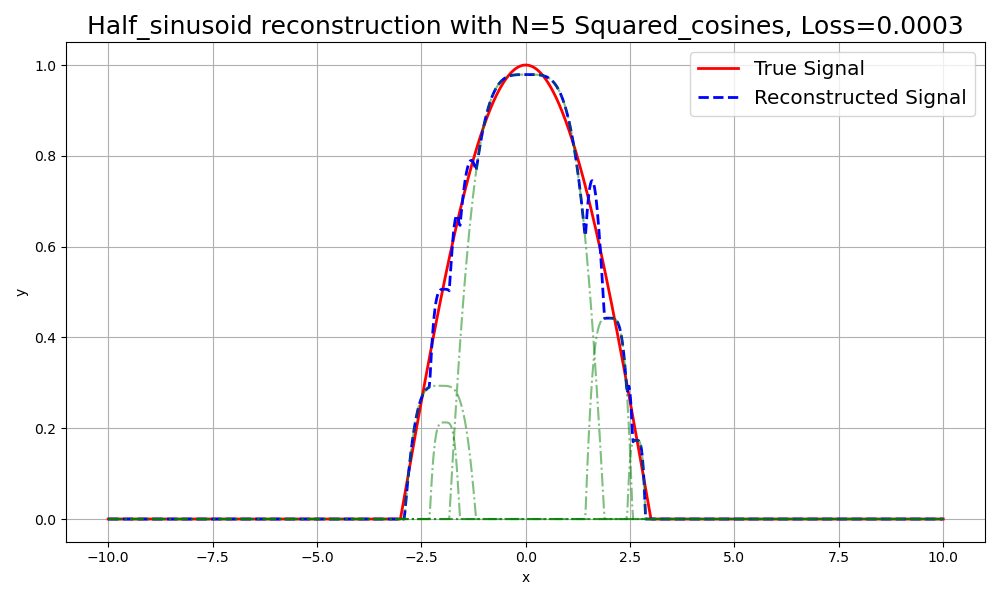} 
    \includegraphics[width=0.33\textwidth]{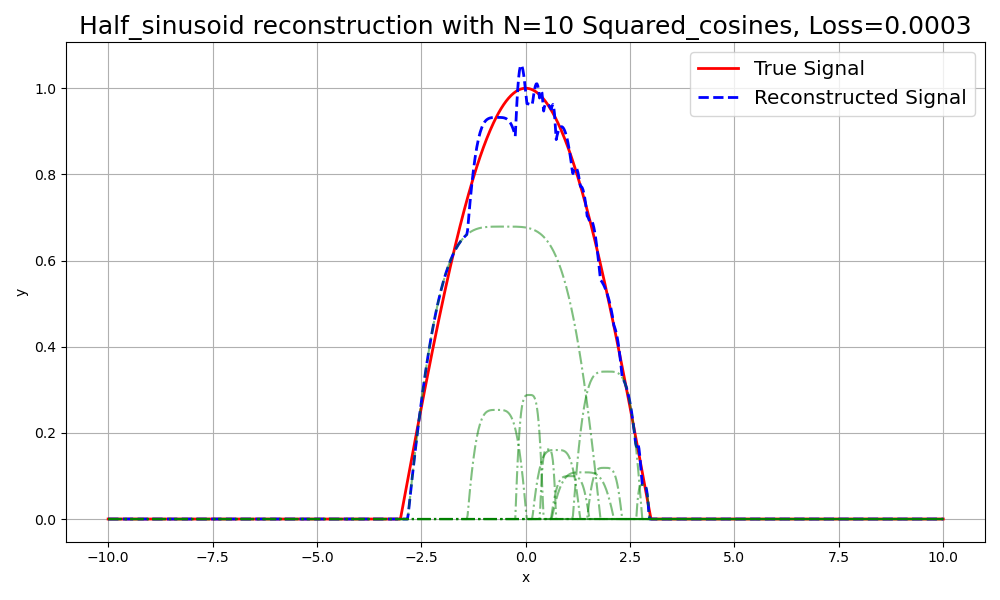} \\
    \includegraphics[width=0.33\textwidth]{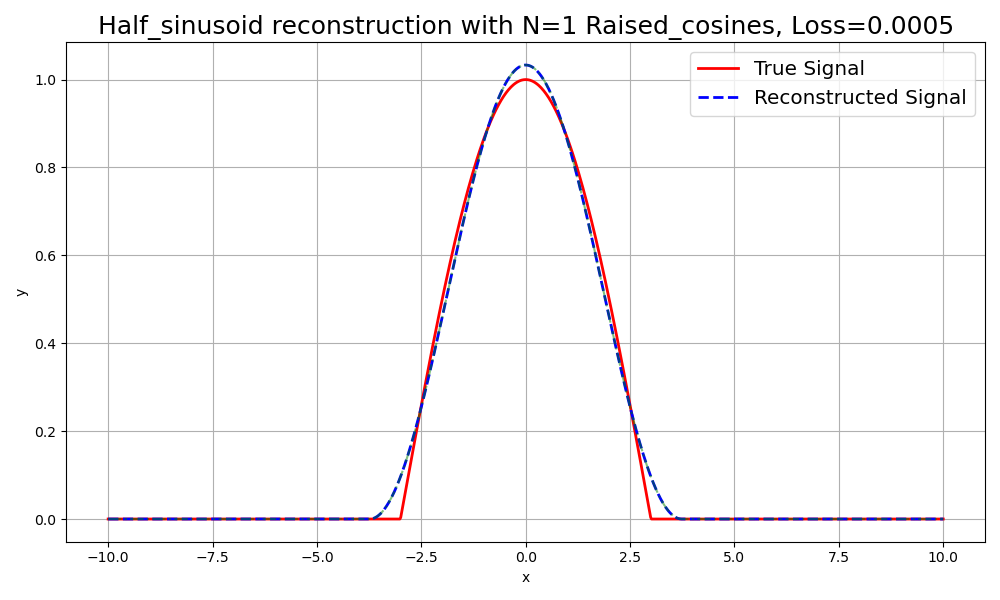} 
    \includegraphics[width=0.33\textwidth]{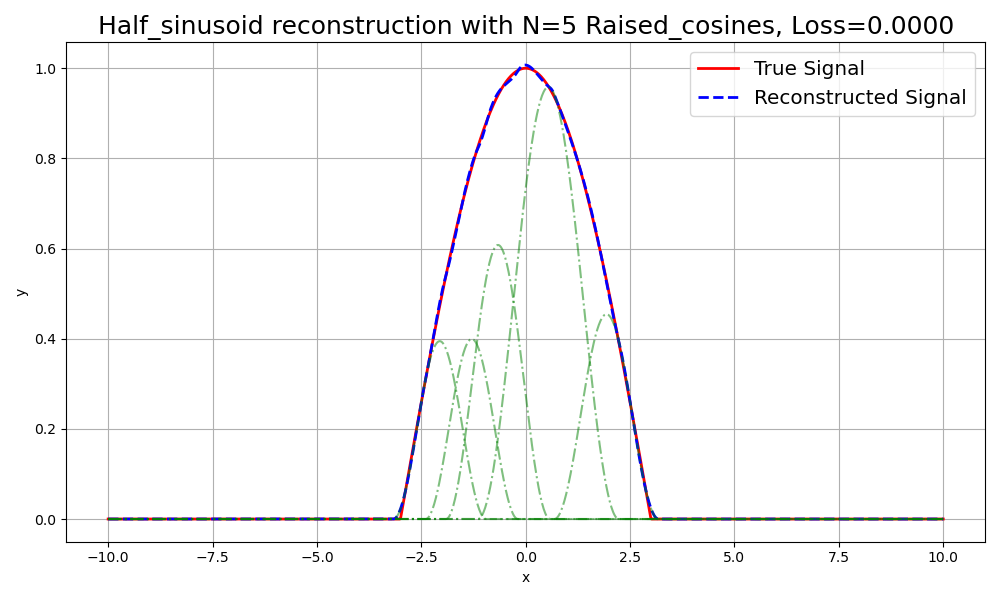} 
    \includegraphics[width=0.33\textwidth]{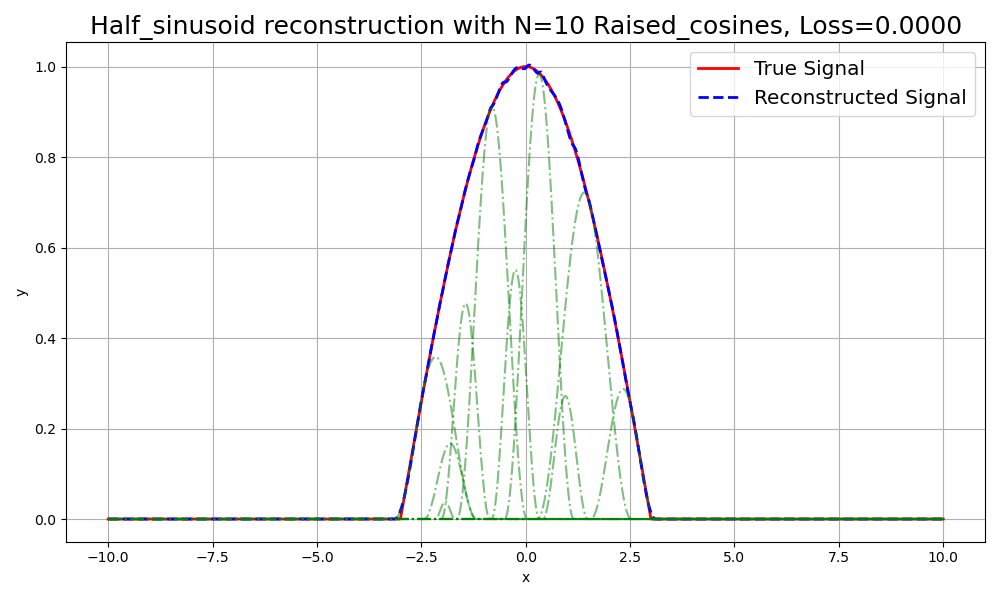} \\
    \includegraphics[width=0.33\textwidth]{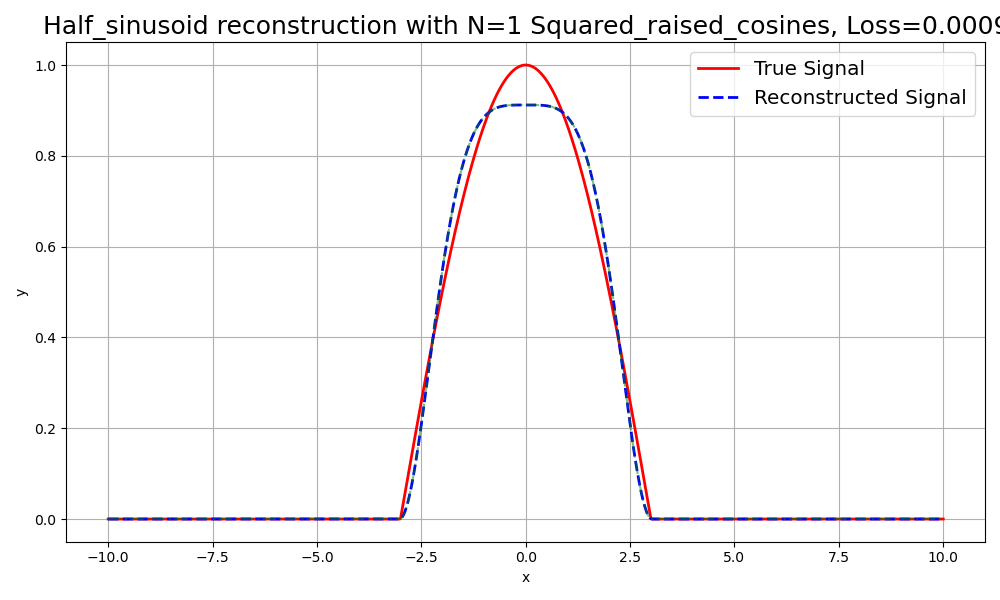} 
    \includegraphics[width=0.33\textwidth]{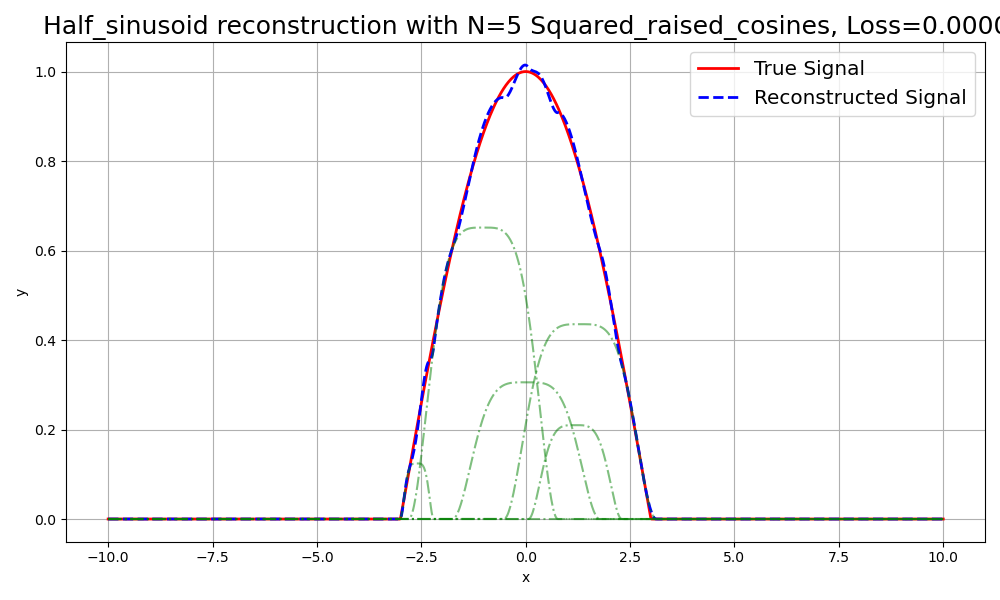} 
    \includegraphics[width=0.33\textwidth]{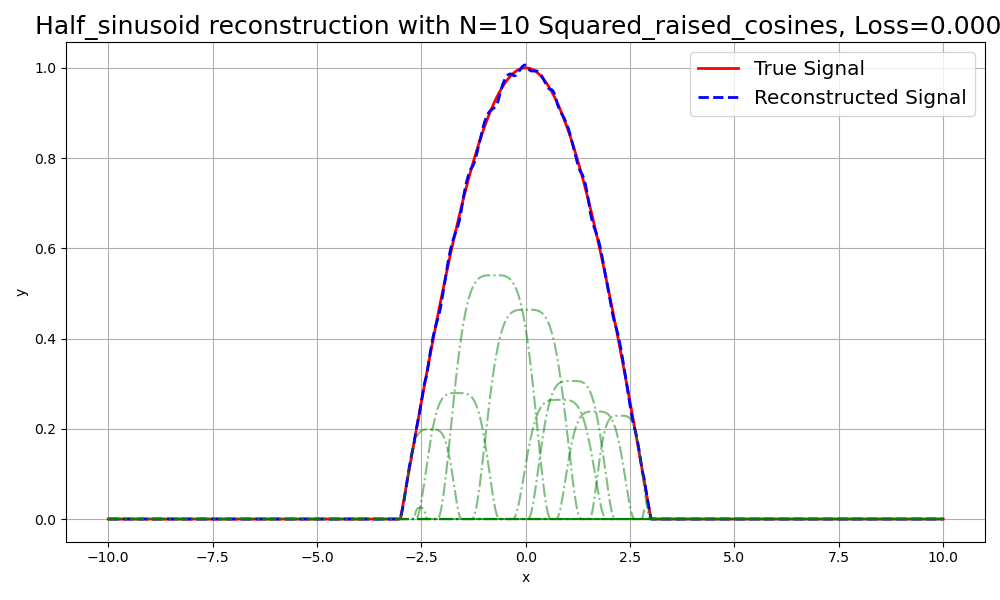} \\
    \includegraphics[width=0.33\textwidth]{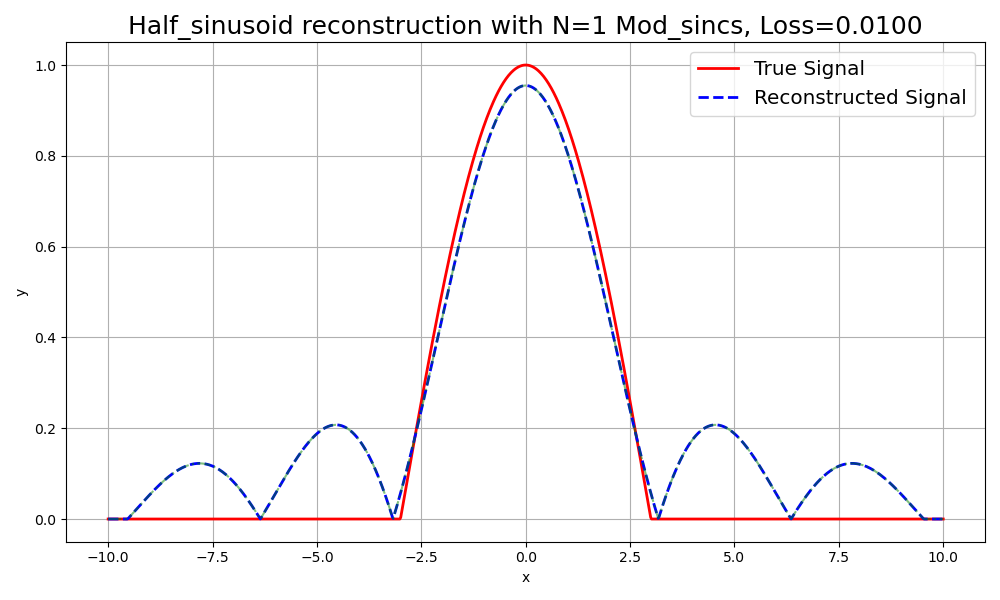} 
    \includegraphics[width=0.33\textwidth]{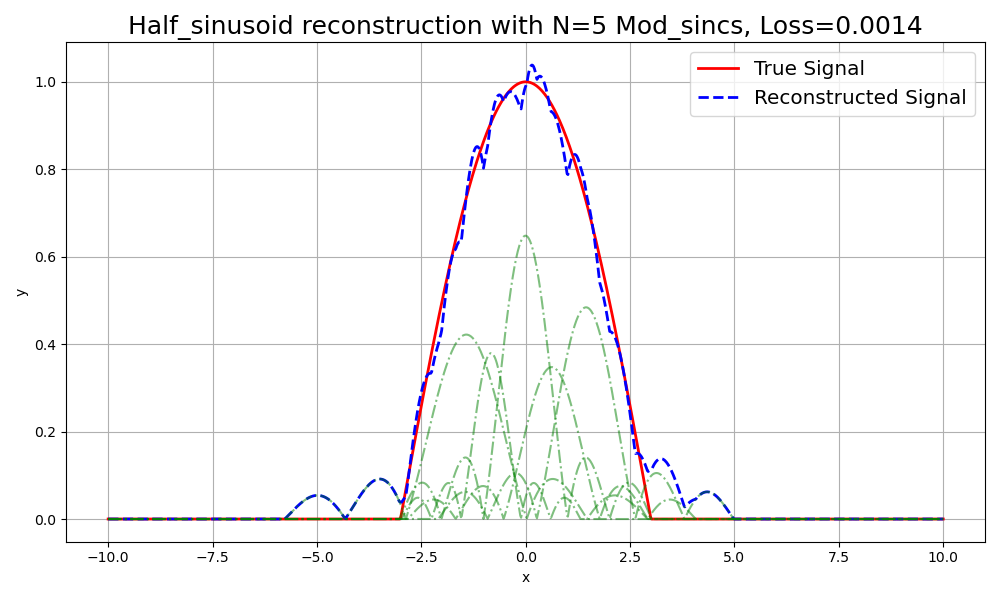} 
    \includegraphics[width=0.33\textwidth]{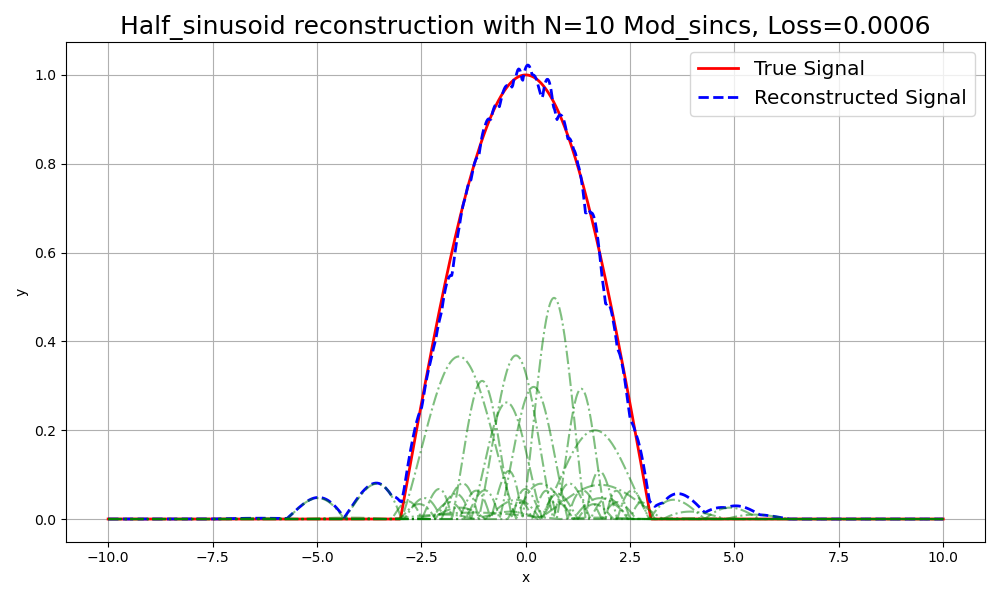} \\    \caption{Visualization of 1D simulations for different splatting methods with varying primitives (N=1, N=5, N=10) for a half-sinusoid single pulse. Each row corresponds to a specific splatting method: Gaussian, Cosine, Squared Cosine, Raised Cosine, Squared Raised Cosine, and Modulated Sinc. The columns represent the number of primitives used.}
    \label{fig:1DSim4}
\end{figure*}

\begin{figure*}
     % Column headings
    \parbox{0.33\textwidth}{\centering {N=1}} 
    \parbox{0.33\textwidth}{\centering {N=5}} 
    \parbox{0.33\textwidth}{\centering {N=10}} \\
    \includegraphics[width=0.33\textwidth]{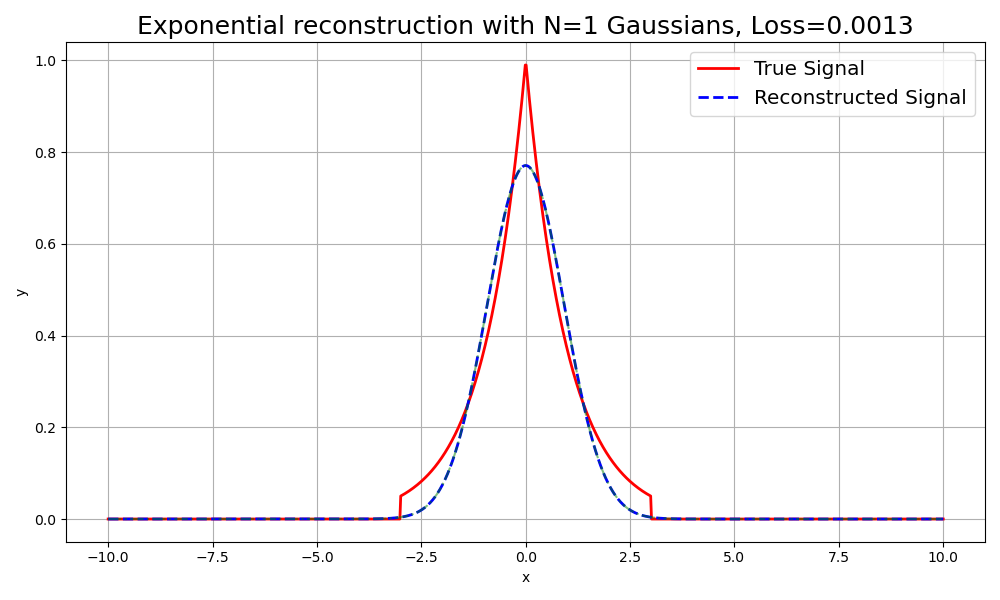} 
    \includegraphics[width=0.33\textwidth]{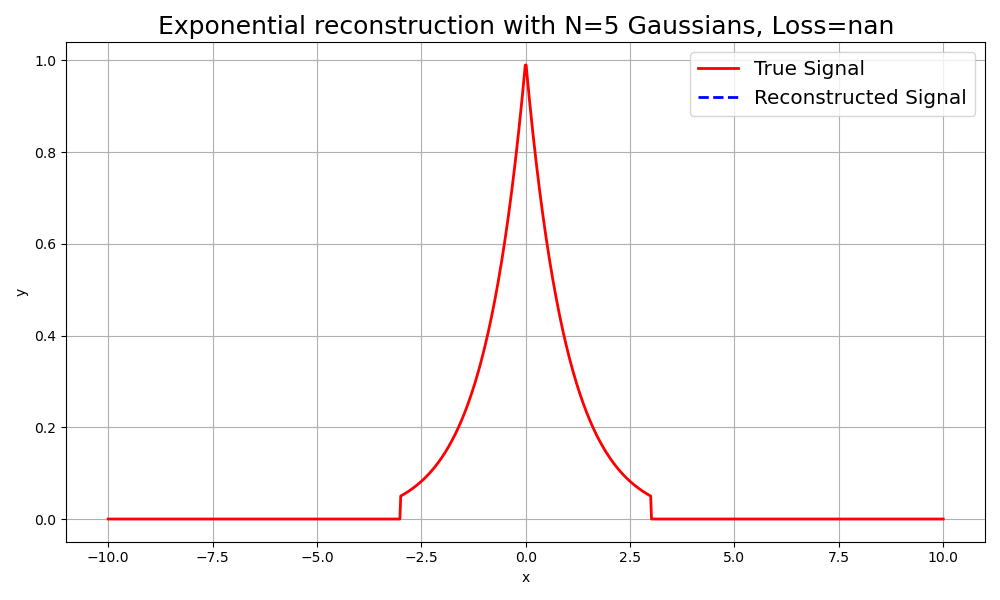} 
    \includegraphics[width=0.33\textwidth]{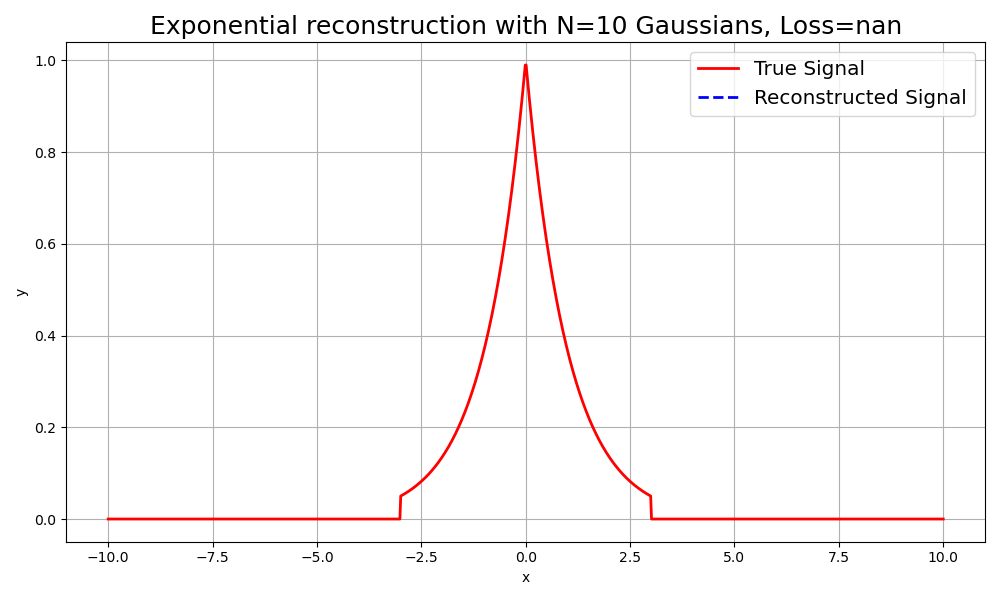} \\
    \includegraphics[width=0.33\textwidth]{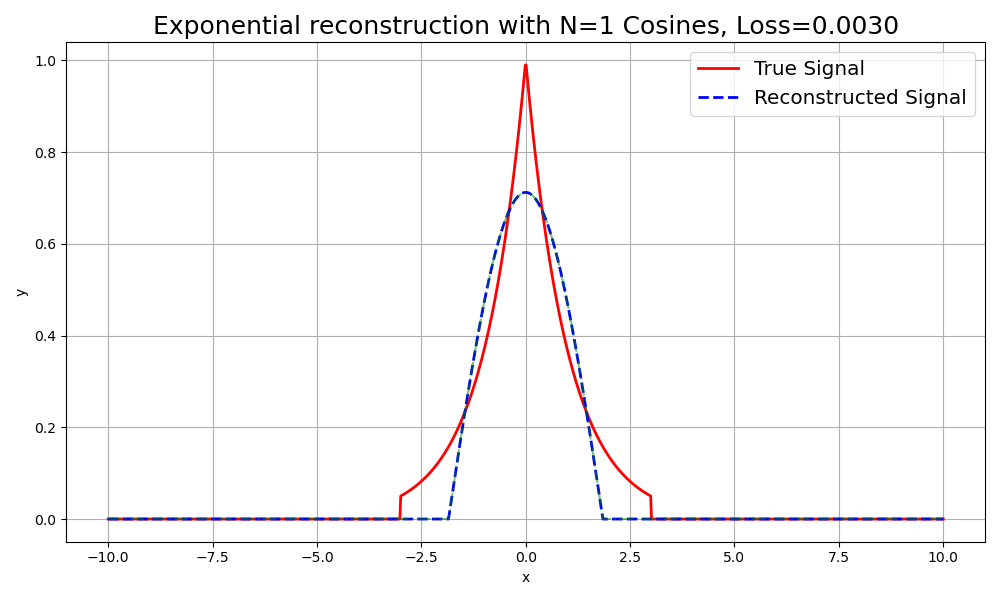} 
    \includegraphics[width=0.33\textwidth]{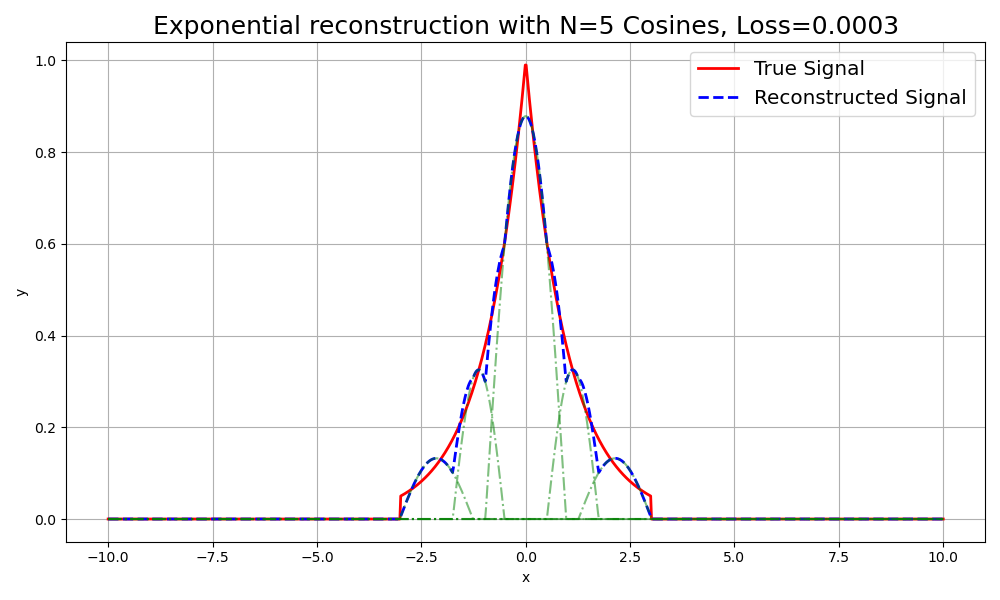} 
    \includegraphics[width=0.33\textwidth]{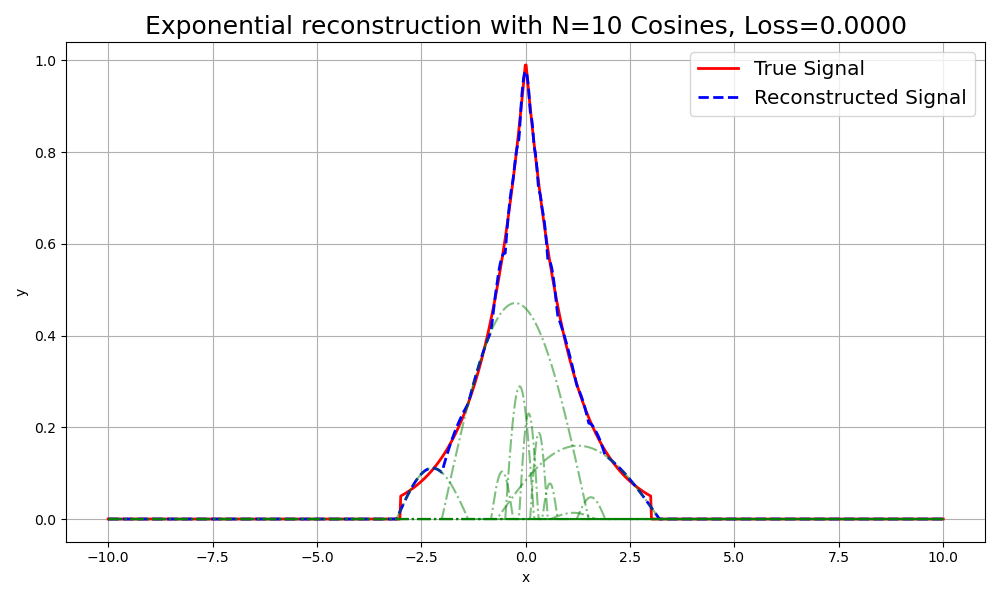} \\
    \includegraphics[width=0.33\textwidth]{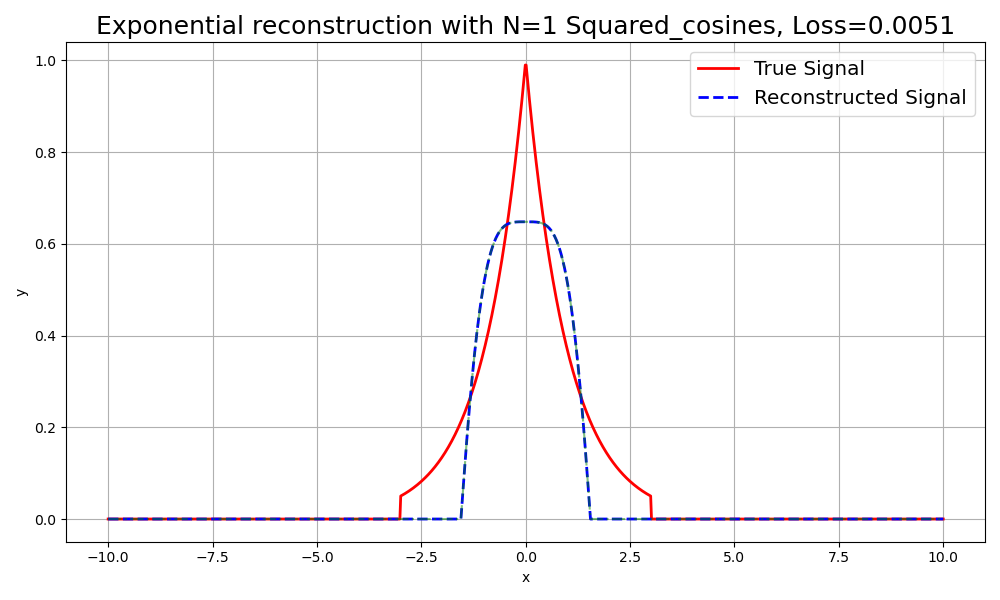} 
    \includegraphics[width=0.33\textwidth]{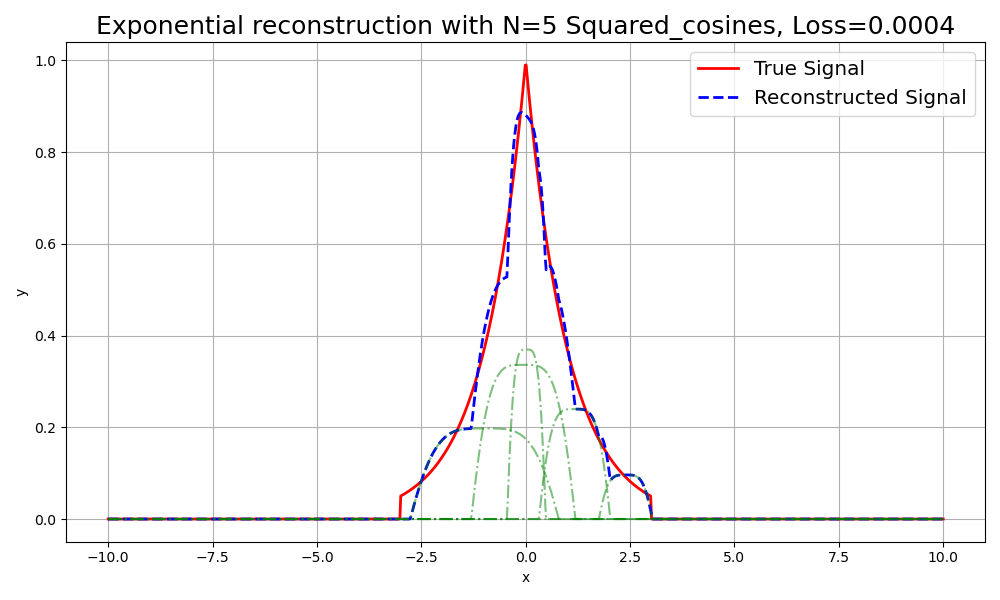} 
    \includegraphics[width=0.33\textwidth]{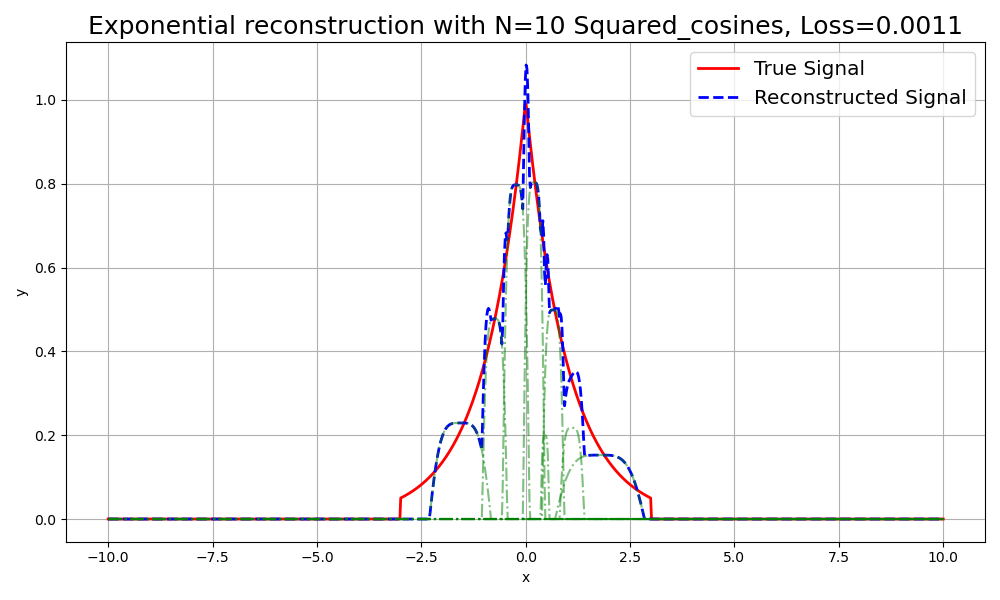} \\
    \includegraphics[width=0.33\textwidth]{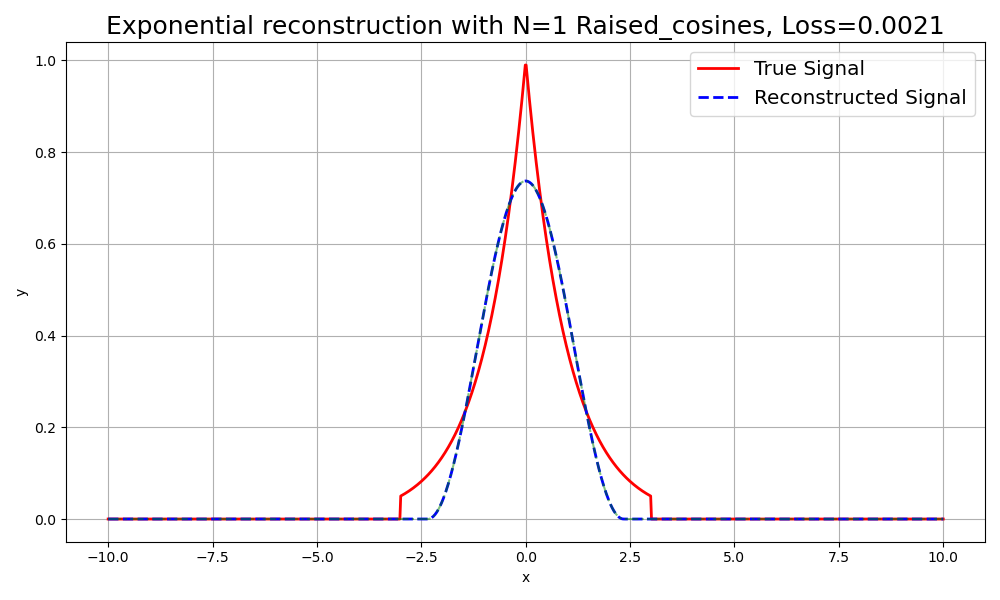} 
    \includegraphics[width=0.33\textwidth]{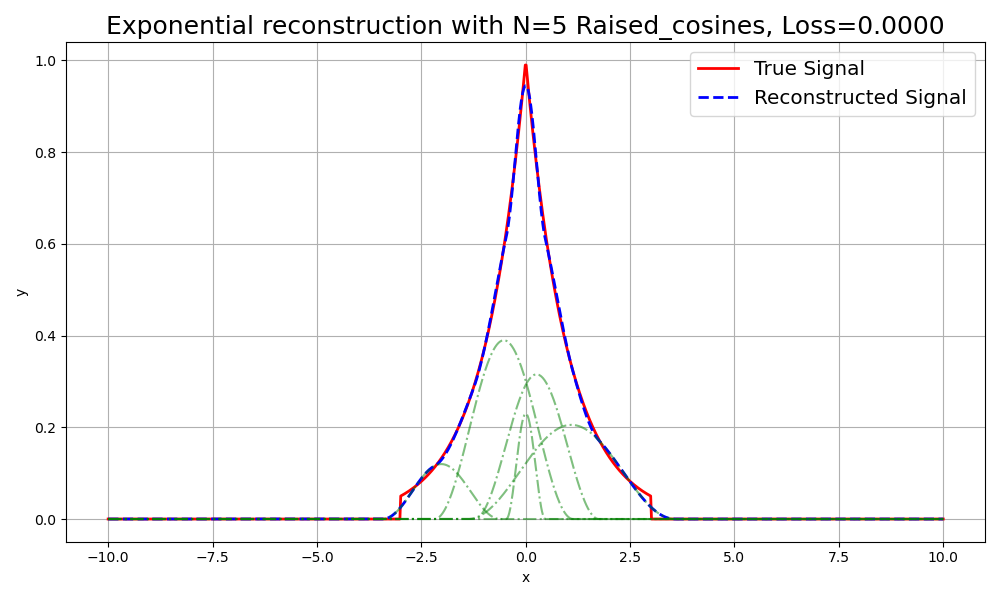} 
    \includegraphics[width=0.33\textwidth]{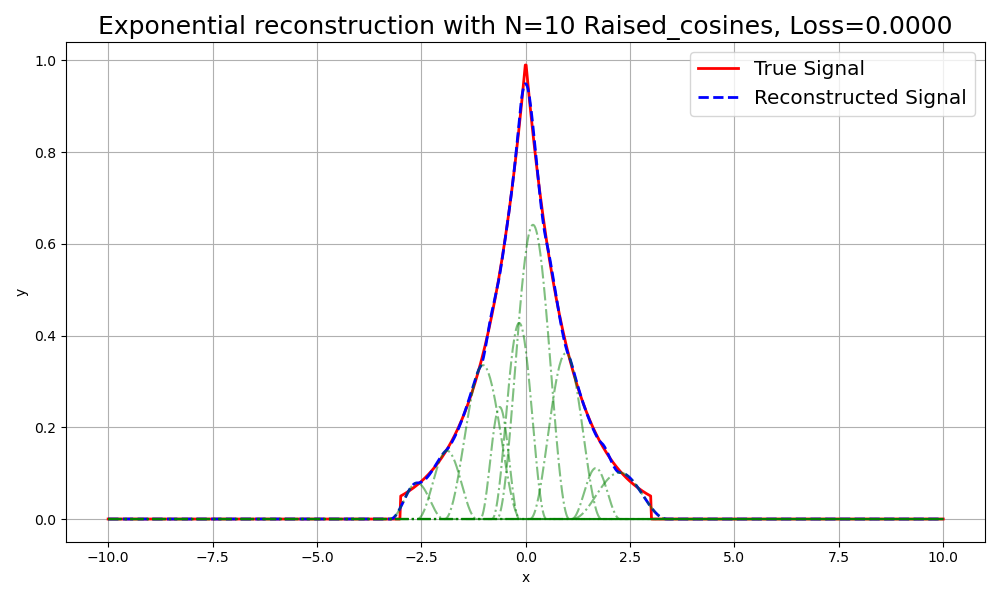} \\
    \includegraphics[width=0.33\textwidth]{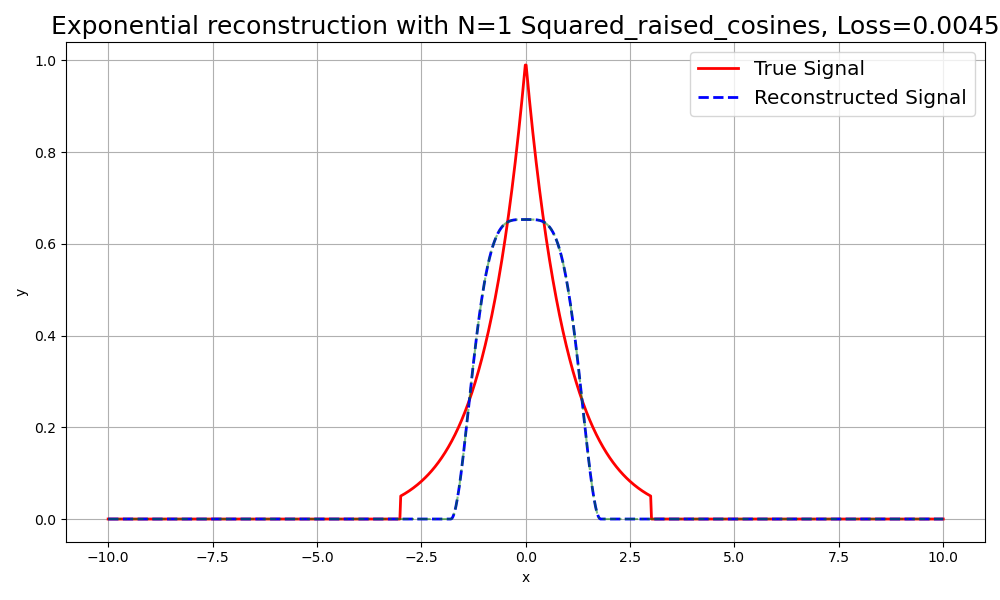} 
    \includegraphics[width=0.33\textwidth]{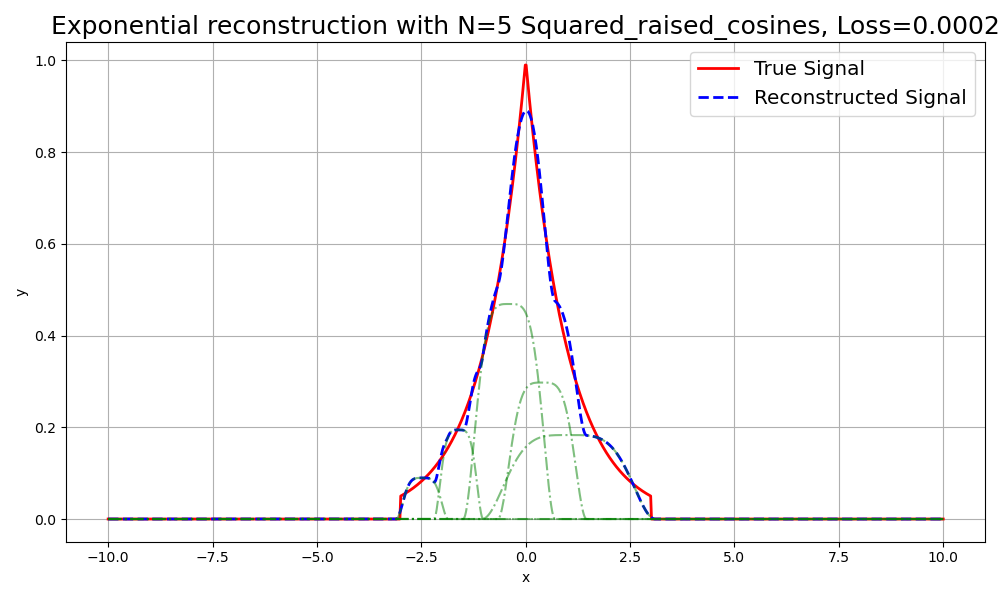} 
    \includegraphics[width=0.33\textwidth]{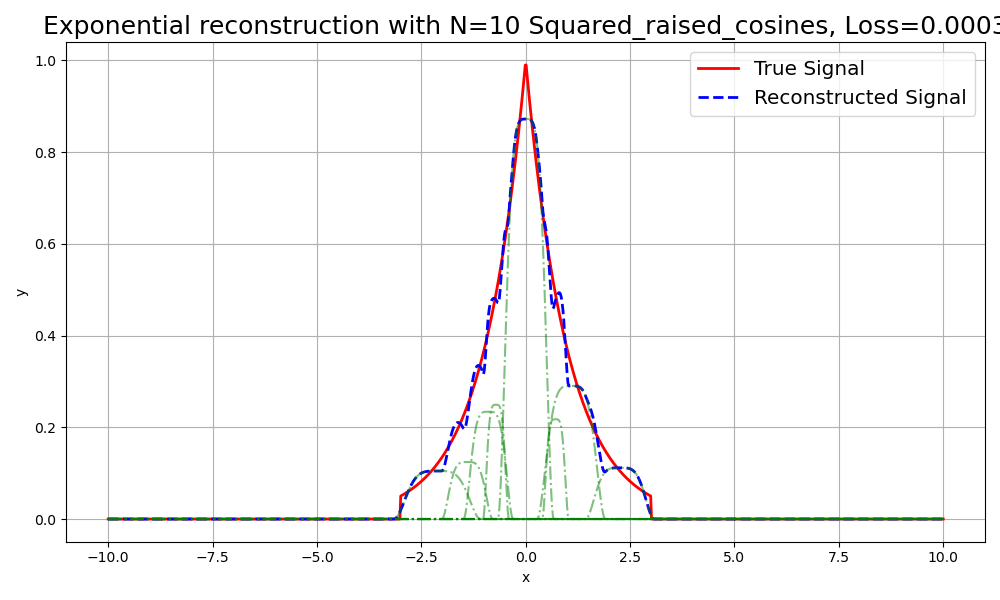} \\
    \includegraphics[width=0.33\textwidth]{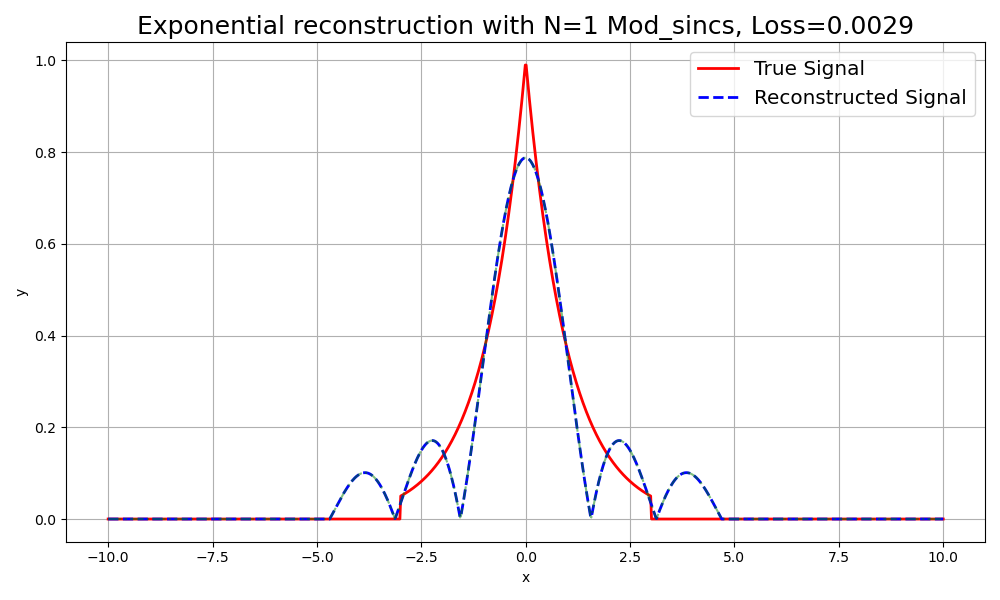} 
    \includegraphics[width=0.33\textwidth]{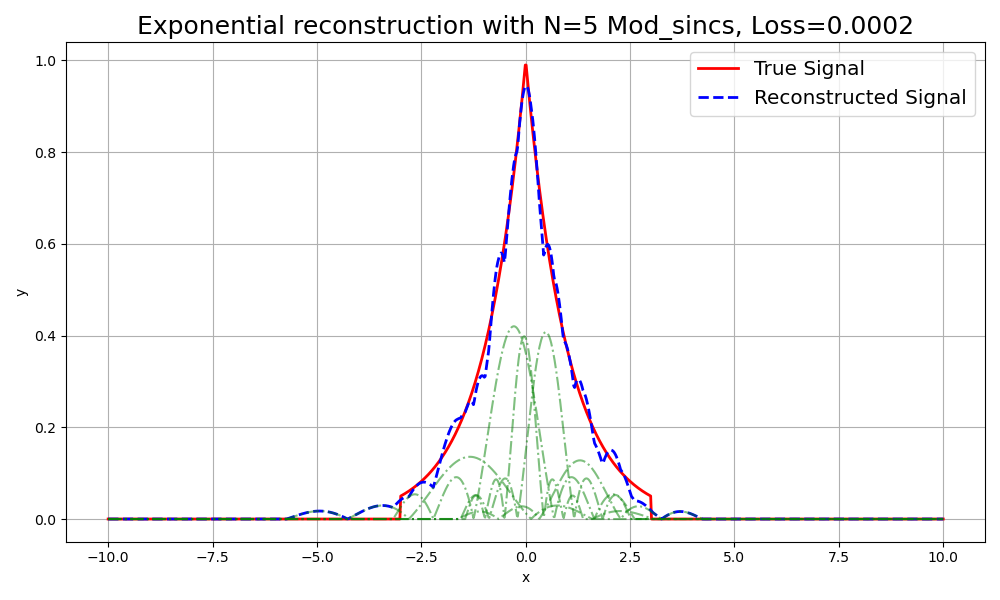} 
    \includegraphics[width=0.33\textwidth]{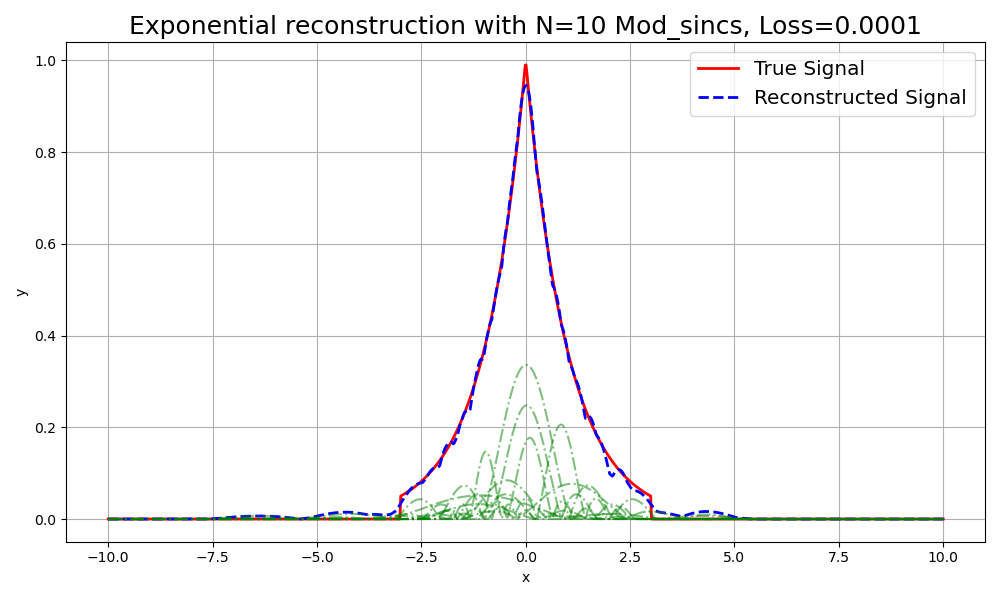} \\    \caption{Visualization of 1D simulations for different splatting methods with varying primitives (N=1, N=5, N=10) for a sharp exponential pulse. Each row corresponds to a specific splatting method: Gaussian, Cosine, Squared Cosine, Raised Cosine, Squared Raised Cosine, and Modulated Sinc. The columns represent the number of primitives used.}
    \label{fig:1DSim5}
\end{figure*}

\begin{figure*}
     % Column headings
    \parbox{0.33\textwidth}{\centering {N=1}} 
    \parbox{0.33\textwidth}{\centering {N=5}} 
    \parbox{0.33\textwidth}{\centering {N=10}} \\
    \includegraphics[width=0.33\textwidth]{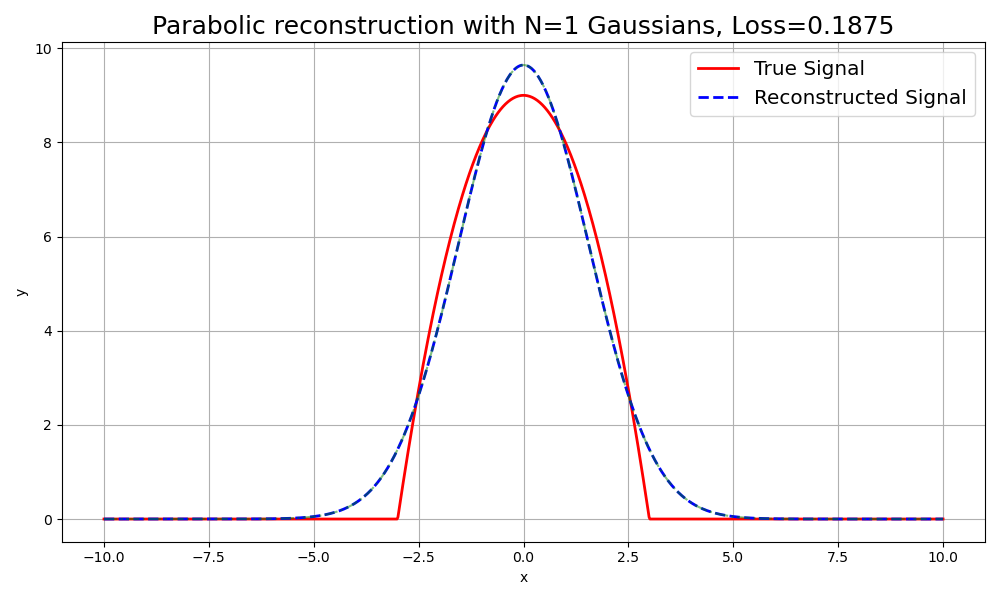} 
    \includegraphics[width=0.33\textwidth]{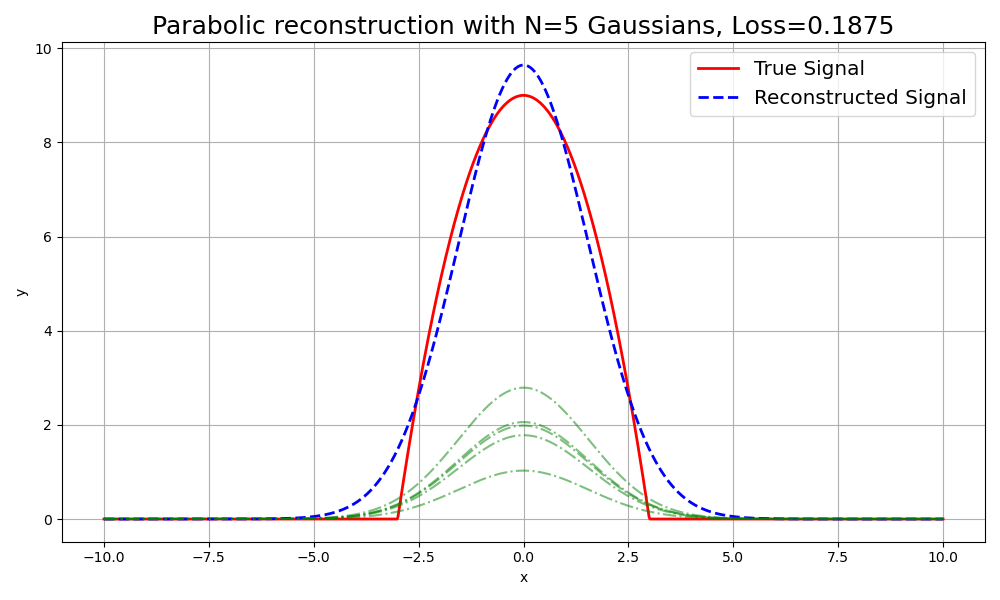} 
    \includegraphics[width=0.33\textwidth]{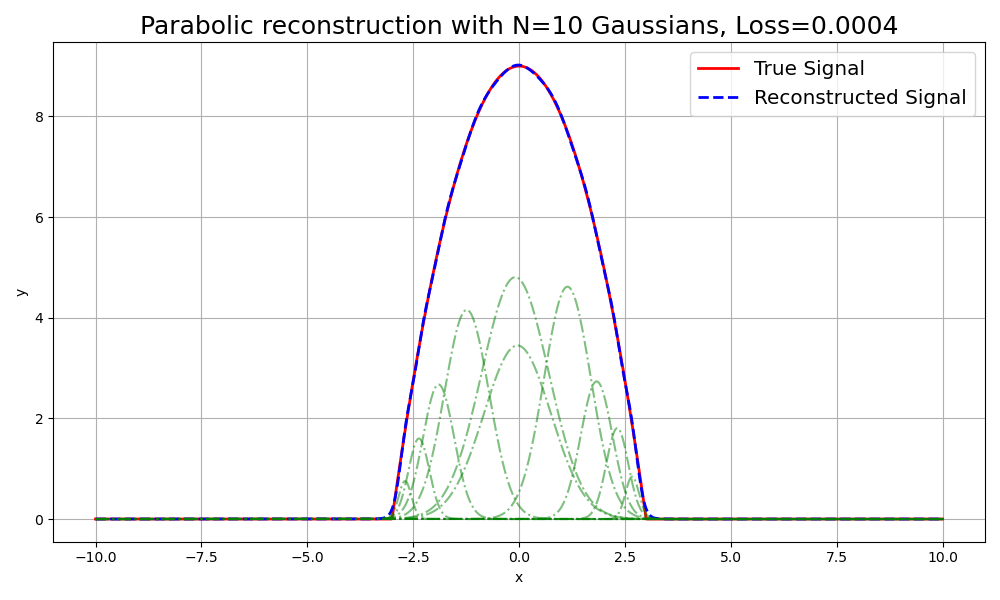} \\
    \includegraphics[width=0.33\textwidth]{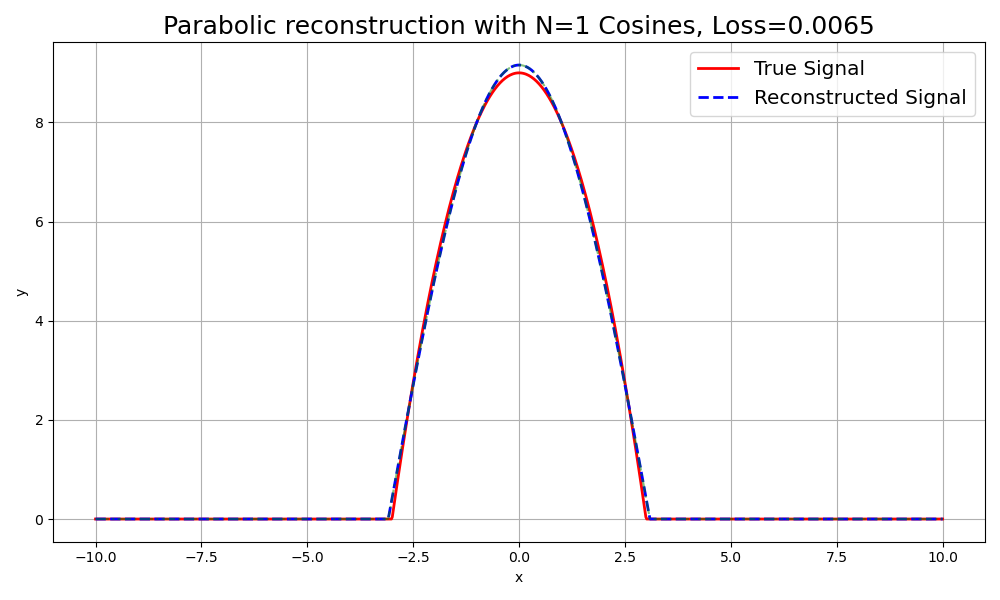} 
    \includegraphics[width=0.33\textwidth]{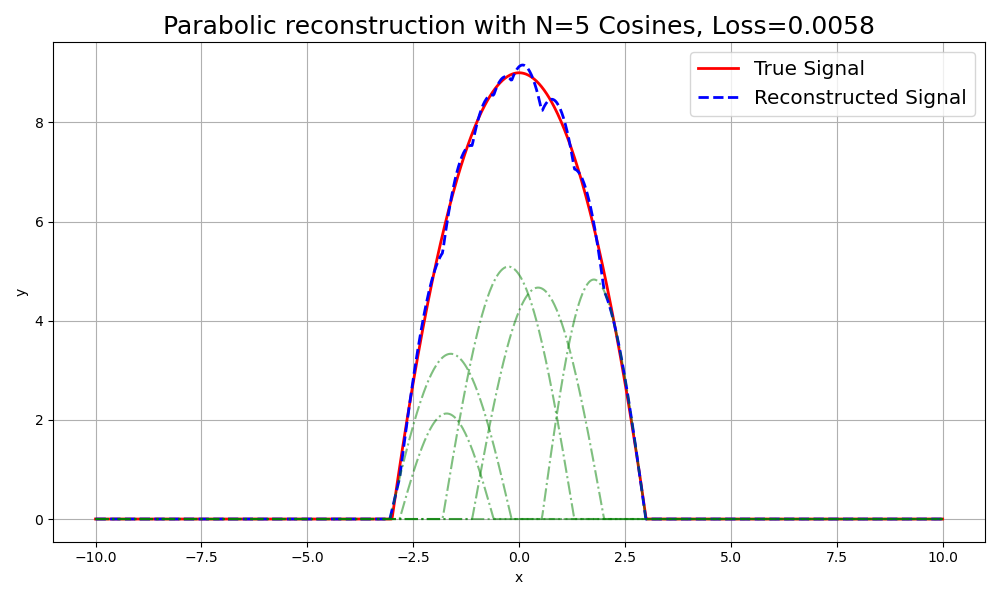} 
    \includegraphics[width=0.33\textwidth]{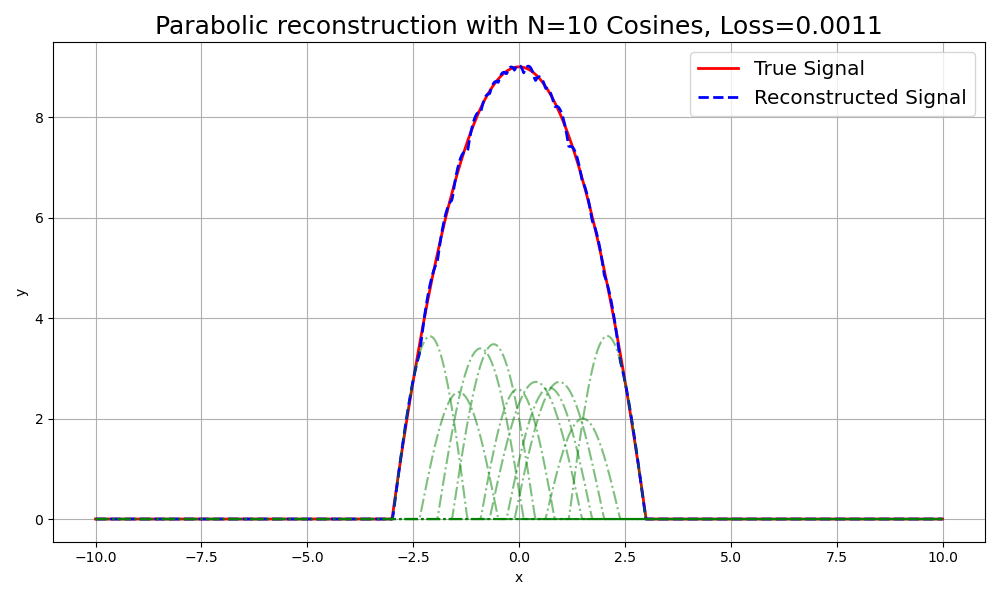} \\
    \includegraphics[width=0.33\textwidth]{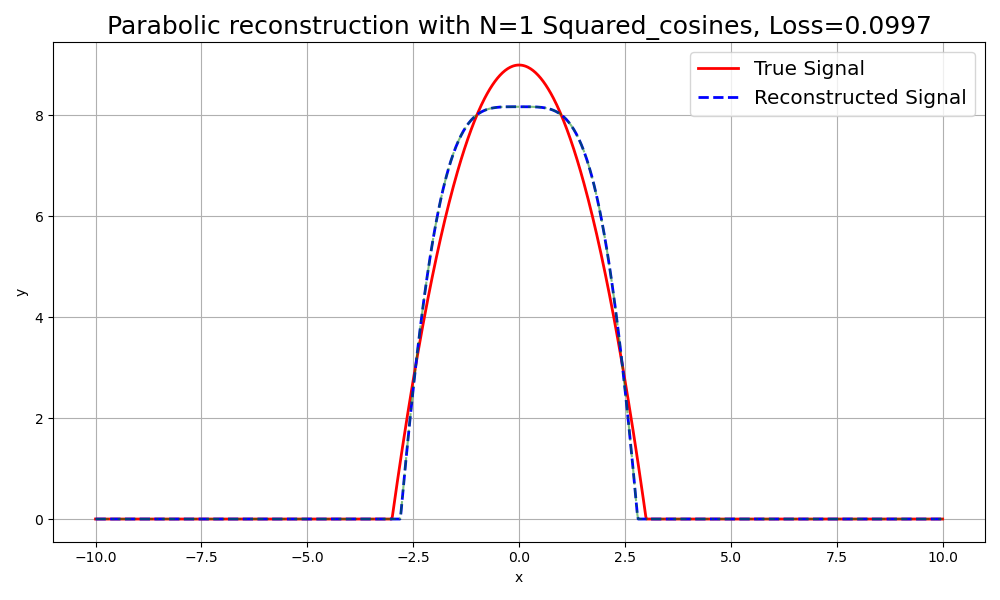} 
    \includegraphics[width=0.33\textwidth]{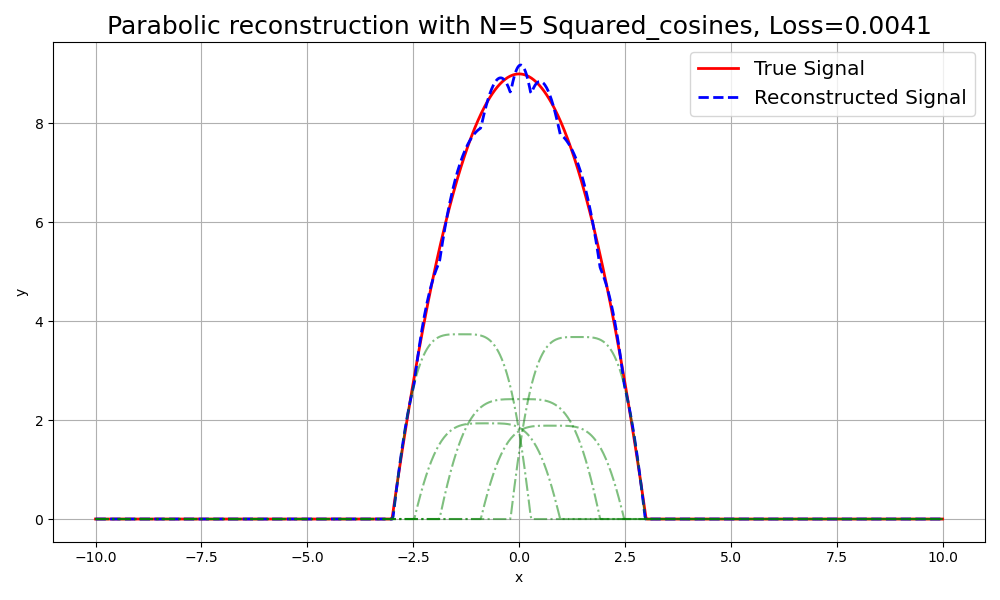} 
    \includegraphics[width=0.33\textwidth]{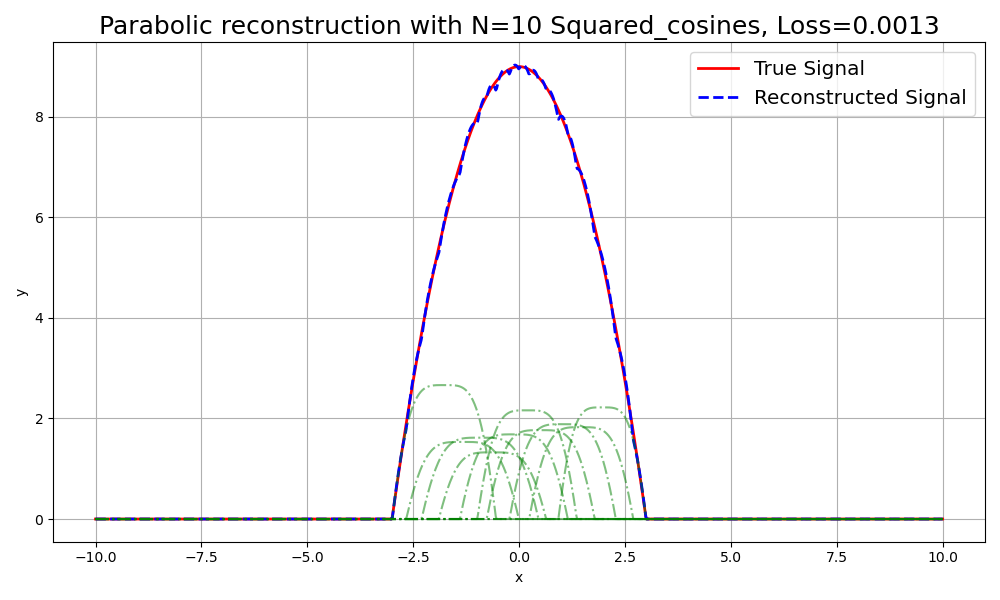} \\
    \includegraphics[width=0.33\textwidth]{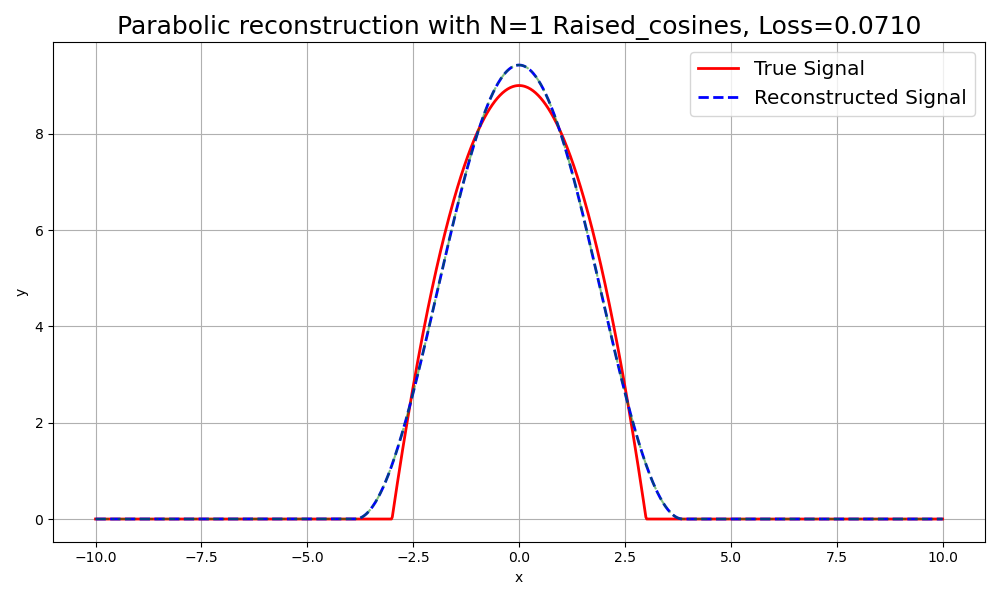} 
    \includegraphics[width=0.33\textwidth]{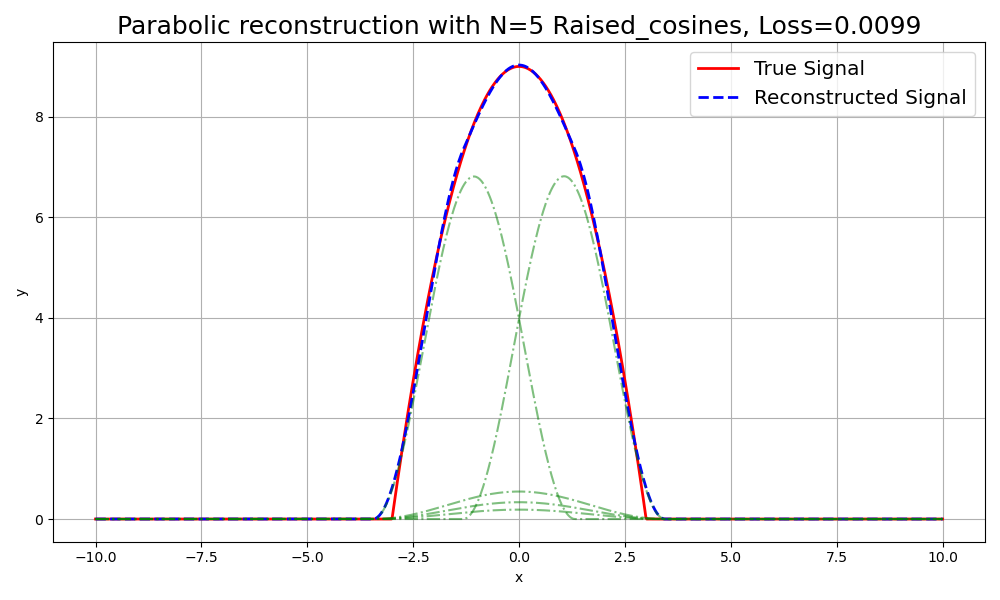} 
    \includegraphics[width=0.33\textwidth]{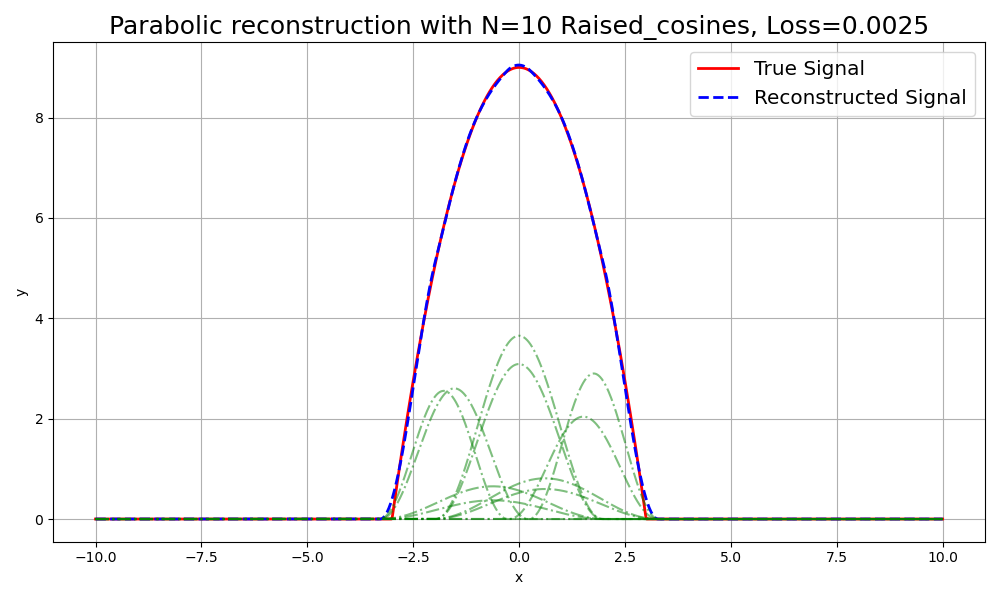} \\
    \includegraphics[width=0.33\textwidth]{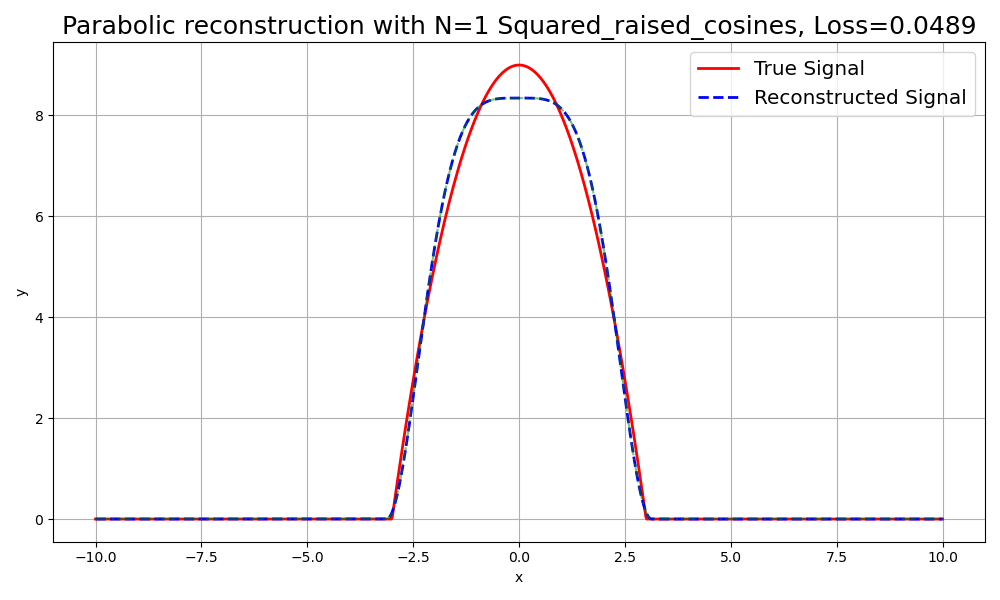} 
    \includegraphics[width=0.33\textwidth]{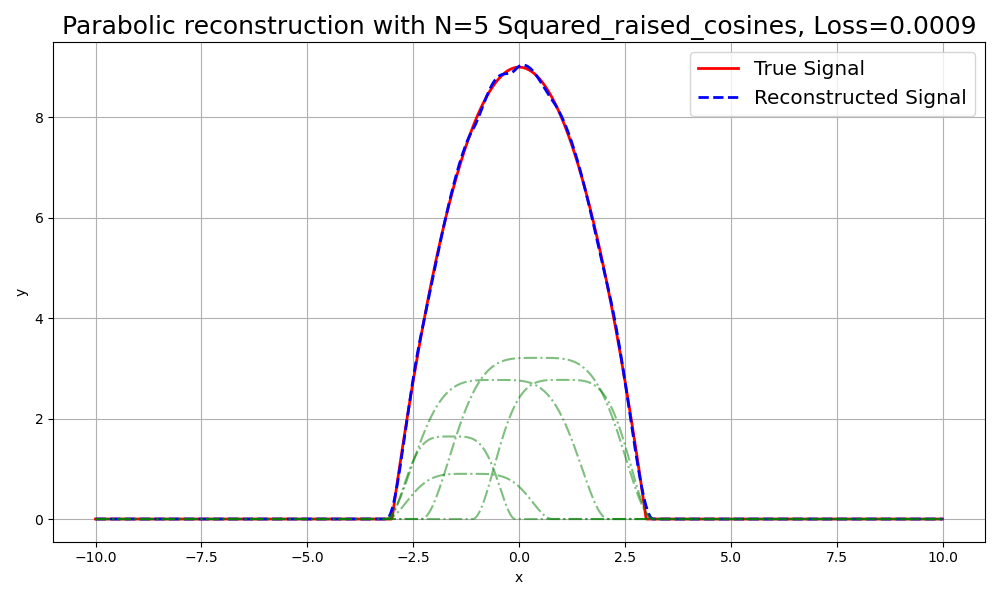} 
    \includegraphics[width=0.33\textwidth]{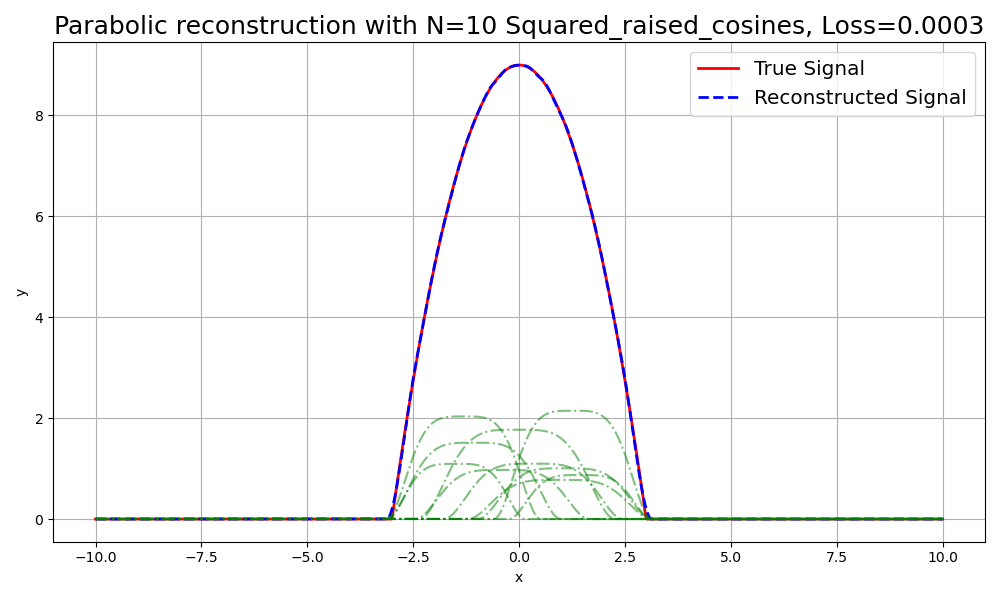} \\
    \includegraphics[width=0.33\textwidth]{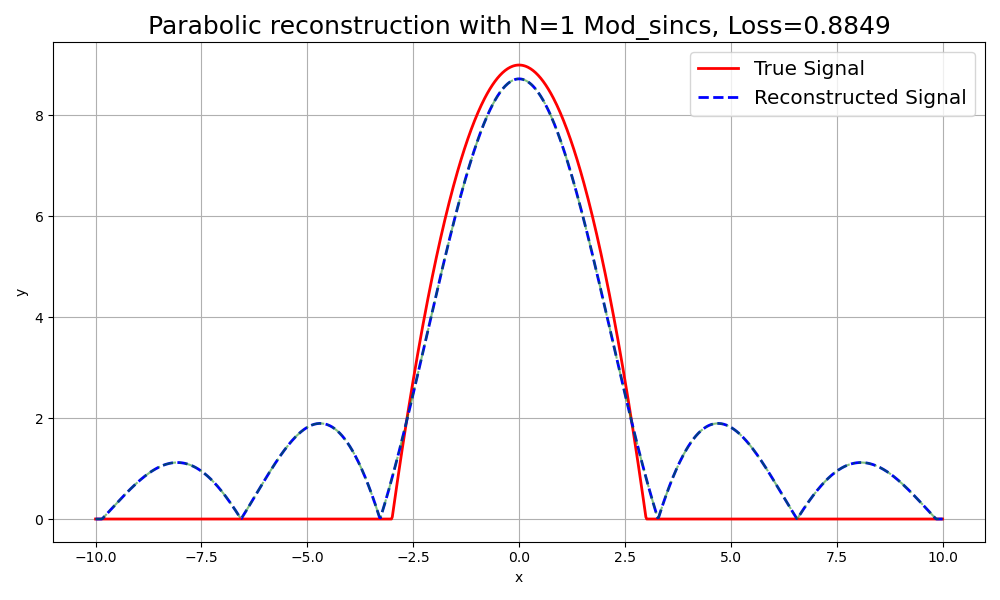} 
    \includegraphics[width=0.33\textwidth]{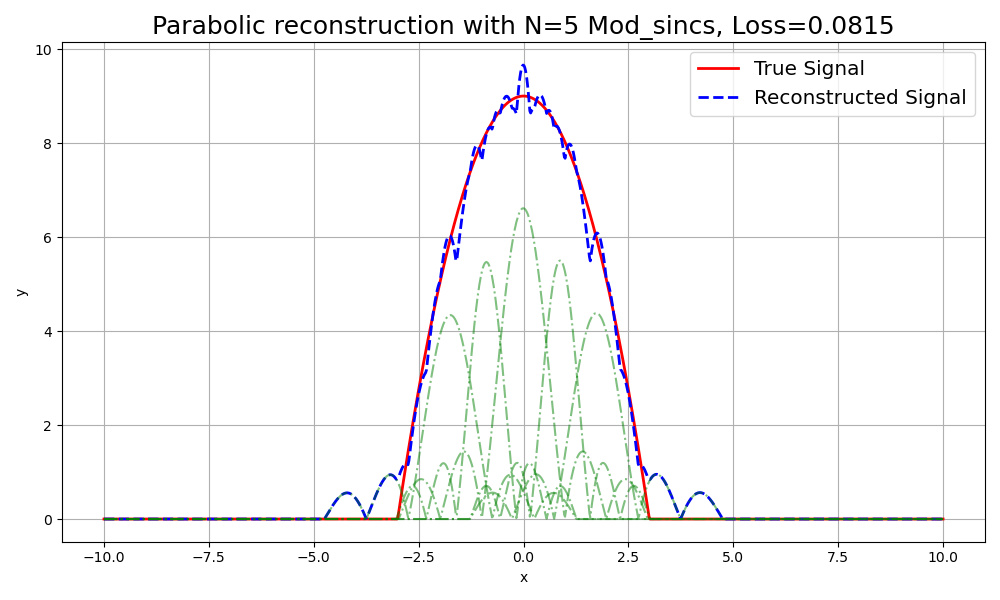} 
    \includegraphics[width=0.33\textwidth]{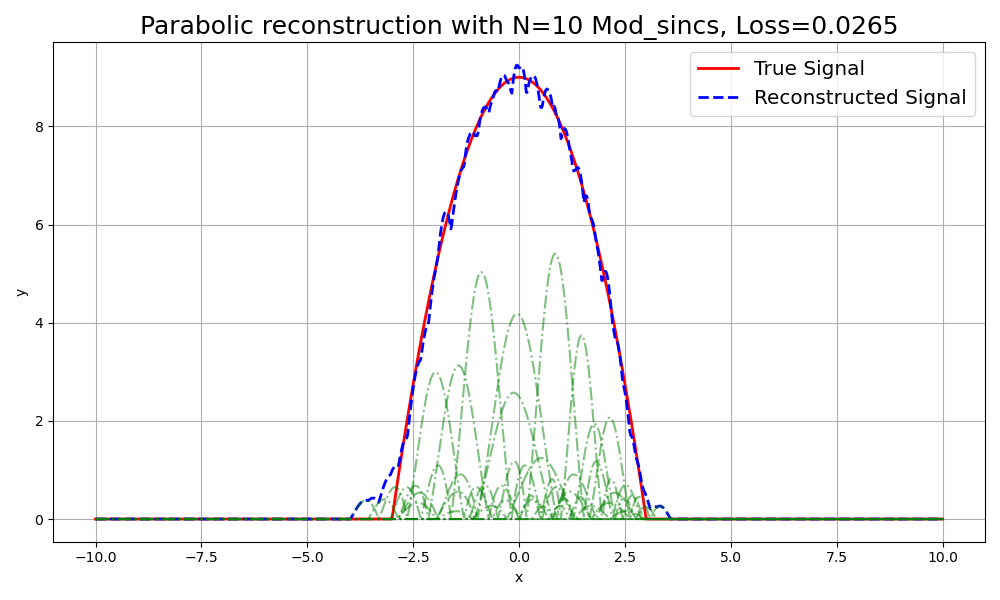} \\    \caption{Visualization of 1D simulations for different splatting methods with varying primitives (N=1, N=5, N=10) for a parabolic pulse. Each row corresponds to a specific splatting method: Gaussian, Cosine, Squared Cosine, Raised Cosine, Squared Raised Cosine, and Modulated Sinc. The columns represent the number of primitives used.}
    \label{fig:1DSim6}
\end{figure*}

\begin{figure*}
     % Column headings
    \parbox{0.33\textwidth}{\centering {N=1}} 
    \parbox{0.33\textwidth}{\centering {N=5}} 
    \parbox{0.33\textwidth}{\centering {N=10}} \\
    \includegraphics[width=0.33\textwidth]{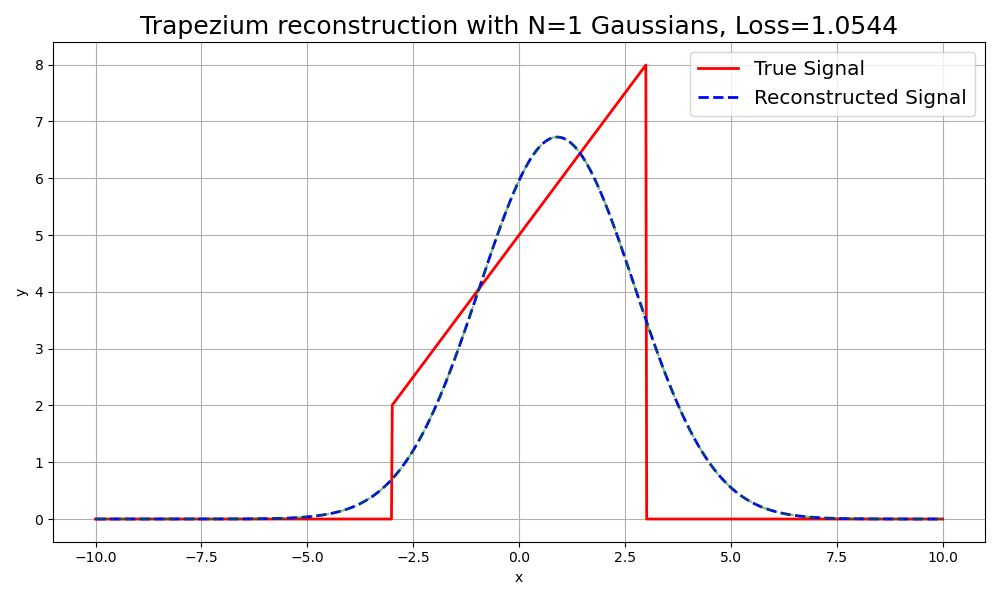} 
    \includegraphics[width=0.33\textwidth]{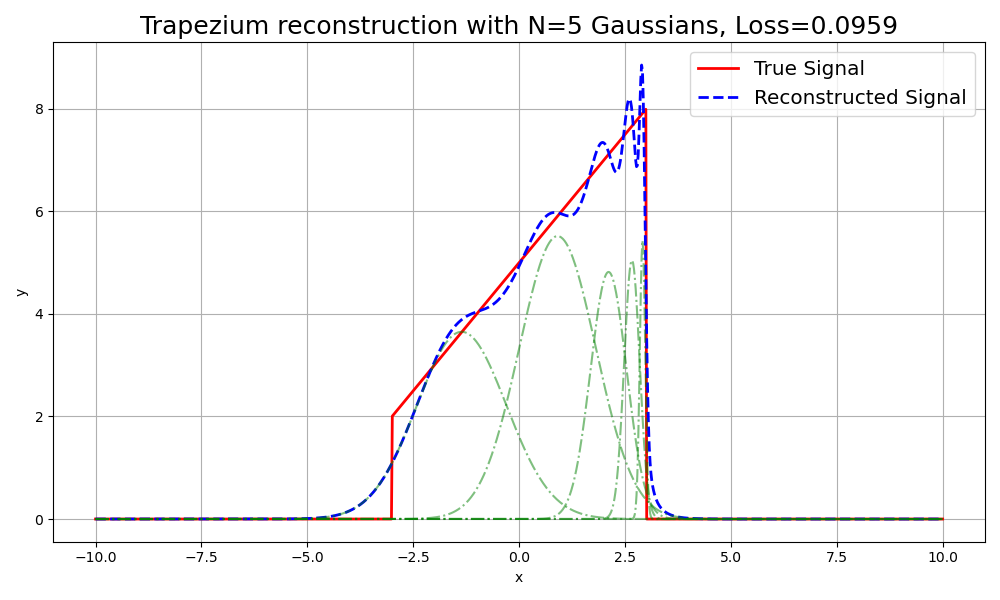} 
    \includegraphics[width=0.33\textwidth]{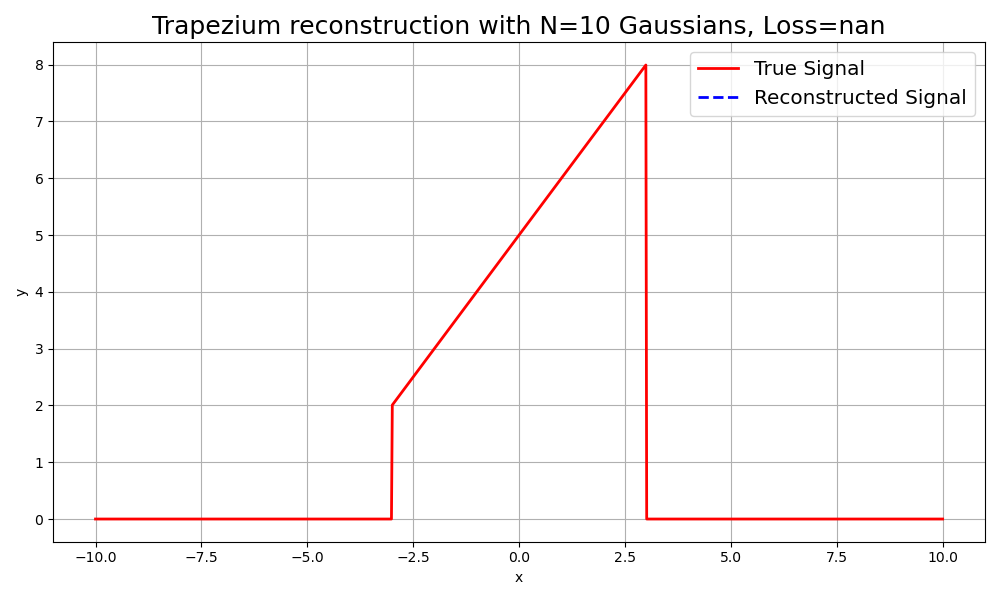} \\
    \includegraphics[width=0.33\textwidth]{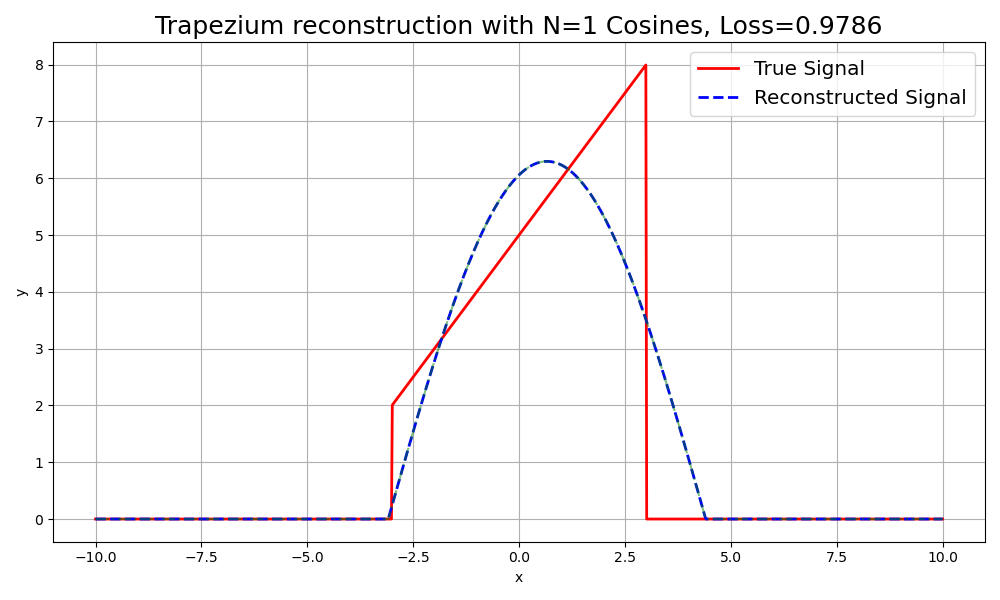} 
    \includegraphics[width=0.33\textwidth]{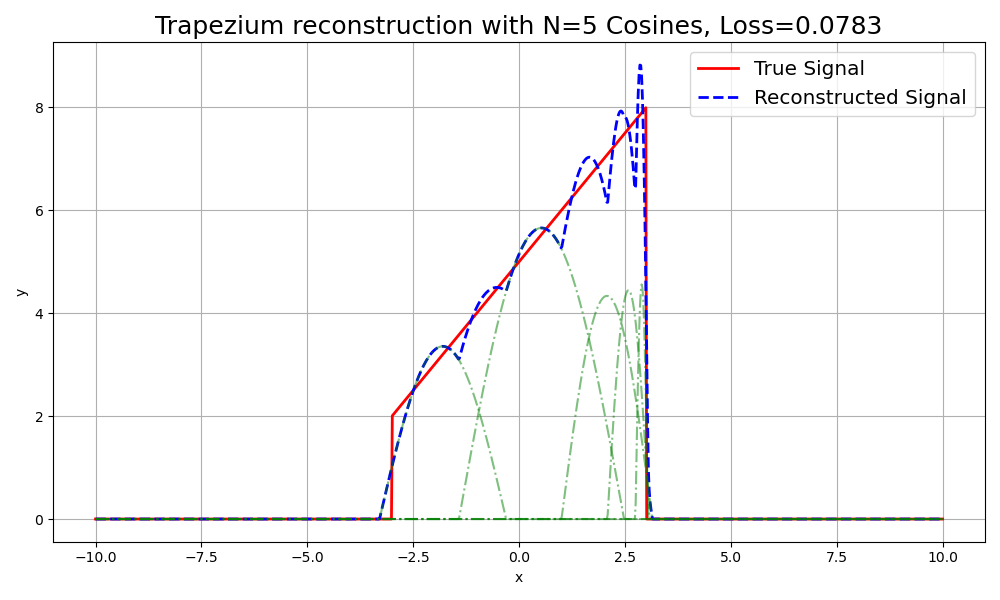} 
    \includegraphics[width=0.33\textwidth]{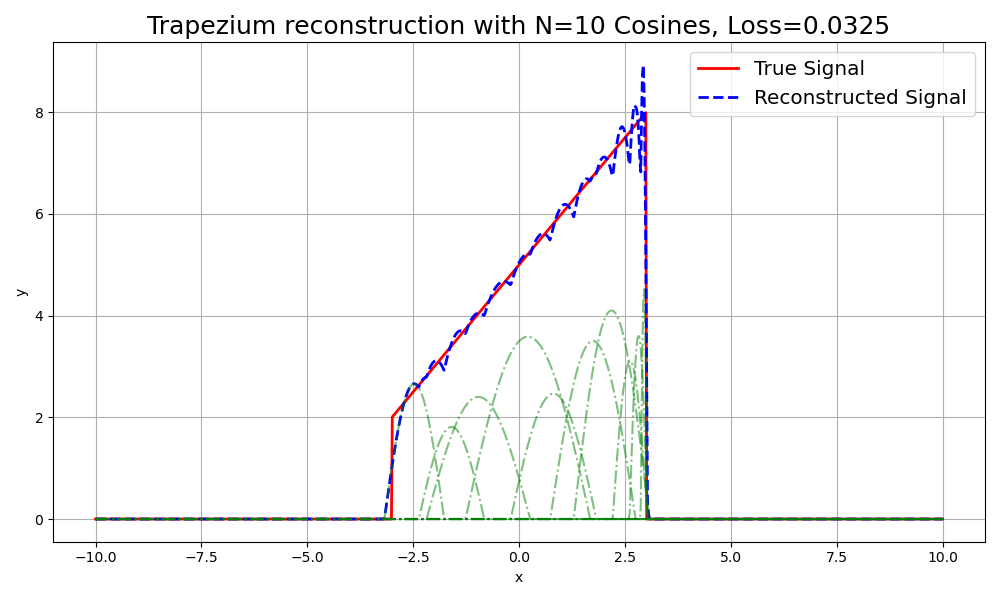} \\
    \includegraphics[width=0.33\textwidth]{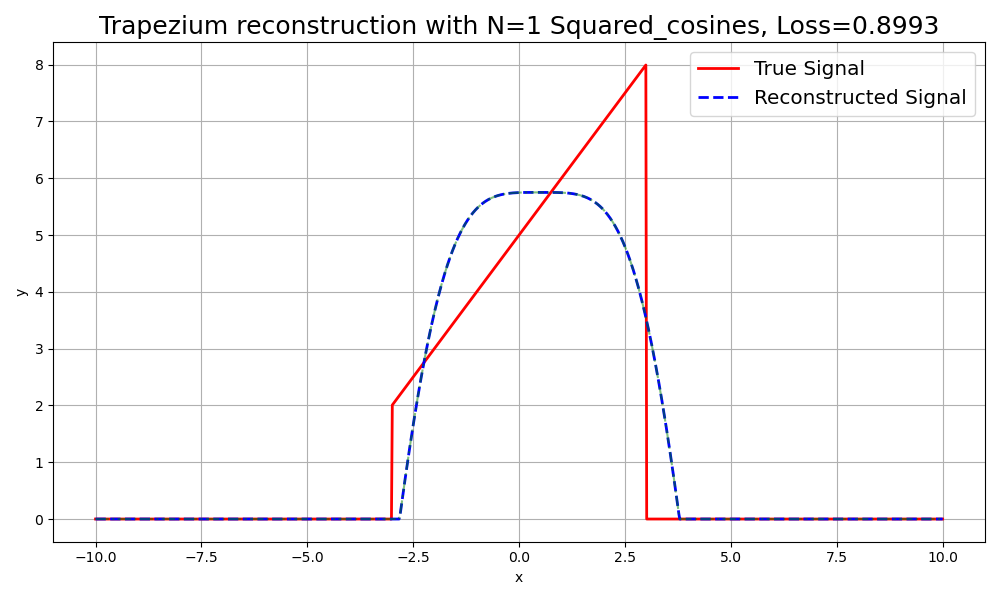} 
    \includegraphics[width=0.33\textwidth]{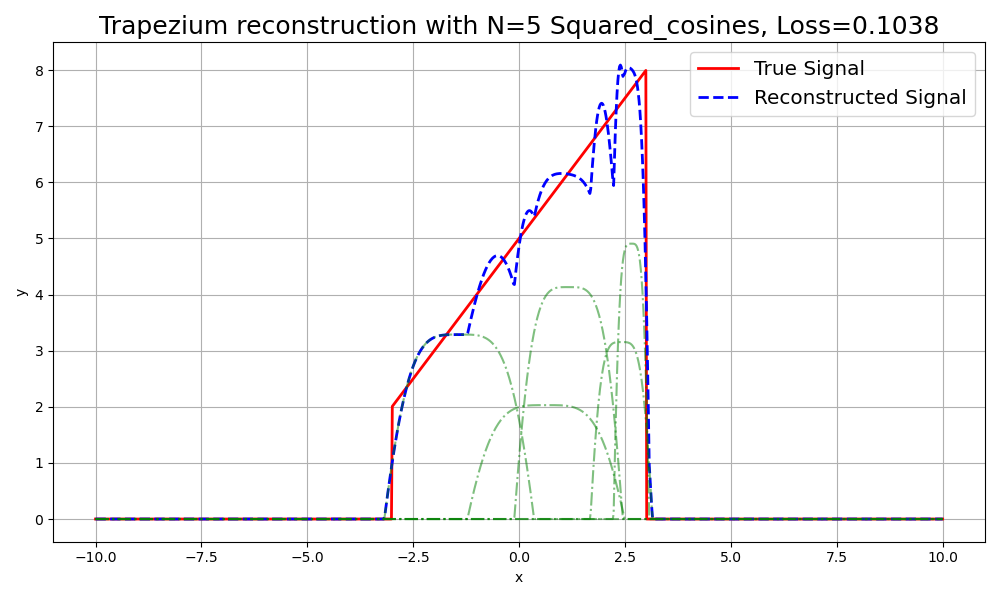} 
    \includegraphics[width=0.33\textwidth]{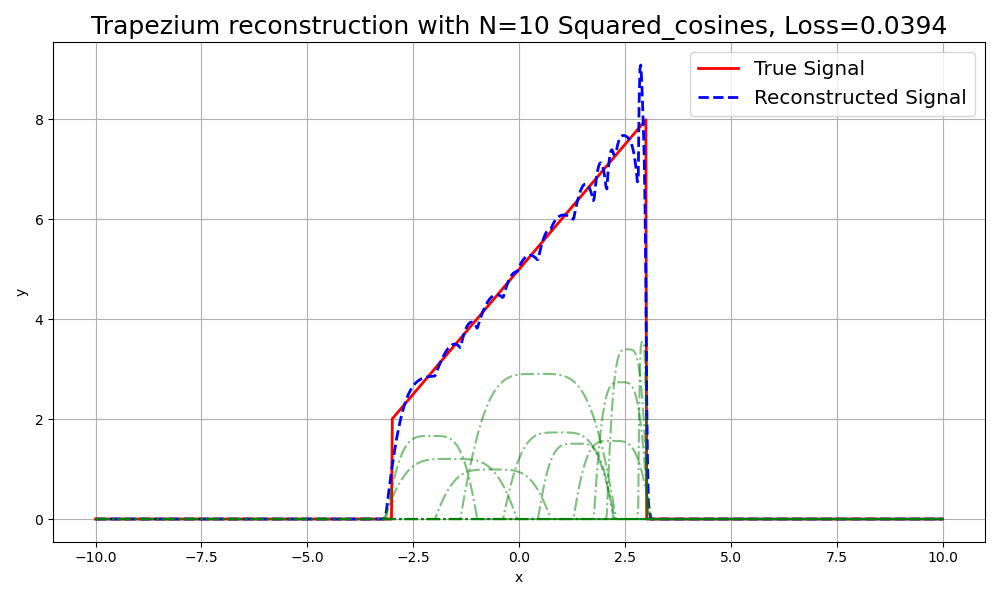} \\
    \includegraphics[width=0.33\textwidth]{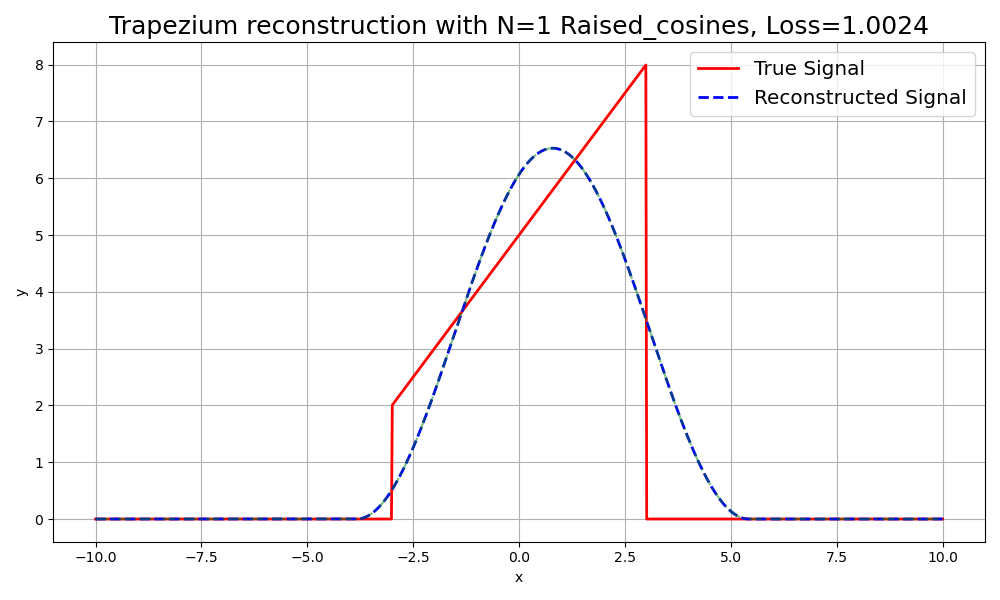} 
    \includegraphics[width=0.33\textwidth]{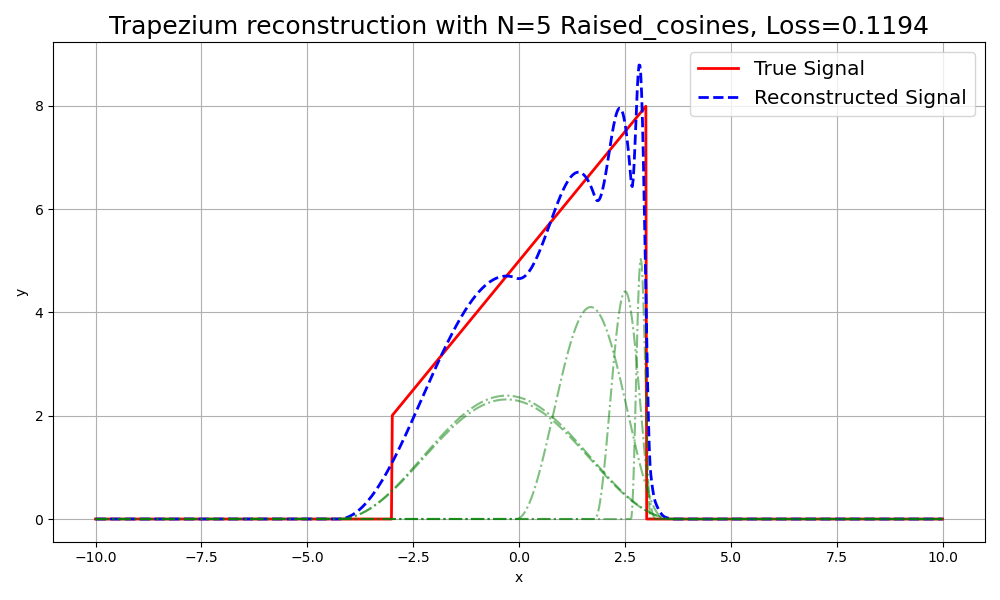} 
    \includegraphics[width=0.33\textwidth]{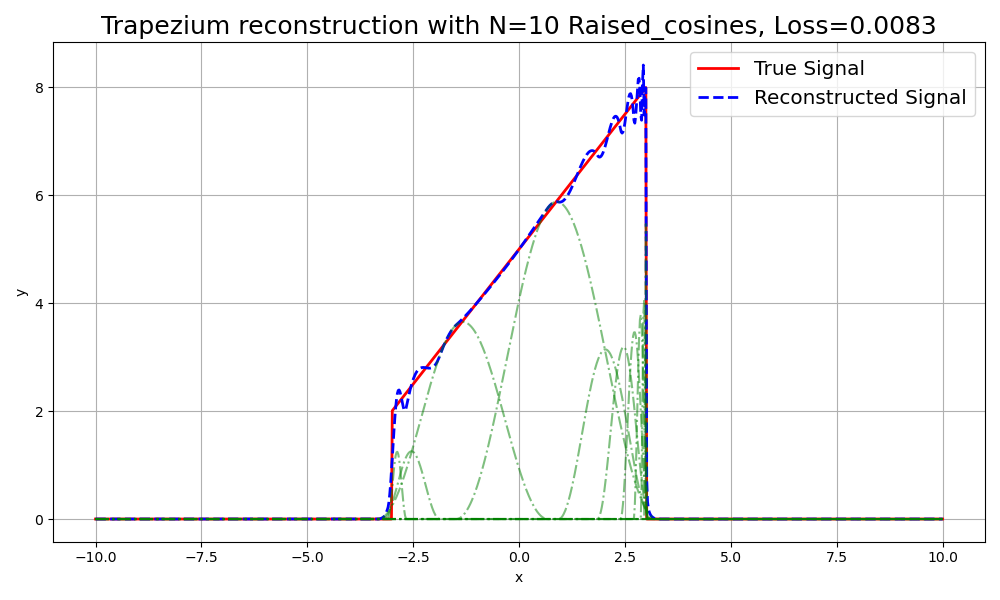} \\
    \includegraphics[width=0.33\textwidth]{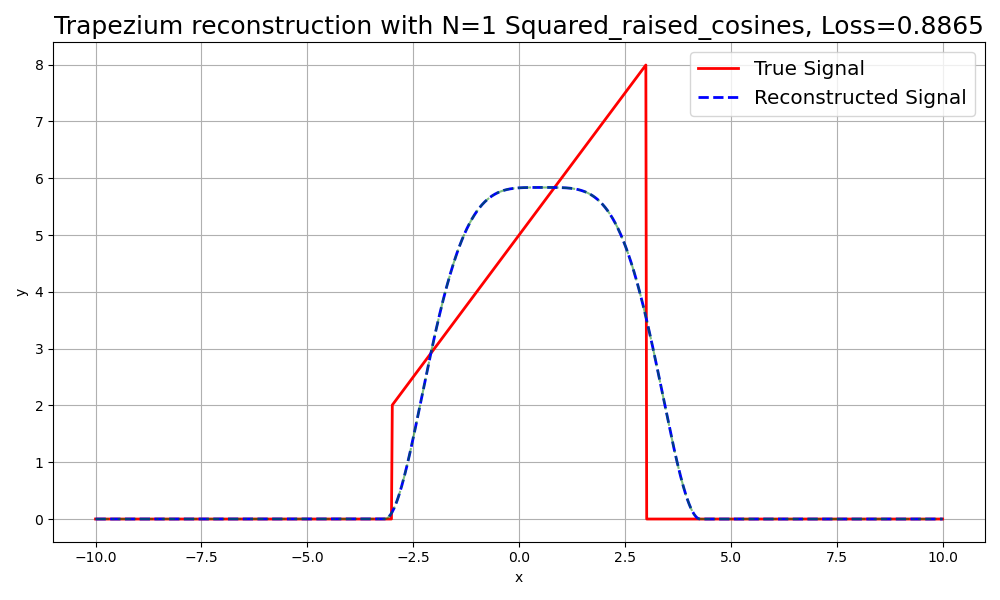} 
    \includegraphics[width=0.33\textwidth]{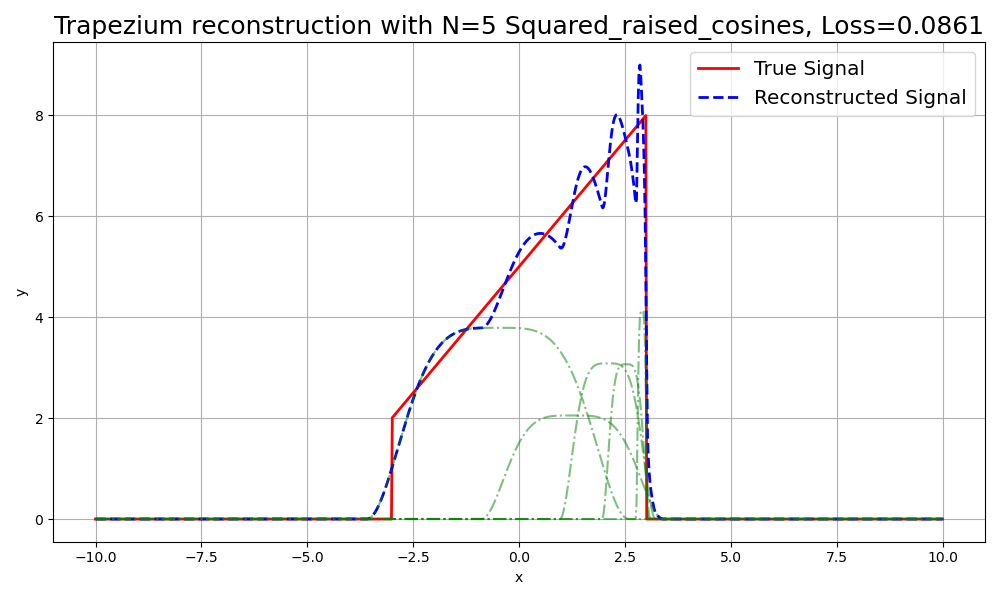} 
    \includegraphics[width=0.33\textwidth]{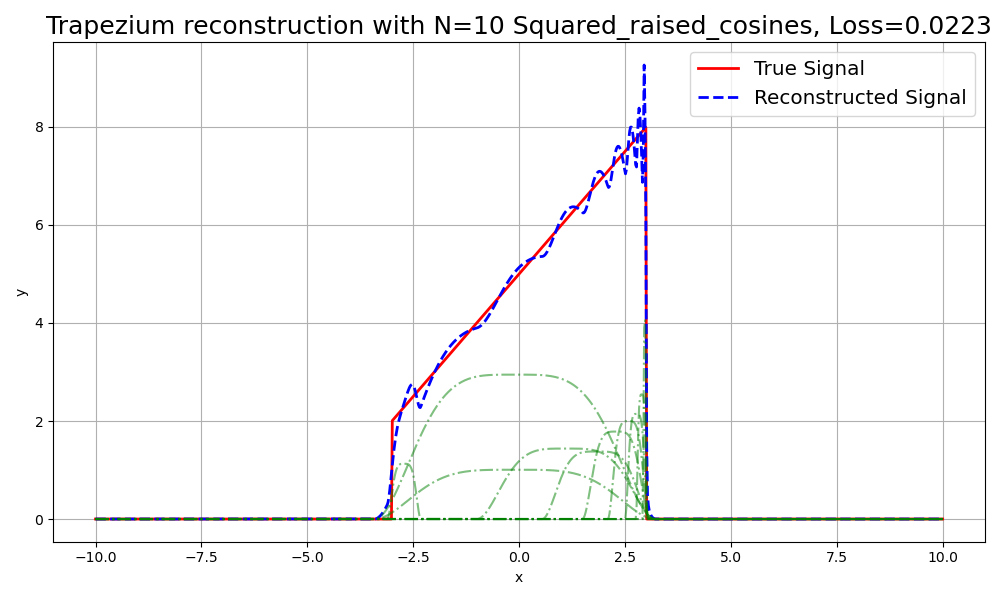} \\
    \includegraphics[width=0.33\textwidth]{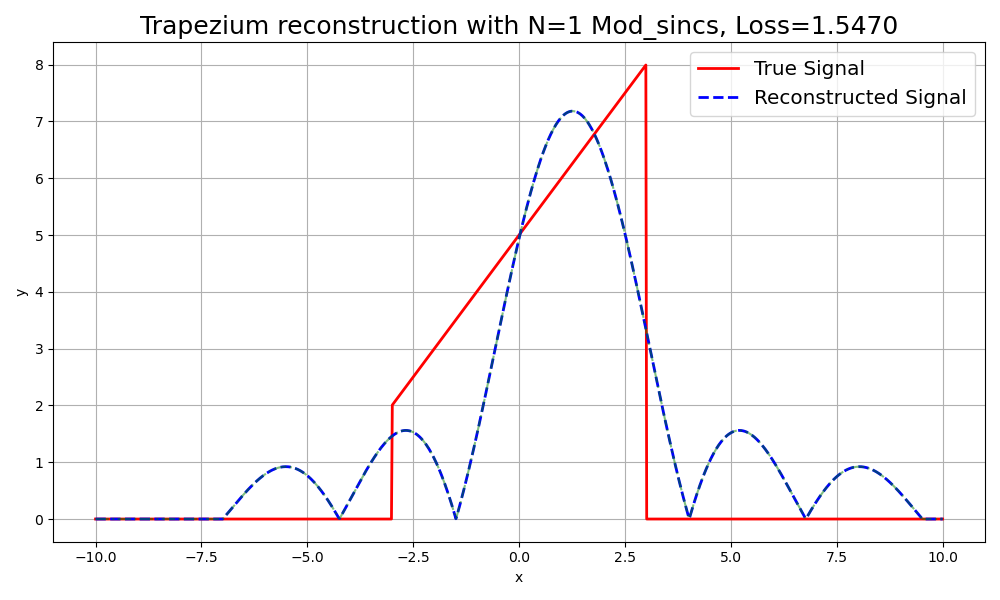} 
    \includegraphics[width=0.33\textwidth]{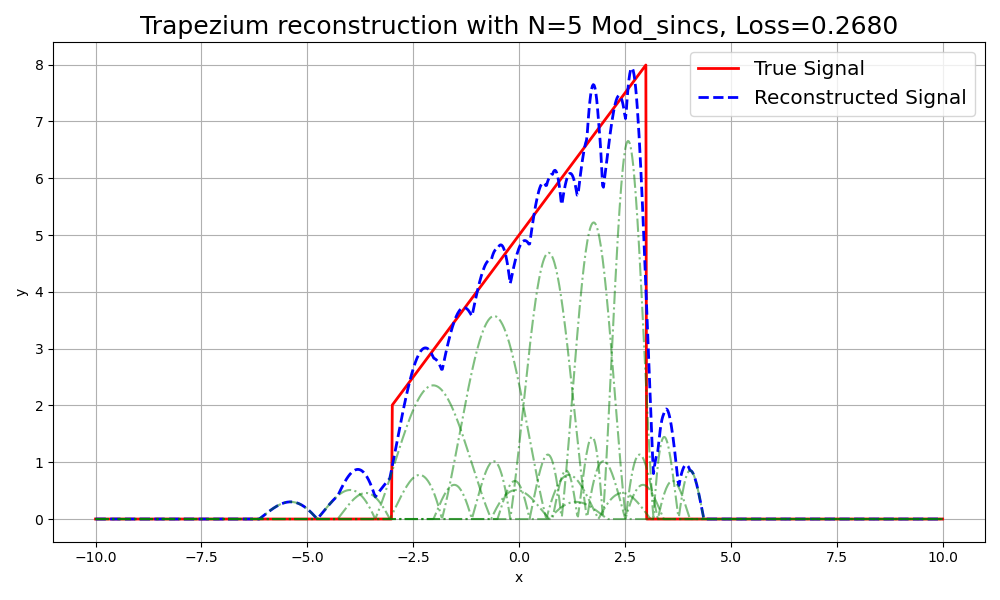} 
    \includegraphics[width=0.33\textwidth]{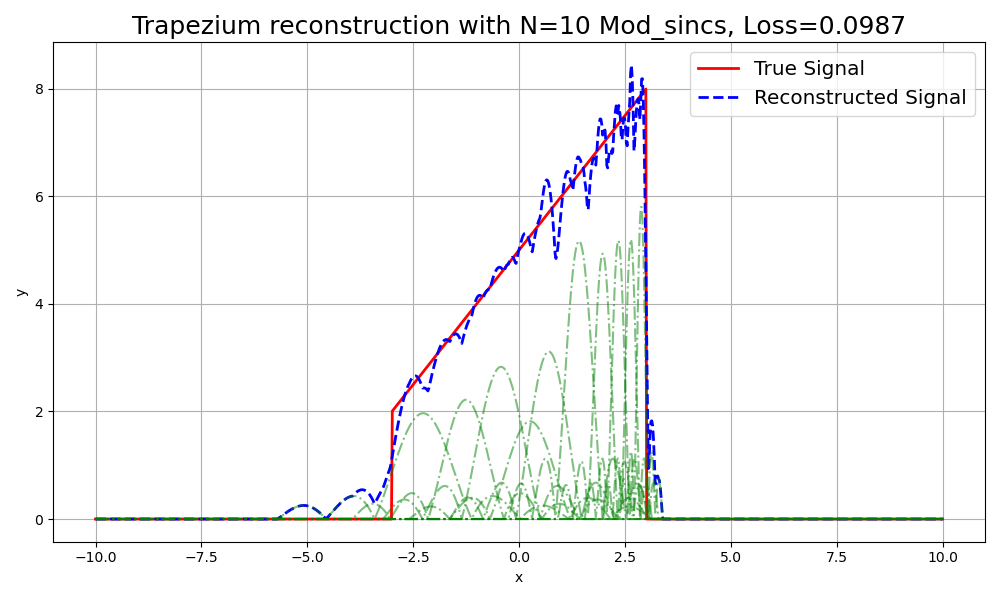} \\    \caption{Visualization of 1D simulations for different splatting methods with varying primitives (N=1, N=5, N=10) for a trapezoid pulse. Each row corresponds to a specific splatting method: Gaussian, Cosine, Squared Cosine, Raised Cosine, Squared Raised Cosine, and Modulated Sinc. The columns represent the number of primitives used.}
    \label{fig:1DSim7}
\end{figure*}

\textbf{1D Simulations. } In Figures ~\ref{fig:1DSim1}, ~\ref{fig:1DSim2}, ~\ref{fig:1DSim3}, ~\ref{fig:1DSim4}, ~\ref{fig:1DSim5}, ~\ref{fig:1DSim6}, ~\ref{fig:1DSim7}, we show an extended version of the initial 1D simulation described in Sec.~\ref{sec:AssessDARB} in our paper. Here, we conduct experiments with various reconstruction kernels in 1D as a toy experiment to understand their signal reconstruction properties. These kernels include Gaussians, cosines, squared cosines, raised cosines, squared raised cosines, and modulus sincs. The reconstruction process is optimized using backpropagation, with different means and variances applied to each kernel. This approach is used to reconstruct various complex signal types, including a square pulse, a triangular pulse, a Gaussian pulse, a half-sinusoid single pulse, a sharp exponential pulse, a parabolic pulse, and a trapezoidal pulse.

We are grateful to the authors of GES \cite{Ges} for open-sourcing their 1D simulation codes, which we have improved upon for this purpose. Expanding beyond GES, here, we also demonstrate the reconstruction of non-symmetric 1D signals to better represent real-world 3D reconstructions and further explore the capabilities of various DARBFs. As shown in the simulations, Gaussians are not the only effective interpolators; other DARBFs can provide improved 1D signal reconstructions in specific cases. We generate synthetic 3D covariance matrices that align with the properties of actual 3D covariances in the Gaussian Splatting pipeline. These matrices satisfy three key properties: (1) they are symmetric, (2) they are positive semi-definite, and (3) they have full rank. From these, we derive the actual 2D covariances, as proposed in \cite{EWA_Splatting}, by removing the last row and column of the 3D covariance matrices. Additionally, we obtain the projected 2D covariances by integrating along one axis using Simpson's rule.

To approximate the relationship between these projected 2D covariances and the actual 2D covariances defined in \cite{EWA_Splatting}, we train a single-layer Multi-Layer Perceptron (MLP). Through extensive experiments, we observed that this MLP model can be closely approximated by a linear regression model. Furthermore, the correction factor $\psi$ can be simplified to a scalar value.We input 6 unique variables ($a$, $b$, $c$, $d$, $e$, and $f$) derived from the 3D covariance and model the 3 unique outputs ($a'$, $b'$ and $d'$) from the 2D covariance.

\begin{equation}
\Sigma' = 
\begin{pmatrix}
a & b & c \\
b & d & e \\
e & d & f \\
\end{pmatrix}
\rightarrow
\begin{pmatrix}
a' & b' \\
b' & d' \\
\end{pmatrix}
= \Sigma'_{2 \times 2}.
\label{eq:2x2covProjection_unique}
\end{equation}

The results we obtained using 1000 random samples, similar to the 3D covariances from 3DGS, are as follows:
\vspace{-10pt}
\begin{equation}
a' = 1.36a -0.2b -0.1c - 0.0d +0.0e +0.0f
\end{equation}
\begin{equation}
b' = 0.0a + 1.36b - 0.0c - 0.0d +0.0e +0.0f
\end{equation}
\begin{equation}
d' = 0.0a + 0.0b -0.2c + 1.36d +0.1e +0.0f
\end{equation}

Experiments show that only the direct 3D covariance elements corresponding to the 2D covariance elements have higher weights, which can be represented as a single factor. A detailed statistical analysis using T-tests revealed distinct differences, highlighting the influence of weights on specific components of the 2D covariance matrix. 
\begin{equation}
\Sigma' = 
\begin{pmatrix}
a & b & c \\
b & d & e \\
e & d & f \\
\end{pmatrix}
\rightarrow
\psi \begin{pmatrix}
a & b \\
b & d \\
\end{pmatrix}
= \Sigma'_{2 \times 2}.
\label{eq:2x2covProjection_supp}
\end{equation}

\begin{table}[h]
    \centering
    \scriptsize
    \begin{tabular}{c|cc|cc|cc}
        \toprule
        \textbf{LR} & \multicolumn{2}{c|}{\textbf{PSNR}} & \multicolumn{2}{c|}{\textbf{Training Time}} & \multicolumn{2}{c}{\textbf{Memory}} \\
        & Gaussian & HCS & Gaussian & HCS & Gaussian & HCS \\
        \midrule
        0.01 & 31.49  & 31.335  & 22:31  & 23:26  & 255.61  & 246.48  \\
        0.02 & 31.75  & 31.34   & 19:48  & 17:32  & 293.61  & 282.65  \\
        0.1  & 31.1   & 29.38   & 13:31  & 11:20  & 364     & 391.76  \\
        0.2  & 30.5   & 26.54   & 10:34  & 8:26   & 241.42  & 315.93  \\
        \bottomrule
    \end{tabular}
    \caption{Comparison of Gaussian and HCS methods based on PSNR, Training Time, and Memory usage for different learning rates.}
    \label{tab:comparisonwithLR}
\end{table}

\begin{table}[h!]
    \centering
    % \caption{Comparison with prior methods in 3D reconstruction methods in PSNR, LPIPS, and training efficiency. Benchmark values for 3DGS and GES are from their respective papers and may not ensure a fully fair comparison.}
    \scriptsize
    \setlength{\tabcolsep}{8pt}
    \begin{tabular}{@{}lccccr@{}}
        \hline
        \rowcolor{white}
        Method & SSIM$\uparrow$ & PSNR$\uparrow$ & LPIPS$\downarrow$ & Train$\downarrow$ \\
        \hline
        Plenoxels~\cite{fridovich2022plenoxels} & 0.626 & 23.08 & 0.463 & 19m$^*$ \\

        INGP~\cite{InstantNGP} &  0.699 & 25.59 & 0.331 &  \cellcolor{red!25}5.5m$^*$ \\
        % BakedSDF~\cite{73} & 0.697 & 24.51 & 0.309 & \cellcolor{orange!30}539 \\
        % SMERF~\cite{10} & \cellcolor{orange!15}0.818 & \cellcolor{orange!15}27.99 & \cellcolor{orange!15}0.211 & 228 \\
        % \hline
        Mip-NeRF360~\cite{Mip-nerf_360} &  0.792 & 27.69 & 0.237 & 48h \\
        3DGS~\cite{kerbl3Dgaussians} &  \cellcolor{red!25}0.815 &  \cellcolor{orange!25}27.21 &  \cellcolor{red!25}0.214 & 30m$^*$ \\
        % EAGLES~\cite{14} & 0.809 & 27.16 & 0.238 & 137 \\
        GES~\cite{Ges} &  0.794 &  26.91 &  0.250 &  23m$^*$ \\
        \hline
        \textbf{DARBS (RC)} & \cellcolor{orange!25}0.813 &\cellcolor{red!25}27.45 &\cellcolor{red!25}0.214 & \cellcolor{yellow!25}19min \\
        \textbf{DARBS (HC)} & \cellcolor{yellow!25}0.790 &  \cellcolor{yellow!25}27.04 &  \cellcolor{orange!25}0.247 & \cellcolor{orange!25}16min \\
        % \textbf{Ours} & \cellcolor{red!15}0.843 & \cellcolor{red!15}28.14 & \cellcolor{red!15}0.171 & 1.924 \\
        % Zip-NeRF~\cite{2} & 0.836 & 28.54 & 0.177 & 0.25 \\
        \hline
    \end{tabular}
    % \caption{Comparison of various existing methods with SSIM, PSNR, LPIPS and training time metrics.}
    \caption{Comparison with prior methods in 3D reconstruction methods in PSNR, LPIPS, and training efficiency. Benchmark values for 3DGS and GES are from their respective papers and may not ensure a fully fair comparison.}
    \label{table:comparison}
\end{table}

\begin{table}[h]
    \centering
    \scriptsize
    \begin{tabular}{c|cc|cc}
        \toprule
         {\textbf{Scene}}& \multicolumn{2}{c|}{\textbf{RTX 3090}} & \multicolumn{2}{c}{\textbf{RTX 4090}} \\
        (Mip-NeRF 360 dataset) & Gaussian & HC & Gaussian & HC \\
        \midrule
        Kitchen & 32m 15s  & 29m 59s  & 21m 53s  & 19m 41s  \\
        Room    & 28m 32s  & 25m 28s  & 19m 48s  & 17m 32s  \\
        \bottomrule
    \end{tabular}
    \caption{Performance comparison of Gaussian and half-cosine (HC) methods on RTX 3090 and RTX 4090 for different environments. Even with upgraded computational capabilities, our results indicate that half-cosine kernels achieve better training speeds compared to Gaussian kernels.}
    \label{tab:performance}
\end{table}

By formulating this as a regression problem and applying closed-form solutions, we determined the scalar correlation factor for each DARBF. The scalar approximations for each kernel are presented in the original paper in Table~\ref{tab:kernels_cf}.

\textbf{Convergence Experiments. }We acknowledge that the objective function remains Lipschitz continuous, as the smoothness induced by each reconstruction kernel may vary. The Table~\ref{tab:comparisonwithLR} above presents the convergence behavior for different learning rates. Neither reconstruction kernel achieves optimal performance at excessively high or low learning rates.

Cosine kernels do not induce smoothness to the same extent as Gaussian kernels. At higher learning rates, cosine-based reconstructions may perform worse than Gaussian-based ones. However, at optimal learning rates, the inherent properties of cosine kernels—such as their steeper roll-off, sparse gradients due to the blunt peak, and spatial constraints—enable them to achieve competitive visual quality and faster convergence compared to Gaussian kernels.

As this is the first study exploring the behavior of non-exponential kernels in this context, a detailed mathematical analysis remains beyond the scope of this work. However, we identify opportunities for future research to further investigate the mathematical properties of these kernels through rigorous analytical methods.

%%%%%%%%%%%%%%%%%%%%%%%%%%%%%%%%%%%%%%%

%%%%%%%%%% Fig. 2 for Train Plots %%%%%%%%%%%

%%%%%%%%%%%%%%%%%%%%%%%%%%%%%%%%%%%%%%%%

\newpage\begin{table*}[t]
\footnotesize
\centering
% \caption{\textbf{Performance metrics across various scenes.} Red color denotes best performance, while yellow denotes third-best. Higher values are better for PSNR, SSIM, and lower values are better for LPIPS, Memory, and Training Time. Here we present a detailed breakdown of results for all scenes from Mip-NeRF360 \cite{Mip-nerf_360} dataset for 7k and 30k iterations. The performance of each model is heavily affected by the nature of the scene. Some values may differ from the main table due to stochastic processes, as these results are from a single instance of a full evaluation experiment series. In contrast, Table~\ref{Table:FinalResultTable} presents the mean results averaged across multiple experiments.}
\begin{tabular}{llc|ccccccccc|c}
\toprule
Metric & Model & Step & Bicycle & Flowers & Garden & Stump & Treehill & Room & Counter & Kitchen & Bonsai & Mean \\
\midrule
\multirow{10}{*}{PSNR (dB)} & \multirow{2}{*}{3DGS}  & 7k  & \cellcolor{orange!35}23.759 & \cellcolor{orange!35}20.4657 & \cellcolor{orange!35}26.21 & \cellcolor{orange!35}25.712 & \cellcolor{orange!35}22.09 & \cellcolor{red!25}29.439 & \cellcolor{orange!35}27.179 & \cellcolor{orange!35}29.213 & \cellcolor{red!25}29.863 & \cellcolor{orange!35}25.9525 \\
                       &                        & 30k & \cellcolor{red!25}25.248 & \cellcolor{red!25}21.519 & \cellcolor{orange!35}27.352 & \cellcolor{red!25}26.562 & \cellcolor{yellow!25}22.554 & \cellcolor{red!25}31.597 & \cellcolor{red!25}29.055 & \cellcolor{yellow!25}31.378 & \cellcolor{red!25}32.316 & \cellcolor{red!25}27.4509 \\
\cmidrule{2-13}
                       & \multirow{2}{*}{3DRCS} & 7k  & \cellcolor{red!25}23.81 & \cellcolor{red!25}20.52 & \cellcolor{red!25}26.274 & \cellcolor{red!25}25.825 & \cellcolor{orange!35}22.09 & \cellcolor{yellow!25}29.3625 & \cellcolor{red!25}27.261 & \cellcolor{red!25}29.312 & \cellcolor{yellow!25}29.635 & \cellcolor{red!25}26.0058 \\
                       &                        & 30k & \cellcolor{orange!35}25.222 & \cellcolor{orange!35}21.492 & \cellcolor{red!25}27.374 & \cellcolor{orange!35}26.542 & 22.456 & \cellcolor{yellow!25}31.342 & \cellcolor{orange!35}29.032 & \cellcolor{orange!35}31.431 & 31.842 & \cellcolor{red!25}27.4547 \\
\cmidrule{2-13}
                       & \multirow{2}{*}{3DHCS} & 7k  & 23.037 & 19.929 & 25.351 & 24.821 & \cellcolor{yellow!25}22.078 & 29.127 & 26.837 & 28.756 & 29.481 & 25.4607 \\
                       &                        & 30k & 24.369 & 21.046 & 26.585 & 25.937 & \cellcolor{orange!35}22.564 & 31.003 & 28.782 & \cellcolor{red!25}31.95 & \cellcolor{yellow!25}31.89 & 27.0451 \\
\cmidrule{2-13}
                       & \multirow{2}{*}{3DSS}  & 7k  & \cellcolor{yellow!25}23.371 & \cellcolor{yellow!25}20.177 & \cellcolor{yellow!25}25.685 & \cellcolor{yellow!25}25.2 & \cellcolor{red!25}22.134 & \cellcolor{orange!35}29.373 & \cellcolor{yellow!25}27.078 & \cellcolor{yellow!25}28.88 & \cellcolor{orange!35}29.805 & \cellcolor{yellow!25}25.7447 \\
                       &                        & 30k & 24.813 & \cellcolor{yellow!25}21.2706 & \cellcolor{yellow!25}26.985 & \cellcolor{yellow!25}26.262 & \cellcolor{red!25}22.619 & \cellcolor{orange!35}31.467 & \cellcolor{yellow!25}28.982 & 31.094 & \cellcolor{orange!35}32.213 & \cellcolor{yellow!25}27.3006 \\
\cmidrule{2-13}
                       & \multirow{2}{*}{3DIMQS}  & 7k  & 23.309 & 19.782 & 25.669 & 24.893 & 21.591 & 28.986 & 26.71 & 28.211 & 28.132 & 25.2536 \\
                       &                        & 30k & \cellcolor{yellow!25}24.885 & 20.838 & 26.838 & 26.169 & 22.389 & 31.185 & 28.547 & 30.34 & 30.173 & 26.8182 \\
\midrule
\multirow{10}{*}{LPIPS} & \multirow{2}{*}{3DGS}  & 7k  & \cellcolor{orange!35}0.328   & \cellcolor{orange!35}0.422   & \cellcolor{orange!35}0.16   & \cellcolor{orange!35}0.294   & \cellcolor{orange!35}0.418   & \cellcolor{orange!35}0.26   &\cellcolor{orange!35} 0.2455   & \cellcolor{orange!35}0.157   & \cellcolor{orange!35}0.236   & \cellcolor{orange!35}0.2801   \\
                       &                        & 30k & \cellcolor{orange!35}0.211   & \cellcolor{yellow!25}0.342   & \cellcolor{red!25}0.108   & \cellcolor{orange!35}0.217   & \cellcolor{orange!35}0.33   & \cellcolor{orange!35}0.22   & \cellcolor{orange!35}0.202   & \cellcolor{red!25}0.126   & \cellcolor{orange!35}0.205   & \cellcolor{orange!35}0.2178   \\
\cmidrule{2-13}
                       & \multirow{2}{*}{3DRCS} & 7k  & \cellcolor{red!25}0.3197   & \cellcolor{red!25}0.41   & \cellcolor{red!25}0.156   & \cellcolor{red!25}0.283   & \cellcolor{red!25}0.406   & \cellcolor{red!25}0.257   & \cellcolor{red!25}0.241   & \cellcolor{red!25}0.155   & \cellcolor{red!25}0.231   & \cellcolor{red!25}0.2732   \\
                       &                        & 30k & \cellcolor{red!25}0.2079   & \cellcolor{orange!35}0.334   & \cellcolor{red!25}0.108   & \cellcolor{red!25}0.212   & \cellcolor{red!25}0.324   & \cellcolor{red!25}0.219   & \cellcolor{red!25}0.2   & \cellcolor{red!25}0.126   & \cellcolor{red!25}0.203   & \cellcolor{red!25}0.2148   \\
\cmidrule{2-13}
                       & \multirow{2}{*}{3DHCS} & 7k  & 0.403   & 0.458   & 0.233   & 0.35   & 0.46   & 0.2787   & 0.256   & 0.168   & 0.238   & 0.3161   \\
                       &                        & 30k & 0.281   & 0.368   & 0.156   & 0.26   & 0.375   & 0.234   & 0.208   & \cellcolor{yellow!25}0.132   & \cellcolor{yellow!25}0.206   & 0.2466   \\
\cmidrule{2-13}
                       & \multirow{2}{*}{3DSS}  & 7k  & 0.374   & \cellcolor{yellow!25}0.444   & 0.206   & \cellcolor{yellow!25}0.328   & 0.444   & 0.271   & \cellcolor{yellow!25}0.25   & \cellcolor{yellow!25}0.164   & \cellcolor{yellow!25}0.237   & 0.302   \\
                       &                        & 30k & 0.249   & 0.359   & \cellcolor{yellow!25}0.135   & \cellcolor{yellow!25}0.241   & \cellcolor{yellow!25}0.356   & 0.229   & \cellcolor{yellow!25}0.205   & \cellcolor{orange!35}0.129   & \cellcolor{yellow!25}0.206   & 0.2343   \\
\cmidrule{2-13}
                       & \multirow{2}{*}{3DIMQS}  & 7k  & \cellcolor{yellow!25}0.355   & 0.462   & \cellcolor{yellow!25}0.18   & 0.335   & \cellcolor{yellow!25}0.443   & \cellcolor{yellow!25}0.268   & 0.253   & 0.17   & 0.242   & \cellcolor{yellow!25}0.3008   \\
                       &                        & 30k & \cellcolor{yellow!25}0.238   & \cellcolor{red!25}0.281   & \cellcolor{orange!35}0.126   & 0.248   & 0.358   & \cellcolor{yellow!25}0.228   & 0.21   & 0.134   & 0.21   & \cellcolor{yellow!25}0.2258   \\
\midrule
\multirow{10}{*}{SSIM}  & \multirow{2}{*}{3DGS}  & 7k  & \cellcolor{orange!35}0.669   & \cellcolor{orange!35}0.523   & \cellcolor{orange!35}0.826   & \cellcolor{orange!35}0.722   & \cellcolor{orange!35}0.586   & \cellcolor{orange!35}0.894   & \cellcolor{orange!35}0.875   & \cellcolor{orange!35}0.903   & \cellcolor{orange!35}0.92   & \cellcolor{orange!35}0.7686   \\
                       &                        & 30k & \cellcolor{orange!35}0.763   & \cellcolor{orange!35}0.6   & \cellcolor{orange!35}0.863   & \cellcolor{orange!35}0.769   & \cellcolor{red!25}0.633   & \cellcolor{orange!35}0.917   & \cellcolor{orange!35}.903   & \cellcolor{red!25}0.9256   & \cellcolor{orange!35}0.939   & \cellcolor{orange!35}0.8125   \\
\cmidrule{2-13}
                       & \multirow{2}{*}{3DRCS} & 7k  & \cellcolor{red!25}0.6735   & \cellcolor{red!25}0.5302   & \cellcolor{red!25}0.8297   & \cellcolor{red!25}0.7283   & \cellcolor{red!25}0.5913   & \cellcolor{red!25}0.8953   & \cellcolor{red!25}0.8786   & \cellcolor{red!25}0.905   & \cellcolor{red!25}0.9214   & \cellcolor{red!25}0.7726   \\
                       &                        & 30k & \cellcolor{red!25}0.7641   & \cellcolor{red!25}0.6039   & \cellcolor{red!25}0.864   & \cellcolor{red!25}0.7703   & \cellcolor{orange!35}0.6325   & \cellcolor{red!25}0.9176   & \cellcolor{red!25}0.906   & \cellcolor{red!25}0.9256   & \cellcolor{red!25}0.94   & \cellcolor{red!25}0.8137   \\
\cmidrule{2-13}
                       & \multirow{2}{*}{3DHCS} & 7k  & 0.602   & 0.478   & 0.771   & 0.668   & 0.557   & 0.883   & 0.866   & 0.894   & 0.915   & 0.7371   \\
                       &                        & 30k & 0.706   & 0.566   & 0.827   & 0.7337   & 0.61   & 0.91   & \cellcolor{yellow!25}0.898   & \cellcolor{yellow!25}0.92   & 0.9355   & 0.7895   \\
\cmidrule{2-13}
                       & \multirow{2}{*}{3DSS}  & 7k  & 0.631   & \cellcolor{yellow!25}0.498   & 0.795   & \cellcolor{yellow!25}0.693   & \cellcolor{yellow!25}0.571   & \cellcolor{yellow!25}0.889   & \cellcolor{yellow!25}0.873   & \cellcolor{yellow!25}0.899   & \cellcolor{yellow!25}0.918   & \cellcolor{yellow!25}0.7508   \\
                       &                        & 30k & 0.736   & \cellcolor{yellow!25}0.583   & \cellcolor{yellow!25}0.845   & \cellcolor{yellow!25}0.752   & \cellcolor{yellow!25}0.623   & \cellcolor{yellow!25}0.914   & \cellcolor{orange!35}0.903   & \cellcolor{orange!35}0.923   & \cellcolor{yellow!25}0.938   & \cellcolor{yellow!25}0.8008   \\
\cmidrule{2-13}
                       & \multirow{2}{*}{3DIMQS}  & 7k  & \cellcolor{yellow!25}0.637   & 0.474   & \cellcolor{yellow!25}0.8   & 0.678   & 0.564   & 0.884   & 0.862   & 0.889   & 0.911   & 0.7443   \\
                       &                        & 30k & \cellcolor{yellow!25}0.739   & 0.555   & 0.843   & 0.742   & 0.615   & 0.909   & 0.894   & 0.914   & 0.928   & 0.7932   \\

\midrule
\multirow{10}{*}{Memory (MB)}  & \multirow{2}{*}{3DGS}  & 7k  & 753   & 503   & 840   & 844   & 502   & 259   & 229   & \cellcolor{yellow!25}358   & 
\cellcolor{yellow!25}252   & 504   \\
                       &                        & 30k & 1135   & 658   & 950   & 1024   & 727   & 313   & \cellcolor{yellow!25}250   & \cellcolor{yellow!25}384   & \cellcolor{yellow!25}258   & 633   \\
\cmidrule{2-13}
                       & \multirow{2}{*}{3DRCS} & 7k  & 789   & 516   & 842   & 879   & 528   & 255   & 238   & 370   & 275   & 521   \\
                       &                        & 30k & 1120   & 668   & 953   & 1010   & 749   & 343   & 276   & 412   & 279   & 645   \\
\cmidrule{2-13}
                       & \multirow{2}{*}{3DHCS} & 7k  & \cellcolor{orange!35}570   & \cellcolor{orange!35}412   & \cellcolor{orange!35}630   & \cellcolor{orange!35}699   & \cellcolor{orange!35}413   & \cellcolor{orange!35}211   & \cellcolor{yellow!25}228   & \cellcolor{orange!35}322   & \cellcolor{orange!35}211   & \cellcolor{orange!35}410  \\
                       &                        & 30k & \cellcolor{orange!35}858   & \cellcolor{orange!35}578   & \cellcolor{orange!35}739   & \cellcolor{orange!35}867   & \cellcolor{orange!35}632   & \cellcolor{orange!35}256   & \cellcolor{orange!35}242   & \cellcolor{orange!35}372   & \cellcolor{orange!35}245   & \cellcolor{orange!35}532   \\
\cmidrule{2-13}
                       & \multirow{2}{*}{3DSS}  & 7k  & \cellcolor{yellow!25}592   & \cellcolor{yellow!25}424   & \cellcolor{yellow!25}672   & \cellcolor{yellow!25}706   & \cellcolor{yellow!25}396   & \cellcolor{yellow!25}240   & \cellcolor{orange!35}223   & 364   & 266   & \cellcolor{yellow!25}431   \\
                       &                        & 30k & \cellcolor{yellow!25}948   & \cellcolor{yellow!25}586   & \cellcolor{yellow!25}800   & \cellcolor{yellow!25}901   & \cellcolor{yellow!25}602   & \cellcolor{yellow!25}281   & \cellcolor{yellow!25}250   & 393   & 272   & \cellcolor{yellow!25}559   \\
\cmidrule{2-13}
                       & \multirow{2}{*}{3DIMQS}  & 7k  & \cellcolor{red!25}394   & \cellcolor{red!25}263   & \cellcolor{red!25}484   & \cellcolor{red!25}478   & \cellcolor{red!25}245   & \cellcolor{red!25}153   & \cellcolor{red!25}140   & \cellcolor{red!25}223   & \cellcolor{red!25}149   & \cellcolor{red!25}281   \\
                       &                        & 30k & \cellcolor{red!25}608   & \cellcolor{red!25}373   & \cellcolor{red!25}518   & \cellcolor{red!25}611   & \cellcolor{red!25}375   & \cellcolor{red!25}182   & \cellcolor{red!25}151   & \cellcolor{red!25}238   & \cellcolor{red!25}153   & \cellcolor{red!25}356   \\

\midrule
\multirow{10}{*}{Training time (s)}  & \multirow{2}{*}{3DGS}  & 7k  & \cellcolor{yellow!25}181   & 160   & 212   & 173   & 163   & \cellcolor{yellow!25}225   & \cellcolor{yellow!25}240   & \cellcolor{yellow!25}265   & 217   & 204   \\
                       &                        & 30k & 1378   & 920   & 1302   & 1149   & 1020   & 1178   & 1107   & 1313   & 949   & 1146   \\
\cmidrule{2-13}
                       & \multirow{2}{*}{3DRCS} & 7k  & 193   & 155   & 218   & 182   & 159   & \cellcolor{red!25}194   & \cellcolor{red!25}205   & \cellcolor{red!25}239   & \cellcolor{orange!35}192   & \cellcolor{orange!35}193   \\
                       &                        & 30k & 1581   & 995   & 1400   & 1292   & 1098   & \cellcolor{orange!35}1087   & \cellcolor{orange!35}1010   & \cellcolor{orange!35}1247   & \cellcolor{orange!35}876   & 1176   \\
\cmidrule{2-13}
                       & \multirow{2}{*}{3DHCS} & 7k  & \cellcolor{orange!35}164   & \cellcolor{red!25}145   & \cellcolor{orange!35}199   & \cellcolor{orange!35}158   & \cellcolor{red!25}145   & \cellcolor{orange!35}217   & \cellcolor{orange!35}223   & \cellcolor{orange!35}252   & \cellcolor{yellow!25}210   & \cellcolor{red!25}172   \\
                       &                        & 30k & \cellcolor{orange!35}1103   & \cellcolor{orange!35}806   & \cellcolor{orange!35}1035   & \cellcolor{orange!35}996   & \cellcolor{orange!35}848   & \cellcolor{red!25}1002   & \cellcolor{red!25}934   & \cellcolor{red!25}1181   & \cellcolor{red!25}870   & \cellcolor{orange!35}975   \\
\cmidrule{2-13}
                       & \multirow{2}{*}{3DSS}  & 7k  & 182   & \cellcolor{yellow!25}153   & \cellcolor{yellow!25}202   & \cellcolor{yellow!25}169   & \cellcolor{yellow!25}157   & 233   & 243   & 281   & \cellcolor{red!25}185   & 201   \\
                       &                        & 30k & \cellcolor{yellow!25}1324   & \cellcolor{yellow!25}879   & \cellcolor{yellow!25}1215   & \cellcolor{yellow!25}1093   & \cellcolor{yellow!25}944   & 1200   & \cellcolor{yellow!25}1106   & 1413   & 1001   & \cellcolor{yellow!25}1130   \\
\cmidrule{2-13}
                       & \multirow{2}{*}{3DIMQS}  & 7k  & \cellcolor{red!25}161   & \cellcolor{orange!35}146   & \cellcolor{red!25}186   & \cellcolor{red!25}152   & \cellcolor{orange!35}155   & 238   & 263   & 269   & 223   & \cellcolor{yellow!25}199   \\
                       &                        & 30k & \cellcolor{red!25}1004   & \cellcolor{red!25}704   & \cellcolor{red!25}993   & \cellcolor{red!25}842   & \cellcolor{red!25}799   & \cellcolor{yellow!25}1105   & 1049   & \cellcolor{yellow!25}1261   & \cellcolor{yellow!25}880   & \cellcolor{red!25}960   \\
\bottomrule
\end{tabular}
\caption{\textbf{Performance metrics across various scenes.} Red color denotes best performance, while yellow denotes third-best. Higher values are better for PSNR, SSIM, and lower values are better for LPIPS, Memory, and Training Time. Here we present a detailed breakdown of results for all scenes from Mip-NeRF360 \cite{Mip-nerf_360} dataset for 7k and 30k iterations. The performance of each model is heavily affected by the nature of the scene. Some values may differ from the main table due to stochastic processes, as these results are from a single instance of a full evaluation experiment series. In contrast, Table~\ref{Table:FinalResultTable} presents the mean results averaged across multiple experiments.}
\label{table:SupTable1}
\end{table*}

%%%%%%%%%%%%%%%%%%%%%%%%%%%%%%%%%%%%%%%%%%%%%%%%%%%%%%

\newpage\begin{table*}[t]
\footnotesize
\centering
% \caption{\textbf{Performance metrics across various scenes.} Red color denotes best performance, while yellow denotes third-best. Higher values are better for PSNR, SSIM, and lower values are better for LPIPS, Memory, and Training Time. Here we present a detailed breakdown of results for all scenes from Tanks\&Temples \cite{Tanks&Temples2017} and Deep Blending \cite{DeepBlending2018} datasets for 7k and 30k iterations. The performance of each model is heavily affected by the nature of the scene. Some values may differ from the main table due to stochastic processes, as these results are from a single instance of a full evaluation experiment series. In contrast, Table~\ref{Table:FinalResultTable} presents the mean results averaged across multiple experiments.}
\begin{tabular}{llc|cc|c|cc|c}
\toprule
% Metric & Model & Step & Truck & Train & Mean & DrJohnson & Playroom & Mean \\
% \midrule
Metric & Model & Step & \multicolumn{3}{c|}{Tanks\&Temples} & \multicolumn{3}{c}{Deep Blending} \\
\cmidrule{4-9}
& & & Truck & Train & Mean & DrJohnson & Playroom & Mean \\
\midrule
\multirow{10}{*}{PSNR (dB)} & \multirow{2}{*}{3DGS}  & 7k  & \cellcolor{orange!35}23.933 & \cellcolor{red!25}19.795 & \cellcolor{orange!35}21.784 & \cellcolor{red!25}27.609 & \cellcolor{orange!35}29.354 & \cellcolor{orange!25}28.4215 \\
                       &                        & 30k & \cellcolor{red!25}25.481 & \cellcolor{red!25}22.201 & \cellcolor{red!25}23.771 & \cellcolor{red!25}29.493 & \cellcolor{orange!35}29.976 & \cellcolor{red!25}29.6645 \\
\cmidrule{2-9}
                       & \multirow{2}{*}{3DRCS} & 7k  & \cellcolor{red!25}24.026 & \cellcolor{orange!35}19.758 & \cellcolor{red!25}21.882 & \cellcolor{orange!35}27.437 & \cellcolor{red!25}29.417 & \cellcolor{red!25}28.477 \\
                       &                        & 30k & \cellcolor{orange!35}25.314 & \cellcolor{orange!35}22.077 & \cellcolor{orange!35}23.6355 & \cellcolor{yellow!25}29.35 & \cellcolor{red!25}29.981 & \cellcolor{yellow!25}29.6355 \\
\cmidrule{2-9}
                       & \multirow{2}{*}{3DHCS} & 7k  & 22.851 & 19.391 & 21.071 & 26.844 & 28.965 & 27.9345 \\
                       &                        & 30k & 24.561 & 21.721 & 23.108 & 29.003 & 29.772 & 29.3875 \\
\cmidrule{2-9}
                       & \multirow{2}{*}{3DSS}  & 7k  & \cellcolor{yellow!25}23.44 & \cellcolor{yellow!25}19.658 & \cellcolor{yellow!25}21.589 & \cellcolor{yellow!25}27.371 & \cellcolor{red!25}29.417 & \cellcolor{yellow!25}28.394 \\
                       &                        & 30k & \cellcolor{yellow!25}25.03 & \cellcolor{yellow!25}21.973 & \cellcolor{yellow!25}23.5065 & \cellcolor{orange!35}29.414 & \cellcolor{yellow!25}29.955 & \cellcolor{orange!35}29.6645 \\
\cmidrule{2-9}
                       & \multirow{2}{*}{3DIMQS}  & 7k  & 23.3 & 19.43 & 21.365 & 27.279 & 28.904 & 28.0915 \\
                       &                        & 30k & 24.83 & 21.856 & 23.343 & 29.239 & 29.769 & 29.504 \\
\midrule
\multirow{10}{*}{LPIPS} & \multirow{2}{*}{3DGS}  & 7k  & \cellcolor{orange!35}0.197 & \cellcolor{orange!35}0.318 & \cellcolor{orange!35}0.2515 & \cellcolor{orange!35}0.318 & \cellcolor{orange!35}0.284 & \cellcolor{orange!35}0.301 \\
                       &                        & 30k & \cellcolor{orange!35}0.144 & \cellcolor{orange!35}0.199 & \cellcolor{orange!35}0.1725 & \cellcolor{red!25}0.237 & \cellcolor{red!25}0.243 & \cellcolor{red!25}0.24 \\
\cmidrule{2-9}
                       & \multirow{2}{*}{3DRCS} & 7k  & \cellcolor{red!25}0.19 & \cellcolor{red!25}0.312 & \cellcolor{red!25}0.251 & \cellcolor{red!25}0.3178 & \cellcolor{red!25}0.282 & \cellcolor{red!25}0.2999 \\
                       &                        & 30k & \cellcolor{red!25}0.1423 & \cellcolor{red!25}0.196 & \cellcolor{red!25}0.1661 & \cellcolor{orange!35}0.238 & \cellcolor{orange!35}0.2435 & \cellcolor{orange!35}0.2407 \\
\cmidrule{2-9}
                       & \multirow{2}{*}{3DHCS} & 7k  & 0.235 & 0.354 & 0.2945 & 0.341 & 0.298 & 0.3195 \\
                       &                        & 30k & 0.169 & 0.231 & 0.2 & 0.25 & 0.256 & 0.253 \\
\cmidrule{2-9}
                       & \multirow{2}{*}{3DSS}  & 7k  & 0.218 & \cellcolor{yellow!25}0.334 & \cellcolor{yellow!25}0.276 & \cellcolor{yellow!25}0.325 & \cellcolor{yellow!25}0.292 & \cellcolor{yellow!25}0.3085 \\
                       &                        & 30k & 0.159 & \cellcolor{yellow!25}0.218 & 0.1885 & 0.243 & 0.25 & 0.2465 \\
\cmidrule{2-9}
                       & \multirow{2}{*}{3DIMQS}  & 7k & \cellcolor{yellow!25}0.213 & 0.34 & 0.2765 & 0.327 & \cellcolor{yellow!25}0.292 & 0.3095 \\
                       &                        & 30k & \cellcolor{yellow!25}0.154 & 0.221 & \cellcolor{yellow!25}0.1875 & \cellcolor{yellow!25}0.24 & \cellcolor{yellow!25}0.247 & \cellcolor{yellow!25}0.2435 \\
\midrule
\multirow{10}{*}{SSIM}  & \multirow{2}{*}{3DGS}  & 7k  & \cellcolor{orange!35}0.848 & \cellcolor{orange!35}0.719 & \cellcolor{orange!35}0.7815 & \cellcolor{red!25}0.87 & \cellcolor{orange!35}0.894 & \cellcolor{red!25}0.882 \\
                       &                        & 30k& \cellcolor{orange!35}0.88 & \cellcolor{orange!35}0.818 & \cellcolor{orange!35}0.851 & \cellcolor{red!25}0.903 & \cellcolor{red!25}0.903 & \cellcolor{red!25}0.903 \\
\cmidrule{2-9}
                       & \multirow{2}{*}{3DRCS} & 7k  & \cellcolor{red!25}0.8527 & \cellcolor{red!25}0.7242 & \cellcolor{red!25}0.7884 & \cellcolor{orange!35}0.8696 & \cellcolor{red!25}0.8942 & \cellcolor{orange!35}0.8819 \\
                       &                        & 30k & \cellcolor{red!25}0.8818 & \cellcolor{red!25}0.8197 & \cellcolor{red!25}0.8507 & \cellcolor{yellow!25}0.9015 & \cellcolor{yellow!25}0.9013 & \cellcolor{yellow!25}0.9014 \\
\cmidrule{2-9}
                       & \multirow{2}{*}{3DHCS} & 7k  & 0.813 & 0.684 & 0.7485 & 0.856 & 0.887 & 0.8715 \\
                       &                        & 30k & 0.858 & 0.791 & 0.8245 & 0.899 & 0.9 & 0.8995 \\
\cmidrule{2-9}
                       & \multirow{2}{*}{3DSS}  & 7k  & \cellcolor{yellow!25}0.831 & \cellcolor{yellow!25}0.704 & \cellcolor{yellow!25}0.7675 & \cellcolor{yellow!25}0.866 & \cellcolor{yellow!25}0.891 & \cellcolor{yellow!25}0.8785 \\
                       &                        & 30k & \cellcolor{yellow!25}0.869 & \cellcolor{yellow!25}0.803 & \cellcolor{yellow!25}0.836 & \cellcolor{orange!35}0.902 & \cellcolor{orange!35}0.902 & \cellcolor{orange!35}0.902 \\
\cmidrule{2-9}
                       & \multirow{2}{*}{3DIMQS}  & 7k  & 0.827 & 0.693 & 0.76 & 0.862 & 0.886 & 0.874 \\
                       &                        & 30k &0.866 & 0.795 & 0.8305 & \cellcolor{orange!35}0.902 & 0.899 & 0.9005 \\

\midrule
\multirow{10}{*}{Memory (MB)}  & \multirow{2}{*}{3DGS}  & 7k  & 406 & 180 & 293 & 462 & 336 & 399 \\
                       &                        & 30k & 485 & 257 & 371 & 742 & \cellcolor{yellow!25}412 & 577 \\
\cmidrule{2-9}
                       & \multirow{2}{*}{3DRCS} & 7k  & 476 & \cellcolor{orange!35}132 & 304 & 491 & 352 & 421.5 \\
                       &                        & 30k & 548 & \cellcolor{orange!35}181 & 364.5 & 767 & 446 & 606.5 \\
\cmidrule{2-9}
                       & \multirow{2}{*}{3DHCS} & 7k  & \cellcolor{orange!35}344 & \cellcolor{yellow!25}145 & \cellcolor{orange!35}244.5 & \cellcolor{orange!35}355 & \cellcolor{yellow!25}331 & \cellcolor{orange!35}343 \\
                       &                        & 30k & \cellcolor{yellow!25}446 & \cellcolor{yellow!25}230 & \cellcolor{yellow!25}338 & \cellcolor{orange!35}634 & 417 & \cellcolor{orange!35}482.5 \\
\cmidrule{2-9}
                       & \multirow{2}{*}{3DSS}  & 7k  & \cellcolor{yellow!25}355 & 156 & \cellcolor{yellow!25}255.5 & \cellcolor{yellow!25}399 & \cellcolor{orange!35}316 & \cellcolor{yellow!25}357.5 \\
                       &                        & 30k & \cellcolor{orange!35}434 & 232 & \cellcolor{orange!35}333 & \cellcolor{yellow!25}682 & \cellcolor{orange!35}400 & \cellcolor{yellow!25}541 \\
\cmidrule{2-9}
                       & \multirow{2}{*}{3DIMQS}  & 7k  & \cellcolor{red!25}236 & \cellcolor{red!25}114 & \cellcolor{red!25}175 & \cellcolor{red!25}299 & \cellcolor{red!25}197 & \cellcolor{red!25}248 \\
                       &                        & 30k & \cellcolor{red!25}291 & \cellcolor{red!25}156 & \cellcolor{red!25}223.5 & \cellcolor{red!25}489 & \cellcolor{red!25}239 & \cellcolor{red!25}364 \\

\midrule
\multirow{10}{*}{Training time (s)}  & \multirow{2}{*}{3DGS}  & 7k  & 132 & \cellcolor{yellow!25}112 & \cellcolor{yellow!25}122 & \cellcolor{orange!35}212 & 176 & 194 \\
                       &                        & 30k & \cellcolor{yellow!25}738 & \cellcolor{yellow!25}631 & \cellcolor{yellow!25}684.5 & 1385 & 1039 & 1212 \\
\cmidrule{2-9}
                       & \multirow{2}{*}{3DRCS} & 7k  & 134 & \cellcolor{orange!35}105 & \cellcolor{orange!35}119.5 & \cellcolor{red!25}193 & \cellcolor{red!25}162 & \cellcolor{red!25}177.5 \\
                       &                        & 30k & 793 & 660 & 726.5 & \cellcolor{yellow!25}1378 & \cellcolor{yellow!25}1023 & \cellcolor{yellow!25}1200.5 \\
\cmidrule{2-9}
                       & \multirow{2}{*}{3DHCS} & 7k  & \cellcolor{red!25}114 & \cellcolor{red!25}103 & \cellcolor{red!25}108.5 & \cellcolor{yellow!25}213 & \cellcolor{yellow!25}169 & \cellcolor{yellow!25}191 \\
                       &                        & 30k & \cellcolor{red!25}604 & \cellcolor{red!25}498 & \cellcolor{red!25}551 & \cellcolor{red!25}1163 & \cellcolor{orange!35}947 & \cellcolor{red!25}1055 \\
\cmidrule{2-9}
                       & \multirow{2}{*}{3DSS}  & 7k  & \cellcolor{yellow!25}131 & 118 & 124.5 & 219 & 176 & 197.5 \\
                       &                        & 30k & 745 & 659 & 702 & 1407 & 1087 & 1247 \\
\cmidrule{2-9}
                       & \multirow{2}{*}{3DIMQS}  & 7k  & \cellcolor{orange!35}129 & 123 & 126 & 216 & \cellcolor{orange!35}164 & \cellcolor{orange!35}190 \\
                       &                        & 30k & \cellcolor{orange!35}624 & \cellcolor{orange!35}602 & \cellcolor{orange!35}613 & \cellcolor{orange!35}1262 & \cellcolor{red!25}918 & \cellcolor{orange!35}1090 \\
\bottomrule
\end{tabular}
\caption{\textbf{Performance metrics across various scenes.} Red color denotes best performance, while yellow denotes third-best. Higher values are better for PSNR, SSIM, and lower values are better for LPIPS, Memory, and Training Time. Here we present a detailed breakdown of results for all scenes from Tanks\&Temples \cite{Tanks&Temples2017} and Deep Blending \cite{DeepBlending2018} datasets for 7k and 30k iterations. The performance of each model is heavily affected by the nature of the scene. Some values may differ from the main table due to stochastic processes, as these results are from a single instance of a full evaluation experiment series. In contrast, Table~\ref{Table:FinalResultTable} presents the mean results averaged across multiple experiments.}
\label{table:SupTable2}
\end{table*}

%%%%%%%%%%%%%%%%%%%%%%%%%%%%%%%%%%%%%%%%%%%%%%%%%%%%%%

%%%%%%%%%%%%%%%%%%%%%%%%%%%%%%%%%%%%%%%%%%%%%%%%%%%%%%
\newpage

% \section{1D Simulations}

\newpage

\newpage

\newpage

\newpage

\newpage

\newpage

\end{document}